\documentclass[a4paper]{book}

\usepackage{fullpage}
\usepackage{amsthm}
\usepackage{amssymb}
\usepackage{amsmath}
\usepackage{hyperref}
\usepackage{cleveref}
\usepackage{xifthen}
\usepackage{xparse}
\usepackage{dsfont}
\usepackage{rotating}
\usepackage{xcolor}
\usepackage{mathtools}
\usepackage{slashbox}
\usepackage{enumitem}

\usepackage{subcaption}

\usepackage{algorithm}
\usepackage{algorithmicx}
\usepackage{algpseudocode}

\usepackage[round,comma]{natbib}
\newcommand{\lr}[1]{\left (#1\right)}
\newcommand{\lrc}[1]{\left \{#1\right\}}
\newcommand{\lrs}[1]{\left [#1 \right]}
\newcommand{\lra}[1]{\left |#1\right|}
\newcommand{\lru}[1]{\left \lceil#1\right\rceil}
\newcommand{\vp}[2]{\left \langle #1 , #2 \right \rangle}
\newcommand{\EEE}[1]{\mathbb E \lrs{#1}}
\newcommand{\EE}{\mathbb E}
\newcommand{\R}{\mathbb R}
\newcommand{\N}{\mathbb N}

\NewDocumentCommand{\E}{o}{\mathbb E\IfValueT{#1}{\lrs{#1}}}
\NewDocumentCommand{\1}{o}{\mathds 1{\IfValueT{#1}{\lr{#1}}}}
\let\P\undefined
\NewDocumentCommand{\P}{o}{\mathbb P{\IfValueT{#1}{\lr{#1}}}}

\newcommand{\Lhat}{\hat L(h,S)}

\newcommand{\ErhoL}{\E_\rho\lrs{L(h)}}
\newcommand{\ErhoLhat}{\E_\rho\lrs{\Lhat}}

\newcommand{\Strain}{S^{\texttt{train}}}
\newcommand{\Sval}{S^{\texttt{val}}}
\newcommand{\Stest}{S^{\texttt{test}}}

\newcommand{\x}{\mathbf x}
\newcommand{\y}{\mathbf y}
\newcommand{\z}{\mathbf z}
\newcommand{\w}{\mathbf w}
\newcommand{\p}{\mathbf p}
\newcommand{\PP}{\mathbf P}
\newcommand{\I}{\mathbf I}
\renewcommand{\u}{\mathbf u}
\renewcommand{\v}{\mathbf v}
\newcommand{\X}{\mathbf X}
\newcommand{\A}{\mathbf A}
\newcommand{\0}{\mathbf 0}

\newcommand{\F}{\mathcal F}
\newcommand{\DD}{\mathcal D}
\newcommand{\HH}{\mathcal H}
\newcommand{\XX}{\mathcal{X}}
\newcommand{\YY}{\mathcal{Y}}
\newcommand{\ZZ}{\mathcal{Z}}
\newcommand{\calS}{\mathcal{S}}
\newcommand{\Z}{\mathcal{Z}}

\newcommand{\hbest}{\hat h_S^*}
\newcommand{\D}{\mathbb{D}}

\newcommand{\nmin}{n_{\texttt{min}}}

\NewDocumentCommand{\hKNN}{o}{\IfValueF{#1}{h_{\texttt{K-NN}}}\IfValueT{#1}{h_{\texttt{#1-NN}}}}

\let\emptyset\varnothing

\newcommand{\Lval}{\hat L^{\text{\normalfont val}}}

\DeclareMathOperator{\sign}{sign}
\DeclareMathOperator{\KL}{KL}
\DeclareMathOperator{\kl}{kl}
\let\H\relax
\DeclareMathOperator{\H}{H}
\DeclareMathOperator{\MV}{MV}
\newcommand{\V}{\mathbb V}
\NewDocumentCommand{\Var}{o}{\V\IfValueT{#1}{\lrs{#1}}}

\newcommand{\FO}{\operatorname{FO}}
\newcommand{\TND}{\operatorname{TND}}
\newcommand{\DIS}{\operatorname{DIS}}

\DeclareMathOperator{\VC}{VC}
\newcommand{\dVC}{d_{\VC}}

\DeclareMathOperator{\FAT}{FAT}
\newcommand{\dfat}{d_{\FAT}}
\newcommand{\lfat}{\ell_{\FAT}}
\newcommand{\Lfat}{L_{\FAT}}
\newcommand{\hatLfat}{\hat L_{\FAT}}

\DeclareMathOperator{\dist}{dist}
\DeclareMathOperator{\diag}{diag}
\newcommand{\ones}{\mathbf 1}

\newcommand{\err}{\ell}
\newcommand{\Err}{\hat L}
\newcommand{\ERR}{L}
\newcommand{\sgn}[1]{\sign\lr{#1}}

\newcommand{\Utrain}{U^{\mathrm{train}}}
\newcommand{\Uval}{U^{\mathrm{val}}}
\newcommand{\nval}{n^{\mathrm{val}}}
\DeclareMathOperator{\B}{B}
\newcommand{\Ex}{\mathcal{E}}
\newcommand{\Pcal}{\mathcal P}

\newtheorem{theorem}{Theorem}[chapter]
\newtheorem{corollary}[theorem]{Corollary}
\newtheorem{lemma}[theorem]{Lemma}
\newtheorem{definition}[theorem]{Definition}

\theoremstyle{definition}
\newtheorem*{example*}{Example}
\newtheorem{example}[theorem]{Example}
\newtheorem*{exercise*}{Exercise}
\newtheorem{exercise}{Exercise}[chapter]

\title{Machine Learning\\The Science of Selection under Uncertainty}

\author{Yevgeny Seldin}

\begin{document}

\frontmatter

\maketitle

\newpage
    \thispagestyle{empty}
    \vspace*{5cm}
\begin{center}
    \emph{To my parents -- Elena Markman and Anatoly Seldin}
\end{center}
\newpage

\tableofcontents

\chapter{Preface}

Learning, whether natural or artificial, is a process of selection. It starts with a set of candidate options and selects the more successful ones. In the case of machine learning the selection is done based on empirical estimates of prediction accuracy of candidate prediction rules on some data. Due to randomness of data sampling the empirical estimates are inherently noisy, leading to selection under uncertainty. The book provides statistical tools to obtain theoretical guarantees on the outcome of selection under uncertainty. We start with concentration of measure inequalities, which are the main statistical instrument for controlling how much an empirical estimate of expectation of a function deviates from the true expectation. The book covers a broad range of inequalities, including Markov's, Chebyshev's, Hoeffding's, Bernstein's, Empirical Bernstein's, Unexpected Bernstein's, kl, and split-kl. We then study the classical (offline) supervised learning and provide a range of tools for deriving generalization bounds, including Occam's razor, Vapnik-Chervonenkis analysis, and PAC-Bayesian analysis. The latter is further applied to derive generalization guarantees for weighted majority votes. After covering the offline setting, we turn our attention to online learning. We present the space of online learning problems characterized by environmental feedback, environmental resistance, and structural complexity. A common performance measure in online learning is regret, which compares performance of an algorithm to performance of the best prediction rule in hindsight, out of a restricted set of prediction rules. We present tools for deriving regret bounds in stochastic and adversarial environments, and under full information and bandit feedback.

\section*{Reading Guide}

The book is used in teaching parts of three courses at the University of Copenhagen: ``Machine Learning A'', ``Machine Learning B'', and ``Online and Reinforcement Learning''. The material is split in the following way:

\textbf{Machine Learning A:} \Cref{ch:Intro}, \Cref{ch:SupervisedLearning}, Sections \ref{sec:Markov}---\ref{sec:Hoeffding}, and Sections \ref{sec:LbS}---\ref{sec:Occam}.

\smallskip

\textbf{Machine Learning B:} Parts of \Cref{ch:CoM} and \Cref{ch:Generalization} not covered in Machine Learning A.

\smallskip

\textbf{Online and Reinforcement Learning:} \Cref{ch:Online}.

\smallskip
Both ``Machine Learning B'' and ``Online and Reinforcement Learning'' assume that the reader is familiar with the material of ``Machine Learning A'', but they are independent of each other and can be read in any order. 

\Cref{ch:Regression} assumes that the reader is familiar with Chapters~\ref{ch:Intro} and \ref{ch:SupervisedLearning}, but otherwise independent of the rest of the material.

\section*{Acknowledgements}

I would like to thank my co-lecturers and students for asking excellent questions and for constantly inspiring me to find better ways to present the material, as well as for fishing out errors and typos in the text. Special thanks goes to my father, Anatoly Seldin, for his lifelong curiosity and engaging discussions, which, among other things, served an inspiration for \Cref{ch:Intro}. 

The readers are more than welcome to propose further suggestions and report any typos to me at \href{mailto:seldin@di.ku.dk}{seldin@di.ku.dk}. Your feedback will serve everyone who will read the book after you.

\mainmatter

\chapter{Introduction}
\label{ch:Intro}

\begin{figure}
    \centering
    \includegraphics[width=0.5\linewidth]{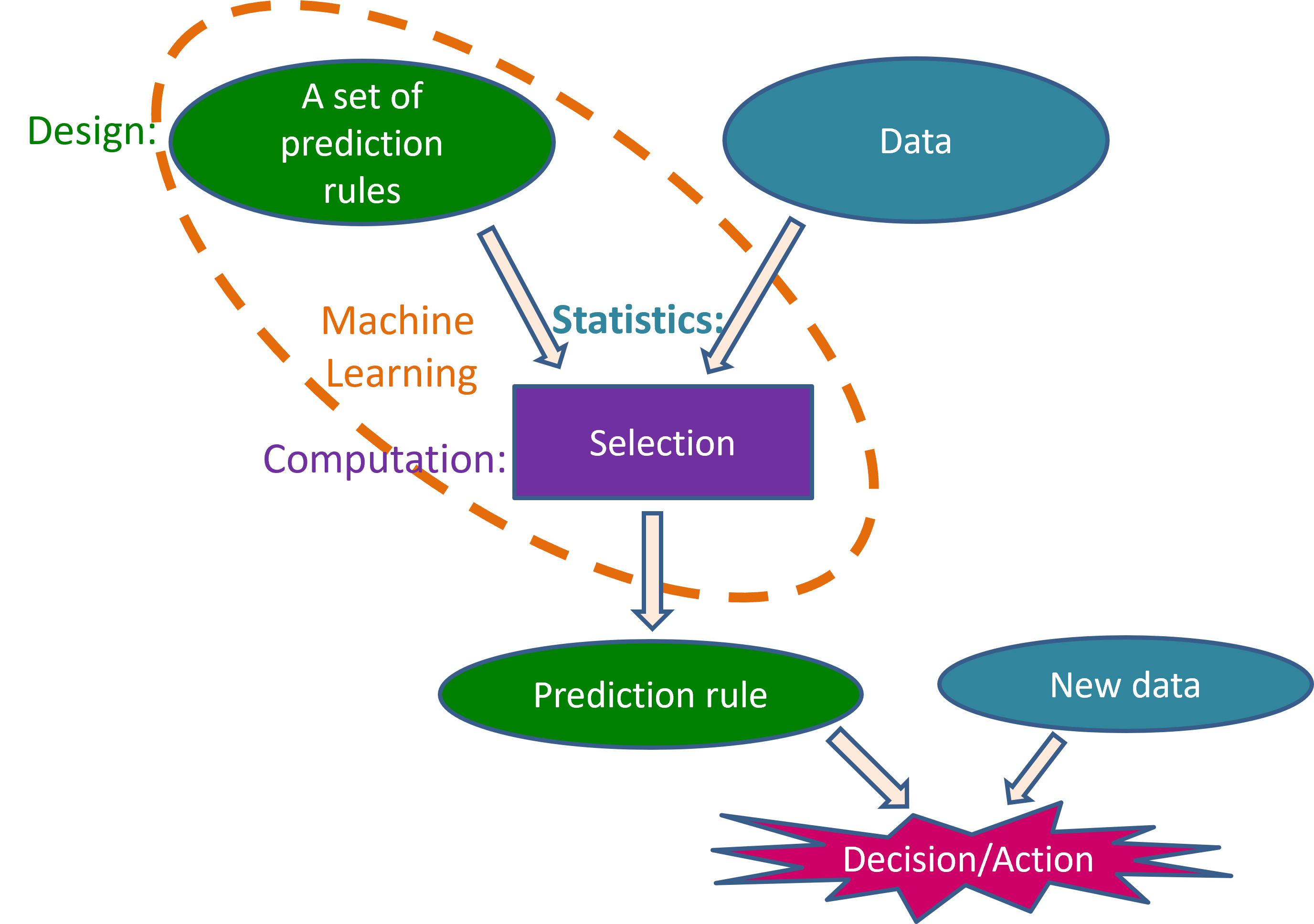}
    \caption{A ``classical'' machine learning process.}
    \label{fig:ML}
\end{figure}

Machine Learning lies at the intersection of Computer Science and Statistics. In a ``classical'' form it starts with a set of potential prediction rules and uses data to select a prediction rule that is expected to provide good predictions on new data, see \Cref{fig:ML}. This process involves three key elements: \emph{design}, \emph{computation}, and \emph{statistics}.

\paragraph{Design} Learning starts with a set of potential prediction rules. It could be a set of parameters defining linear separators, a set of neural networks, decision trees, or anything else, as well as any combination of the above. The set may come with a prioritization (a ``prior knowledge''), which would say that some rules are more likely than others to be good prediction rules. It is obviously desirable for the set to contain a good prediction rule, because it is the set that learning procedure selects from. However, there are additional desiderata dictated by computation and statistics, as discussed next. Design of a good set of candidate prediction rules is normally based on domain knowledge and experience.

\paragraph{Statistics} Having the set of candidate prediction rules the next task is to estimate their quality, and this is where statistics comes in. The estimation is done by using annotated data given to the algorithm. The algorithm can take any prediction rule and calculate how well it preforms on the data provided to the algorithm, and use the outcome as an estimate of how well the same rule would preform on new data. The challenge is that the estimates based on the data are random samples of the expected performance of the algorithm on new data, and so they are inherently uncertain (which is why the book is called ``The science of selection under uncertainty''). They may happen to be better or worse than their expected value, and thus confuse the selection process. Imagine that you had a set of archers and you had to select an archer to send to an arch shooting competition. And imagine that you would ask each archer to shoot once and send the one with the best shot to the competition. For each of the archers the trial shot may happen to be better or worse than their expected performance. So, on the one hand, you want the selection set to be sufficiently large, so that with sufficiently large probability you would have some good archers in the set. But, on the other hand, for any poor archer there is a small probability that on the selection day they would accidentally hit the bull eye and get selected. And if the selection set contains too many poor archers, with \emph{high} probability at least one of them would have a lucky selection day and get selected instead of a good archer, and ruin the final competition. In the context of a learning process, the archers are the prediction rules, their trial shots are their predictions on the training data, and the competition is the prediction of the selected rule on a new data point. And there is a statistical trade-off between starting with a sufficiently rich set of candidate prediction rules in the design phase to have a good candidate in the set, but at the same time not an overly large to control the probability of weak candidates accidentally wining the selection. In other words, even for a poor prediction rule, with small probability the training data may happen to fit the rule well and get predicted better than expected, and if there are too many poor prediction rules, with high probability at least one of them may happen to be ``lucky'' and confuse the selection process.

(We note that it is not the plain number of prediction rules that makes a difference, because if they are sufficiently similar their probabilities of being ``lucky'' may be correlated, but it is ``richness'' of the set, which is discussed in details in \Cref{ch:Generalization}.)

\paragraph{Computation} Finally comes the computational aspect of the selection process. If the set of candidate prediction rules is finite and small, it may be computationally feasible to estimate the quality of all of them. But if the set is large, or even potentially infinite, there is a need in computationally efficient procedures to find candidates that are estimated to deliver good predictions. And this is where Computer Science comes into play.

\paragraph{Design + Statistics + Computation} At the end, all the three elements should fit together. Learning needs to start from a candidate set of prediction rules that is sufficiently rich to include good prediction rules, but, at the same time, not overly rich, so that excessive amount of poor prediction rules would not confuse the selection process. Alternatively, the candidate set may come with prioritization that would give sufficient preference to potentially good prediction rules. And the candidate set has to be sufficiently structured or small to support computationally efficient selection.

\paragraph{} The book focuses primarily on the statistical aspect of the selection process. We provide tools for quantifying, controlling, and reducing the uncertainty, as well as tools for prioritizing prediction rules in the selection process. \Cref{ch:Online} departs from the ``classical'' batch learning paradigm and moves into the realm of online learning, while still maintaining a strong focus on the statistical aspect of taking actions (making selections) under uncertainty.

\chapter{Supervised Learning}
\label{ch:SupervisedLearning}

The most basic and widespread form of machine learning is supervised learning. In the classical batch supervised learning setting the learner is given an annotated sample, which is used to derive a prediction rule for annotating new samples. We start with a simple informal example and then formalize the problem.

Let's say that we want to build a prediction rule that will use the average grade of a student in home assignments, say on a 100-points scale, to predict whether the student will pass the final exam. Such a prediction rule could be used for preliminary filtering of students to be allowed to take the final exam. The annotated sample could be a set of average grades of students from the previous year with indications of whether they have passed the final exam. The prediction rule could take a form of a threshold grade (a.k.a.\ decision stump), above which the student is expected to pass and below fail.

Now assume that we want to take a more refined approach and look into individual grades in each assignment, say, 5 assignments in total. For example, different assignments may have different relevance for the final exam or, maybe, some students may demonstrate progression throughout the course, which would mean that their early assignments should not be weighted equally with the later ones. In the refined approach each student can be represented by a point in a 5-dimensional space. The one-dimensional threshold could be replaced by a separating hyperplane, which separates the 5-dimensional space of grades into a linear subspace, where most students are likely to pass, and the complement, where they are likely to fail. An alternative approach is to look at ``nearest neighbors'' of a student in the space of grades. Given a grade profile of a student (the point representing the student in the 5-dimensional space) we look at students with the closest grade profile and see whether most of them passed or failed. This is known as the \emph{$K$ Nearest Neighbors} algorithm, where $K$ is the number of neighbors we look at. But how many neighbors $K$ should we look at? Considering the extremes gives some intuition about the problem. Taking just one nearest neighbor may be unreliable. For example, we could have a good student that accidentally failed the final exam and then all the neighbors will be marked as ``expected to fail''. Going in the other extreme and taking all the students in as neighbors is also undesirable, because effectively it will ignore the individual profile altogether. So a good value of $K$ should be somewhere between 1 and $n$, where $n$ is the size of the annotated set. But how to find it? Well, read on and you will learn how to approach this question formally.

\section{The Supervised Learning Setting}

We start with a bunch of notations and then illustrate them with examples.
\begin{itemize}
	\item ${\cal X}$ - the sample space.
	\item ${\cal Y}$ - the label space.
	\item $X \in {\cal X}$ - unlabeled sample.
	\item $(X,Y) \in \lr{{\cal X} \times {\cal Y}}$ - labeled sample.
	\item $S = \lrc{(X_1,Y_1),\dots,(X_n,Y_n)}$ - a training set. We assume that $(X_i,Y_i)$ pairs in $S$ are sampled i.i.d.\ according to an unknown, but fixed distribution $p(X,Y)$.
	\item $h:{\cal X} \rightarrow {\cal Y}$ - a hypothesis, which is a function from ${\cal X}$ to ${\cal Y}$.
	\item ${\cal H}$ - a hypothesis set.
	\item $\err(Y',Y)$ - the loss function for predicting $Y'$ instead of $Y$.
	\item $\Err(h,S) = \frac{1}{n} \sum_{i=1}^n \err(h(X_i), Y_i)$ - the empirical loss (a.k.a.\ error or risk) of $h$ on $S$. (In many textbooks $S$ is omitted from the notation and $\Err(h)$ or $\hat L_n(h)$ is used to denote $\Err(h,S)$.)
	\item $\ERR(h) = \EEE{\err(h(X),Y)}$ - the expected loss (a.k.a.\ error or risk) of $h$, where the expectation is taken with respect to $p(X,Y)$.
\end{itemize}

\subsubsection{The Learning Protocol} 
The classical supervised learning acts according to the following protocol:
\begin{enumerate}
	\item The learner gets a training set $S$ of size $n$ sampled i.i.d.\ according to $p(X,Y)$.
	\item The learner returns a prediction rule $h$.
	\item New instances $(X,Y)$ are sampled according to $p(X,Y)$, but only $X$ is observed and $h$ is used to predict the unobserved $Y$.
\end{enumerate}

\noindent
\emph{\textbf{The goal of the learner is to return $h$ that minimizes $\ERR(h)$, which is the expected error on new samples.}
}

\subsubsection{Examples - Sample and Label Spaces}

Let's say that we want to predict person's height based on age, gender, and weight. Then $\mathcal{X} = \N \times \lrc{\pm 1} \times \R$ and $\mathcal{Y} = \R$. If we want to predict gender based on age, weight, and height, then $\mathcal{X} = \N \times \R \times \R$ and $\mathcal{Y} = \lrc{\pm 1}$. If we want to predict the height of a baby at the age of 4 years based on his or her height at the ages of 1, 2, and 3 years, then $\mathcal{X} = \R^3$ and $\mathcal{Y} = \R$.

\subsection{Classification, Regression, and Other Supervised Learning Problems}

The most widespread forms of supervised learning are classification and regression. We also mention a few more, mainly to show that the supervised learning setting is much richer.

\paragraph{Classification} A supervised learning problem is a classification problem when the output (label) space ${\cal Y}$ is binary. The goal of the learning algorithm is to separate between two classes: yes or no; good or bad; healthy or sick; male or female; etc. Most often the translation of the binary label into numerical representation is done by either taking ${\cal Y} = \lrc{\pm 1}$ or ${\cal Y} = \lrc{0,1}$. Sometimes the setting is called \emph{binary classification} to emphasize that ${\cal Y}$ takes just two values.

\paragraph{Regression} A supervised learning problem is a regression problem when the output space $\mathcal{Y} = \R$. For example, prediction of person's height would be a regression problem.

\paragraph{Multiclass Classification} When ${\cal Y}$ consists of a finite and typically unordered and relatively small set of values, the corresponding supervised learning problem is called multiclass classification. For example, prediction of a study program a student will apply for based on his or her grades would be a multiclass classification problem. Finite ordered output spaces, for example, prediction of age or age group can also be modeled as multiclass classification, but it may be possible to exploit the structure of $\mathcal Y$ to obtain better solutions. For example, it may be possible to exploit the fact that ages 22 and 23 are close together, whereas 22 and 70 are far apart; therefore, it may be possible to share some information between close ages, as well as exploit the fact that predicting 22 instead of 23 is not such a big mistake as predicting 22 instead of 70. Depending on the setting, it may be preferable to model prediction of ordered sets as regression rather than multiclass classification.

\paragraph{Structured Prediction} Consider the problem of machine translation. An algorithm gets a sentence in English as an input and should produce a sentence in Danish as an output. In this case the output (the sentence in Danish) is not merely a number, but a structured object and such prediction problems are known as structured prediction. 

\subsection{The Loss Function $\ell(Y',Y)$}

The loss function (a.k.a.\ the error function) encodes how much the user of an algorithm cares about various kinds of mistakes. Most literature on binary classification, including this book, uses the \emph{zero-one loss} defined by 
\[
\ell(Y',Y) = \1[Y'\neq Y] = \begin{cases}1,&\text{if } Y'\neq Y\\0,&\text{otherwise,}\end{cases}
\]
where $\1$ is the indicator function. Common loss functions in regression are the \emph{square loss}
\[
\ell(Y',Y) = (Y'-Y)^2
\]
and the \emph{absolute loss}
\[
\ell(Y',Y) = |Y' - Y|.
\]

The above loss functions are convenient general choices, but not necessarily the right choice for a particular application. For example, imagine that you design an algorithm for fire alarm that predicts ``fire / no fire". Assume that the cost of a house is 3,000,000 DKK and the cost of calling in a fire brigade is 2,000 DKK. Then the loss function would be
\begin{center}
\begin{tabular}{cc|c|c}
& \backslashbox{Y'}{Y} & no fire & fire\\
\cline{2-4}
$\ell(Y',Y) = $~~~~ & no fire & 0 & 3,000,000\\
\cline{2-4}
& fire & 2,000 & 0
\end{tabular}
\end{center}
The loss for making the correct prediction is zero, but the loss of \emph{false positive} (predicting fire when in reality there is no fire) and \emph{false negative} (predicting no fire when the reality is fire) are not symmetric anymore.

Pay attention that the loss depends on how the predictions are used and the loss table depends on the user. For example, if the same alarm is installed in a house that is worth 10,000,000 DKK, the ratio between the cost of false positives and false negatives will be very different and, as a result, the optimal prediction strategy will not necessarily be the same.

\section{$K$ Nearest Neighbors for Binary Classification}

One of the simplest algorithms for binary classification is $K$ Nearest Neighbors ($K$-NN). The algorithm is based on an externally provided distance function $d(\x,\x')$ that computes distances between pairs of points $\x$ and $\x'$. For example, for points in $\R^d$ the distance could be the Euclidean distance $d(\x,\x') = \|\x - \x'\| = \sqrt{\sum_{j=1}^d (x_j - x'_j)^2} = \sqrt{(\x-\x')^T (\x-\x')}$, where $\x = (x_1,\dots,x_d)$ and $x_j$ is the $j$-th coordinate of vector $\x$. Other choices of distance measures are possible and, in general, lead to different predictions. The choice of the distance measure $d(\x,\x')$ is the key for success or failure of $K$-NN, but we leave the topic of selection of $d$ outside the scope of the book.

$K$-NN algorithm takes as input a set of training points $S = \lrc{(\x_1,y_1),\dots,(\x_n,y_n)}$ and predicts the label of a target point $\x$ based on the majority vote of $K$ points from $S$, which are the closest to $\x$ in terms of the distance measure $d(\x_i,\x)$.

\begin{algorithm}
\caption{$K$ Nearest Neighbors ($K$-NN) for Binary Classification with $\mathcal Y = \lrc{\pm 1}$}
\begin{algorithmic}[1]
\State {\bf Input:} A set of labeled points $\lrc{(\x_1,y_1),\dots,(\x_n,y_n)}$ and a target point $\x$ that has to be classified.

\State Calculate the distances $d_i = d(\x_i, \x)$.

\State\label{step:order}Sort $d_i$-s in ascending order and let $\sigma: \lrc{1,\dots,n} \to \lrc{1,\dots,n}$ be the corresponding permutation of indices. In other words, for any pair of indices $i < j$ we should have $d_{\sigma(i)} \leq d_{\sigma(j)}$.

\State The output of $K$-NN is $y = \sgn{\sum_{i=1}^K y_{\sigma(i)}}$. It is the majority vote of $K$ points that are the closest to $\x$. Note that we can calculate the output of $K$-NN for all $K$ in one shot.
\end{algorithmic}
\end{algorithm}

The ordering of $d_i$-s in Step \ref{step:order} is identical to the ordering of $d_i^2$ and for the Euclidean distance we can save the computation of the square root by working with squared distances.

The hypothesis space ${\cal H}$ is implicit in the $K$-NN algorithm. It is the space of all possible partitions of the sample space ${\cal X}$. The output hypothesis $h$ is parametrized by all training points $h_S = h_{\lrc{(\x_1,y_1),\dots,(\x_n,y_n)}}$. In the sequel we will see other prediction rules that operate with more explicit hypothesis spaces, for example, a space of all linear separators.

\subsection{How to Pick $K$ in $K$-NN?}

One of the key questions in $K$-NN is how to pick $K$. It is instructive to consider the extreme cases to gain some intuition. In $1$-NN the prediction is based on a single sample $\lr{\x_i, y_i}$ which happens to be closest to the target point $\x$. This may not be the best thing to do. Imagine that you are admitted to a hospital and a diagnostic system determines whether you are healthy or sick based on a single annotated patient that has the symptoms closest to yours (in distance measure $d$). You would likely prefer to be diagnosed based on the majority of diagnoses of several patients with similar symptoms. At the other extreme, in $n$-NN, where $n$ is the number of samples in $S$, the prediction is based on the majority of labels $y_i$ within the sample $S$, without even taking any particular $\x$ into account. So the desirable $K$ is somewhere between $1$ and $n$, but how to find it?

Let $\hKNN$ denote the prediction rule of $K$-NN. As $K$ goes from 1 to $n$, $K$-NN provides $n$ different prediction rules, $\hKNN[1]$, $\hKNN[2]$, $\dots$, $\hKNN[n]$ (or half of that if we only take the odd values of $K$). Recall that we are interested in finding $K$ that minimizes the expected loss $L(\hKNN)$ and that $L(\hKNN)$ is unobserved. We can calculate the empirical loss $\hat L(\hKNN,S)$ for any $K$. However, $\hat L(\hKNN[1],S)$ is always zero\footnote{This is because the closest point in $S$ to a sample point $\x_i$ is $\x_i$ itself and we assume that $S$ includes no identical points with dissimilar labels, which is a reasonable assumption if $\mathcal X = \R^d$.} and in general the empirical error of $K$-NN is an underestimate of its expected error and we need other tools to estimate $L(\hKNN)$. We start developing these tools in the next section and continue throughout the book.

\section{Validation}

Whenever we select a hypothesis $\hbest$ out of a hypothesis set $\HH$ based on empirical performances $\hat L(h,S)$, the empirical performance $\hat L(\hbest,S)$ becomes a biased estimate of $L(\hbest)$. This is clearly observed in 1-NN, where $\hat L(\hKNN[1],S) = 0$, but $L(\hKNN[1])$ is most often not zero (we remind that the hypothesis space in 1-NN is the space of all possible partitions of the sample space $\mathcal{X}$ and $\hKNN[1]$ is the hypothesis that achieves the minimal empirical error in this space). The reason is that when we do the selection we pick $\hbest$ that is best suited for $S$ (it achieves the minimal $\hat L(h,S)$ out of all $h$ in $\HH$). Therefore, from the perspective of $\hbest$ the new samples $(X,Y)$ are not ``similar'' to the samples $(X_i,Y_i)$ in $S$. A bit more precisely, $(X,Y)$ is not exchangeable with $(X_i,Y_i)$, because if we would exchange $(X_i,Y_i)$ with $(X,Y)$ it is likely that $\hbest$, the hypothesis that minimizes $\hat L(h,S)$, would be different. Again, this is very clear in 1-NN: if we change one sample $(X_i,Y_i)$ in $S$ we get a different prediction rule $\hKNN[1]$. We get back to this topic in much more details in Chapter~\ref{ch:Generalization} after we develop some mathematical tools for analyzing the bias in Chapter~\ref{ch:CoM}. For now we present a simple solution for estimating $L(\hbest)$ and motivate why we need the tools from Chapter~\ref{ch:CoM}.

The solution is to split the sample set $S$ into training set $\Strain$ and validation set $\Sval$. We can then find the best hypothesis for the training set, $h^*_{\Strain}$, and validate it on the validation set by computing $\hat L(h^*_{\Strain}, \Sval)$. Note that from the perspective of $h^*_{\Strain}$ the samples in $\Sval$ are exchangeable with any new samples $(X,Y)$. If we exchange $(X_i,Y_i) \in \Sval$ with another sample $(X,Y)$ coming from the same distribution, $h^*_{\Strain}$ will stay the same and in expectation $\E[\ell(h^*_{\Strain}(X_i),Y_i)] = \E[\ell(h^*_{\Strain}(X),Y)]$, meaning that on average $\hat L(h^*_{\Strain}, \Sval)$ will also stay the same (only on average, the exact value may change). Therefore, $\hat L(h^*_{\Strain}, \Sval)$ is an unbiased estimate of $L(h^*_{\Strain})$. (We get back to this point in much more details in Chapter~\ref{ch:Generalization}.)

Now we get to the question of how to split $S$ into $\Strain$ and $\Sval$, and again it is very instructive to consider the extreme cases. Imagine that we keep a single sample for validation and use the remaining $n-1$ samples for training. Let's say that we keep the last sample, $(X_n,Y_n)$, for validation, then $\hat L(h^*_{\Strain}, \Sval) = \ell(h^*_{\Strain}(X_n),Y_n)$ and in the case of zero-one loss it is either zero or one. Even though $\hat L(h^*_{\Strain}, \Sval)$ is an unbiased estimate of $L(h^*_{\Strain})$, it clearly does not represent it well. At the other extreme, if we keep $n-1$ points for validation and use the single remaining point for training we run into a different kind of problem: a classifier trained on a single point is going to be extremely weak. Let's say that we have used the first point, $(X_1,Y_1)$, for training. In the case of $K$-NN classifier, as well as most other classifiers, $h^*_{\Strain}$ will always predict $Y_1$, no matter what input it gets. The validation error $\hat L(h^*_{\Strain}, \Sval)$ will be a very good estimate of $L(h^*_{\Strain})$, but this is definitely not a classifier we want.

So how many samples from $S$ should go into $\Strain$ and how many into $\Sval$? Currently there is no ``gold answer'' to this question, but in Chapters~\ref{ch:CoM} and \ref{ch:Generalization} we develop mathematical tools for intelligent reasoning about it. An important observation to make is that for $h$ independent of $(X,Y)$ the zero-one loss $\ell(h(X),Y)$ is a Bernoulli random variable with bias $\P[\ell(h(X),Y) = 1] = L(h)$. Furthermore, when $h$ is independent of a set of samples $\lrc{(X_1,Y_1),\dots,(X_m,Y_m)}$ (i.e., these samples are not used for selecting $h$), the losses $\ell(h(X_i),Y_i)$ are independent identically distributed (i.i.d.)\ Bernoulli random variables with bias $L(h)$. Therefore, when $\Sval$ is of size $m$, the validation loss $\hat L(h^*_{\Strain}, \Sval)$ is an average of $m$ i.i.d.\ Bernoulli random variables with bias $L(h^*_{\Strain})$. The validation loss $\hat L(h^*_{\Strain}, \Sval)$ is observed, but the expected loss that we are actually interested in is unobserved. One of the key questions that we are interested in is how far $\hat L(h^*_{\Strain}, \Sval)$ can be from $L(h^*_{\Strain})$. We have already seen that $m=1$ is too little. But how large should it be, 10, 100, 1000? Essentially this question is equivalent to asking how many times do we need to flip a biased coin in order to get a satisfactory estimate of its bias. In Chapter~\ref{ch:CoM} we develop concentration of measure inequalities that answer this question.

Another technical question is which samples should go into $\Strain$ and which into $\Sval$? From the theoretical perspective we assume that $S$ is sampled i.i.d.\ and, therefore, it does not matter. We can take the first $n-m$ samples into $\Strain$ and the last $m$ into $\Sval$ or split in any other way. From a practical perspective the samples may actually not be i.i.d.\ and there could be some parameter that has influenced their order in $S$. For example, they could have been ordered alphabetically. Therefore, from a practical perspective it is desirable to take a random permutation of $S$ before splitting, unless the order carries some information we would like to preserve. For example, if $S$ is a time-ordered series of product reviews and we would like to build a classifier that classifies them into positive and negative, we may want to get an estimate of temporal variation and keep the order when we do the split, i.e., train on the earlier samples and validate on the later.

\subsection{Test Set: It's not about how you call it, it's about how you use it!}

Assume that we have split $S$ into $\Strain$ and $\Sval$; we have trained $\hKNN[1],\dots,\hKNN[n]$ on $\Strain$; we calculated $\hat L(\hKNN[1],\Sval), \dots, \hat L(\hKNN[n],\Sval)$ and picked the value $K^*$ that minimizes $\hat L(\hKNN,\Sval)$. Is $\hat L(\hKNN[K$^*$],\Sval)$ an unbiased estimate of $L(\hKNN[K$^*$])$?

This is probably one of the most conceptually difficult points about validation, at least when you encounter it for the first time. While for each $\hKNN$ individually $\hat L(\hKNN,\Sval)$ is an unbiased estimate of $L(\hKNN[K])$, the validation loss $\hat L(\hKNN[K$^*$],\Sval)$ is a \emph{biased} estimate of $L(\hKNN[K$^*$])$. This is because $\Sval$ was used for selection of $K^*$ and, therefore, $\hKNN[K$^*$]$ depends on $\Sval$. So if we want to get an unbiased estimate of $L(\hKNN[K$^*$])$ we have to reserve some ``fresh'' data for that. So we need to split $S$ into $\Strain$, $\Sval$, and $\Stest$; train the $K$-NN classifiers on $\Strain$; pick the best $K^*$ based on $\hat L(\hKNN,\Sval)$; and then compute $\hat L(\hKNN[K$^*$],\Stest)$ to get an unbiased estimate of $L(\hKNN[K$^*$])$.

\paragraph{It's not about how you call it, it's about how you use it!} Some people think that if you call some data a test set it automatically makes loss estimates on this set unbiased. This is not true. Imagine that you have split $S$ into $\Strain$, $\Sval$, and $\Stest$; you trained $K$-NN on $\Strain$, picked the best value $K^*$ using $\Sval$, and estimated the loss of $\hKNN[K$^*$]$ on $\Stest$. And now you are unhappy with the result and you want to try a different learning method, say a neural network. You go through the same steps: you train networks with various parameter settings on $\Strain$, you validate them on $\Sval$, and you pick the best parameter set $\theta^*$ based on the validation loss. Finally, you compute the test loss of the neural network parametrized by $\theta^*$ on $\Stest$. It happens to be lower than the test loss of $K^*$-NN and you decide to go with the neural network. Does the empirical loss of the neural network on $\Stest$ represent an unbiased estimate of its expected loss? No! Why? Because our choice to pick the neural network was based on its superior performance relative to $\hKNN[K$^*$]$ on $\Stest$, so $\Stest$ was used in selection of the neural network. Therefore, there is dependence between $\Stest$ and the hypothesis we have selected, and the loss on $\Stest$ is biased. If we want to get an unbiased estimate of the loss we have to find new ``fresh'' data or reserve such data from the start and keep it in a locker until the final evaluation moment. Alternatively, we can correct for the bias and in Chapter~\ref{ch:Generalization} we will learn some tools for making the correction. The main take-home message is: \textbf{\emph{It is not about how you call a data set, $\Strain$, $\Sval$, or $\Stest$, it is the way you use it which determines whether you get unbiased estimates or not!}} In some cases it is possible to get unbiased estimates or to correct for the bias already with $\Strain$, and sometimes there is bias even on $\Stest$ and we need to correct for that.

\subsection{Cross-Validation}

Sometimes it feels wasteful to use only part of the data for training and part for validation. A \emph{heuristic} way around it is cross-validation. In the standard $N$-fold cross-validation setup the data $S$ are split into $N$ non-overlapping folds $S_1,\dots,S_N$. Then for $i \in \lrc{1,\dots,N}$ we train on all folds except the $i$-th and validate on $S_i$. We then take the average of the $N$ validation errors and pick the parameter that achieves the minimum (for example, the best $K$ in $K$-NN). Finally, we train a model with the best parameter we have selected in the cross-validation procedure (for example the best $K^*$ in $K$-NN) using all the data $S$.

The standard cross-validation procedure described above is a heuristic and has no theoretical guarantees. It is fairly robust and widely used in practice, but it is possible to construct examples, where it fails. In Chapter~\ref{ch:Generalization} we describe a modification of the cross-validation procedure, which comes with theoretical generalization guarantees and is empirically competitive with the standard cross-validation procedure.

\section{Perceptron - Basic Algorithm for Linear Classification}

Linear classification is another basic family of classification strategies. Let ${\cal X} = \R^d$ and ${\cal Y} = \lrc{\pm 1}$. A hyperplane in $\R^d$ is described by a tuple $(\w,b)$, where $\w \in \R^d$ and $b \in \R$. The points $\x$ on the hyperplane are described by the equation 
\[
\w^T \x + b = 0.
\]
A linear classifier $h = (\w,b)$ assigns label $+1$ to all points on the ``positive'' side of the hyperplane and $-1$ on the ``negative'' side of the hyperplane. Specifically,
\[
h(\x) = \sgn{\w^T \x + b}.
\]

\paragraph{Homogeneous classifiers} We distinguish between \emph{homogeneous} linear classifiers and non-homogeneous linear classifiers. A homogeneous linear classifier is described by a hyperplane passing through the origin. From the mathematical point of view it means that $b=0$.

We note that any linear classifier in $\R^d$ can be transformed into a homogeneous linear classifier in $\R^{d+1}$ by the following transformation
\begin{align*}
\x &\to (\x; 1)\\
\lrc{\w, b} &\to (\w; b)
\end{align*}
(where by ``;'' we mean that we append a row to a column vector). In other words, we append ``1'' to the $\x$ vector and combine $\w$ and $b$ into one vector in $\R^{d+1}$. Note that $\w^T \x + b = (\w;b)^T (\x;1)$ and, therefore, the predictions of the transformed model are identical to predictions of the original model. Through this transformation any learning algorithm for homogeneous classifiers can be directly applied to learning non-homogeneous classifiers.

\paragraph{Hypothesis space} The hypothesis space in linear classification is the space of all possible separating hyperplanes. If we are talking about homogeneous linear classifiers then it is restricted to hyperplanes passing through the origin. Thus, for homogeneous linear classifiers ${\cal H} = \R^d$ and for general linear classifiers ${\cal H} = \R^{d+1}$.

\paragraph{Perceptron algorithm} Perceptron is the simplest algorithm for learning homogeneous separating hyperplanes. It operates under the \textbf{\emph{assumption that the data are separable by a homogeneous hyperplane}}, meaning that there exists a hyperplane passing through the origin that perfectly separates positive points from negative.

\begin{algorithm}
\caption{Perceptron}
\begin{algorithmic}[1]
\State {\bf Input:} A training set $\lrc{(\x_1,y_1),\dots,(\x_n,y_n)}$
\State \textbf{Initialization:} $\w_1 = \0$ (where $\0$ is the zero vector)
\State $t = 1$
\While{exists $(\x_{i_t}, y_{i_t})$, such that $y_{i_t} (\w_t^T \x_{i_t}) \leq 0$}\label{step:selection}
\State\label{step:perceptron} $\w_{t+1} = \w_t + y_{i_t} \x_{i_t}$
\State $t = t+1$
\EndWhile
\State \textbf{Return:} $\w_t$
\end{algorithmic}
\end{algorithm}

Note that a point $(\x,y)$ is classified correctly if $y \w^T \x > 0$ and misclassified if $y \w^T \x \leq 0$. Thus, the selection step (line \ref{step:selection} in the pseudocode) picks a misclassified point, as long as there exists such. The update step (line \ref{step:perceptron} in the pseudocode) rotates the hyperplane $\w$, so that the classification is ``improved''. Specifically, the following property is satisfied: if $(\x_{i_t},y_{i_t})$ is the point selected at step $t$ then $y_{i_t} \w_{t+1}^T \x_{i_t} > y_{i_t} \w^T_t \x_{i_t}$ (verification of this property is left as an exercise to the reader). Note this property does not guarantee that after the update $\w_{t+1}$ will classify $(\x_{i_t},y_{i_t})$ correctly. But it will rotate in the right direction and after sufficiently many updates $(\x_{i_t},y_{i_t})$ will end up on the right side of the hyperplane. Also note that while the classification of $(\x_{i_t},y_{i_t})$ is improved, it may go the opposite way for other points. As long as the data are linearly separable, the algorithm will eventually find the separation.

The algorithm does not specify the order in which misclassified points are selected. Two natural choices are sequential and random. We leave it as an exercise to the reader to check which of the two choices leads to faster convergence of the algorithm.

\section{Exercises}

\begin{exercise}[\textit{Make your own}] Imagine that you would like to write a learning algorithm that would predict the final grade of a student in the Machine Learning course based on their profile, for example, their grades in prior courses, their study program, etc. Such an algorithm would have been extremely useful: we would save significant time on grading and predict the final grade when the student just signs up for the course. We expect that the students would also appreciate such service and avoid all the worries about their grades. Anyhow, if you were to make such an algorithm.
\begin{enumerate}
	\item What profile information would you collect and what would be the sample space $\mathcal{X}$?
	\item What would be the label space $\mathcal{Y}$?
	\item How would you define the loss function $\ell(y,\hat{y})$?
	\item Assuming that you want to apply $K$-Nearest-Neighbors, how would you define the distance measure $d(x,x')$?
	\item How would you evaluate the performance of your algorithm? (In terms of the loss function you have defined earlier.)
	\item Assuming that you have achieved excellent performance and decided to deploy the algorithm, would you expect any issues coming up? How could you alleviate them?
\end{enumerate}

There is no single right answer to the question. The main purpose is to help you digest the definitions we are working with. Your answer should be short, no more than 2-3 sentences for each bullet point. For example, it is sufficient to mention 2-3 items for the profile information, you should not make a page-long list.

\end{exercise}

\begin{exercise}[\textit{Digits Classification with $K$ Nearest Neighbors}]
\newcommand{\xtrain}{\x^{\texttt{train}}}
\newcommand{\xtest}{\x^{\texttt{targ}}}
\newcommand{\Xtrain}{\X^{\texttt{train}}}
\newcommand{\Xtest}{\X^{\texttt{targ}}}

In this question you will implement and apply the $K$ Nearest Neighbors learning algorithm to classify handwritten digits. You should make your own implementation (rather than use libraries), but it is allowed to use library functions for vector and matrix operations. Apart from implementation of the $K$-NN algorithm, the question aims to improve your skills of working with vector operations in Python.

\paragraph{Preparation}
\begin{itemize}
	\item Download \texttt{MNIST-5-6-Subset.zip} file.\footnote{The file can be downloaded at \url{https://drive.google.com/file/d/1ztcOra97a6-udEvOB1QbgFXIMypcNXbs/view}. It is a subset of digits '5' and '6' from the famous MNIST dataset \citep{MNIST}.} \\ 
        The file contains:
        \begin{itemize}
            \item \texttt{MNIST-5-6-Subset.txt}
            \item \texttt{MNIST-5-6-Subset-Labels.txt}
            \item \texttt{MNIST-5-6-Subset-Light-Corruption.txt}
            \item \texttt{MNIST-5-6-Subset-Moderate-Corruption.txt}
            \item \texttt{MNIST-5-6-Subset-Heavy-Corruption.txt}
        \end{itemize}
	\item \texttt{MNIST-5-6-Subset.txt} is a space-separated file of real numbers (written as      text).  It contains a $784\times1877$ matrix, written column-by-column (the first 784 numbers in the file correspond to the first column; the next 784 numbers are the second column, and so on).
        \begin{itemize}
            \item Each column in the matrix is a $28\times28$ grayscale image of a digit, stored column-by-column (the first 28 out of 784 values correspond to the first column of the $28\times28$ image, the next 28 values correspond to the second column, and so on). In the appendix you can find a Python script that serves as an illustration of one way to load and visualize the data.
        \end{itemize}
	\item \texttt{MNIST-5-6-Subset-Labels.txt} is a space-separated file of 1877 integers. The numbers label the images in \texttt{MNIST-5-6-Subset.txt} file: the first number (``5'') is the number drawn in the image corresponding to the first column; the second number corresponds to the second column, and so on.
	\item \texttt{Light-Corruption}, \texttt{Moderate-Corruption}, and \texttt{Heavy-Corruption} are corrupted versions of the digits in \texttt{MNIST-5-6-Subset.txt}, the order is preserved. It is a good idea to visualize the corrupted images to get a feeling of the corruption magnitude.
\end{itemize}


\paragraph{Detailed Instructions}
We pursue several goals in this question:
\begin{itemize}
    \item Get your hands on implementation of $K$-NN and practice vector operations in Python.
    \item  Explore fluctuations of the validation error as a function of the size of a validation set. (\textbf{Task\#1})
    \item Explore the impact of data corruption on the optimal value of $K$. (\textbf{Task\#2}, optional)
\end{itemize}

\textbf{IMPORTANT: Please, remember to include axis labels, legends and appropriate titles in your plots!}

\paragraph{Task \#1} 



In order to explore fluctuations of the validation error as a function of the size of the validation set, we use the following construction:
\begin{itemize}
    \item Implement a Python function \texttt{knn(training\_points, training\_labels, test\_points, test\_labels)} that takes as input a $d\times m$ matrix of training points \texttt{training\_points}, where $m$ is the number of training points and $d$ is the dimension of each point ($d=784$ in the case of digits), a vector \texttt{training\_labels} of the corresponding $m$ training labels, a $d\times n$ matrix \texttt{test\_points} of $n$ test points, and their labels \texttt{test\_labels} (you will need to convert the labels from $\lrc{5,6}$ to $\lrc{-1,1}$). The function should return a vector of length $m$, where each element represents the average error of K-NN on the test points for the corresponding value of $K$ for $K\in\lrc{1,\dots,m}$. \textbf{Include a printout of your implementation of the function in the report.} (Only this function, not all of your code, the complete code should be included in the \texttt{.zip} file.) Ideally, the function should have no for-loops, check the practical advice at the end of the question.
    \item Use the first $m$ digits for training the $K$-NN model. Take $m=50$.
    \item Consider five validation sets, where for $i\in\{1,\dots,5\}$ the set $i$ consists of digits $m+(i-1)\times n+1,\dots,m+i\times n$, and where $n$ is the size of each of the five validation sets (we will specify $n$ in a moment). The data split is visualized below.
    
    \includegraphics[width=.95\textwidth]{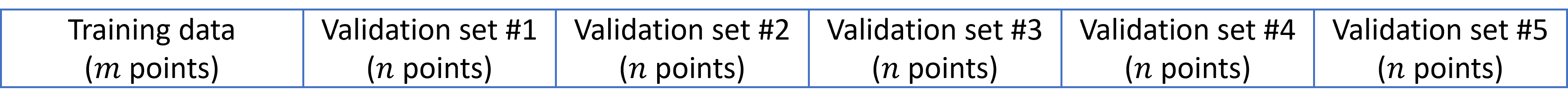}
    \item Calculate the validation error for each of the sets as a function of $K$, for $K\in\{1,\dots,m\}$. Plot the validation error for each of the five validation sets as a function of $K$ in the same figure (you will get five lines in the figure).
    \item Execute the experiment above with $n\in\{10,20,40,80\}$. You will get four figures for the four values of $n$, with five lines in each figure. \textbf{Include these four figures in your report.}
    \item Create a figure where for each $n\in\{10,20,40,80\}$ you plot the variance of the validation error over the five validation sets, as a function of $K$. You will get four lines in this figure, one for each $n$. \textbf{Include this figure in your report.} (Clarification, in case you got confused: fix $n$ and $K$, then you have five numbers corresponding to validation errors on the five validation sets. You should compute the variance of these five values. Now keep $n$ fixed, take $K\in\lrc{1,\dots,m}$, and compute the variance as a function of $K$, i.e., compute it for each $K$ separately. This gives you one line. And then each $n\in\lrc{10,20,40,80}$ gives you a line, so you get four lines.)
    \item What can you say about fluctuations of the validation error as a function of $n$? \textbf{Answer in the report.}
    \item What can you say about the prediction accuracy of $K$-NN as a function of $K$? \textbf{Answer in the report.}
    \item A high-level comment: a more common way of visualizing variation of outcomes of experiment repetitions is to plot the mean and error bars, but this form of visualization makes it too easy for humans to ignore the error bars and concentrate just on the mean, see the excellent book of \citet{Kah11}. The visualization you are asked to provide in this question makes it hard to ignore the variation.
\end{itemize}

\paragraph{Task \#2 (optional, not for submission)} 

In order to explore the influence of corruptions on the performance of $K$-NN and on the optimal value of $K$, we use this construction:
\begin{itemize}
    \item Take the uncorrupted set, take $m$ as before and $n=80$, and construct training and validation sets as above. Plot five lines for the five validation sets, as a function of $K$, for $K\in\{1,\dots,m\}$. \textbf{Include this figure in your report.}
    \item Repeat the experiment with the \texttt{Light-Corruption} set (both training and test images should be taken from the lightly corrupted set), then with the \texttt{Moderate-Corruption} set, and then with the \texttt{Heavy-Corruption} set. \textbf{Include one figure for each of the corrupted sets in your report.}
    \item Discuss how corruption magnitude influences the prediction accuracy of $K$-NN and the optimal value of $K$. \textbf{Answer in the report.}
\end{itemize}

\noindent
\textit{Optional, not for submission: You are very welcome to experiment further with the data.}

\paragraph{Practical Advice} Check \Cref{app:KNN} for practical advice on how to make a vectorized implementation of $K$-NN. In interpreter programming languages like Python using vectorized implementations is much more efficient than using for-loops.
\end{exercise}

\chapter{Concentration of Measure Inequalities}
\label{ch:CoM}

Concentration of measure inequalities are one of the main tools for analyzing learning algorithms. This chapter is devoted to a number of concentration of measure inequalities that form the basis for the results discussed in later chapters.

\section{Markov's Inequality}
\label{sec:Markov}

Markov's Inequality is the simplest and relatively weak concentration inequality. Nevertheless, it forms the basis for many much stronger inequalities that we will see in the sequel, and for some distributions it is actually tight (see \Cref{ex:Markovs-tightness}).

\begin{theorem}[Markov's Inequality]
For any non-negative random variable $X$ and $\varepsilon > 0$:
\[
\P[X \geq \varepsilon] \leq \frac{\EEE{X}}{\varepsilon}.
\]
\end{theorem}

\begin{figure}%
\centering
\includegraphics[width=.5\columnwidth]{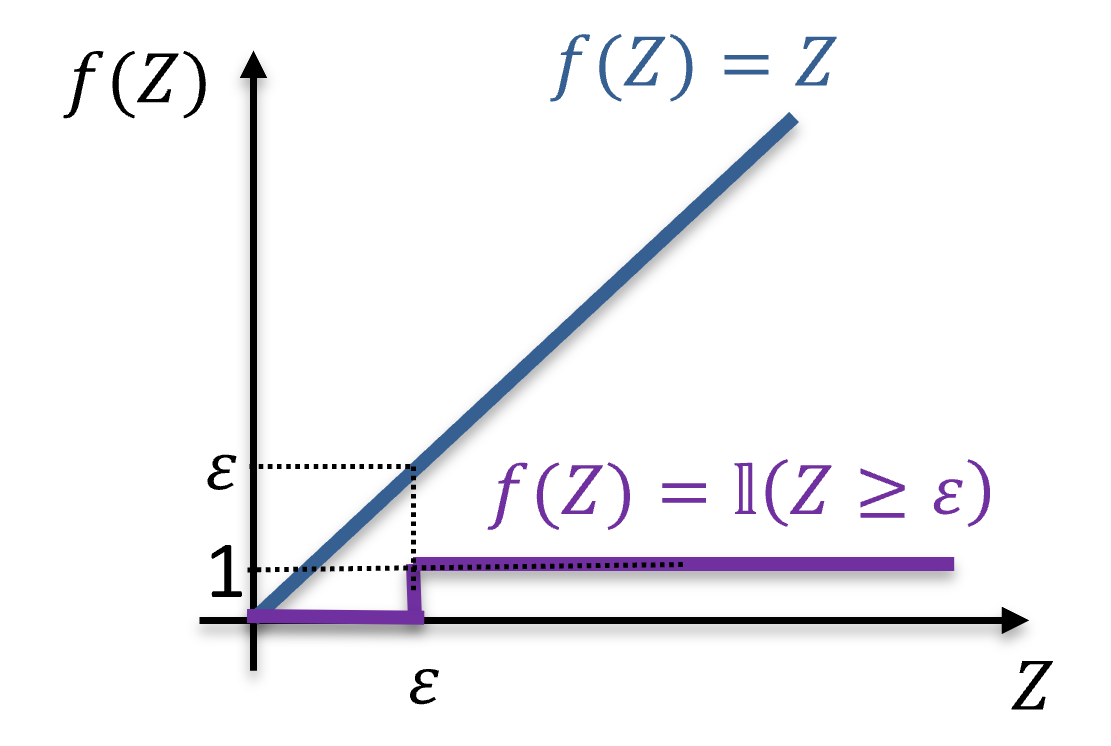}%
\caption{Relation between the identity function and the indicator function.}%
\label{fig:indicator-bound}%
\end{figure}

\begin{proof}
Define a random variable $Y = \1[X\geq \varepsilon]$ to be the indicator function of whether $X$ exceeds $\varepsilon$. Then $Y\leq \frac{X}{\varepsilon}$ (see Figure~\ref{fig:indicator-bound}). Since $Y$ is a Bernoulli random variable, $\E[Y] = \P[Y=1]$ (see Appendix~\ref{app:prob-theory}). We have:
\[
\P[X\geq \varepsilon] = \P[Y=1] = \E[Y] \leq \E[\frac{X}{\varepsilon}] = \frac{\E[X]}{\varepsilon}.
\]
Check yourself: where in the proof do we use non-negativity of $X$ and strict positiveness of $\varepsilon$?
\end{proof}

By denoting the right hand side of Markov's inequality by $\delta$ we obtain the following equivalent statement. For any non-negative random variable $X$:
\[
\P[X \geq \frac{1}{\delta}\E[X]] \leq \delta.
\]

\begin{example*}
We would like to bound the probability that we flip a fair coin 10 times and obtain 8 or more heads. Let $X_1,\dots,X_{10}$ be i.i.d.\ Bernoulli random variables with bias $\frac{1}{2}$. The question is equivalent to asking what is the probability that $\sum_{i=1}^{10} X_i \geq 8$. We have $\E[\sum_{i=1}^{10} X_i] = 5$ (the reader is invited to prove this statement formally) and by Markov's inequality
\[
\P[\sum_{i=1}^{10} X_i \geq 8] \leq \frac{\E[\sum_{i=1}^{10} X_i]}{8} = \frac{5}{8}.
\]
\end{example*}

We note that even though Markov's inequality is weak, there are situations in which it is tight. We invite the reader to construct an example of a random variable for which Markov's inequality is tight.

\section{Chebyshev's Inequality}

Our next stop is Chebyshev's inequality, which exploits variance to obtain tighter concentration.

\begin{theorem}[Chebyshev's inequality]
For any $\varepsilon > 0$
\[
\P[\lra{X - \E[X]} \geq \varepsilon] \leq \frac{\Var[X]}{\varepsilon^2}.
\]
\end{theorem}

\begin{proof}
The proof uses a transformation of a random variable. We have that $\P[\lra{X - \E[X]} \geq \varepsilon] = \P[\lr{X - \E[X]}^2 \geq \varepsilon^2]$, because the first statement holds if and only if the second holds. In addition, using Markov's inequality and the fact that $\lr{X - \E[X]}^2$ is a non-negative random variable we have
\[
\P[\lra{X - \E[X]} \geq \varepsilon] = \P[\lr{X - \E[X]}^2 \geq \varepsilon^2] \leq \frac{\E[\lr{X - \E[X]}^2]}{\varepsilon^2} = \frac{\Var[X]}{\varepsilon^2}.
\]
Check yourself: where in the proof did we use the positiveness of $\varepsilon$?
\end{proof}

In order to illustrate the relative advantage of Chebyshev's inequality compared to Markov's consider the following example. Let $X_1,\dots,X_n$ be $n$ independent identically distributed Bernoulli random variables and let $\hat \mu_n = \frac{1}{n}\sum_{i=1}^n X_i$ be their average. We would like to bound the probability that $\hat \mu_n$ deviates from $\E[\hat \mu_n]$ by more than $\varepsilon$ (this is the central question in machine learning). We have $\E[\hat \mu_n] = \E[X_1] = \mu$ and by independence of $X_i$-s and Theorem~\ref{thm:independent-var} we have $\Var[\hat \mu_n] = \frac{1}{n^2}\Var[n\hat \mu_n] = \frac{1}{n^2} \sum_{i=1}^n \Var[X_i] = \frac{1}{n} \Var[X_1]$. By Markov's inequality
\[
\P[\hat \mu_n - \E[\hat \mu_n] \geq \varepsilon] = \P[\hat \mu_n \geq \E[\hat \mu_n] + \varepsilon] \leq \frac{\E[\hat \mu_n]}{\E[\hat \mu_n] + \varepsilon} = \frac{\E[X_1]}{\E[X_1] + \varepsilon}.
\]
Note that as $n$ grows the inequality stays the same. By Chebyshev's inequality we have
\[
\P[\hat \mu_n - \E[\hat \mu_n] \geq \varepsilon] \leq \P[\lra{\hat \mu_n - \E[\hat \mu_n]} \geq \varepsilon] \leq \frac{\Var[\hat \mu_n]}{\varepsilon^2} = \frac{\Var[X_1]}{n\varepsilon^2}.
\]
Note that as $n$ grows the right hand side of the inequality decreases at the rate of $\frac{1}{n}$. Thus, in this case Chebyshev's inequality is much tighter than Markov's and it illustrates that as the number of random variables grows the probability that their average significantly deviates from the expectation decreases. In the next section we show that this probability actually decreases at an exponential rate.

\section{Hoeffding's Inequality}
\label{sec:Hoeffding}

Hoeffding's inequality is a much more powerful concentration result.

\begin{theorem}[Hoeffding's Inequality]
\label{thm:Hoeffding}
Let $X_1,\dots,X_n$ be independent real-valued random variables, such that for each $i \in \lrc{1,\dots,n}$ there exist $a_i \leq b_i$, such that $X_i \in [a_i,b_i]$. Then for every $\varepsilon > 0$:
\begin{equation}
\label{eq:Hoeff+}
\P[\sum_{i=1}^n X_i - \EEE{\sum_{i=1}^n X_i} \geq \varepsilon] \leq e^{- 2 \varepsilon^2 / \sum_{i=1}^n (b_i - a_i)^2}
\end{equation}
and 
\begin{equation}
\label{eq:Hoeff-}
\P[\sum_{i=1}^n X_i - \EEE{\sum_{i=1}^n X_i} \leq -\varepsilon] \leq e^{- 2 \varepsilon^2 / \sum_{i=1}^n (b_i - a_i)^2}.
\end{equation}
\end{theorem}

By taking a union bound of the events in \eqref{eq:Hoeff+} and \eqref{eq:Hoeff-} we obtain the following corollary.
\begin{corollary}
Under the assumptions of Theorem \ref{thm:Hoeffding}:
\begin{equation}
\label{eq:Hoeff+-}
\P[\lra{\sum_{i=1}^n X_i - \EEE{\sum_{i=1}^n X_i}} \geq \varepsilon] \leq 2e^{- 2 \varepsilon^2 / \sum_{i=1}^n (b_i - a_i)^2}.
\end{equation}
\end{corollary}
Equations \eqref{eq:Hoeff+} and \eqref{eq:Hoeff-} are known as ``one-sided Hoeffding's inequalities'' and \eqref{eq:Hoeff+-} is known as ``two-sided Hoeffding's inequality''.

If we assume that $X_i$-s are identically distributed and belong to the $[0,1]$ interval we obtain the following corollary (see \Cref{ex:Hoeffding-normalization}).
\begin{corollary}
\label{cor:Hoeffding}
Let $X_1,\dots,X_n$ be independent random variables, such that $X_i \in [0,1]$ and $\EEE{X_i} = \mu$ for all $i$, then for every $\varepsilon > 0$:
\begin{equation}
\label{eq:Hoeff01+}
\P[\frac{1}{n}\sum_{i=1}^n X_i - \mu \geq \varepsilon] \leq e^{- 2 n\varepsilon^2}
\end{equation}
and 
\begin{equation}
\label{eq:Hoeff01-}
\P[\mu - \frac{1}{n} \sum_{i=1}^n X_i \geq \varepsilon] \leq e^{- 2 n \varepsilon^2}.
\end{equation}
\end{corollary}
Recall that by Chebyshev's inequality $\hat \mu_n = \frac{1}{n} \sum_{i=1}^n X_i$ converges to $\mu$ at the rate of $n^{-1}$. Hoeffding's inequality demonstrates that the convergence is actually much faster, at least at the rate of $e^{-n}$.

The proof of Hoeffding's inequality is based on Hoeffding's lemma.
\begin{lemma}[Hoeffding's Lemma]
\label{lem:Hoeff}
Let $X$ be a random variable, such that $X \in [a,b]$. Then for any $\lambda \in \R$:
\[
\EEE{e^{\lambda X}} \leq e^{\lambda \EEE{X} + \frac{\lambda^2 (b-a)^2}{8}}.
\]
\end{lemma}
The function $f(\lambda) = \EEE{e^{\lambda X}}$ is known as the \emph{moment generating function} of $X$, since $f'(0) = \EEE{X}$, $f''(0) = \EEE{X^2}$, and, more generally, $f^{(k)}(0) = \EEE{X^k}$. We provide a proof of the lemma immediately after the proof of Theorem \ref{thm:Hoeffding}.

\begin{proof}[Proof of Theorem \ref{thm:Hoeffding}]
We prove the first inequality in Theorem \ref{thm:Hoeffding}. The second inequality follows by applying the first inequality to $-X_1,\dots,-X_n$. The proof is based on Chernoff's bounding technique. For any $\lambda > 0$ the following holds:
\[
\P[\sum_{i=1}^n X_i - \EEE{\sum_{i=1}^n X_i} \geq \varepsilon] = \P[e^{\lambda \lr{\sum_{i=1}^n X_i - \EEE{\sum_{i=1}^n X_i}}} \geq e^{\lambda \varepsilon}] \leq \frac{\EEE{e^{\lambda \lr{\sum_{i=1}^n X_i - \EEE{\sum_{i=1}^n X_i}}}}}{e^{\lambda \varepsilon}},
\]
where the first step holds since $e^{\lambda x}$ is a monotonically increasing function for $\lambda > 0$ and the second step holds by Markov's inequality. We now take a closer look at the nominator:
\begin{align}
\EEE{e^{\lambda \lr{\sum_{i=1}^n X_i - \EEE{\sum_{i=1}^n X_i}}}} &= \EEE{e^{\lr{ \sum_{i=1}^n \lambda \lr{X_i - \EEE{X_i}}}}}\notag\\
&= \EEE{\prod_{i=1}^n e^{\lambda \lr{X_i - \EEE{X_i}}}}\notag\\
&= \prod_{i=1}^n \EEE{e^{\lambda \lr{X_i - \EEE{X_i}}}}\label{eq:1}\\
&\leq \prod_{i=1}^n e^{\lambda^2 (b_i - a_i)^2 / 8}\label{eq:2}\\
&= e^{(\lambda^2 / 8) \sum_{i=1}^n (b_i - a_i)^2},\notag
\end{align}
where \eqref{eq:1} holds since $X_1,\dots,X_n$ are independent and \eqref{eq:2} holds by Hoeffding's lemma applied to a random variable $Z_i = X_i - \EEE{X_i}$ (note that $\EEE{Z_i} = 0$ and that $Z_i \in [a_i - \mu_i, b_i - \mu_i]$ for $\mu_i = \EEE{X_i}$). \emph{Pay attention to the crucial role that independence of $X_1,\dots,X_n$ plays in the proof! Without independence we would not have been able to exchange the expectation with the product and the proof would break down! And it is not just that the proof would break down, but it is actually possible to construct examples of dependent random variables for which the empirical mean does not converge to its expectation, see \Cref{ex:independence}.} To complete the proof we substitute the bound on the expectation into the previous calculation and obtain:
\[
\P[\sum_{i=1}^n X_i - \EEE{\sum_{i=1}^n X_i} \geq \varepsilon] \leq e^{(\lambda^2 / 8) \lr{\sum_{i=1}^n (b_i - a_i)^2} - \lambda \varepsilon}.
\]
This expression is minimized by 
\[\lambda^* = \arg\min_\lambda e^{(\lambda^2 / 8) \lr{\sum_{i=1}^n (b_i - a_i)^2} - \lambda \varepsilon} = \arg \min_\lambda \lr{(\lambda^2 / 8) \lr{\sum_{i=1}^n (b_i - a_i)^2} - \lambda \varepsilon} = \frac{4 \varepsilon}{\sum_{i=1}^n (b_i - a_i)^2}.
\]
\emph{It is important to note that the best choice of $\lambda$ does not depend on the sample. In particular, it allows to fix $\lambda$ before observing the sample.} By substituting $\lambda^*$ into the calculation we obtain the result of the theorem.
\end{proof}

\begin{proof}[Proof of Lemma \ref{lem:Hoeff}]
Note that
\[
\EEE{e^{\lambda X}} = \EEE{e^{\lambda \lr{X - \EEE{X}} + \lambda \EEE{X}}} = e^{\lambda \EEE{X}} \times \EEE{e^{\lambda \lr{X - \EEE{X}}}}.
\]
Hence, it is sufficient to show that for any random variable $Z$ with $\EEE{Z} = 0$ and $Z \in [a,b]$ we have:
\[
\EEE{e^{\lambda Z}} \leq e^{\lambda^2 (b-a)^2 / 8}.
\]
By convexity of the exponential function, for $z \in [a,b]$ we have:
\[
e^{\lambda z} \leq \frac{z - a}{b-a} e^{\lambda b} + \frac{b-z}{b-a} e^{\lambda a}.
\]
Let $p = -a/(b-a)$. Then:
\begin{align*}
\EEE{e^{\lambda Z}} &\leq \EEE{\frac{Z - a}{b-a} e^{\lambda b} + \frac{b-Z}{b-a} e^{\lambda a}}\\
&= \frac{\EEE{Z} - a}{b-a} e^{\lambda b} + \frac{b-\EEE{Z}}{b-a} e^{\lambda a}\\
&= \frac{-a}{b-a} e^{\lambda b} + \frac{b}{b-a} e^{\lambda a}\\
&= \lr{1 - p + p e^{\lambda (b-a)}}e^{-p \lambda (b-a)}\\
&= e^{\phi(u)},
\end{align*}
where $u = \lambda (b-a)$ and $\phi(u) = - pu + \ln \lr{1 - p + p e^u}$ and we used the fact that $\EEE{Z} = 0$. It is easy to verify that the derivative of $\phi$ is
\[
\phi'(u) = -p + \frac{p}{p+(1-p)e^{-u}}
\]
and, therefore, $\phi(0) = \phi'(0) = 0$. Furthermore,
\[
\phi''(u) = \frac{p(1-p)e^{-u}}{\lr{p + (1-p)e^{-u}}^2} \leq \frac{1}{4}.
\]
By Taylor's theorem, $\phi(u) = \phi(0) + u \phi'(0) + \frac{u^2}{2} \phi''(\theta)$ for some $\theta \in [0,u]$. Thus, we have:
\[
\phi(u) = \phi(0) + u \phi'(0) + \frac{u^2}{2} \phi''(\theta) = \frac{u^2}{2} \phi''(\theta) \leq \frac{u^2}{8} = \frac{\lambda^2 \lr{b-a}^2}{8}.
\]
\end{proof}

\subsection{Understanding Hoeffding's Inequality}
\label{sec:understand-Hoeffding}

Hoeffding's inequality involves three interconnected terms: $n$, $\varepsilon$, and $\delta = 2e^{-2n\varepsilon^2}$, which is the bound on the probability that the event under $\P[]$ holds (for the purpose of the discussion we consider two-sided Hoeffding's inequality for random variables bounded in $[0,1]$). We can fix any two of the three terms $n$, $\varepsilon$, and $\delta$ and then the relation $\delta = e^{-2n\varepsilon^2}$ provides the value of the third. Thus, we have
\begin{align*}
\delta &= 2e^{-2n\varepsilon^2},\\
\varepsilon &= \sqrt{\frac{\ln \frac{2}{\delta}}{2n}},\\
n &= \frac{\ln \frac{2}{\delta}}{2\varepsilon^2}.
\end{align*}

Overall, Hoeffding's inequality tells by how much the empirical average $\frac{1}{n} \sum_{i=1}^n X_i$ can deviate from its expectation $\mu$, but the interplay between the three parameters provides several ways of seeing and using Hoeffding's inequality. For example, if the number of samples $n$ is fixed (we have made a fixed number of experiments and now analyze what we can get from them), there is an interplay between the precision $\varepsilon$ and confidence $\delta$. We can request higher precision $\varepsilon$, but then we have to compromise on the confidence $\delta$ that the desired bound $\lra{\frac{1}{n} \sum_{i=1}^n X_i - \mu} \leq \varepsilon$ holds. And the other way around: we can request higher confidence $\delta$, but then we have to compromise on precision $\varepsilon$, i.e., we have to increase the allowed range $\pm\varepsilon$ around $\mu$, where we expect to find the empirical average $\frac{1}{n}\sum_{i=1}^n X_i$.

As another example, we may have target precision $\varepsilon$ and confidence $\delta$ and then the inequality provides us the number of experiments $n$ that we have to perform in order to achieve the target.

It is often convenient to write the inequalities \eqref{eq:Hoeff01+} and \eqref{eq:Hoeff01-} with a fixed confidence in mind, thus we have
\begin{align*}
\P[\frac{1}{n}\sum_{i=1}^n X_i - \mu \geq \sqrt{\frac{\ln \frac{1}{\delta}}{2n}}] &\leq \delta,\\
\P[\mu - \frac{1}{n} \sum_{i=1}^n X_i \geq \sqrt{\frac{\ln \frac{1}{\delta}}{2n}}] &\leq \delta,\\
\P[\lra{\frac{1}{n}\sum_{i=1}^n X_i - \mu} \geq \sqrt{\frac{\ln \frac{2}{\delta}}{2n}}] &\leq \delta.\\
\end{align*}
(Pay attention that the $\ln 2$ factor in the last inequality comes from the union bound over the first two inequalities: if we want to keep the same confidence we have to compromise on precision.)

In many situations we are interested in the complimentary events. Thus, for example, we have
\[
\P[\mu - \frac{1}{n} \sum_{i=1}^n X_i \leq \sqrt{\frac{\ln \frac{1}{\delta}}{2n}}] \geq 1-\delta.
\]
Careful reader may point out that the inequalities above should be strict (``$<$'' and ``$>$''). This is true, but if it holds for strict inequalities it also holds for non-strict inequalities (``$\leq$'' and ``$\geq$''). Since strict inequalities provide no practical advantage we will use the non-strict inequalities to avoid the headache of remembering which inequalities should be strict and which should not.

The last inequality essentially says that with probability at least $1-\delta$ we have 
\begin{equation}   
\label{eq:Hoeffding-direct}
\mu \leq \frac{1}{n} \sum_{i=1}^n X_i + \sqrt{\frac{\ln \frac{1}{\delta}}{2n}}
\end{equation}
and this is how we will occasionally use it. Note that the random variable is $\frac{1}{n} \sum_{i=1}^n X_i$ and the right way of interpreting the above inequality is actually that with probability at least $1-\delta$
\[
\frac{1}{n} \sum_{i=1}^n X_i \geq \mu - \sqrt{\frac{\ln \frac{1}{\delta}}{2n}},
\]
i.e., the probability is over $\frac{1}{n} \sum_{i=1}^n X_i$ and not over $\mu$. However, many generalization bounds that we study in Chapter~\ref{ch:Generalization} are written in the first form in the literature and we follow the tradition.

\section{Sampling Without Replacement}

Let $X_1,\dots,X_n$ be a sequence of random variables \emph{sampled without replacement} from a finite set of values $\XX = \lrc{x_1,\dots,x_N}$ of size $N$. The random variables $X_1,\dots,X_n$ are \emph{dependent}. For example, if $\XX = \lrc{-1, +1}$ and we sample two values then $X_1 = -X_2$. Since $X_1,\dots,X_n$ are dependent, the concentration results from previous sections do not apply directly. However, the following result by \citet[Theorem 4]{Hoe63}, which we cite without a proof, allows to extend results for sampling with replacement to sampling without replacement.

\begin{lemma} Let $X_1,\dots,X_n$ denote a random sample without replacement from a finite set $\XX = \lrc{x_1,\dots,x_N}$ of $N$ real values. Let $Y_1,\dots,Y_n$ denote a random sample with replacement from $\XX$. Then for any continuous and convex function $f:\R\to\R$
\[
\E[f\lr{\sum_{i=1}^n X_i}] \leq \E[f\lr{\sum_{i=1}^n Y_i}].
\]
\label{lem:swr}
\end{lemma}

In particular, the lemma can be used to prove Hoeffding's inequality for sampling without replacement.

\begin{theorem}[Hoeffding's inequality for sampling without replacement] 
Let $X_1,\dots,X_n$ denote a random sample without replacement from a finite set $\XX = \lrc{x_1,\dots,x_N}$ of $N$ values, where each element $x_i$ is in the $[0,1]$ interval. Let $\mu = \frac{1}{N}\sum_{i=1}^N x_i$ be the average of the values in $\XX$. Then for all $\varepsilon > 0$
\begin{align*}
\P[\frac{1}{n} \sum_{i=1}^n X_i - \mu \geq \varepsilon] &\leq e^{-2n\varepsilon^2},\\
\P[\mu - \frac{1}{n} \sum_{i=1}^n X_i \geq \varepsilon] &\leq e^{-2n\varepsilon^2}.
\end{align*}
\label{thm:Hoeffding-wr}
\end{theorem}

The proof is a minor adaptation of the proof of Hoeffding's inequality for sampling with replacement using Lemma~\ref{lem:swr} and is left as an exercise. (Note that it requires a small modification inside the proof, because Lemma~\ref{lem:swr} cannot be applied directly to the statement of Hoeffding's inequality.)

While formal proof requires a bit of work, intuitively the result is quite expected. Imagine the process of sampling without replacement. If the average of points sampled so far starts deviating from the mean of the values in $\XX$, the average of points that are left in $\XX$ deviates in the opposite direction and ``applies extra force'' to new samples to bring the average back to $\mu$. In the limit when $n=N$ we are guaranteed to have the average of $X_i$-s being equal to $\mu$.

\section{Basics of Information Theory: Entropy, Relative Entropy, and the Method of Types}

In this section we briefly introduce a number of basic concepts from information theory that are very useful for deriving concentration inequalities. Specifically, we introduce the notions of entropy and relative entropy \citep[Chapter 2]{CT06} and some basic tools from the method of types \citep[Chapter 11]{CT06}. 

\subsection{Entropy}

We start with the definition of entropy.
\begin{definition}[Entropy] Let $p(x)$ be a distribution of a discrete random variable $X$ taking values in a finite set ${\cal X}$. We define the \emph{entropy} of $p$ as:
\[
\H(p) = -\sum_{x \in {\cal X}} p(x) \ln p(x).
\]
We use the convention that $0 \ln 0 = 0$ (which is justified by continuity of $z \ln z$, since $z \ln z \to 0$ as $z \to 0$).
\end{definition}

We have special interest in Bernoulli random variables.

\begin{definition}[Bernoulli random variable] $X$ is a Bernoulli random variable with bias $p$ if $X$ accepts values in $\lrc{0,1}$ with $\P[X=0] = 1-p$ and $\P[X=1] = p$.
\end{definition}

Note that expectation of a Bernoulli random variable is equal to its bias:
\[
\EEE{X} = 0 \times \P[X=0] + 1 \times \P[X=1] = \P[X=1] = p.
\]

With a slight abuse of notation we specialize the definition of entropy to Bernoulli random variables.
\begin{definition}[Binary entropy] Let $p$ be a bias of Bernoulli random variable $X$. We define the entropy of $p$ as
\[
\H(p) = -p \ln p - (1-p) \ln (1-p).
\]
\end{definition}
Note that when we talk about Bernoulli random variables $p$ denotes the bias of the random variable and when we talk about more general random variables $p$ denotes the complete distribution.

Entropy is one of the central quantities in information theory and it has numerous applications. We start by using binary entropy to bound binomial coefficients.
\begin{lemma}
\label{lem:Binomial}
\[
\frac{1}{n+1} e^{n \H\lr{\frac{k}{n}}} \leq \binom{n}{k} \leq e^{n \H\lr{\frac{k}{n}}}.
\]
\end{lemma}
(Note that $\frac{k}{n} \in [0,1]$ and $\H\lr{\frac{k}{n}}$ in the lemma is the binary entropy.)
\begin{proof}
By the binomial formula we know that for any $p \in [0,1]$:
\begin{equation}
\label{eq:binom}
\sum_{i=0}^n \binom{n}{i} p^i (1-p)^{n-i} = 1.
\end{equation}
We start with the upper bound. Take $p = \frac{k}{n}$. Since the sum is larger than any individual term, for the $k$-th term of the sum we get:
\begin{align*}
1 &\geq\binom{n}{k} p^{k} \lr{1 - p}^{n-k}\\
&= \binom{n}{k} \lr{\frac{k}{n}}^{k} \lr{1 - \frac{k}{n}}^{n-k}\\
& = \binom{n}{k} \lr{\frac{k}{n}}^{k} \lr{\frac{n-k}{n}}^{n-k}\\
& = \binom{n}{k} e^{k\ln\frac{k}{n} + (n-k)\ln\frac{n-k}{n}}\\
& = \binom{n}{k} e^{n \lr{\frac{k}{n}\ln\frac{k}{n} + \frac{n-k}{n}\ln\frac{n-k}{n}}}\\
& = \binom{n}{k} e^{-n \H\lr{\frac{k}{n}}}.
\end{align*}
By changing sides of the inequality we obtain the upper bound.

For the lower bound it is possible to show that if we fix $p = \frac{k}{n}$ then $\binom{n}{k} p^k (1-p)^{n-k} \geq \binom{n}{i} p^{i} (1-p)^{n-i}$ for any $i \in \lrc{0,\dots,n}$, see \citet[Example 11.1.3]{CT06} for details. We also note that there are $n+1$ elements in the sum in equation \eqref{eq:binom}. Again, take $p = \frac{k}{n}$, then
\[
1 \leq (n+1) \max_i \binom{n}{i} \lr{\frac{k}{n}}^i \lr{\frac{n-k}{n}}^{n-i} = (n+1) \binom{n}{k} \lr{\frac{k}{n}}^k \lr{\frac{n-k}{n}}^{n-k} = (n+1) \binom{n}{k} e^{-n \H\lr{\frac{k}{n}}},
\]
where the last step follows the same steps as in the derivation of the upper bound.
\end{proof}

Lemma \ref{lem:Binomial} shows that the number of configurations of chosing $k$ out of $n$ objects is directly related to the entropy of the imbalance $\frac{k}{n}$ between the number of objects that are selected ($k$) and the number of objects that are left out ($n-k$). 

\subsection{The Kullback-Leibler ($\KL$) Divergence (Relative Entropy)}

We now introduce an additional quantity, the \emph{Kullback-Leibler} (\emph{KL}) \emph{divergence}, also known as \emph{Kullback-Leibler distance} and as \emph{relative entropy}.

\begin{definition}[Relative entropy or Kullback-Leibler divergence] 
Let $p(x)$ and $q(x)$ be two probability distributions of a random variable $X$ (or two probability density functions, if $X$ is a continuous random variable), the \emph{Kullback-Leibler divergence} or \emph{relative entropy} is defined as:
\[
\KL(p\|q) = \mathbb E_{p} \lrs{\ln \frac{p(X)}{q(X)}} = \begin{cases}\sum_{x \in {\cal X}} p(x) \ln \frac{p(x)}{q(x)}, & \text{if ${\cal X}$ is discrete}\\ \int_{x \in {\cal X}} p(x) \ln \frac{p(x)}{q(x)} dx, & \text{if ${\cal X}$ is continuous} \end{cases}.
\]
We use the convention that $0 \ln \frac{0}{0} = 0$ and $0 \ln \frac{0}{q} = 0$ and $p \ln \frac{p}{0} = \infty$.
\end{definition}

We specialize the definition to Bernoulli distributions.

\begin{definition}[Binary $\kl$-divergence] Let $p$ and $q$ be biases of two Bernoulli random variables. The \emph{binary $\kl$ divergence} is defined as:
\[
\kl(p\|q) = \KL([1-p, p]\|[1-q,q]) = p \ln \frac{p}{q} + (1-p) \ln \frac{1-p}{1-q}.
\]
\end{definition}

KL divergence is the central quantity in information theory. Although it is not a distance measure, because it does not satisfy the triangle inequality, it is the right way of measuring distances between probability distributions. This is illustrated by the following example.
\begin{example}
\label{ex:kl}
Let $X_1,\dots,X_n$ be an i.i.d.\ sample of $n$ Bernoulli random variables with bias $p$ and let $\frac{1}{n} \sum_{i=1}^n X_i$ be the empirical bias of the sample. (Note that $\frac{1}{n} \sum_{i=1}^n X_i \in \lrc{0, \frac{1}{n}, \frac{2}{n}, \dots, \frac{n}{n}}$.) Then for $p\in(0,1)$
\begin{align}
\P[\frac{1}{n} \sum_{i=1}^n X_i = \frac{k}{n}] &= \binom{n}{k} p^k (1-p)^{n-k}\notag\\
&=\binom{n}{k} e^{-n \H\lr{\frac{k}{n}}} e^{n \H\lr{\frac{k}{n}}} e^{n \lr{\frac{k}{n} \ln p + \frac{n-k}{n} \ln (1-p)}}\notag\\
&= \binom{n}{k} e^{-n \H\lr{\frac{k}{n}}} e^{-n\kl\lr{\frac{k}{n} \middle \|p}}\label{eq:Pphat}
\end{align}
By Lemma \ref{lem:Binomial} we have $\frac{1}{n+1}\leq \binom{n}{k} e^{-n \H\lr{\frac{k}{n}}} \leq 1$, which gives
\begin{equation}
\label{eq:phatp}
\frac{1}{n+1} e^{-n\kl\lr{\frac{k}{n} \middle \|p}} \leq \P[\frac{1}{n} \sum_{i=1}^n X_i = \frac{k}{n}] \leq e^{-n\kl\lr{\frac{k}{n} \middle \|p}}.
\end{equation}
Thus, $\kl\lr{\frac{k}{n}\|p}$ governs the probability of observing empirical bias $\frac{k}{n}$ when the true bias is $p$. It is easy to verify that $\kl(p\|p) = 0$, and it is also possible to show that $\kl(\hat p\|p)$ is convex in $\hat p$, and that $\kl(\hat p\|p) \geq 0$ \citep{CT06}. And so, the probability of empirical bias is maximized when it coincides with the true bias.
\end{example}

\subsubsection{Properties of the $\KL$ and $\kl$ Divergences}

The $\KL$ divergence between two probability distributions is always non-negative.
\begin{theorem}[{Nonnegativity of $\KL$ \citep[Theorem 2.3.6]{CT06}}]
Let $p(x)$ and $q(x)$ be two probability distributions. Then 
\[
\KL(p\|q) \geq 0
\]
with equality if and only if $p(x)=q(x)$ for all $x$.
\end{theorem}

\begin{corollary}[Nonnegativity of $\kl$]
For $p,q \in [0,1]$
\[
\kl(p\|q) \geq 0
\]
with equality if and only if $p=q$.
\end{corollary}
The $\KL$ divergence is also convex.
\begin{theorem}[{Convexity of $\KL$ \citep[Theorem 2.7.2]{CT06}}]
$\KL(p\|q)$ is convex in the pair $(p, q)$; that is, if $(p_1, q_1)$ and $(p_2, q_2)$ are two pairs of probability
mass functions, then 
\[
\KL(\lambda p_1 + (1 - \lambda)p_2 \|\lambda q_1 + (1 - \lambda) q_2) \leq \lambda \KL(p_1\|q_1) + (1 - \lambda)\KL(p_2\|q_2)
\]
for all $0 \leq \lambda \leq 1$.
\end{theorem}

\begin{corollary}[Convexity of $\kl$]
\label{cor:kl-convexity}
For $p_1,q_1,p_2,q_2\in[0,1]$
\[
\kl(\lambda p_1 + (1 - \lambda)p_2 \|\lambda q_1 + (1 - \lambda) q_2) \leq \lambda \kl(p_1\|q_1) + (1 - \lambda)\kl(p_2\|q_2)
\]
for all $0 \leq \lambda \leq 1$.
\end{corollary}

\section{The $\kl$ Inequality}

Example \ref{ex:kl} shows that $\kl$ can be used to bound the empirical bias when the true bias is known. But in machine learning we are usually interested in the inverse problem - how to infer the true bias $p$ when the empirical bias $\hat p$ is known. Next we demonstrate that this is also possible and that it leads to an inequality, which is tighter than Hoeffding's inequality. We start with a simple version of $\kl$ lemma based on one-line derivation. Then we provide a tight version of the $\kl$ lemma and a lower bound showing that it cannot be improved any further. We then use the $\kl$ lemma to derive a $\kl$ inequality, and also provide a tighter version of the $\kl$ inequality, which is not based on the $\kl$ lemma. And we finish with relaxations of the $\kl$ inequality, which provide an intuitive interpretation of its implication.

\subsection{A Simple Version of the $\kl$ Lemma}

\begin{lemma}[Simple $\kl$ Lemma]
\label{lem:Ekl}
Let $X_1,\dots,X_n$ be i.i.d.\ Bernoulli with bias $p$ and let $\hat p = \frac{1}{n} \sum_{i=1}^n X_i$ be the empirical bias. Then
\[
\E[e^{n \kl(\hat p\|p)}] \leq n+1.
\]
\end{lemma}

\begin{proof}
For $p\in(0,1)$
\[
\E[e^{n \kl(\hat p\|p)}] = \sum_{k=0}^n \P[\hat p = \frac{k}{n}] e^{n \kl\lr{\frac{k}{n}\middle \|p}} \leq \sum_{k=0}^n e^{-n \kl\lr{\frac{k}{n}\middle \|p}} e^{n \kl\lr{\frac{k}{n}\middle\|p}} = n+1,
\]
where the inequality was derived in equation \ref{eq:phatp}. For $p\in\lrc{0,1}$ we have $\E[e^{n\kl(\hat p\|o)}]=1$, so the inequality is satisfied trivially.
\end{proof}

\subsection{A Tight Version of the $\kl$ Lemma}
\label{sec:klTight}

In this section we provide a tight versions of the $\kl$ lemma. The improvement is based on a tighter control of the binomial coefficients, which is achieved by using Stirling's approximation of the factorial, $\sqrt{2\pi n}\lr{\frac{n}{e}}^n \leq n!\leq \sqrt{2\pi n}\lr{\frac{n}{e}}^n e^{\frac{1}{12n}}$. The result is due to \citet{WR61}.
\begin{lemma}
\label{lem:BinomialTight}
For $1 \leq k \leq n-1$
\[
\frac{1}{2}\sqrt{\frac{n}{2k(n-k)}}e^{n\H\lr{\frac{k}{n}}} \leq \binom{n}{k} \leq \frac{e^{\frac{1}{12n}}}{\sqrt{2\pi}}\sqrt{\frac{n}{k(n-k)}} e^{n\H\lr{\frac{k}{n}}}.
\]
The upper bound can be simplified using $\frac{e^{\frac{1}{12n}}}{\sqrt{2\pi}} < \frac{1}{2}$ for $n\geq 1$.
\end{lemma}
\begin{proof}
By Stirling's approximation
\begin{align*}
\binom{n}{k} &\leq \frac{\sqrt{2\pi n}\lr{\frac{n}{e}}^n e^{\frac{1}{12n}}}{\sqrt{2\pi k}\lr{\frac{k}{e}}^k \sqrt{2\pi (n-k)}\lr{\frac{n-k}{e}}^{n-k}} \\
&= \frac{e^{\frac{1}{12n}}}{\sqrt{2\pi}}\sqrt{\frac{n}{k(n-k)}} \frac{1}{\lr{\frac{k}{n}}^k \lr{\frac{n-k}{n}}^{n-k}}\\
&= \frac{e^{\frac{1}{12n}}}{\sqrt{2\pi}}\sqrt{\frac{n}{k(n-k)}} e^{n\H\lr{\frac{k}{n}}}.
\end{align*}
The lower bound is derived in a similar way, see \citet[Lemma 17.5.1]{CT06}.
\end{proof}

By combining \Cref{lem:BinomialTight} with Equation \eqref{eq:Pphat}, for $1 \leq k \leq n-1$
\[
\frac{1}{2}\sqrt{\frac{n}{2 k(n-k)}} e^{-n\kl\lr{\frac{k}{n} \middle \|p}} \leq \P[\frac{1}{n} \sum_{i=1}^n X_i = \frac{k}{n}] \leq \frac{e^{\frac{1}{12n}}}{\sqrt{2\pi}}\sqrt{\frac{n}{k(n-k)}} e^{-n\kl\lr{\frac{k}{n} \middle \|p}}.
\]
The refinement can be used to tighten \Cref{lem:Ekl}.
\begin{lemma}[$\kl$ Lemma {\citep{Mau04}}]
\label{lem:klRef}
Let $X_1,\dots,X_n$ be i.i.d.\ with $X_1\in [0,1]$, $\E[X_1] = p$, and $\hat p = \frac{1}{n} \sum_{i=1}^n X_i$. Then
\[
\E[e^{n\kl(\hat p\|p)}]\leq 2\sqrt n.
\]
\end{lemma}
The proof of \Cref{lem:klRef} is based on two auxiliary results. The first is a technical bound on a summation.
\begin{lemma}[{\citep[Lemma 4]{Mau04}}]
\label{lem:suminvsqrt}
For $n\geq 2$
\[
1 \leq \sum_{k=1}^{n-1} \frac{1}{\sqrt{k(n-k)}} \leq \pi.
\] 
\end{lemma}
The second result allows to extend results for Bernoulli random variables to random variables bounded in the $[0,1]$ interval.
\begin{lemma}[{\citep[Lemma 3]{Mau04}}]
\label{lem:ConvexExtension}
Let $X_1,\dots,X_n$ be i.i.d.\ with $X_1\in [0,1]$, and let $Y_1,\dots,Y_n$ be i.i.d.\ Bernoulli, such that $\E[Y_1]=\E[X_1]$. Then for any convex function $f:[0,1]^n\to\R$
\[
\E[f(X_1,\dots,X_n)] \leq \E[f(Y_1,\dots,Y_n)].
\]  
\end{lemma}
The proof of \Cref{lem:klRef} acts the same way as the proof of \Cref{lem:Ekl}, just using the tighter bound on the binomial coefficients.
\begin{proof}[Proof of \Cref{lem:klRef}]
We prove the result for Bernoulli random variables. The extension to general random variables bounded in $[0,1]$ follows by \Cref{lem:ConvexExtension}. For Bernoulli random variables and $p\in(0,1)$ we have
\begin{equation}   
\label{eq:Ekl}
\E[e^{n\kl(\hat p\|p)}] = \sum_{k=0}^n \P[\hat p = \frac{k}{n}] e^{n\kl\lr{\frac{k}{n}\middle\|p}} = \sum_{k=0}^n \binom{n}{k} e^{-n\H\lr{\frac{k}{n}}}e^{-n\kl\lr{\frac{k}{n}\middle\|p}}e^{n\kl\lr{\frac{k}{n}\middle\|p}}=\sum_{k=0}^n \binom{n}{k} e^{-n\H\lr{\frac{k}{n}}},
\end{equation}
where the middle equality is by \eqref{eq:Pphat}. By \Cref{lem:BinomialTight}, for $n\geq 3$ we have
\[
\sum_{k=0}^n \binom{n}{k} e^{-n\H\lr{\frac{k}{n}}} = 2 + \sum_{k=1}^{n-1} \binom{n}{k} e^{-n\H\lr{\frac{k}{n}}} \leq 2 + e^{\frac{1}{12n}}\sqrt\frac{n}{2\pi}\sum_{k=1}^{n-1}\frac{1}{\sqrt{k(n-k)}} \leq 2 + e^{\frac{1}{12n}}\sqrt{\frac{\pi n}{2}},
\]
where the last inequality is by \Cref{lem:suminvsqrt}. For $n\geq 8$ we have $2 + e^{\frac{1}{12n}}\sqrt{\frac{\pi n}{2}} \leq 2\sqrt n$, whereas for $n\in\lrc{1,\dots,7}$ a direct calculation confirms that we also have $\sum_{k=0}^n \binom{n}{k} e^{-n\H\lr{\frac{k}{n}}} \leq 2\sqrt n$. For $p\in\lrc{0,1}$ we have $\E[e^{n\kl(\hat p\|p)}] = 1$, so the lemma holds trivially.
\end{proof}

The next result shows that the $\kl$ Lemma (\Cref{lem:klRef}) cannot be improved much further.
\begin{lemma}[$\kl$ Lemma - Lower Bound {\citep{Mau04}}] Let $X_1,\dots,X_n$ be i.i.d.\ Bernoulli with $\E[X_1]=p$ and $\hat p = \frac{1}{n}\sum_{i=1}^n X_i$. Then for $p\in(0,1)$
\[
\E[e^{n\kl(\hat p\|p)}] \geq \sqrt n.
\]   
\end{lemma}
\begin{proof}
Starting from \eqref{eq:Ekl} and applying \Cref{lem:BinomialTight}, for $p\in(0,1)$ and $n\geq 3$ we have
\[
\E[e^{n\kl(\hat p\|p)}] = \sum_{k=0}^n \binom{n}{k}e^{-n\H\lr{\frac{k}{n}}} = 2 + \sum_{k=1}^{n-1} \binom{n}{k}e^{-n\H\lr{\frac{k}{n}}} \geq \frac{1}{2}\sqrt{\frac{n}{2}}\sum_{k=1}^{n-1} \frac{1}{\sqrt{k(n-k)}}.
\]
The function $f(n) = \sum_{k=1}^{n-1} \frac{1}{\sqrt{k(n-k)}}$ is monotonically increasing with $n$. For $n\geq 88$ we have $f(n) > \sqrt 8$, thus $f(n)\sqrt{\frac{n}{8}} \geq \sqrt n$. For $n \in \lrc{1,\dots,87}$ a direct calculation confirms that $\sum_{k=0}^n \binom{n}{k}e^{-n\H\lr{\frac{k}{n}}}\geq \sqrt n$.
\end{proof}

\subsection{$\kl$ Inequality}

By combining the $\kl$ lemma with Markov's inequality, we obtain the $\kl$ inequality.
\begin{theorem}[$\kl$ Inequality via $\kl$ Lemma]
\label{thm:klsqrtn}
    Let $X_1,\dots,X_n$ be i.i.d.\ with $X_1\in [0,1]$, $\E[X_1] = p$, and $\hat p = \frac{1}{n} \sum_{i=1}^n X_i$. Then
    \[
\P[\kl(\hat p\|p) \geq \frac{\ln\frac{2\sqrt n}{\delta}}{n}] \leq \delta.    
    \]
\end{theorem}

\begin{proof}
\[
\P[\kl(\hat p\|p) \geq \frac{\ln\frac{2\sqrt n}{\delta}}{n}] = \P[e^{n\kl(\hat p\|p)} \geq \frac{2\sqrt n}{\delta}] \leq \frac{\delta}{2\sqrt n} \E[e^{n\kl(\hat p\|p)}] \leq \delta,
\]
where the first inequality is by Markov's inequality, and the second inequality is by the $\kl$ lemma (\Cref{lem:klRef}).
\end{proof}

Even though \Cref{lem:klRef} cannot be improved much further, it is possible to improve the $\kl$ inequality through a direct derivation that does not go through $\E[e^{n\kl(\hat p\|p)}]$.
\begin{theorem}[$\kl$ Inequality~\citep{Lan05,FBBT21,FBB22}]\label{thm:kl1}
Let $X_1,\dots,X_n$ be i.i.d.\ with $X_1\in[0,1]$, $\E[X_1] = p$, and $\hat{p}=\frac{1}{n}\sum_{i=1}^n X_i$. Then, for any $\delta\in(0,1)$:
\[
\P[\kl(\hat p\|p)\geq \frac{\ln\frac{1}{\delta}}{n}] \leq \delta.
\]
\end{theorem}
We note that the direct derivation of the $\kl$ inequality that is behind \Cref{thm:kl1} cannot be combined with PAC-Bayesian analysis that we study in \Cref{sec:PAC-Bayes}. There we need to use \Cref{lem:klRef} and pay the cost of $\ln 2\sqrt n$, as in \Cref{thm:klsqrtn}. But in direct applications of the $\kl$ inequality, for example, in combination of the $\kl$ inequality with the Occam's razor, this cost can be avoided (see \Cref{ex:Occam-kl}).

\subsection{Relaxations of the $\kl$-inequality: Pinsker's and refined Pinsker's inequalities}

\Cref{thm:kl1} implies that with probability at least $1-\delta$
\begin{equation}
\label{eq:kl-delta}
\kl(\hat p\|p) \leq \frac{\ln \frac{1}{\delta}}{n}.
\end{equation}
This leads to an implicit bound on $p$, which is not very intuitive and not always convenient to work with. In order to understand the behavior of the $\kl$ inequality better we use a couple of its relaxations. The first relaxation is known as Pinsker's inequality, see \citet[Lemma 11.6.1]{CT06}.

\begin{lemma}[Pinsker's inequality]
\label{lem:Pinsker}
\[
\KL(p\|q) \geq \frac{1}{2} \|p-q\|_1^2,
\]
where $\|p-q\|_1 = \sum_{x\in{\cal X}} |p(x) - q(x)|$ is the $L_1$-norm.
\end{lemma}

\begin{corollary}[Pinsker's inequality for the binary $\kl$ divergence]
\label{cor:kl}
\begin{equation}
\label{eq:klL1}
\kl(p\|q) \geq \frac{1}{2} \lr{|p-q| + |(1 - p) - (1-q)|}^2 = 2 (p-q)^2.
\end{equation}
\end{corollary}

By applying Corollary~\ref{cor:kl} to inequality \eqref{eq:kl-delta}, we obtain that with probability at least $1-\delta$
\begin{equation}
\label{eq:klPinsker}
p\leq \hat p + \sqrt{\frac{\kl(\hat p\|p)}{2}} \leq \hat p + \sqrt{\frac{\ln \frac{1}{\delta}}{2n}},
\end{equation}
where the first inequality is a deterministic inequality following by \eqref{eq:klL1}, and the second inequality holds with probability at least $1-\delta$ by \eqref{eq:kl-delta}. Note that inequality \eqref{eq:klPinsker} is exactly the same as Hoeffding's inequality in Equation \eqref{eq:Hoeffding-direct} (in fact, one way of proving Hoeffding's inequality is by deriving it via the $\kl$ divergence). Therefore, the $\kl$ inequality is always at least as tight as Hoeffding's inequality. But since \eqref{eq:klPinsker} was achieved by Pinsker's relaxation of the $\kl$ inequality, the unrelaxed $\kl$ inequality can be tighter than Hoeffding's inequality.

Next we show that for small values of $\hat p$ the $\kl$ inequality is significantly tighter than Hoeffding's inequality. For this we use refined Pinsker's inequality \citep[Lemma 8.4]{Mar96,Mar97,Sam00,BLM13}.
\begin{lemma}[Refined Pinsker's inequality]
\label{lem:RefinedPinsker}
\[
\kl(p\|q) \geq \frac{(p-q)^2}{2 \max\lrc{p,q}} + \frac{(p-q)^2}{2 \max\lrc{(1-p),(1-q)}}.
\]
\end{lemma}
\begin{corollary}[Refined Pinsker's inequality]
\label{cor:RefPin2}
If $q > p$ then
\[
\kl(p\|q) \geq \frac{(p-q)^2}{2q}.
\]
\end{corollary}
\begin{corollary}[Refined Pinsker's inequality - upper bound]
\label{cor:kl+}
If $\kl(p\|q) \leq \varepsilon$ then
\[
q \leq p + \sqrt{2 p \varepsilon} + 2 \varepsilon.
\]
\end{corollary}
\begin{corollary}[Refined Pinsker's inequality - lower bound]
\label{cor:kl-}
If $\kl(p\|q) \leq \varepsilon$ then
\[
q \geq p - \sqrt{2 p \varepsilon}.
\]
\end{corollary}
By applying Corollary~\ref{cor:kl+} to inequality \eqref{eq:kl-delta}, we obtain that with probability at least $1-\delta$
\begin{equation}
\label{eq:klRefPinsker}
p \leq \hat p + \sqrt{\frac{2 \hat p \ln \frac{1}{\delta}}{n}} + \frac{2 \ln \frac{1}{\delta}}{n}.
\end{equation}
When $\hat p$ is close to zero, the latter inequality is significantly tighter than Hoeffding's inequality. It exhibits what is know as ``fast convergence rate'', where for small values of $\hat p$ it approaches $p$ at the rate of $\frac{1}{n}$ rather than $\frac{1}{\sqrt n}$, as in Hoeffding's inequality. Similarly, \Cref{cor:kl-} clearly illustrates that when $\hat p$ is close to zero, the convergence of $\hat p$ to $p$ from below also has ``fast convergence rate''.

We note that the $\kl$ inequality is always at least as tight as any of its relaxations, and that although there is no analytic inversion of $\kl(\hat p\|p)$, it is possible to invert it numerically to obtain even tighter bounds than the relaxations above. We use $\kl^{-1^+}(\hat{p},\varepsilon):=\max\lrc{p: p\in[0,1] \text{ and }\kl(\hat{p}\|p)\leq \varepsilon}$ to denote the upper inverse of $\kl$ and $\kl^{-1^-}(\hat{p},\varepsilon):=\min\lrc{p: p\in[0,1] \text{ and }\kl(\hat{p}\|p)\leq \varepsilon}$ to denote the lower inverse of $\kl$. Then under the conditions of \Cref{thm:kl1}
\begin{align}
\P[p\geq \kl^{-1^+}\lr{\hat p, \frac{1}{n}\ln\frac{1}{\delta}}
] &\leq \delta,\label{eq:kl-upper}\\
\P[p \leq \kl^{-1^-}\lr{\hat p, \frac{1}{n}\ln\frac{1}{\delta}}]&\leq \delta.\label{eq:kl-lower}
\end{align}
Since $\kl(\hat p\|p)$ is convex in $p$, the inverses can be found using binary search.

Finally, we remind the reader that the random variable in the $\kl$ inequality is $\hat p$. Therefore, the correct way to see all the inequalities above is as inequalities on $\hat p$ rather than $p$. I.e., it is $\hat p$ that does not deviate a lot from $p$ with high probability, rather than $p$ staying close to $\hat p$ with high probability.

\section{Split-$\kl$ Inequality}
\label{sec:split-kl}

The $\kl$ inequality in \Cref{thm:kl1} is almost the tightest that can be achieved for sums of Bernoulli random variables. (It is possible to obtain a bit tighter bounds by analyzing the binomial distribution directly, but the extra gains are minor \citep{Lan05}.). However, it can potentially be very lose for sums of random variables taking values within the $[0,1]$ interval. The reason is that the $\kl$ inequality first maps any random variable taking values in the $[0,1]$ interval to a Bernoulli random variable (a random variable taking values $\lrc{0,1}$) with identical expectation, and then bounds the concentration of the original random variables by concentration of the Bernoulli random variables, as in \Cref{lem:ConvexExtension}. Whenever the original random variables have small variance, and the corresponding Bernoulli random variables have large variance, this approach is unable to exploit the small variance. (See \Cref{ex:Bernoulli-max-variance}, where you are asked to prove that if we fix expectation of a random variable, then Bernoulli random variable has the highest variance out of all random variables taking values in the $[0,1]$ interval and having a target expectation.) As an extreme example, imagine random variables $X_1,\dots,X_n$, which all take value $p=\frac{1}{2}$ with probability $1$. Then for any $\varepsilon > 0$ we have $\P[\lra{p - \frac{1}{n}\sum_{i=1}^n X_i} > \varepsilon] = 0$, whereas the $\kl$ inequality only guarantees convergence of $\hat p=\frac{1}{n}\sum_{i=1}^n X_i$ to $p$ at the rate of $\sqrt{\frac{\ln\frac{1}{\delta}}{n}}$.

\subsection{Split-$\kl$ Inequality for Discrete Random Variables}

In order to address the issue above, \citet{WZCAS24} have proposed a way to represent discrete random variables as a superposition of Bernoulli random variables, and then apply the $\kl$ inequality to the Bernoulli elements in the decomposition. This approach preserves the $\kl$ tightness for the decomposition elements, and through it provides the combinatorial tightness of $\kl$ for general discrete random variables. The approach builds on an earlier work by \citet{WS22} for ternary random variables.

\begin{figure}
    \centering
    \includegraphics[width=\textwidth]{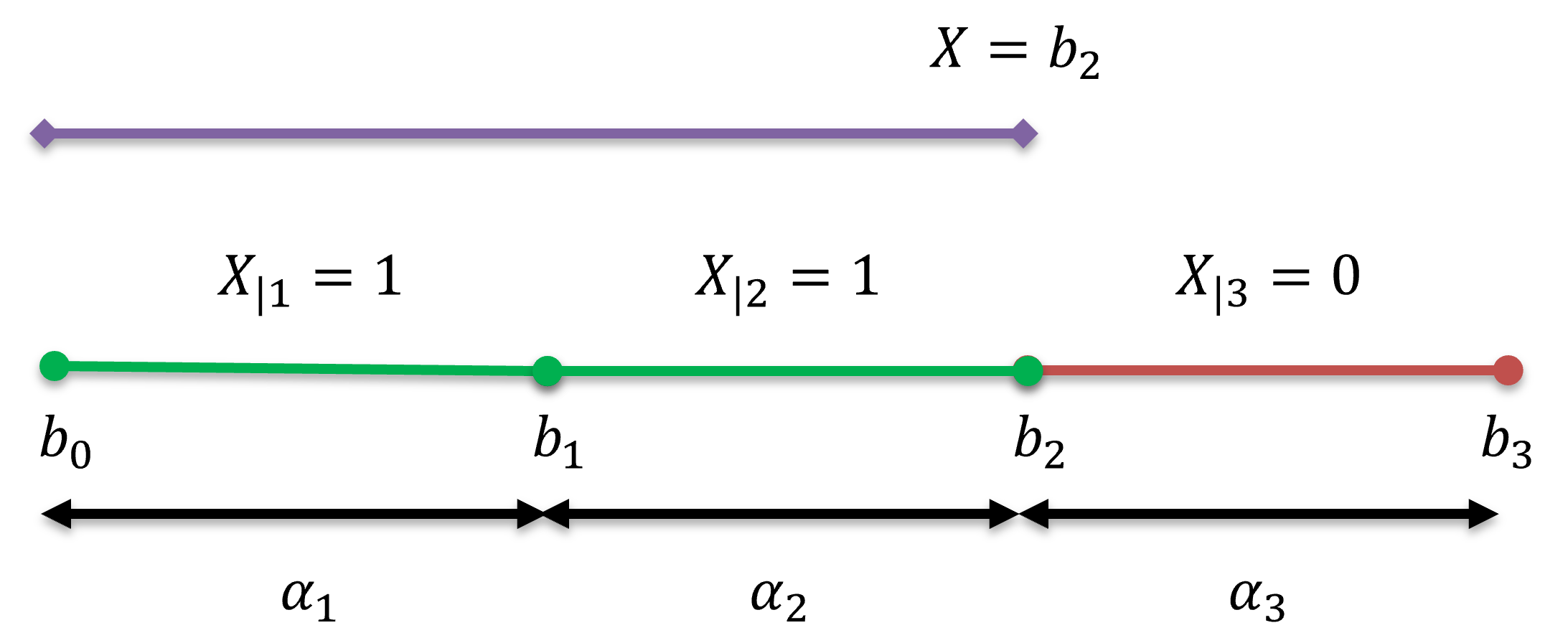}
    \caption{\textbf{Decomposition of a discrete random variable into a superposition of binary random variables.} The figure illustrates a decomposition of a discrete random variable $X$ with domain of four values $b_0 < b_1 < b_2 < b_3$ into a superposition of three binary random variables, $X = b_0 + \sum_{j=1}^3 \alpha_j X_{|j}$. A way to think about the decomposition is to compare it to a progress bar. In the illustration $X$ takes value $b_2$, and so the random variables $X_{|1}$ and $X_{|2}$ corresponding to the first two segments ``light up'' (take value 1), whereas the random variable $X_{|3}$ corresponding to the last segment remains ``turned off'' (takes value 0). The value of $X$ equals the sum of the lengths $\alpha_j$ of the ``lighted up'' segments. (The figure is borrowed from \citet{WZCAS24}.)}
    \label{fig:decomposition}
\end{figure}

We now describe the decomposition. Let $X \in \lrc{b_0,\dots,b_K}$ be a $(K+1)$-valued random variable, where $b_0 < b_1 < \cdots < b_K$. For $j\in\lrc{1,\dots,K}$ define $X_{|j} = \1[X\geq b_j]$ and $\alpha_j = b_j - b_{j-1}$. Then $X = b_0+\sum_{j=1}^K \alpha_j X_{|j}$, see \Cref{fig:decomposition} for an illustration. 

For a sequence $X_1,\dots,X_n$ of $(K+1)$-valued random variables with the same support, we let $X_{i|j} = \1[X_i\geq b_j]$ denote the elements of binary decomposition of $X_i$. 

\begin{theorem}[Split-$\kl$ inequality for discrete random variables {\citep{WZCAS24}}]\label{thm:Split_kl}
Let $X_1,\dots,X_n$ be i.i.d.\ random variables taking values in $\lrc{b_0,\dots,b_K}$ with $\E[X_i] = p$ for all $i$. Let $\hat p_{|j} = \frac{1}{n}\sum_{i=1}^n X_{i|j}$. Then for any $\delta\in(0,1)$:
\[
\P[p\geq b_0+ \sum_{j=1}^K \alpha_j \kl^{-1,+}\lr{\hat p_{|j},\frac{1}{n}\ln\frac{K}{\delta}}]\leq \delta.
\]
\end{theorem}

\begin{proof}
    Let $p_{|j} = \E[\hat p_{|j}]$, then $p = b_0+\sum_{j=1}^K \alpha_j p_{|j}$ and
    \[\P[p\geq b_0+ \sum_{j=1}^K \alpha_j \kl^{-1,+}\lr{\hat p_{|j},\frac{1}{n}\ln\frac{K}{\delta}}] \leq \P[\exists j: p_{|j}\geq \kl^{-1,+}\lr{\hat p_{|j},\frac{1}{n}\ln\frac{K}{\delta}}] \leq \delta,
    \]
    where the first inequality is by the decomposition of $p$ and the second inequality is by the union bound and \eqref{eq:kl-upper}.
\end{proof}

Since the $\kl$ inequalities provide almost the tightest bounds on the deviations of $\hat p_{|j}$ from $p_{|j}$ for each $j$ individually, the split-$\kl$ inequality is almost the tightest that can be achieved for discrete random variables overall, as long as $K$ and the corresponding $\ln K$ cost in the bound is not too large.

\subsection{Split-$\kl$ Inequality for Bounded Continuous Random Variables}

The split-$\kl$ inequality can also be applied to continuous random variables. Let $b_0 < b_1 < \dots < b_K$ be an arbitrary split of an interval $[b_0,b_K]$ into $K$ segments with $\alpha_j = b_j-b_{j-1}$ being the length of segment $j$, and let $X \in [b_0,b_K]$ be a continuous random variable. Let 
\[
X_{|j} = \begin{cases}
0, &\text{if $X < b_{j-1}$,}\\
\frac{X - b_{j-1}}{\alpha_j}, &\text{if $b_{j-1}\leq X\leq b_j$,}\\
1, &\text{if $X > b_j$.}
\end{cases}
\]
Then $X=b_0+\sum_{j=1}^K \alpha_j X_{|j}$, and the split-$\kl$ inequality can be applied in exactly the same way as in the discrete case. The tightness of split-$\kl$ for continuous random variables depends on whether the probability mass is concentrated on or between the segment boundaries $b_0,\dots,b_K$ and on the magnitude of the $\ln K$ cost of the union bound.

\section{Bernstein's Inequality}

Bernstein's inequality is one of the most broadly known tools that exploit small variance to obtain tighter concentration. As most concentration of measure inequalities we have seen so far, it is based on a bound on a moment generating function.

\begin{lemma}[Bernstein's Lemma]
\label{lem:Bernstein}
Let $Z$ be a random variable, such that $\E[Z] = 0$, $\E[Z^2]\leq \nu$, and $Z \leq b$. Then for any $\lambda\in\lr{0,\frac{3}{b}}$
\[
\E[e^{\lambda Z}] \leq \exp\lr{\frac{\lambda^2 \nu}{2\lr{1-\frac{b\lambda}{3}}}}.
\]
\end{lemma}
\begin{proof}
For $x\leq 0$ we have $e^x\leq 1+x+\frac{1}{2}x^2$ and, therefore, for $Z\leq 0$ we have
\[
e^{\lambda Z}\leq 1 + \lambda Z + \frac{\lambda^2 Z^2}{2}  \leq 1 + \lambda Z + \frac{\lambda^2 Z^2}{2\lr{1-\frac{b\lambda}{3}}},
\]
where the last inequality holds because $\lambda\in\lr{0,\frac{3}{b}}$, and so $\lr{1-\frac{b\lambda}{3}} \in (0,1)$.

For $Z>0$ we use Taylor's expansion of the exponent, $e^x = 1 + x + \frac{x^2}{2} + \sum_{i=3}^\infty \frac{1}{i!}x^i$, which gives
\begin{align*}
e^{\lambda Z} &= 1 + \lambda Z + \frac{\lambda^2 Z^2}{2} + \sum_{i=3}^\infty \frac{1}{i!}(\lambda Z)^i\\
&\leq 1 + \lambda Z + \frac{\lambda^2 Z^2}{2} + \frac{\lambda^2 Z^2}{2} \sum_{i=3}^\infty \lr{\frac{1}{3}\lambda Z}^{i-2}\\
&= 1 + \lambda Z + \frac{\lambda^2 Z^2}{2}\sum_{i=0}^\infty \lr{\frac{1}{3}\lambda Z}^i\\
&\leq 1 + \lambda Z + \frac{\lambda^2 Z^2}{2}\sum_{i=0}^\infty \lr{\frac{1}{3}\lambda b}^i\\
&= 1 + \lambda Z + \frac{\lambda^2 Z^2}{2\lr{1-\frac{b\lambda}{3}}}.
\end{align*}
By combining this with the earlier inequality, we obtain that for all $Z$
\[
e^{\lambda Z} \leq 1 + \lambda Z + \frac{\lambda^2 Z^2}{2\lr{1-\frac{b\lambda}{3}}}.
\]
And, therefore,
\[
\E[e^{\lambda Z}] \leq 1 + \lambda \E[Z] + \frac{\lambda^2 \E[Z^2]}{2\lr{1-\frac{b\lambda}{3}}} \leq 1 + \frac{\lambda^2 \nu}{2\lr{1-\frac{b\lambda}{3}}} \leq \exp\lr{\frac{\lambda^2 \nu}{2\lr{1-\frac{b\lambda}{3}}}},
\]
where in the second step we used the facts that $\E[Z] = 0$ and $\E[Z^2]\leq \nu$, and the last step is based on the inequality $1+x \leq e^x$ that holds for all $x$.
\end{proof}

Now we need a couple of technical results, which we leave as an exercise.

\begin{lemma}
\label{lem:Bernstein-dual}
For $x>0$ let $f(x) = 1 + x - \sqrt{1+2x}$, then for any $\varepsilon > 0$
\[
\sup_{\lambda\in\lr{0,\frac{1}{c}}} \lr{\varepsilon \lambda - \frac{\lambda^2 \nu}{2(1-c\lambda)}} = \frac{\nu}{c^2}f\lr{\frac{c\varepsilon}{\nu}}.
\]
\end{lemma}

\begin{lemma}
\label{lem:Bernstein-inverse}
For $x>0$ let $f(x) = 1 + x - \sqrt{1+2x}$, then $f^{-1}(x) = x + \sqrt{2x}$.
\end{lemma}

And now we are ready to present Bernstein's inequality.
\begin{theorem}[Bernstein's Inequality]
\label{thm:Bernstein}
Let $X_1,\dots,X_n$ be independent random variables, such that for all $i$ we have $\E[X_i] - X_i \leq b$ and $\Var[X_i] \leq \nu$. Then
\[
\P[\E[\frac{1}{n}\sum_{i=1}^n X_i] \geq \frac{1}{n}\sum_{i=1}^n X_i + \sqrt{\frac{2\nu \ln\frac{1}{\delta}}{n}} + \frac{b\ln\frac{1}{\delta}}{3n}]\leq \delta.
\]
\end{theorem}

Note that if $\nu$ is close to zero, Bernstein's inequality provides ``fast convergence rate'', meaning that $\frac{1}{n}\sum_{i=1}^n X_i$ converges to $\E[\frac{1}{n}\sum_{i=1}^n X_i]$ at the rate of $\frac1n$ rather than at the rate of $\frac{1}{\sqrt n}$. 
\begin{proof}
The proof is based on Chernoff's bounding technique and follows the same strategy as earlier proofs of Hoeffding's and $\kl$ inequalities, just now using Bernstein's lemma instead of Hoeffding's or $\kl$ lemma. Let $Z_i = \E[X_i] - X_i$, then $\E[Z_i] = 0$, $\E[Z_i^2] = \Var[X_i] \leq \nu$, and $Z_i \leq b$. For any $\lambda \in \lr{0,\frac{b}{3}}$ we have
\begin{align*}
\P[\E[\frac{1}{n}\sum_{i=1}^n X_i] \geq \frac{1}{n}\sum_{i=1}^n X_i + \varepsilon] &= \P[\sum_{i=1}^n Z_i \geq n \varepsilon]\\
&= \P[e^{\lambda \sum_{i=1}^n Z_i} \geq e^{\lambda n \varepsilon}]\\
&\leq e^{-\lambda n\varepsilon} \E[e^{\lambda \sum_{i=1}^n Z_i}]\\
&= e^{-\lambda n\varepsilon} \prod_{i=1}^n \E[e^{\lambda Z_i}]\\
&\leq e^{-\lambda n\varepsilon} \prod_{i=1}^n \exp\lr{\frac{\lambda^2 \nu}{2\lr{1-\frac{b\lambda}{3}}}}\\
&= \exp\lr{-n\lr{\lambda \varepsilon - \frac{\lambda^2  \nu}{2\lr{1-\frac{b\lambda}{3}}}}},
\end{align*}
where the first inequality is by Markov's inequality and the second inequality is by Bernstein's lemma.

Since the bound holds for any $\lambda \in \lr{0,\frac{b}{3}}$, we have
\[
\P[\E[\frac{1}{n}\sum_{i=1}^n X_i] \geq \frac{1}{n}\sum_{i=1}^n X_i + \varepsilon] \leq \exp\lr{-n \sup_{\lambda\in\lr{0,\frac{3}{b}}}\lr{\lambda \varepsilon - \frac{\lambda^2  \nu}{2\lr{1-\frac{b\lambda}{3}}}}} = \exp\lr{-n a f(u)},
\]
where $a = \frac{9\nu}{b^2}$, $u = \frac{b\varepsilon}{3\nu}$, $f(u) = 1 + u - \sqrt{1+2u}$, and the inequality follows by \Cref{lem:Bernstein-dual}. Note that the right hand side of the inequality above is deterministic (independent of the random variable $\frac{1}{n}\sum_{i=1}^n X_i$), meaning that the optimal $\lambda$ can be selected deterministically before observing the sample.

Finally, taking $\exp\lr{-naf(u)} = \delta$ and using \Cref{lem:Bernstein-inverse} to express $\varepsilon$ in terms of $\delta$, we obtain the statement in the theorem.
\end{proof}

\section{Empirical Bernstein's Inequality}

Bernstein's inequality (\Cref{thm:Bernstein}) assumes access to an upper bound $\nu$ on the variance. Empirical Bernstein's inequality presented in this section constructs a high-probability upper bound on the variance based on the sample, and then applies it within Bernstein's inequality, i.e., replaces $\nu$ with a high-probability upper bound on $\V[X]$.

\begin{theorem}[Empirical Bernstein's Inequality \citep{MP09}]
\label{thm:EmpiricalBernstein}
Let $X_1,\dots,X_n$ be independent identically distributed random variables taking values in $[0,1]$. Let $\hat \nu_n = \frac{1}{n(n-1)}\sum_{1\leq i < j\leq n} (X_i - X_j)^2$. Then for any $\delta \in (0,1]$:
\[
\P[\E[\frac{1}{n}\sum_{i=1}^n X_i] \geq \frac{1}{n}\sum_{i=1}^n X_i + \sqrt{\frac{2\hat \nu_n \ln \frac{2}{\delta}}{n}} + \frac{7\ln \frac{2}{\delta}}{3(n-1)}] \leq \delta.
\]
\end{theorem}

The $\ln\frac2\delta$ term in the bound comes from a union bound over empirical bound on the variance and a bound on expectation of $X$. Note that the overall cost of replacing the true variance with its estimate is relatively small. \citet{GZ13} offer a slight improvement of the bound, showing that for $n\geq 5$ the denominator in the last term can be increased from $n-1$ to $n$. We omit a proof of the theorem, but in \Cref{ex:EmpiricalBernstein} you are given a chance to prove a slightly weaker version of the theorem yourself.

\section{Unexpected Bernstein's Inequality}

Empirical Bernstein's inequality (\Cref{thm:EmpiricalBernstein}) proceeds by bounding the variance using empirical variance estimate, and then using Bernstein's inequality to bound the expectation using the variance estimate. Unexpected Bernstein's inequality proceeds by bounding the expectation directly via the first and the second empirical moments. It is based on the following inequality due to \citet[Equation (4.12)]{FGL15}: for $z\leq 1$ and $\lambda \in [0,1)$
\begin{equation}
\label{eq:UnexpectedBernsteinBasis}
e^{-\lambda z + z^2(\lambda + \ln(1-\lambda))} \leq 1 - \lambda z.
\end{equation}
The inequality can be used to prove the Unexpected Bernstein's lemma.
\begin{lemma}[Unexpected Bernstein's lemma \citep{FGL15}]
\label{lem:UnexpectedBernstein}
Let $X$ be a random variable bounded from above by $b>0$. Then for all $\lambda \in \left [0,\frac{1}{b}\right)$
\[
\E[e^{\lambda \lr{\E[X] - X} + \frac{b\lambda + \ln(1-b\lambda)}{b^2}X^2}]\leq 1.
\]
\end{lemma}
A proof is left as \Cref{ex:UnexpectedBernsteinLemma}. Given the Unexpected Bernstein's lemma, we can use already well-established pipeline to prove the Unexpected Bernstein's inequality.
\begin{theorem}[Unexpected Bernstein's Inequality \citep{FGL15,MGG20,WS22}]
\label{thm:UnexpectedBernstein}
Let $X_1,\dots,X_n$ be independent identically distributed random variables bounded from above by $b$ for $b>0$. Let $\mu = \E[X_1]$, $\hat \mu_n = \frac1n\sum_{i=1}^n X_i$, and $\hat s_n = \frac1n\sum_{i=1}^n X_i^2$. Let $\psi(u) = u - \ln(1+u)$. Then for any $\lambda \in [0,1/b)$ and $\delta\in(0,1]$:
\[
\P[\mu \geq \hat \mu_n + \frac{\psi(-\lambda b)}{\lambda b^2}\hat s_n + \frac{\ln\frac1\delta}{\lambda n}] \leq \delta.
\]
\end{theorem}
A proof is left as \Cref{ex:UnexpectedBernstein} and follows exactly the same steps as the proofs of Hoeffding's, $\kl$, and Bernstein's inequalities. As already mentioned, the Unexpected Bernstein's inequality goes in one step from empirical first and second moments, $\hat \mu_n$ and $\hat \nu_n$, to a bound on $\mu$. Note that in contrast to Hoeffding's and Bernstein's inequalities, the value of $\lambda$ that minimizes the bound in \Cref{thm:UnexpectedBernstein} depends on $\hat s_n$, which is a random variable. Therefore, we cannot plug it into the bound. Instead, we can take a grid of $\lambda$ values, a union bound over the grid, and then pick the best $\lambda$ from the grid. \citet{MGG20} proposed to use the grid $\Lambda = \lrc{1/2b, \dots,1/(2^k b)}$ for $k=\lru{\log_2\lr{\sqrt{n/\ln(1/\delta)}/2}}$, which works reasonably well in practice \citep{WS22}. Formally, the bound then becomes:
\[
\P[\mu \geq \hat \mu_n + \min_{\lambda\in\Lambda}\lr{\frac{\psi(-\lambda b)}{\lambda b^2}\hat s_n + \frac{\ln\frac k\delta}{\lambda n}}] \leq \sum_{\lambda\in\Lambda} \P[\mu \geq \hat \mu_n + \frac{\psi(-\lambda b)}{\lambda b^2}\hat s_n + \frac{\ln\frac k\delta}{\lambda n}] \leq \delta.
\]
(Note that $\ln\frac1\delta$ is replaced by $\ln\frac k\delta$ due to the union bound.)

\section{Exercises}

\begin{exercise}[\textit{Tightness of Markov's inequality}]
\label{ex:Markovs-tightness}
Let $\varepsilon^*$ be fixed. Design an example of a random variable $X$ for which
\[
\P[X \geq \varepsilon^*] = \frac{\E[X]}{\varepsilon^*}.
\]
Prove that the above equality holds for your random variable.

\emph{Guidance:} ``Design a random variable'' means design a distribution by which the random variable is distributed. To design a distribution you should say what values the random variable can take and with what probabilities. You should construct an example, where the random variable can take strictly more than one value, because otherwise the example is trivial. But two values are actually sufficient to make a valid non-trivial example.
\end{exercise}

\begin{exercise}[\textit{The effect of normalization in Hoeffding's inequality}]
\label{ex:Hoeffding-normalization}
Prove that \Cref{cor:Hoeffding} (simplified Hoeffding's inequality for random variables in the $[0,1]$ interval) follows from \Cref{thm:Hoeffding} (general Hoeffding's inequality). [Showing this for one of the two inequalities is sufficient.]

\emph{Remark:} Some literature sources present Hoeffding's inequality in the normalized form of \Cref{cor:Hoeffding}, whereas other sources present it in the unnormalized form of \Cref{thm:Hoeffding}. The exercise aims to illustrate how to go from one to the other and back.
\end{exercise}

\begin{exercise}[\textit{Numerical comparison of Markov's, Chebyshev's, and Hoeffding's inequalities}]~
\begin{enumerate}[label = {\Alph*.}]
\item Make 1,000,000 repetitions of the experiment of drawing 20 i.i.d.\ Bernoulli random variables $X_1,\dots,X_{20}$ with mean $0.5$ and answer the following questions.

\begin{enumerate}
	\item Plot the empirical frequency of observing $\frac{1}{20} \sum_{i=1}^{20} X_i \geq \alpha$ for $\alpha \in \lrc{0.5, 0.55, 0.6, \dots, 0.95, 1}$.
	\item Explain why the above granularity of $\alpha$ is sufficient. I.e., why, for example, taking $\alpha = 0.51$ will not provide any extra information about the experiment.
	\item Use Markov's inequality to compute a bound on $\P[\frac{1}{20} \sum_{i=1}^{20} X_i \geq \alpha]$ and plot the bound in the same figure.
	\item Use Chebyshev's inequality to compute a bound on $\P[\frac{1}{20} \sum_{i=1}^{20} X_i \geq \alpha]$ and plot the bound in the same figure. (You may have a problem calculating the bound for some values of $\alpha$. If it happens and whenever the bound exceeds 1, replace it with the trivial bound of 1, because we know that probabilities are always bounded by 1.)
	\item Use Hoeffding's inequality to compute a bound on $\P[\frac{1}{20} \sum_{i=1}^{20} X_i \geq \alpha]$ and plot the bound in the same figure.
	\item Compare the four plots.
	\item For $\alpha = 1$ and $\alpha = 0.95$ calculate the exact probability $\P[\frac{1}{20} \sum_{i=1}^{20} X_i \geq \alpha]$ and compare it with the Hoeffding's bound. 
	(No need to add this one to the plot.)
\end{enumerate}

\item Repeat the question with $X_1,\dots,X_{20}$ with mean $0.1$ (i.e., $\E[X_1] = 0.1$) and $\alpha \in \lrc{0.1, 0.15, \dots, 1}$.

\item Discuss the results.
\end{enumerate}
Do not forget to put axis labels and a legend in your plots. 
\end{exercise}

\begin{exercise}[\textit{The role of independence}] 
\label{ex:independence}
Design an example of identically distributed, but \emph{dependent} Bernoulli random variables $X_1,\dots,X_n$ (i.e., $X_i \in \{0,1\}$), such that
\[
\P[\lra{\mu - \frac{1}{n} \sum_{i=1}^n X_i} \geq \frac{1}{2}] = 1,
\]
where $\mu = \E[X_i]$. 

Note that in this case $\frac{1}{n} \sum_{i=1}^n X_i$ does not converge to $\mu$ as $n$ goes to infinity. The example shows that independence is crucial for convergence of empirical means to the expected values.    
\end{exercise}

\begin{exercise}[\textit{From a lower bound on the expectation to a lower bound on the probability}]
\label{ex:E-P-lower}
The question shows that if the expectation of a \emph{bounded} random variable $X$ is large, then with high probability $X$ should take large values.
\begin{itemize}
    \item Let $X$ be a random variable that is always upper bounded by $b$. Let $c < a < b$. Prove that if $\E[X] \geq a$, then $\P[X \geq c] \geq \frac{a-c}{b-c}$.

    \item Indicate which part of the proof relies on the assumption that $X\leq b$.
\end{itemize}

\emph{Hint:} see the illustration in Figure~\ref{fig:E-P}. The claim can be proved directly or by using Markov's inequality.

\emph{Optional:} Let $c < a$. Construct an example of an \emph{unbounded} random variable, such that $\E[X]\geq a$, but $\P(X\geq c)$ is small. The example shows that the boundedness assumption is crucial.

\begin{figure}
    \centering
    \includegraphics[width=.5\textwidth]{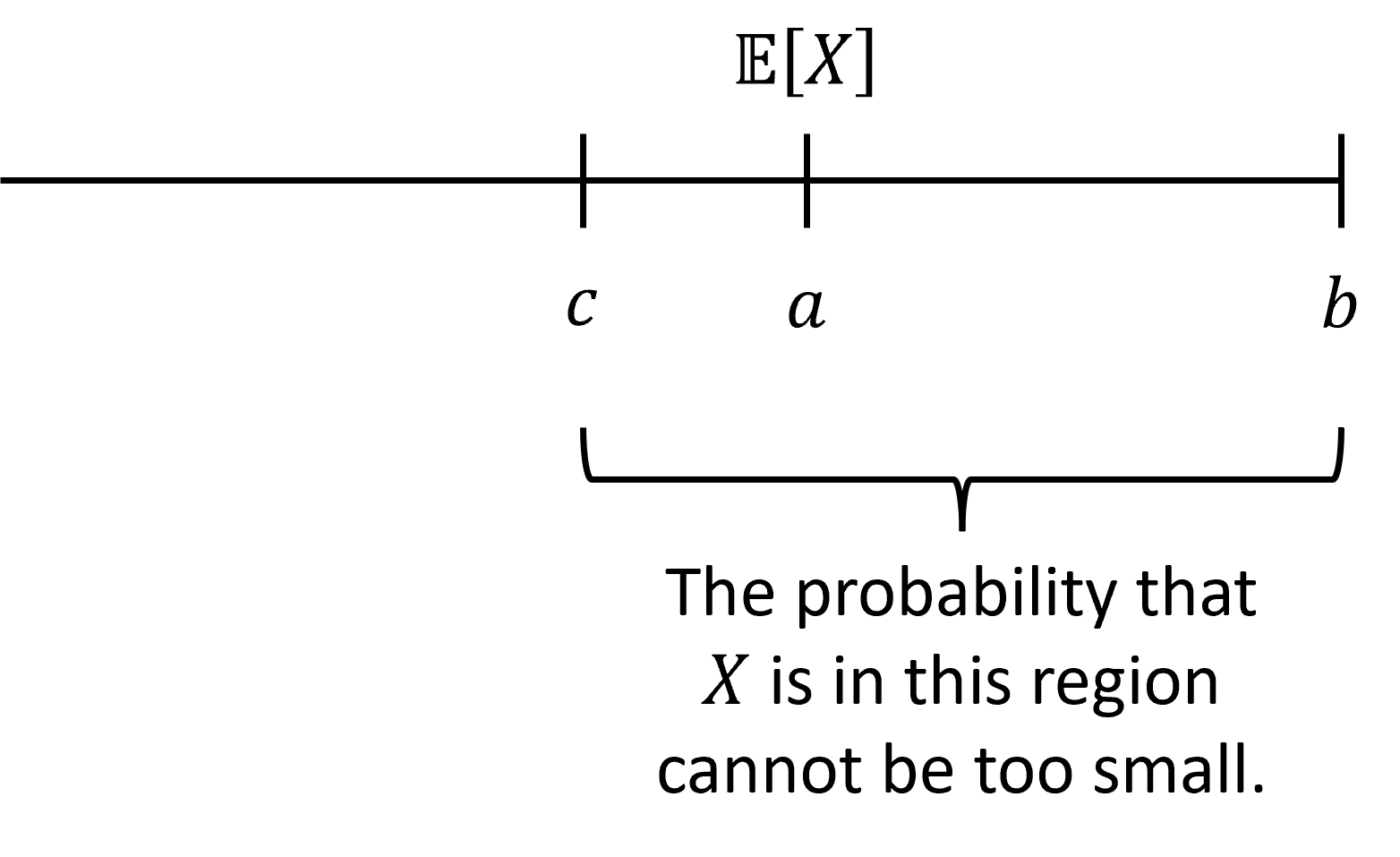}
    \caption{Illustration to \Cref{ex:E-P-lower}.}
    \label{fig:E-P}
\end{figure}
    
\end{exercise}

\begin{exercise}[\textit{Bernoulli Distribution Maximizes the Variance}]
\label{ex:Bernoulli-max-variance}
Let $X$ be a random variable taking values in the $[0,1]$ interval, and let $Y$ be a Bernoulli random variable (taking values $\lrc{0,1}$), such that $\E[X] = \E[Y]$. Prove that $\Var[Y] \geq \Var[X]$.
\end{exercise}

\begin{exercise}[\textit{Asymmetry of the $\kl$ divergence}]
Prove that $\kl$ is asymmetric in its arguments by providing an example of $p$ and $q$ for which $\kl(p\|q) \neq \kl(q\|p)$.    
\end{exercise}

\begin{exercise}[\textit{Numerical comparison of $\kl$ inequality with its relaxations and with Hoeffding's inequality}]
\label{ex:numerical-kl}
Let $X_1,\dots,X_n$ be a sample of $n$ independent Bernoulli random variables with bias $p = \P[X=1]$. Let $\hat p_n = \frac{1}{n} \sum_{i=1}^n X_i$ be the empirical average. In this question you make a numerical comparison of the relative power of various bounds on $p$ we have studied. Specifically, we consider the following bounds:
\begin{enumerate}[label = {\Alph*.}]
	\item \textbf{Hoeffding's inequality}: by Hoeffding's inequality, with probability at least $1-\delta$:
	\[
	p \leq \hat p_n + \sqrt{\frac{\ln \frac{1}{\delta}}{2n}}.
	\]
	(``The Hoeffding's bound'', which you are asked to plot, refers to the right hand side of the inequality above.)
    
	\item \textbf{The $\kl$ inequality}: The bound on $p$ that follows by the $\kl$ inequality (\Cref{thm:kl1}).
    
	\emph{Some guidance:} There is no closed-form expression for computing $\kl^{-1^+}(\hat p_n,\varepsilon)$, so it has to be computed numerically. The function $\kl(\hat p_n\|p)$ is convex in $p$ (you are very welcome to pick some value of $\hat p_n$ and plot $\kl(\hat p_n\|p)$ as a function of $p$ to get an intuition about its shape). We also have $\kl(\hat p_n\|\hat p_n)=0$, which is the minimum, and $\kl(\hat p_n\|p)$ monotonically increases in $p$, as $p$ grows from $\hat p_n$ up to 1. So you need to find the point $p\in[\hat p_n, 1]$ at which the value of $\kl(\hat p_n\|p)$ grows above $\varepsilon$. You could do it inefficiently by linear search or, exponentially more efficiently, by binary search.
	
	\emph{A technicality:} In the computation of $\kl$ we define $0 \ln 0 = 0$. In numerical calculations $0 \ln 0$ is undefined. So you should treat $0 \ln 0$ operations separately, either by directly assigning the zero value or by replacing them with $0 \ln 1 = 0$.
	
	\item \textbf{Pinsker's relaxation of the $\kl$ inequality}: the bound on $p$ that follows from $\kl$-inequality by Pinsker's inequality (\Cref{lem:Pinsker}).
	\item \textbf{Refined Pinsker's relaxation of the $\kl$ inequality}: the bound on $p$ that follows from $\kl$-inequality by refined Pinsker's inequality (\Cref{cor:kl+}).
\end{enumerate}

\noindent
In this task you should do the following:
\begin{enumerate}
	\item Write down explicitly the four bounds on $p$ you are evaluating.
	\item Plot the four bounds on $p$ as a function of $\hat p_n$ for $\hat p_n \in [0,1]$, $n = 1000$, and $\delta = 0.01$. You should plot all the four bounds in one figure, so that you can directly compare them. Clip all the bounds at 1, because otherwise they are anyway meaningless and will only destroy the scale of the figure.
	\item Generate a ``zoom in'' plot for $\hat p_n \in [0,0.1]$.
	\item Compare Hoeffding's lower bound on $p$ with $\kl$ lower bound on $p$ for the same values of $\hat p_n, n, \delta$ in a separate figure (no need to consider the relaxations of the $\kl$). 
	
	\emph{Some guidance:} For computing the ``lower inverse'' $\kl^{-1^-}(\hat p_n,\varepsilon)$ you can either adapt the function for computing the ``upper inverse'' you wrote earlier (and we leave it to you to think how to do this), or implement a dedicated function for computing the ``lower inverse''. Direct computation of the ``lower inverse'' works the same way as the computation of the ``upper inverse''. The function $\kl(\hat p_n\|p)$ is convex in $p$ with minimum $\kl(\hat p_n\|\hat p_n)=0$ achieved at $p=\hat p_n$, and monotonically decreasing in $p$, as $p$ increases from 0 to $\hat p_n$. So you need to find the point $p\in[0,\hat p_n]$ at which the value of $\kl(\hat p_n\|p)$ decreases below $\varepsilon$. You can do it by linear search or, more efficiently, by binary search. And, as mentioned earlier, you can save all the code writing if you find a smart way to reuse the function for computing the ``upper inverse'' to compute the ``lower inverse''. Whatever way you chose you should explain in your main \texttt{.pdf} submission file how you computed the upper and the lower bound.
	
	\item Write down your conclusions from the experiment. For what values of $\hat p_n$ which bounds are tighter and is the difference significant?
	\item {} [Optional, not for submission.] You are welcome to experiment with other values of $n$ and $\delta$.
\end{enumerate}
\end{exercise}

\begin{exercise}[\textit{Refined Pinsker's Upper Bound}]
Prove \Cref{cor:kl+}. You are allowed to base the proof on \Cref{lem:RefinedPinsker}.
\end{exercise}

\begin{exercise}[\textit{Refined Pinsker's Lower Bound}]
Prove \Cref{cor:kl-}. You are allowed to base the proof on \Cref{lem:RefinedPinsker}.
\end{exercise}

\begin{exercise}[\textit{Numerical comparison of the $\kl$ and split-$\kl$ inequalities}]
Compare the $\kl$ and split-$\kl$ inequalities. Take a ternary random variable (a random variable taking three values) $X \in \lrc{0,\frac{1}{2},1}$. Let $p_0=\P[X=0]$, $p_{\frac{1}{2}}=\P[X=\frac{1}{2}]$, and $p_1=\P[X=1]$. Set $p_0=p_1=(1-p_{\frac{1}{2}})/2$, i.e., the probabilities of $X=0$ and $X=1$ are equal, and there is just one parameter $p_\frac{1}{2}$, which controls the probability mass of the central value. Compare the two bounds as a function of $p_\frac{1}{2} \in [0,1]$. Let $p=\E[X]$ (in the constructed example, for any value of $p_\frac{1}{2}$ we have $p=\frac{1}{2}$, because $p_0=p_1$). For each value of $p_\frac{1}{2}$ in a grid covering the $[0,1]$ interval draw a random sample $X_1,\dots,X_n$ from the distribution we have constructed and let $\hat p_n = \frac{1}{n} \sum_{i=1}^n X_i$. Generate a figure, where you plot the $\kl$ and the split-$\kl$ bounds on $p - \hat p_n$ as a function of $p_\frac{1}{2}$ for $p_\frac{1}{2}\in[0,1]$. For the $\kl$ bound, the bound on $p-\hat p_n$ is $\kl^{-1^+}\lr{\hat p_n,\frac{\ln\frac{1}{\delta}}{n}} - \hat p_n$; pay attention that in contrast to \Cref{ex:numerical-kl} we subtract the value of $\hat p_n$ after inversion of $\kl$ to get a bound on the difference $p-\hat p_n$ rather than on $p$. For the split-$\kl$ bound you subtract $\hat p_n$ from the right hand side of the expression inside the probability in \Cref{thm:Split_kl}. Take $n=100$ and $\delta = 0.05$. Briefly reflect on the outcome of the comparison.
\end{exercise}

\begin{exercise}[\textit{A Simple Version of Empirical Bernstein's Inequality}]
\label{ex:EmpiricalBernstein}
In this exercise you will derive a bit weaker form of Empirical Bernstein's inequality through a relatively straightforward derivation.
\begin{enumerate}
    \item Let $X$ and $X'$ be two independent identically distributed random variables. Prove that $\E[(X-X')^2] = 2\Var[X]$.
    \item Let $X_1,\dots,X_n$ be independent identically distributed random variables taking values in the $[0,1]$ interval, and assume that $n$ is even. Let $\hat \nu_n = \frac{1}{n}\sum_{i=1}^{n/2} (X_{2i} - X_{2i-1})^2$ and let $\nu = \Var[X_1]$. Prove that
    \[
    \P[\nu \geq \hat \nu_n + \sqrt{\frac{\ln\frac1\delta}{n}}]\leq \delta.
    \]
    \item Let $X_1,\dots,X_n$, $n$, $\nu$, and $\hat \nu_n$ as before, and let $\mu=\E[X_1]$. Prove that
    \begin{equation}
    \label{eq:WeakEmpiricalBernstein}
    \P[\mu \geq \frac{1}{n}\sum_{i=1}^n X_i + \sqrt{\frac{2\hat \nu_n\ln \frac2\delta}{n}} + \sqrt2\lr{\frac{\ln\frac2\delta}{n}}^{\frac34}+\frac{\ln\frac2\delta}{3n}]\leq \delta.
    \end{equation}
    \emph{Hint: The proof uses the inequalities $\sqrt{a+b}\leq \sqrt a + \sqrt b$ for $a,b\geq 0$, and $\P[A] \leq \P[A|B] + \P[\bar B]$ for any pair of events $A$ and $B$, where $\bar B$ is the complement of $B$.} 
\end{enumerate}
\emph{Discussion: since for $x\leq 1$ we have $x^\frac34\geq x$, the bound in \eqref{eq:WeakEmpiricalBernstein} is slightly weaker than the bound in \Cref{thm:EmpiricalBernstein}, but the proof is much simpler. The work of \citet{MP09} and \citet{GZ13} removes the $\lr{\frac{\ln\frac2\delta}{n}}^{\frac34}$ term.}
\end{exercise}

\begin{exercise}
\label{ex:UnexpectedBernsteinLemma}
Prove \Cref{lem:UnexpectedBernstein}.
\end{exercise}

\begin{exercise}
\label{ex:UnexpectedBernstein}
Prove \Cref{thm:UnexpectedBernstein}.    
\end{exercise}

\chapter{Generalization Bounds for Classification}
\label{ch:Generalization}

One of the most central questions in machine learning is: ``How much can we trust the predictions of a learning algorithm?''. A way of answering this question is by providing generalization bounds on the expected performance of the algorithm on new data points. In this chapter we derive a number of generalization bounds for supervised classification.

\section{Overview: Learning by Selection}
\label{sec:LbS}

\begin{figure}%
\centering
\includegraphics[width=.8\columnwidth]{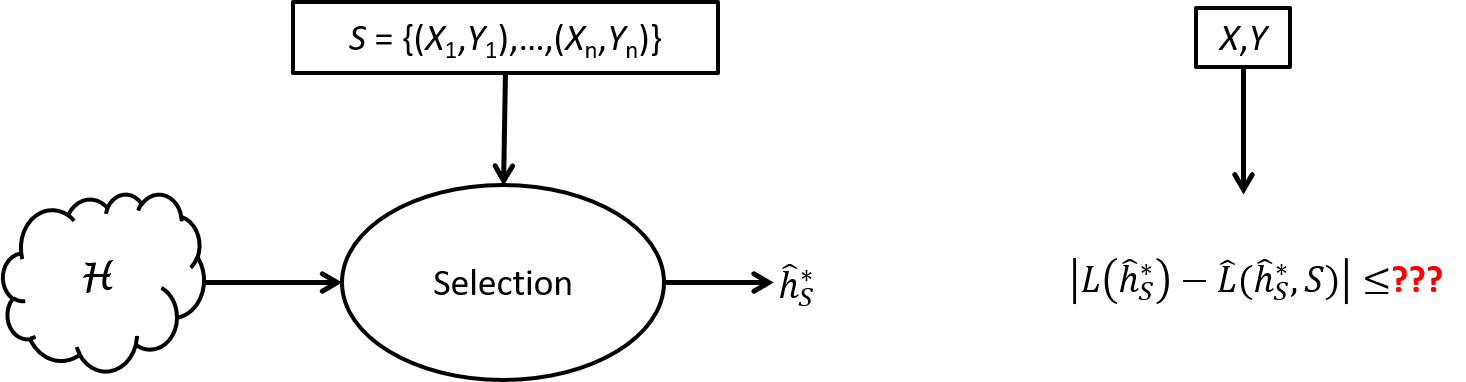}%
\caption{Learning by Selection.}%
\label{fig:selection}%
\end{figure}

The classical process of learning can be seen as a selection process (see Figure~\ref{fig:selection}):
\begin{enumerate}
	\item We start with a hypothesis set $\HH$, which is a set of plausible prediction rules (for example, linear separators).
	\item We observe a sample $S$ sampled i.i.d.\ according to a fixed, but unknown distribution $p(X,Y)$.
	\item Based on the empirical performances $\hat L(h,S)$ of the hypotheses in $\HH$, we \emph{select} a prediction rule $\hbest$, which we consider to be the ``best'' in $\HH$ in some sense. Typically, $\hbest$ is either the \emph{empirical risk minimizer} (ERM), $\hbest = \displaystyle \arg\min_h \hat L(h,S)$, or a regularized empirical risk minimizer.
	\item $\hbest$ is then applied to predict labels for new samples $X$.
\end{enumerate} 

In this chapter we are concerned with the question of what can be said about the expected loss $L(\hbest)$, which is the error we are expected to make on new samples. More precisely, we provide tools for bounding the probability that $\hat L(\hbest,S)$ is significantly smaller than $L(\hbest)$. Recall that $\hat L(\hbest,S)$ is observed and $L(\hbest)$ is unobserved. Having small $\hat L(\hbest,S)$ and large $L(\hbest)$ is undesired, because it means that based on $\hat L(\hbest,S)$ we believe that $\hbest$ performs well, but in reality it does not.

\paragraph{Assumtions} \textbf{\emph{
There are two key assumptions we make throughout the chapter:
\begin{enumerate}
	\item The samples in $S$ are i.i.d..
	\item The new samples $(X,Y)$ come from the same distribution as the samples in $S$.
\end{enumerate}}}
These are the assumptions behind concentration of measure inequalities developed in Chapter~\ref{ch:CoM} and it is important to remember that if they are not satisfied the results derived in this chapter are not valid.

In a sense, it is intuitive why we have to make these assumptions. For example, if we train a language model using data from The Wall Street Journal and then apply it to Twitter the change in prediction accuracy can be very dramatic. Even though both are written in English and comprehensible by humans, the language used by professional journalists writing for The Wall Street Journal is very different from the language used in the short tweets. 

The two assumptions are behind most supervised learning algorithms that you can meet in practice and, therefore, it is important to keep them in mind. In Chapter~\ref{ch:Online} we discuss how to depart from them, but for now we stick with them.

Given the assumptions above, for any fixed prediction rule that is independent of $S$, the empirical loss is an unbiased estimate of the true loss, $\E[\hat L(h,S)] = L(h)$. An intuitive way to see it is that under the assumptions that the samples in $S$ are i.i.d.\ and coming from the same distribution as new samples $(X,Y)$, from the perspective of $h$ the new samples $(X,Y)$ are in no way different from the samples in $S$: any new sample $(X,Y)$ could have happened to be in $S$ instead of some other sample $(X_i,Y_i)$ (they are ``exchangeable''). Formally,
\begin{align*}
\E_{(X_1,Y_1),\dots,(X_n,Y_n)}\lrs{\hat L(h,S)} &= \E_{(X_1,Y_1),\dots,(X_n,Y_n)}\lrs{\frac{1}{n} \sum_{i=1}^n \ell(h(X_i),Y_i)}\\
&= \frac{1}{n} \sum_{i=1}^n \E_{(X_1,Y_1),\dots,(X_n,Y_n)}\lrs{\ell(h(X_i),Y_i)}\\
&= \frac{1}{n} \sum_{i=1}^n \E_{(X_i,Y_i)}\lrs{\ell(h(X_i),Y_i)}\\
&= \frac{1}{n} \sum_{i=1}^n L(h)\\
&= L(h).
\end{align*}
However, when we make the selection of $\hbest$ based on $S$ the ``exchangeability'' argument no longer applies and $\E[\hat L(\hbest,S)] \neq \E[L(\hbest)]$ (note that $\hbest$ is a random variable depending on $S$ and we take expectation with respect to this randomness). This is because $\hbest$ is tailored to $S$ (for example, it minimizes $\hat L(h,S)$) and from the perspective of selection process the samples in $S$ are not exchangeable with new samples $(X,Y)$. If we exchange the samples we may end up with a different $\hbest$. In the extreme case when the hypothesis space $\HH$ is so rich that it can fit any possible labeling of the data (for example, the hypothesis space corresponding to 1-nearest-neighbor prediction rule) we may end up in a situation, where $\hat L(\hbest, S)$ is always zero, but $\E[L(\hbest)] \geq \frac{1}{4}$, as in the following informal example. 

\paragraph{Informal Lower Bound} Imagine that we want to learn a classifier that predicts whether a student's birthday is on an even or odd day based on student's id. Assume that the total number of students is $2n$, that the hypothesis class $\HH$ includes all possible mappings from student id to even/odd, so that $|\HH| = 2^{2n}$, and that we observe a sample of $n$ uniformly sampled students (potentially with repetitions). Since all possible mappings are within $\HH$, we have $\hbest \in \HH$ for which $\hat L(\hbest,S) = 0$. However, $\hbest$ is guaranteed to make zero error only on the samples that were observed, which constitute at most half of the total number of students. For the remaining students $\hbest$ can, at the best, make a random guess which will succeed with probability $\frac{1}{2}$. Therefore, the expected loss of $\hbest$ is $L(\hbest) \geq \frac{1}{2} \cdot 0 + \frac{1}{2} \cdot \frac{1}{2} = \frac{1}{4}$, where the first term is an upper bound on the probability of observing an already seen student times the expected error $\hbest$ makes in this case and the second term is a lower bound on the probability of observing a new student times the expected error $\hbest$ makes in this case. For a more formal treatment see the lower bounds in Chapter~\ref{sec:lower}.

Considering it from the perspective of expectations, we have:
\begin{align*}
\E_{(X_1,Y_1),\dots,(X_n,Y_n)}\lrs{\hat L(\hbest,S)} = &~\E_{(X_1,Y_1),\dots,(X_n,Y_n)}\lrs{\frac{1}{n} \sum_{i=1}^n \ell(\hbest(X_i),Y_i)}\\
= &~\frac{1}{n} \sum_{i=1}^n \E_{(X_1,Y_1),\dots,(X_n,Y_n)}\lrs{\ell(\hbest(X_i),Y_i)}\\
= &~\frac{1}{n} \sum_{i=1}^n \E_{(X_1,Y_1),\dots,(X_n,Y_n)}\lrs{\ell(\hbest(X_1),Y_1)}\\
= &~\E_{(X_1,Y_1),\dots,(X_n,Y_n)}\lrs{\ell(\hbest(X_1),Y_1)}\\
 {\color{red}{\neq}} &~\E_{(X,Y)}\lrs{\E_{(X_1,Y_1),\dots,(X_n,Y_n)}\lrs{\ell(\hbest(X),Y)}}\\
= &~\E_{(X_1,Y_1),\dots,(X_n,Y_n)}\lrs{\E_{(X,Y)}\lrs{\ell(\hbest(X),Y)}}\\
= &~\E_{(X_1,Y_1),\dots,(X_n,Y_n)}\lrs{L(\hbest)}.
\end{align*}

\begin{figure}%
\centering
\includegraphics[width=.8\columnwidth]{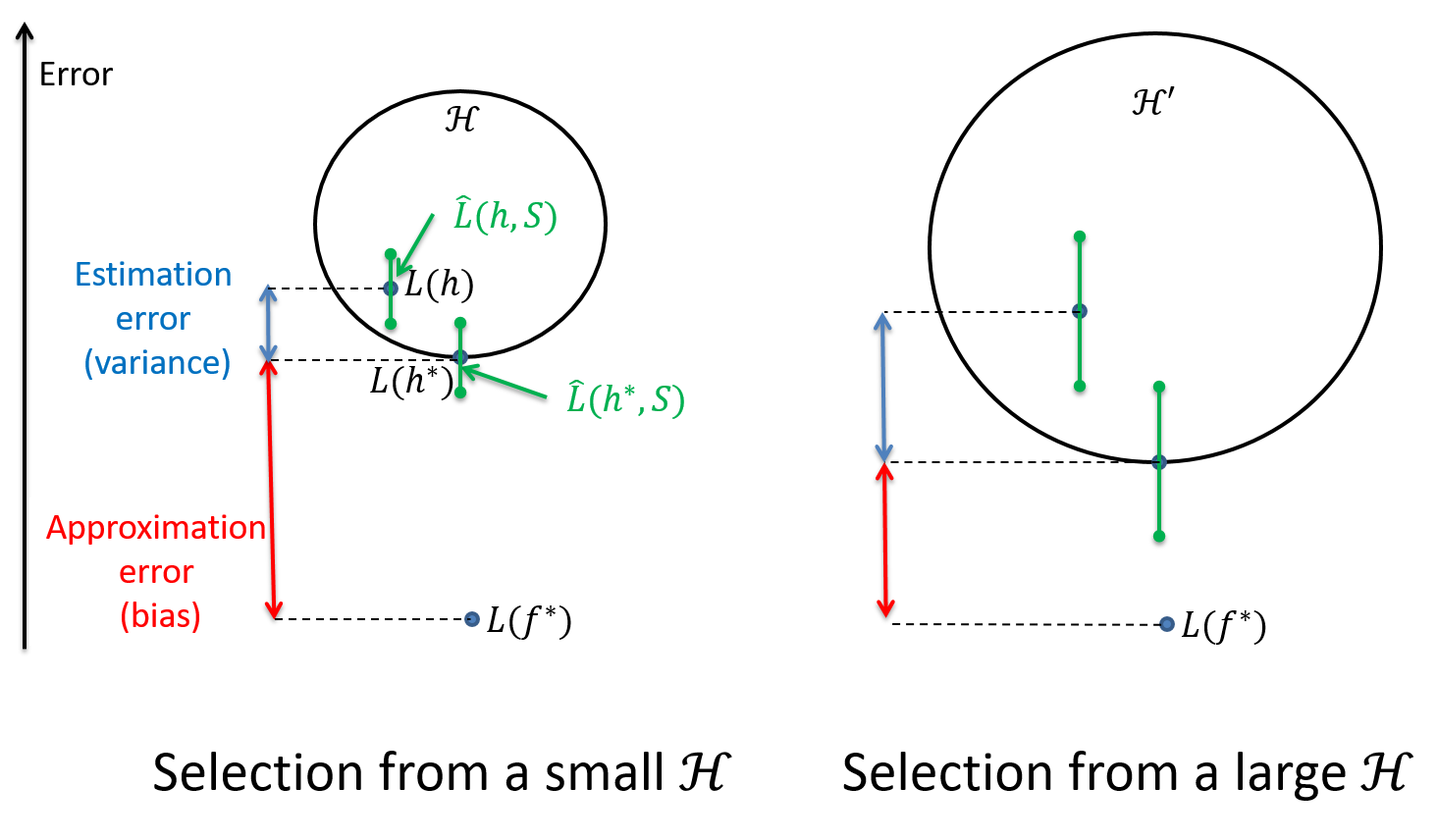}%
\caption{Learning by Selection.}%
\label{fig:bias-variance}%
\end{figure}

The selection leads to the approximation-estimation trade-off (a.k.a.\ bias-variance trade-off), see Figure~\ref{fig:bias-variance}. If the hypothesis class $\HH$ is small it is easy to identify a good hypothesis $h$ in $\HH$, but since $\HH$ is small it is likely that all the hypotheses in $\HH$ are weak. On the other hand, if $\HH$ is large it is more likely to contain stronger hypotheses, but at the same time the probability of confusion with a poor hypothesis grows. This is because there is always a small chance that the empirical loss $\hat L(h,S)$ does not represent the true loss $L(h)$ faithfully. The more hypotheses we take, the higher is the chance that $\hat L(h,S)$ is misleading for some of them, which increases the chance of confusion.

\begin{figure}%
\centering
\includegraphics[width=\columnwidth]{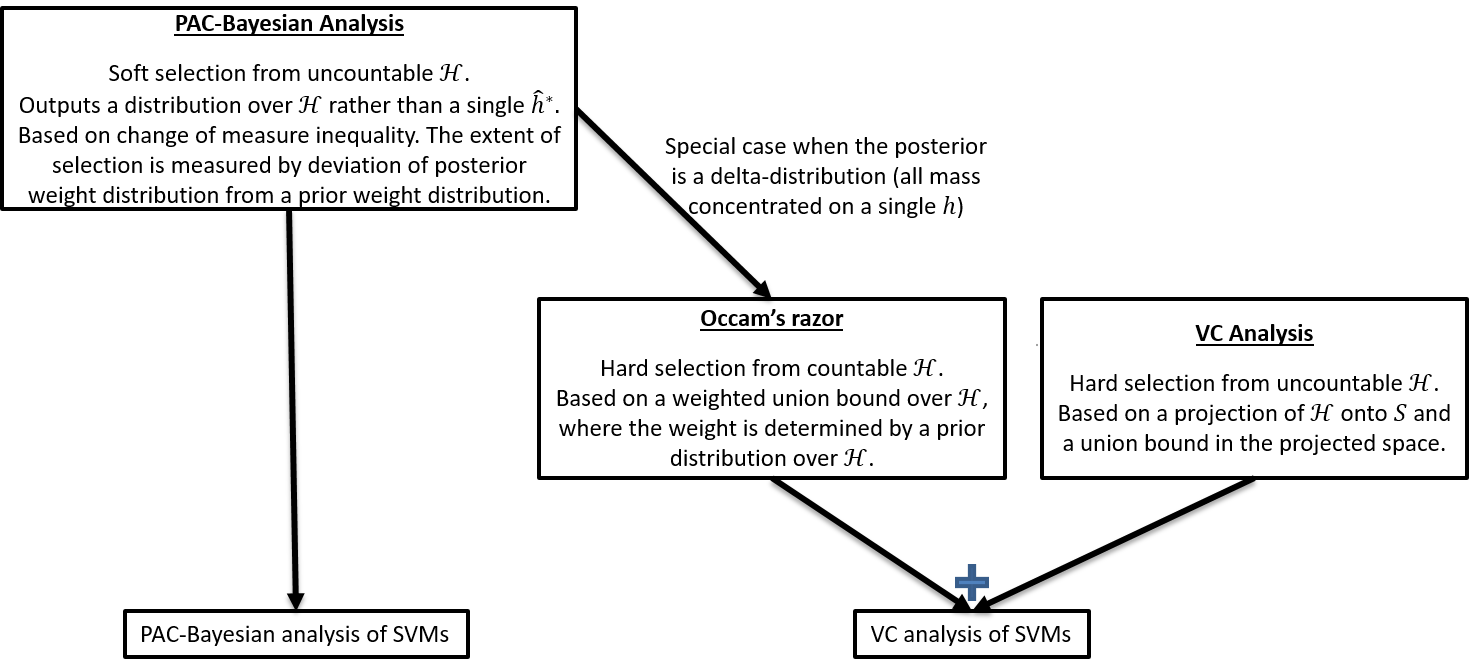}%
\caption{Overview of the major approaches to derivation of generalization bounds considered in this chapter.}%
\label{fig:generalization-bounds-overview}%
\end{figure}

Finding a good balance between approximation and estimation errors is one of the central questions in machine learning. The main tool for analyzing the trade-off from the theoretical perspective are concentration of measure inequalities. Since concentration of measure inequalities do not apply when the prediction rule $\hbest$ depends on $S$, the main approach to analyzing the prediction power of $\hbest$ is to consider cases with no dependency and then take a union bound over selection from these cases. In this chapter we study three different ways of implementing this idea, see Figure~\ref{fig:generalization-bounds-overview} for an overview. We distinguish between \emph{hard selection}, where the learning procedure returns a single hypothesis $h$ and \emph{soft selection}, where the learning procedure returns a distribution over $\HH$.
\begin{enumerate}
	\item \emph{Occam's razor} applies to \emph{hard selection} from a \emph{countable} hypothesis space $\HH$ and it is based on a weighted union bound over $\HH$. We know that for every fixed $h$ the expected loss is close to the empirical loss, meaning that $\lra{L(h) - \hat L(h,S)}$ is small. When $\HH$ is countable we can take a weighted union bound and obtain that $\lra{L(h) - \hat L(h,S)}$ is ``small'' for all $h\in\HH$ (where the magnitude of ``small'' is inversely proportional to the weight of $h$ in the union bound) and thus it is ``small'' for $\hbest$.
	\item \emph{Vapnik-Chervonenkis (VC) analysis} applies to \emph{hard selection} from an \emph{uncountable} hypothesis space $\HH$ and it is based on projection of $\HH$ onto $S$ and a union bound over what we obtain after the projection. The idea is that even when $\HH$ is uncountably infinite, there is only a finite number of ``behaviors'' (ways to label $S$) we can observe on a finite sample $S$. In other words, when we look at $\HH$ through the lens of $S$, we can only distinguish a finite number of subsets of $\HH$, whereas everything that falls within the subsets is equivalent in terms of $\hat L(h,S)$. Therefore, $S$ only serves for a (finite) selection of a subset of $\HH$ out of a finite number of subsets, whereas the (infinite) selection from within the subset is independent of $S$. Selection that is independent of $S$ introduces no bias. As before, the VC analysis exploits the fact that for any fixed $h$ the distance $\lra{L(h) - \hat L(h,S)}$ is small and then takes a union bound over the potential dependencies, which are the dependencies between the subsets (the projections) and $S$.
	\item \emph{PAC-Bayesian analysis} applies to \emph{soft selection} from an \emph{uncountable} hypothesis space $\HH$ and it is based on \emph{change of measure inequality}, which can be seen as a refinement of the union bound. Unlike the preceding two approaches, which return a single classifier $\hbest$, PAC-Bayesian analysis returns a \emph{randomized classifier} defined by a distribution $\rho$ over $\HH$. The actual classification then happens by drawing a new classifier $h$ from $\HH$ according to $\rho$ at each prediction round and applying it to make a prediction. When $\HH$ is countable, $\rho$ can (but does not have to) be a delta-distribution allocating all the mass to a single hypothesis $\hbest$, and in this case the generalization guarantees recover those obtained by the Occam's razor approach. The amount of selection is measured by deviation of $\rho$ from a prior distribution $\pi$, where $\pi$ is selected independently of $S$. It is natural to put more of $\rho$-mass on hypotheses that perform well on $S$, but the more we skew $\rho$ toward well-performing hypotheses the more it deviates from $\pi$. This provides a more refined way of measuring the amount of selection compared to the other two approaches. Furthermore, randomization allows to avoid selection when it is not necessary. The avoidance of selection reduces the variance without impairing the bias. For example, when two hypotheses have similar empirical performance we do not have to commit to one of them, but can instead distribute $\rho$ equally among them. The analysis then provides a certain ``bonus'' for avoiding commitment.
\end{enumerate}

\section{Generalization Bound for a Single Hypothesis}
\label{sec:single-h}

We start with the simplest case, where ${\cal H}$ consists of a single prediction rule $h$. We are interested in the quality of $h$, measured by $\ERR(h)$, but all we can measure is $\Err(h, S)$. What can we say about $\ERR(h)$ based on $\Err(h,S)$? Note that the samples $(X_i,Y_i) \in S$ come from the same distribution as any future samples $(X,Y)$ we will observe. Therefore, $\ell(h(X_i),Y_i)$ has the same distribution as $\ell(h(X),Y)$ for any future sample $(X,Y)$. Let $Z_i = \err(h(X_i),Y_i)$ be the loss of $h$ on $(X_i,Y_i)$. Then $\Err(h,S) = \frac{1}{n} \sum_{i=1}^n Z_i$ is an average of $n$ i.i.d.\ random variables with $\EEE{Z_i} = \EEE{\ell(h(X),Y)} = L(h)$. The distance between $\Err(h,S)$ and $\ERR(h)$ can thus be bounded by application of Hoeffding's inequality.
\begin{theorem}
\label{thm:Single}
Assume that $\err$ is bounded in the $[0,1]$ interval (i.e., $\err(Y',Y) \in [0,1]$ for all $Y'$, $Y$), then for a single $h$ and any $\delta \in (0,1)$ we have:
\begin{equation}
\label{eq:SingleBound}
\P[\ERR(h) \geq \Err(h,S) + \sqrt{\frac{\ln \frac{1}{\delta}}{2n}}] \leq \delta
\end{equation}
and
\begin{equation}
\label{eq:SingleBound+-}
\P[\lra{\ERR(h) - \Err(h,S)} \geq \sqrt{\frac{\ln \frac{2}{\delta}}{2n}}] \leq \delta.
\end{equation}
\end{theorem}
\begin{proof}
For \eqref{eq:SingleBound} take $\varepsilon = \sqrt{\frac{\ln \frac{1}{\delta}}{2n}}$ in \eqref{eq:Hoeff01-} and rearrange the terms. Equation \eqref{eq:SingleBound+-} follows in a similar way from the two-sided Hoeffding's inequality. Note that in \eqref{eq:SingleBound} we have $\frac{1}{\delta}$ and in \eqref{eq:SingleBound+-} we have $\frac{2}{\delta}$.
\end{proof}

There is an alternative way to read equation \eqref{eq:SingleBound}: with probability at least $1-\delta$ we have
\[
\ERR(h) \leq \Err(h,S) + \sqrt{\frac{\ln \frac{1}{\delta}}{2n}}.
\]
We remind the reader that the above inequality should actually be interpreted as 
\[
\hat L(h,S) \geq L(h) - \sqrt{\frac{\ln \frac{1}{\delta}}{2n}}
\]
and it means that with probability at least $1-\delta$ the empirical loss $\hat L(h,S)$ does not underestimate the expected loss $L(h)$ by more than $\sqrt{\ln(1/\delta) / 2n}$. However, it is customary to write the inequality in the first form (as an upper bound on $L(h)$ and we follow the tradition (see the discussion at the end of Section~\ref{sec:understand-Hoeffding}).

Theorem \ref{thm:Single} is analogous to the problem of estimating a bias of a coin based on coin flip outcomes. There is always a small probability that the flip outcomes will not be representative of the coin bias. For example, it may happen that we flip a fair coin 1000 times (without knowing that it is a fair coin!) and observe ``all heads'' or some other misleading outcome. And if this happens we are doomed - there is nothing we can do when the sample does not represent the reality faithfully. Fortunately for us, this happens with a small probability that decreases exponentially with the sample size $n$.

Whether we use the one-sided bound \eqref{eq:SingleBound} or the two-sided bound \eqref{eq:SingleBound+-} depends on the situation. In most cases we are interested in the upper bound on the expected performance of the prediction rule given by \eqref{eq:SingleBound}.

\section{Generalization Bound for Finite Hypothesis Classes}
\label{sec:finite-H}

\begin{figure}%
\includegraphics[width=\columnwidth]{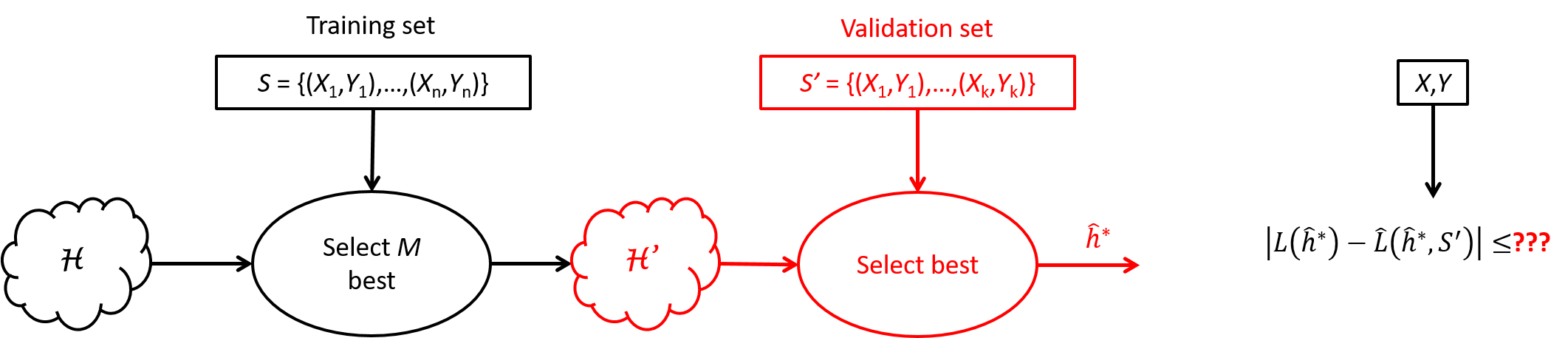}%
\caption{Validation (the red part in the figure) is identical to learning with a reduced hypothesis set ${\cal H}'$ (most often ${\cal H}'$ is finite).}%
\label{fig:validation}%
\end{figure}

A hypothesis set ${\cal H}$ containing a single hypothesis is a very boring set. In fact, we cannot learn in this case, because we end up with the same single hypothesis no matter what the sample $S$ is. Learning becomes interesting when training sample $S$ helps to improve future predictions or, equivalently, decrease the expected loss $\ERR(h)$. In this section we consider the simplest non-trivial case, where ${\cal H}$ consists of a finite number of hypotheses $M$. There are at least two cases, where we meet a finite ${\cal H}$ in real life. The first is when the input space ${\cal X}$ is finite. This case is relatively rare. The second and much more frequent case is when ${\cal H}$ itself is an outcome of a learning process. For example, this is what happens in a validation procedure, see Figure \ref{fig:validation}. In validation we are using a validation set in order to select the best hypothesis out of a finite number of candidates corresponding to different parameter values and/or different algorithms. 

And now comes the delicate point. Let $\hbest$ be a hypothesis with minimal empirical risk, $\displaystyle \hbest = \arg \min_h \hat L(h,S)$ (it is natural to pick the empirical risk minimizer $\hbest$ to make predictions on new samples, but the following discussion equally applies to any other selection rule that takes sample $S$ into account; note that there may be multiple hypotheses that achieve the minimal empirical error and in this case we can pick one arbitrarily). While for each $h$ individually $\EEE{\Err(h,S)} = L(h)$, this is not true for $\EEE{\Err(\hbest,S)}$. In other words, $\EEE{\Err(\hbest,S)} \neq \EEE{L(\hbest)}$ (we have to put expectation on the right hand side, because $\hbest$ depends on the sample). The reason is that when we pick $\hbest$ that minimizes the empirical error on $S$, from the perspective of $\hbest$ the samples in $S$ no longer look identical to future samples $(X,Y)$. This is because $\hbest$ is selected in a very special way - it is selected to minimize the empirical error on $S$ and, thus, it is tailored to $S$ and most likely does better on $S$ than on new random samples $(X,Y)$. One way to handle this issue is to apply a union bound.

\begin{theorem}
\label{thm:Finite}
Assume that $\err$ is bounded in the $[0,1]$ interval and that $|{\cal H}| = M$. Then for any $\delta \in (0,1)$ we have:
\begin{equation}
\label{eq:Finite}
\P[\exists h \in {\cal H} : \ERR(h) \geq \Err(h,S) + \sqrt{\frac{\ln \frac{M}{\delta}}{2n}}] \leq \delta.
\end{equation}
\end{theorem}

\begin{proof}
\[
\P[\exists h \in {\cal H} : \ERR(h) \geq \Err(h,S) + \sqrt{\frac{\ln \frac{M}{\delta}}{2n}}] \leq \sum_{h \in {\cal H}} \P[\ERR(h) \geq \Err(h,S) + \sqrt{\frac{\ln \frac{M}{\delta}}{2n}}] \leq \sum_{h \in {\cal H}} \frac{\delta}{M} = \delta,
\]
where the first inequality is by the union bound and the second is by Hoeffding's inequality.
\end{proof}

Another way of reading Theorem \ref{thm:Finite} is: with probability at least $1-\delta$ for all $h \in {\cal H}$
\begin{equation}
\label{eq:FiniteBound}
\ERR(h) \leq \Err(h,S) + \sqrt{\frac{\ln \frac{M}{\delta}}{2n}}.
\end{equation}
It means that no matter which $h$ from ${\cal H}$ is returned by the algorithm, with high probability we have the guarantee \eqref{eq:FiniteBound}. In particular, it holds for $\hbest$. Again, remember that the random quantity is actually $\hat L(h,S)$ and the right way to read the bound is $\hat L(h,S) \geq L(h) - \sqrt{\ln(M/\delta)/2n}$, see the discussion in the previous section.

The price for considering $M$ hypotheses instead of a single one is $\ln M$. Note that it grows only logarithmically with $M$! Also note that there is no contradiction between the upper bound and the lower bound we have discussed in Section~\ref{sec:LbS}. In the construction of the lower bound we took $M = |\HH| = 2^{2n}$. If we substitute this value of $M$ into \eqref{eq:FiniteBound} we obtain $\sqrt{\ln(M/\delta)/2n} \geq \sqrt{\ln(2)} \geq 0.8$, which has no contradiction with $L(h) \geq 0.25$.

Similar to theorem \ref{thm:Single} it is possible to derive a two-sided bound on the error. It is also possible to derive a lower bound by using the other side of Hoeffding's inequality \eqref{eq:Hoeff01+}: with probability at least $1-\delta$, for all $h\in\HH$ we have $L(h) \geq \hat L(h,S) - \sqrt{\ln(M/\delta)/2n}$. Typically we want the upper bound on $L(h)$, but if we want to compare two prediction rules, $h$ and $h'$, we need an upper bound for one and a lower bound for the other. The ``lazy'' approach is to take the two-sided bound for everything, but sometimes it is possible to save the factor of $\ln(2)$ by carefully considering which hypotheses require the lower bound and which require the upper bound and applying the union bound correspondingly (we are not getting into the details).

\section{Occam's Razor Bound}
\label{sec:Occam}

Now we take a closer look at Hoeffding's inequality. It says that
\[
\P[\ERR(h) \geq \Err(h,S) + \sqrt{\frac{\ln \lr{\frac{1}{\delta}}}{2n}}] \leq \delta,
\]
where $\delta$ is the probability that things go wrong and $\Err(h,S)$ happens to be far away from $\ERR(h)$ because $S$ is not representative for the performance of $h$. There is a dependence between the probability that things go wrong and the requirement on the closeness between $\ERR(h)$ and $\Err(h,S)$. If we want them to be very close (meaning that $\ln\lr{\frac{1}{\delta}}$ is small) then $\delta$ has to be large, but if we can allow larger distance then $\delta$ can be smaller.

So, $\delta$ can be seen as our ``confidence budget'' (or, more precisely, ``uncertainty budget'') - the probability that we allow things to go wrong. The idea behind Occam's Razor bound is to distribute this budget unevenly among the hypotheses in ${\cal H}$. We use $\pi(h) \geq 0$, such that $\sum_{h\in\HH} \pi(h) \leq 1$ as our distribution of the confidence budget $\delta$, where each hypothesis $h$ is assigned $\pi(h)$ fraction of the budget. This means that for every hypothesis $h\in\HH$ the sample $S$ is allowed to be ``non representative'' with probability at most $\pi(h)\delta$, so that the probability that there exists any $h\in\HH$ for which $S$ is not representative is at most $\delta$ (by the union bound). The price that we pay is that the precision (the closeness of $\hat L(h,S)$ to $L(h)$) now differs from one hypothesis to another and depends on the confidence budget $\pi(h)\delta$ that was assigned to it. More precisely, $\hat L(h,S)$ is allowed to underestimate $L(h)$ by up to $\sqrt{\ln(1/(\pi(h)\delta))/2n}$. The precision increases when $\pi(h)$ increases, but since $\sum_{h\in\HH} \pi(h) \leq 1$ we cannot afford high precision for every $h$ and have to compromise. More on this in the next theorem and its applications that follow.

\begin{theorem}[Occam's razor]
\label{thm:Occam}
Let $\ell$ be bounded in $[0,1]$, let ${\cal H}$ be a countable hypothesis set and let $\pi(h)$ be independent of the sample and satisfying $\pi(h) \geq 0$ for all $h$ and $\displaystyle \sum_{h \in {\cal H}} \pi(h) \leq 1$. Then:
\[
\P[\exists h \in {\cal H}: \ERR(h) \geq \Err(h,S) + \sqrt{\frac{\ln\lr{\frac{1}{\pi(h)\delta}}}{2n}}] \leq \delta.
\]
\end{theorem}

\begin{proof}
\begin{align*}
\P[\exists h \in {\cal H}: \ERR(h) \geq \Err(h,S) + \sqrt{\frac{\ln\lr{\frac{1}{\pi(h)\delta}}}{2n}}] &\leq \sum_{h \in {\cal H}} \P[\ERR(h) \geq \Err(h,S) + \sqrt{\frac{\ln\lr{\frac{1}{\pi(h)\delta}}}{2n}}]\\
&\leq \sum_{h \in {\cal H}} \pi(h) \delta\\
&\leq \delta,
\end{align*}
where the first inequality is by the union bound, the second inequality is by Hoeffding's inequality, and the last inequality is by the assumption on $\pi(h)$. Note that $\pi(h)$ has to be selected before we observe the sample (or, in other words, independently of the sample), otherwise the second inequality does not hold. More explicitly, in Hoeffding's inequality $\P[\E[Z_1] - \frac{1}{n} \sum_{i=1}^n Z_i \geq \sqrt{\ln(1/\delta')/2n}] \leq \delta'$ the parameter $\delta'$ has to be independent of $Z_1,\dots,Z_n$. For $\pi(h)$ independent of $S$ we take $\delta' = \pi(h)\delta$ and apply the inequality. But if $\pi(h)$ would be dependent on $S$ we would not be able to apply it.
\end{proof}

Another way of reading Theorem \ref{thm:Occam} is that with probability at least $1-\delta$, for all $h \in {\cal H}$:
\[
\ERR(h) \leq \Err(h,S) + \sqrt{\frac{\ln\lr{\frac{1}{\pi(h)\delta}}}{2n}}.
\]
Again, refer back to the discussion in Section~\ref{sec:single-h} regarding the correct interpretation of the inequality. Note that the bound on $\ERR(h)$ depends both on $\Err(h,S)$ and on $\pi(h)$. Therefore, according to the bound, the best generalization is achieved by $h$ that optimizes the trade-off between empirical performance $\Err(h,S)$ and $\pi(h)$, where $\pi(h)$ can be interpreted as a complexity measure or a prior belief. Also, note that $\pi(h)$ can be designed arbitrarily, but it should be independent of the sample $S$. If $\pi(h)$ happens to put more mass on $h$-s with low $\Err(h,S)$ the bound will be tighter, otherwise the bound will be looser, but it will still be a valid bound. But we cannot readjust $\pi(h)$ after observing $S$! Some considerations behind the choice of $\pi(h)$ are provided in Section~\ref{sec:DecisionTrees}.

Also note that while we can select $\pi(h)$ such that $\sum_{h\in\HH} \pi(h) = 1$ and interpret $\pi$ as a probability distribution over $\HH$, it is not a requirement (we may have $\sum_{h\in\HH} \pi(h) < 1$) and $\pi$ is used as an auxiliary construction for derivation of the bound rather than the prior distribution in the Bayesian sense (for readers who are familiar with Bayesian learning). However, we can use $\pi$ to incorporate prior knowledge into the learning procedure.

\subsection{Applications of Occam's Razor bound}
\label{sec:Occam-apps}

We consider two applications of Occam's Razor bound.

\subsubsection{Generalization bound for finite hypotheses spaces}

An immediate corollary of Occam's razor bound is the generalization bound for finite hypotheses classes that we have already seen in Section~\ref{sec:finite-H}. 

\begin{corollary}
Let ${\cal H}$ be a finite hypotheses class of size $M$, then
\[
\P[\exists h \in {\cal H}: \ERR(h) \geq \Err(h,S) + \sqrt{\frac{\ln \lr{M/\delta}}{2n}}] \leq \delta.
\]
\end{corollary}

\begin{proof}
We set $\pi(h) = \frac{1}{M}$ (which means that we distribute the confidence budget $\delta$ uniformly among the hypotheses in ${\cal H}$) and apply Theorem \ref{thm:Occam}.
\end{proof}

\subsubsection{Generalization bound for binary decision trees}
\label{sec:DecisionTrees}

\begin{theorem}
\label{thm:decision-trees}
Let ${\cal H}_d$ be the set of binary decision trees of depth $d$ and let ${\cal H} = \bigcup_{d=0}^\infty {\cal H}_d$ be the set of binary decision trees of unlimited depth. Let $d(h)$ be the depth of tree (hypothesis) $h$. Then 
\[
\P[\exists h \in {\cal H}: \ERR(h) \geq \Err(h,S) + \sqrt{\frac{\ln \lr{2^{2^{d(h)}} 2^{d(h)+1} /\delta}}{2n}}] \leq \delta.
\]
\end{theorem}

\begin{proof}
We first note that $|{\cal H}_d| = 2^{2^{d}}$. We define $\pi(h) = \frac{1}{2^{d(h)+1}}\frac{1}{2^{2^{d(h)}}}$. The first part of $\pi(h)$ distributes the confidence budget $\delta$ among ${\cal H}_d$-s (we can see it as $\pi({\cal H}_d) = \frac{1}{2^{d+1}}$, the share of confidence budget that goes to ${\cal H}_d$) and the second part of $\pi(h)$ distributes the confidence budget uniformly within ${\cal H}_d$. Since $\sum_{d=0}^\infty \frac{1}{2^{d+1}} = 1$, the assumption $\sum_{h \in {\cal H}} \pi(h) \leq 1$ is satisfied. The result follows by application of Theorem~\ref{thm:Occam}.
\end{proof}
Note that the bound depends on $\ln \lr{\frac{1}{\pi(h)\delta}}$ and the dominating term in $\frac{1}{\pi(h)}$ is $2^{2^{d(h)}}$. 
It comes from the uniform distribution of confidence within ${\cal H}_d$, which makes sense unless we have some prior information about the problem. In absence of such information there is no reason to give preference to any of the trees within ${\cal H}_d$, because ${\cal H}_d$ is symmetric.

Thus, the prior in the proof of Theorem~\ref{thm:decision-trees} exploits structural symmetries within the hypothesis subclasses $\HH_d$ and assigns equal weight to hypotheses that are symmetric under permutation of names of the input variables. While we want $\pi(h)$ to be as large as possible for every $h$, the number of such permutation symmetric hypotheses is the major barrier dictating how large $\pi(h)$ can be (because $\pi$ has to satisfy $\sum_{h\in\HH} \pi(h) \leq 1$). Deeper trees have more symmetric permutations and, therefore, get smaller $\pi(h)$ compared to shallower trees. If there is prior information that breaks the permutation symmetry, it can be used to assign higher prior to the corresponding trees, and if it correctly reflects the true data distribution it will also lead to tighter bounds. If the prior information does not match the true data distribution such adjustments may have the opposite effect.

Concerning the top-level distribution of confidence budget over $\HH_d$-s, the $\pi(\HH_d) = \frac{1}{2^{d+1}}$ part, we could have selected a different series to work with. For example, we could have used $\pi(\HH_d) = \frac{1}{(d+1)(d+2)}$ (for which we have $\sum_{d=0}^\infty \frac{1}{(d+1)(d+2)} = \sum_{d=1}^\infty \frac{1}{d(d+1)} = \sum_{d=1}^\infty \lr{\frac{1}{d} - \frac{1}{d+1}} = 1$) or any other series that sums up to 1. In the case of binary decision trees the dominating complexity term is $\ln \lr{2^{2^{d(h)}}}$, and the choice of the top-level prior has a small impact. However, more generally in absence of prior knowledge ``flatter'' priors, like the one based on $\frac{1}{(d+1)(d+2)}$ series, make more sense, and for some problems it makes a big difference, see \Cref{ex:early-stopping} for an example. 

Note that for a countably infinite set $\lrc{1,2,\dots}$ assigning $\pi(i) = \frac{1}{i(i+1)}$ or $\pi(i)=\frac{6}{\pi^2 i^2}$ (we have $\sum_{i=1}^\infty \frac{1}{i^2} = \frac{\pi^2}{6}$) is ``almost as flat as it can get'' in the following sense. If we would have had just $i$ items, $\lrc{1,\dots,i}$, we could have assigned $\pi(j) = \frac{1}{j}$ for all $1\leq j\leq i$, which would have been a ``flat'' prior. However, for a countably infinite set of items the series $\sum_{i=1}^\infty \frac1i$ diverges, and so $\pi(i)$ has to be smaller than $\frac1i$. Since $\pi(i)$ enters the bound as $\ln\frac{1}{\pi(i)}$ and $\ln(i(i+1)) \approx 2\ln i$, the difference between using $\pi(i)=\frac1i$ (which we cannot do!) and $\pi(i)=\frac{1}{i(i+1)}$ is relatively small, and in this sense $\pi(i)=\frac{1}{i(i+1)}$ is ``almost as flat as it can be''.



\section{Vapnik-Chervonenkis (VC) Analysis}

\begin{figure}%
\centering
\begin{subfigure}{0.47\textwidth}
\centering
        \includegraphics[width=.8\textwidth]{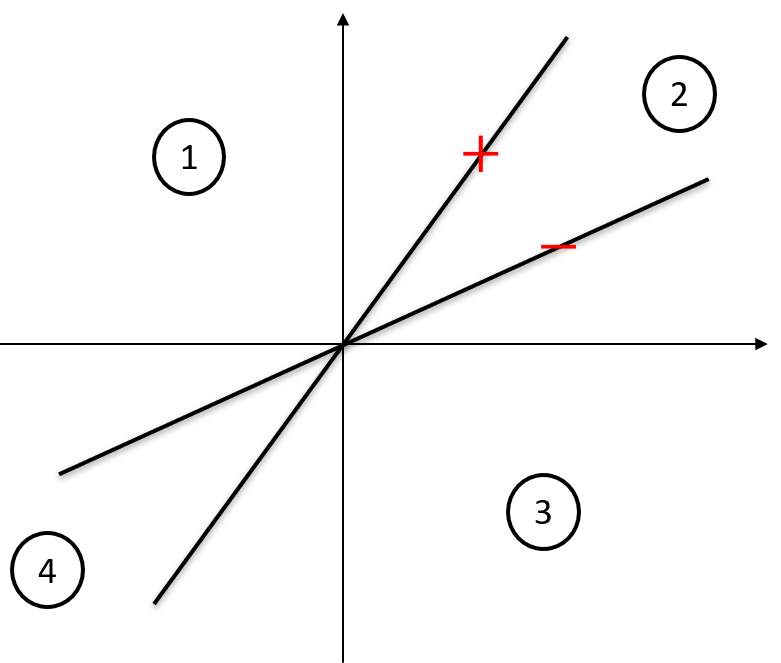}
        \caption{Subsets of linear homogeneous separators defined by two sample points.}
        \label{fig:VC-projection-2}
\end{subfigure}
\qquad
\begin{subfigure}{0.47\textwidth}
\centering
        \includegraphics[width=.8\textwidth]{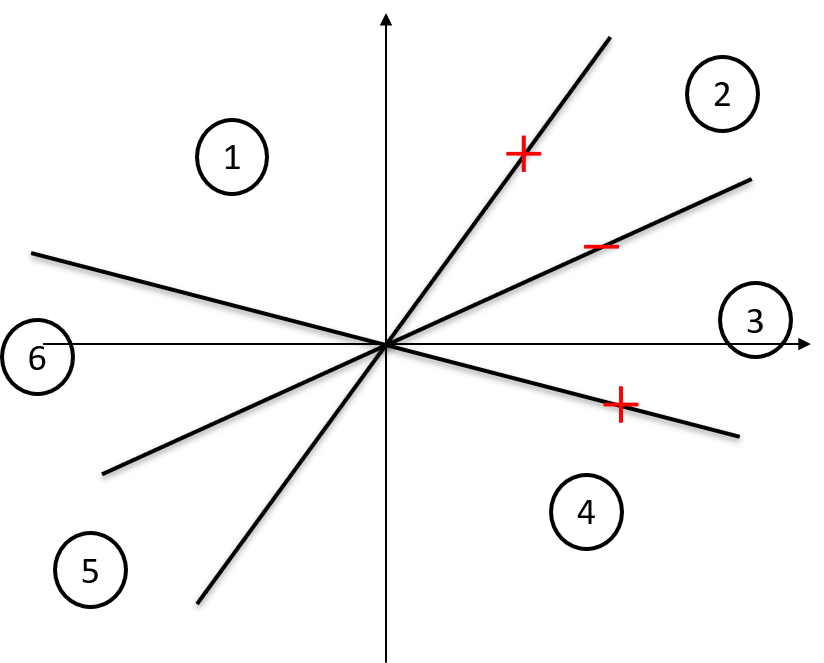}
        \caption{Subsets of linear homogeneous separators defined by three sample points.}
        \label{fig:VC-projection-3}
\end{subfigure}
\caption{\textbf{Subsets of homogeneous linear separators in $\R^2$ formed by \ref{fig:VC-projection-2} two and \ref{fig:VC-projection-3} three sample points.} A homogeneous linear separator in $\R^2$ is defined by a vector $w \in \R^2$. The sample points define a number of regions in $\R^2$ that are shown by the numbers in circles. We say that a linear separator falls within a certain region when the vector $w$ defining it falls within that region. All homogeneous linear separators falling within the same region have the same empirical loss $\hat L(h,S)$ and, therefore, any selection among them is not based on the sample $S$ and introduces no bias. The sample only discriminates between the subsets.}%
\label{fig:VC-projection}%
\end{figure}

Now we present Vapnik-Chervonenkis (VC) analysis of generalization when a hypothesis is selected from an uncountably infinite hypothesis class $\HH$. The reason that we are able to derive a generalization bound even though we are selecting from an uncountably large set is that only a finite part of this selection is based on the sample $S$, whereas the remaining uncountable selection is not based on the sample and, therefore, introduces no bias. Since the sample is finite, the number of distinct labeling patterns, also called dichotomies, $\lr{h(X_1),\dots,h(X_n)}$ is also finite. When two hypotheses, $h$ and $h'$, produce the same labeling pattern, $\lr{h(X_1),\dots,h(X_n)} = \lr{h'(X_1),\dots,h'(X_n)}$, the sample does not discriminate between them and the selection between $h$ and $h'$ is based on some other considerations rather than the sample. Therefore, the sample defines a finite number of (typically uncountably infinite) subsets of the hypothesis space $\HH$, where hypotheses within the same subset produce the same labeling pattern $\lr{h(X_1),\dots,h(X_n)}$. The sample then allows selection of the ``best'' subset, for example, the subset that minimizes the empirical error. All prediction rules within the same subset have the same empirical error $\hat L(h,S)$ and selection among them is independent of $S$. See Figure~\ref{fig:VC-projection} for an illustration.

The \emph{effective selection} based on the sample $S$ depends on the number of subsets of $\HH$ with distinct labeling patterns on $S$. When the number of such subsets is exponential in the size of the sample $n$, the selection is too large and leads to overfitting, as we have already seen for selection from large finite hypothesis spaces in the earlier sections. I.e., we cannot guarantee closeness of $\hat L(\hbest,S)$ to $L(\hbest)$. However, if the number of subsets is subexponential in $n$, we can provide generalization guarantees for $L(\hbest)$. In Figure~\ref{fig:VC-projection} we illustrate (informally) that at a certain point the number of subsets of the class of homogeneous linear separators in $\R^2$ stops growing exponentially with $n$.\footnote{Homogeneous linear separators are linear separators passing through the origin.} For $n=2$ the sample defines $4 = 2^n$ subsets, but for $n=3$ the sample defines $6 < 2^n$ subsets. It can be formally shown that no 3 sample points can define more than 6 subsets of the space of homogeneous linear separators in $\R^2$ (some may define less, but that is even better for us) and that for $n>2$ the number of subsets grows polynomially rather than exponentially with $n$.

In what follows we first bound the distance between $\hat L(h,S)$ and $L(h)$ for all $h\in\HH$ in terms of the number of subsets using symmetrization (Section~\ref{sec:symmetrization}) and then bound the number of subsets (Section~\ref{sec:growth}).

\subsection{The VC Analysis: Symmetrization}
\label{sec:symmetrization}

We start with a couple of definitions.

\begin{definition}[Dichotomies] Let $x_1,\dots,x_n \in \XX$. The set of \emph{dichotomies} (the labeling patterns) generated by $\HH$ on $x_1,\dots,x_n$ is defined by 
\[
\HH\lr{x_1,\dots,x_n} = \lrc{h(x_1),\dots,h(x_n) : h \in \HH}.
\]
\end{definition}

\begin{definition}[The Growth Function] The \emph{growth function} of $\HH$ is the maximal number of dichotomies it can generate on $n$ points:
\[
m_\HH(n) = \max_{x_1,\dots,x_n} \lra{\HH\lr{x_1,\dots,x_n}}.
\]
\end{definition}

Pay attention that $m_\HH(n)$ is defined by the ``worst-case'' configuration of points $x_1,\dots,x_n$, for which $\lra{\HH\lr{x_1,\dots,x_n}}$ is maximized. Thus, for lower bounding $m_\HH(n)$ (i.e., for showing that $m_\HH(n) \geq v$ for some value $v$) we have to find a configuration of points $x_1,\dots,x_n$ for which $\lra{\HH\lr{x_1,\dots,x_n}} \geq v$ or, at least, prove that such configuration exists. For upper bounding $m_\HH(n)$ (i.e., for showing that $m_\HH(n) \leq v$) we have to show that for any possible configuration of points $x_1,\dots,x_n$ we have $\lra{\HH\lr{x_1,\dots,x_n}} \leq v$. In other words, coming up with an example of a particular configuration $x_1,\dots,x_n$ for which $\lra{\HH\lr{x_1,\dots,x_n}} \leq v$ is insufficient for proving that $m_\HH(n) \leq v$, because there may potentially be an alternative configuration of points achieving a larger number of labeling configurations. To be concrete, the illustration in Figure~\ref{fig:VC-projection-3} shows that for the hypothesis space $\HH$ of homogeneous linear separators in $\R^2$ we have $m_\HH(3) \geq 6$, but it does not show that $m_\HH(3) \leq 6$. If we want to prove that $m_\HH(3) \leq 6$ we have to show that no configuration of 3 sample points can differentiate between more than 6 distinct subsets of the hypothesis space. More generally, if we want to show that $m_\HH(n) = v$ we have to show that $m_\HH(n) \geq v$ and $m_\HH(n) \leq v$. I.e., the only way to show equality is by proving a lower and an upper bound.

The following theorem uses the growth function to bound the distance between empirical and expected loss for all $h\in\HH$.
\begin{theorem}
\label{thm:Growth}
Assume that $\ell$ is bounded in the $[0,1]$ interval. Then for any $\delta \in (0,1)$
\[
\P[\exists h\in\HH: L(h) \geq \hat L(h,S) + \sqrt{\frac{8\ln\frac{2 m_\HH(2n)}{\delta}}{n}}] \leq \delta.
\]
\end{theorem}
The result is useful when $m_\HH(2n) \ll e^n$. In Section~\ref{sec:growth} we discuss when we can and cannot expect to have it, but for now we concentrate on the proof of the theorem.

The proof of the theorem is based on three ingredients. First we introduce a ``ghost sample'' $S'$, which is an imaginary sample of the same size as $S$ (i.e., of size $n$). We do not need to have this sample at hand, but we ask what would have happened if we had such sample. Then we apply symmetrization: we show that the probability that for any $h$ the empirical loss $\hat L(h,S)$ is far from $L(h)$ by more than $\varepsilon$ is bounded by twice the probability that $\hat L(h,S)$ is far from $\hat L(h,S')$ by more than $\varepsilon/2$. This allows us to consider the behavior of $\HH$ on the two samples, $S$ and $S'$, instead of studying it over all $\XX$ (because the definition of $L(h)$ involves all $\XX$, whereas the definition of $\hat L(h,S')$ involves only $S'$). In the third step we project $\HH$ onto the two samples, $S$ and $S'$. Even though $\HH$ is uncountably infinite, when we look at it through the prism of $S\cup S'$ we can only observe a finite number of distinct behaviors. More precisely, the number of different ways $\HH$ can label $S\cup S'$ is at most $m_\HH(2n)$. We show that the probability that for any of the possible ways to label $S\cup S'$ the empirical losses $\hat L(h,S)$ and $\hat L(h,S')$ diverge by more than $\varepsilon/2$ decreases exponentially with $n$.

Now we do this formally.

\paragraph{Step 1} We introduce a ghost sample $S' = \lrc{(X_1',Y_1'), \dots, (X_n',Y_n')}$ of size $n$.

\paragraph{Step 2 [Symmetrization]} We prove the following result.
\begin{lemma}
\label{lem:symmetrization}
Assuming that $e^{-n\varepsilon^2/2} \leq \frac{1}{2}$ we have
\begin{equation}
\label{eq:symmetrization}
\P[\exists h\in\HH: L(h) - \hat L(h,S) \geq \varepsilon] \leq 2 \P[\exists h\in\HH: \hat L(h,S') - \hat L(h,S) \geq \frac{\varepsilon}{2}].
\end{equation}
\end{lemma}

\begin{figure}%
\centering
\includegraphics[width=.8\columnwidth]{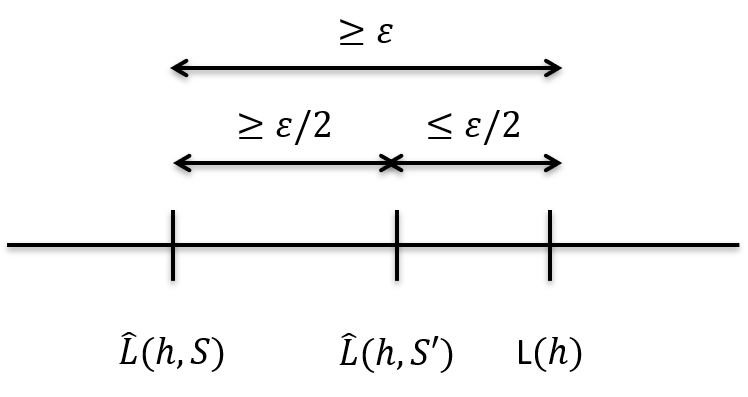}%
\caption{\textbf{Illustration for Step 2 of the proof of Theorem~\ref{thm:Growth}.}}%
\label{fig:VC-distances}%
\end{figure}

The illustration in Figure~\ref{fig:VC-distances} should be helpful for understanding the proof. The distance $L(h) - \hat L(h,S)$ can be expressed as $L(h) - \hat L(h,S) = (L(h) - \hat L(h,S')) + (\hat L(h,S') - \hat L(h,S))$. We remind that in general empirical losses are likely to be close to their expected values. More explicitly, under the mild assumption that $e^{-n\varepsilon^2/2} \leq 1/2$ we have that $L(h) - \hat L(h,S') \leq \varepsilon/2$ with probability greater than $1/2$ . If $L(h) - \hat L(h,S) \geq \varepsilon$ and $L(h) - \hat L(h,S') \leq \varepsilon/2$ we must have $\hat L(h,S') - \hat L(h,S) \geq \varepsilon/2$ (see the illustration). The proof is based on a careful exploitation of this observation. 

\begin{proof}[Proof of Lemma~\ref{lem:symmetrization}] We start from the right hand side of \eqref{eq:symmetrization}.
\begin{align}
&\P[\exists h\in\HH: \hat L(h,S') - \hat L(h,S) \geq \frac{\varepsilon}{2}]\notag\\
&\qquad\geq
\P[\lr{\exists h\in\HH: \hat L(h,S') - \hat L(h,S) \geq \frac{\varepsilon}{2}} \text{ AND } \lr{\exists h\in\HH: L(h) - \hat L(h,S) \geq \varepsilon}]\notag\\
&\qquad= \P[\exists h\in\HH: L(h) - \hat L(h,S) \geq \varepsilon] \P[\exists h\in\HH: \hat L(h,S') - \hat L(h,S) \geq \frac{\varepsilon}{2} \middle| \exists h\in\HH: L(h) - \hat L(h,S) \geq \varepsilon].\label{eq:sym0}
\end{align}
The inequality follows by the fact that for any two events $A$ and $B$ we have $\P[A] \geq \P[A\text{ AND } B]$ and the equality by $\P[A\text{ AND } B] = \P[B]\P[A|B]$. The first term in \eqref{eq:sym0} is the term we want and we need to lower bound the second term. We let $h^*$ be any $h$ for which, by conditioning, we have $L(h^*) - \hat L(h^*,S) \geq \varepsilon$. With high probability we have that $\hat L(h^*,S')$ is close to $L(h^*)$ up to $\varepsilon/2$. And since we are given that $\hat L(h,S)$ is far from $L(h^*)$ by more than $\varepsilon$ it must also be far from $\hat L(h^*,S')$ by more than $\varepsilon/2$ with high probability, see the illustration in Figure~\ref{fig:VC-distances}. Formally, we have:
\begin{align}
&\P[\exists h\in\HH: \hat L(h,S') - \hat L(h,S) \geq \frac{\varepsilon}{2} \middle| \exists h\in\HH: L(h) - \hat L(h,S) \geq \varepsilon]\notag\\
&\hspace{6cm}\geq \P[\hat L(h^*,S') - \hat L(h^*,S) \geq \frac{\varepsilon}{2} \middle | L(h^*) - \hat L(h^*,S) \geq \varepsilon]\label{eq:sym1}\\
&\hspace{6cm}\geq \P[L(h^*) - \hat L(h^*,S') \leq \frac{\varepsilon}{2} \middle | L(h^*) - \hat L(h^*,S) \geq \varepsilon]\label{eq:sym2}\\
&\hspace{6cm}= \P[L(h^*) - \hat L(h^*,S') \leq \frac{\varepsilon}{2}]\label{eq:sym3}\\
&\hspace{6cm}\geq 1 - \P[L(h^*) - \hat L(h^*,S') \geq \frac{\varepsilon}{2}]\notag\\
&\hspace{6cm}\geq 1 - e^{-2 n (\varepsilon/2)^2}\label{eq:sym4}\\
&\hspace{6cm}\geq \frac{1}{2}.\label{eq:sym5}
\end{align}
Explanation of the steps: in \eqref{eq:sym1} the event on the left hand side includes the event on the right hand side; in \eqref{eq:sym2} we have $\hat L(h,S') - \hat L(h,S) = \lr{L(h) - \hat L(h,S)} - \lr{L(h) - \hat L(h,S')}$ and since we are given that $L(h) - \hat L(h,S) \geq \varepsilon$ the event $\hat L(h,S') - \hat L(h,S) \geq \varepsilon/2$ follows from $L(h) - \hat L(h,S') \leq \varepsilon/2$, see Figure~\ref{fig:VC-distances}; in \eqref{eq:sym3} we can remove the conditioning on $S$, because the event of interest concerns $S'$, which is independent of $S$; \eqref{eq:sym4} follows by Hoeffding's inequality; and \eqref{eq:sym5} follows by the lemma's assumption on $e^{-n \varepsilon^2/2}$.

By plugging the result back into \eqref{eq:sym0} and multiplying by 2 we obtain the statement of the lemma.
\end{proof}

\begin{figure}%
\centering
\begin{subfigure}[b]{0.6\textwidth}
\centering
\includegraphics[width=\textwidth]{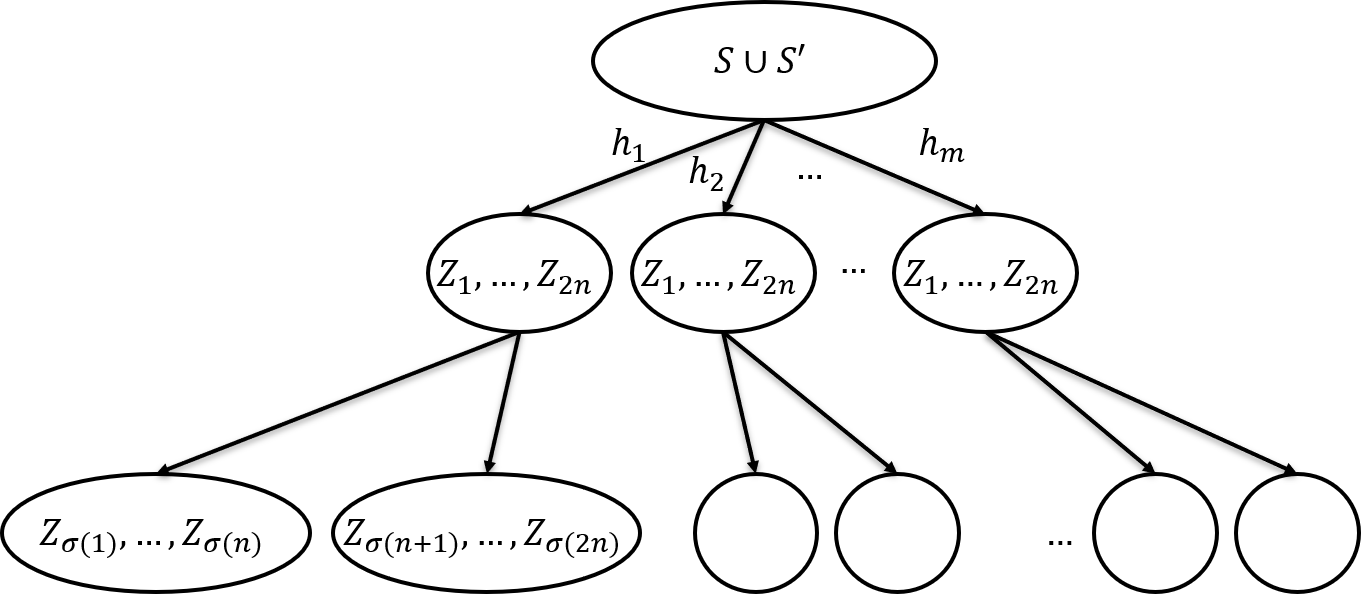}%
\caption{Illustration of the split.}
\end{subfigure}
\qquad
\begin{subfigure}[b]{0.34\textwidth}
\centering
\includegraphics[width=.9\textwidth]{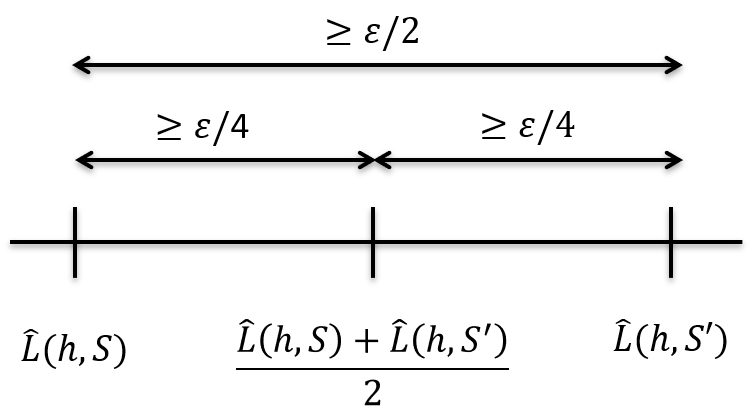}%
\caption{Illustration of the distances.}
\label{fig:sample-split-distances}
\end{subfigure}
\caption{\textbf{Illustration of the split of $S\cup S'$ into $S$ and $S'$.} On the left: First we sample the joint sample $S\cup S'$. Then each hypothesis $h_j$ produces a ``big bag'' of losses $\lrc{Z_1,\dots,Z_{2n}}$, where $Z_i = \ell(h_j(X_i), Y_i)$. Even though $\HH$ is uncountably infinite, the number of different ways to label $S\cup S'$ is at most $m_\HH(2n)$ by the definition of the growth function and thus the number of different ``big bags'' of losses is at most $m_\HH(2n)$ (in the illustration we have $m \leq m_\HH(2n)$). Finally, we split $S\cup S'$ into $S$ and $S'$, which corresponds to splitting the ``big bags'' of $2n$ losses into pairs of ``small bags'' of $n$ losses, corresponding to $\hat L(h_j,S)$ and $\hat L(h_j,S')$. On the right: we illustrate the distances between the average losses in a pair of ``small bags'' and the corresponding ``big bag'', which is the average of the two ``small bags''.}%
\label{fig:sample-split}%
\end{figure}

\paragraph{Step 3 [Projection]} Now we focus on $\P[\exists h\in\HH: \hat L(h,S') - \hat L(h,S) \geq \frac{\varepsilon}{2}]$, which concerns the behavior of $\HH$ on two finite samples, $S$ and $S'$. There are two possible ways to sample $S$ and $S'$. The first is to sample $S$ and then $S'$. An alternative way is to sample a joint sample $S_{2n} = S \cup S'$ and then split it into $S$ and $S'$ by randomly assigning half of the samples into $S$ and half into $S'$. The two procedures are equivalent and lead to the same distribution over $S$ and $S'$. We focus on the second procedure. Its advantage is that once we have sampled $S\cup S'$ the number of ways to label it with hypotheses from $\HH$ is finite, even though $\HH$ is uncountably infinite. This way we turn an uncountably infinite problem into a finite problem. The number of different sequences of losses on $S \cup S'$ is at most the number of different ways to label it, which is at most the growth function $m_\HH(2n)$ by definition. The probability of having $\hat L(h,S') - \hat L(h,S) \geq \varepsilon/2$ for a fixed $h$ reduces to the probability of splitting a sequence of $2n$ losses into $n$ and $n$ losses and having more than $\varepsilon/2$ difference between the average of the two. The latter reduces to the problem of sampling $n$ losses without replacement from a bag of $2n$ losses and obtaining an average which deviates from the bag's average by more than $\varepsilon/4$, see Figure~\ref{fig:sample-split}. This probability can be bounded by Hoeffding's inequality for sampling without replacement and decreases as $e^{-n \varepsilon^2/8}$. Putting this together we obtain the following result.
\begin{lemma}
\label{lem:VC-projection}
\[
\P[\exists h\in\HH: \hat L(h,S') - \hat L(h,S) \geq \frac{\varepsilon}{2}] \leq m_\HH(2n) e^{-n\varepsilon^2/8}.
\]
\end{lemma}
As you may guess, $m_\HH(2n)$ comes from a union bound over the number of possible sequences of losses we may obtain with hypotheses from $\HH$ on $S\cup S'$. We now prove the lemma formally.

\begin{proof}[Proof of Lemma~\ref{lem:VC-projection}]
\begin{align*}
\P[\exists h\in\HH: \hat L(h,S') - \hat L(h,S) \geq \frac{\varepsilon}{2}] &=
\sum_{S\cup S'} \P[S\cup S'] \P[\exists h\in\HH: \hat L(h,S') - \hat L(h,S) \geq \frac{\varepsilon}{2} \middle | S\cup S']\\
&\leq \sup_{S\cup S'}\P[\exists h\in\HH: \hat L(h,S') - \hat L(h,S) \geq \frac{\varepsilon}{2} \middle | S\cup S'].
\end{align*}
Pay attention that the conditional probabilities are with respect to the splitting of $S\cup S'$ into $S$ and $S'$.

Let $\Z(S\cup S') = \lrc{Z_1,\dots,Z_{2n}: Z_i = \ell(h(X_i),Y_i), h \in \HH}$ be the set of all possible sequences of losses that can be obtained by applying $h \in \HH$ to $S \cup S'$. Since there are at most $m_\HH(2n)$ distinct ways to label $S \cup S'$ we have $|\Z(S\cup S')| \leq m_\HH(2n)$. Let $\sigma:\lrc{1,\dots,2n}\to\lrc{1,\dots,2n}$ denote a permutation of indexes. We have
\begin{align}
&\sup_{S\cup S'}\P[\exists h\in\HH: \hat L(h,S') - \hat L(h,S) \geq \frac{\varepsilon}{2} \middle | S\cup S'] \notag\\
&\qquad=
\sup_{S\cup S'} \P[\exists \lrc{Z_1,\dots,Z_{2n}} \in \Z(S\cup S'): \frac{1}{n}\sum_{i=1}^n Z_{\sigma(i)} - \frac{1}{n}\sum_{i=n+1}^{2n} Z_{\sigma(i)} \geq \frac{\varepsilon}{2}]\label{eq:VC-proj-1}\\
&\qquad\leq \sup_{S\cup S'} \sum_{\lrc{Z_1,\dots,Z_{2n}} \in \Z(S\cup S')} \P[\frac{1}{n}\sum_{i=1}^n Z_{\sigma(i)} - \frac{1}{n}\sum_{i=n+1}^{2n} Z_{\sigma(i)} \geq \frac{\varepsilon}{2}]\label{eq:VC-proj-2}\\
&\qquad= \sup_{S\cup S'} \sum_{\lrc{Z_1,\dots,Z_{2n}} \in \Z(S\cup S')} \P[\frac{1}{n}\sum_{i=1}^n Z_{\sigma(i)} - \frac{1}{2n}\sum_{i=1}^{2n} Z_i \geq \frac{\varepsilon}{4}]\label{eq:VC-proj-2.5}\\
&\qquad\leq \sup_{S\cup S'} \sum_{\lrc{Z_1,\dots,Z_{2n}} \in \Z(S\cup S')} e^{-n\varepsilon^2/8}\label{eq:VC-proj-3}\\
&\qquad\leq \sup_{S\cup S'} m_\HH(2n) e^{-n\varepsilon^2/8}\label{eq:VC-proj-4}\\
&\qquad= m_\HH(2n) e^{-n\varepsilon^2/8},\notag
\end{align}
where \eqref{eq:VC-proj-1} follows by the fact that $\Z(S\cup S')$ is the set of all possible losses on $S\cup S'$ and in the step of splitting $S\cup S'$ into $S$ and $S'$ and computing $\hat L(h,S')$ and $\hat L(h,S)$ we are splitting a ``big bag'' of $2n$ losses into two ``small bags'' of $n$ and $n$; all that is left from $\HH$ in the splitting process is $\Z(S\cup S')$; the probability in \eqref{eq:VC-proj-1} is over the split of $S\cup S'$ into $S$ and $S'$, which is expressed by taking the first $n$ elements of a random permutation $\sigma$ of indexes into $S'$ and the last $n$ elements into $S$ and the probability is over $\sigma$; in \eqref{eq:VC-proj-2} we apply the union bound; for \eqref{eq:VC-proj-2.5} see the illustration in Figure~\ref{fig:sample-split-distances}; in \eqref{eq:VC-proj-3} we apply Hoeffding's inequality for sampling without replacement (Theorem~\ref{thm:Hoeffding-wr}) to the process of randomly sampling $n$ losses out of $2n$ and observing $\varepsilon/4$ deviation from the average; in \eqref{eq:VC-proj-4} we apply the bound on $|\Z(S\cup S')|$.
\end{proof}

\paragraph{Step 4 [Putting Everything Together]} All that is left for the proof of Theorem~\ref{thm:Growth} is to put Lemmas~\ref{lem:symmetrization} and \ref{lem:VC-projection} together.

\begin{proof}[Proof of Theorem~\ref{thm:Growth}]
Assuming that $e^{-n\varepsilon^2/2} \leq 1/2$ we have by Lemmas~\ref{lem:symmetrization} and \ref{lem:VC-projection}:
\begin{align*}
\P[\exists h\in\HH: L(h) - \hat L(h,S) \geq \varepsilon] &\leq 2 \P[\exists h\in\HH: \hat L(h,S') - \hat L(h,S) \geq \frac{\varepsilon}{2}]\\
&\leq 2 m_\HH(2n) e^{-n\varepsilon^2/8}.
\end{align*}
Note that if $e^{-n\varepsilon^2/2} > 1/2$ then $2 m_\HH(2n) e^{-n\varepsilon^2/8} > 1$ and the inequality is satisfied trivially (because probabilities are always upper bounded by 1).

By denoting the right hand side of the inequality by $\delta$ and solving for $\varepsilon$ we obtain the result.
\end{proof}

\subsection{Bounding the Growth Function: The VC-dimension}
\label{sec:growth}

In Theorem~\ref{thm:Growth} we relate the distance between the expected and empirical losses to the growth function of $\HH$. Our next goal is to bound the growth function. In order to do so we introduce the concept of shattering and the VC dimension.

\begin{definition} A set of points $x_1,\dots,x_n$ is \emph{shattered} by $\HH$ if functions from $\HH$ can produce all possible binary labellings of $x_1,\dots,x_n$ or, in other words, if 
\[
\|\HH(x_1,\dots,x_n)\| = 2^n.
\]
\end{definition}

For example, the set of homogeneous linear separators in $\R^2$ shatters the two points in Figure~\ref{fig:VC-projection-2}, but it does not shatter the three points in Figure~\ref{fig:VC-projection-3}. Note that if two points lie on one line passing through the origin, they are not shattered by the set of homogeneous linear separators, because they always get the same label. Thus, we may have two sets of points of the same size, where one is shattered and the other is not.

\begin{definition} The \emph{Vapnik-Chervonenkis} (\emph{VC}) \emph{dimension} of $\HH$, denoted by $\dVC(\HH)$ is the maximal number of points that can be shattered by $\HH$. In other words,
\[
\dVC(\HH) = \max \lrc{n \middle | m_\HH(n) = 2^n}.
\]
If $m_\HH(n) = 2^n$ for all $n$, then $\dVC(\HH) = \infty$.
\end{definition}

Similar to the growth function, if we want to show that $\dVC(\HH) = d$ we have to show that $\dVC(\HH) \geq d$ and $\dVC(\HH) \leq d$. For example, the illustration in Figure~\ref{fig:VC-projection-2} provides a configuration of points that are shattered by homogeneous separating hyperplanes in $\R^2$ and thus shows that the VC-dimension of homogeneous separating hyperplanes in $\R^2$ is at least 2. However, the illustration in Figure~\ref{fig:VC-projection-3} \emph{does not} demonstrate that the VC-dimension of homogeneous separating hyperplanes in $R^2$ is smaller than 3. If we want to show that the VC-dimension of homogeneous separating hyperplanes in $\R^2$ is smaller than 3 we have to prove that no configuration of 3 points can be shattered. It is not sufficient to show that one particular configuration of points cannot be shattered. In the same spirit, two points lying on the same line passing through the origin cannot be shattered by homogeneous linear separators, but this does not tell anything about the VC-dimension, because we have another configuration of two points in Figure~\ref{fig:VC-projection-2} that can be shattered. It is possible to show that the VC-dimension of homogeneous separating hyperplanes in $\R^d$ is $d$ and the VC-dimension of general separating hyperplanes in $\R^d$ (not necessarily passing through the origin) is $d+1$, see \citet[Exercise 2.4]{AML12}.

The next theorem bounds the growth function in terms of the VC-dimension.

\begin{theorem}[Sauer's Lemma]
\label{thm:Sauer}
\begin{equation}
\label{eq:Sauer}
m_\HH(n) \leq \sum_{i=0}^{\dVC(\HH)} \binom{n}{i}.
\end{equation}
\end{theorem}
We remind that the binomial coefficient $\binom{n}{k}$ counts the number of ways to pick $k$ elements out of $n$ and that for $n < k$ it is defined as $\binom{n}{k} = 0$. Thus, equation \eqref{eq:Sauer} is well-defined even when $n < \dVC(\HH)$. We also remind that $\sum_{i=0}^n \binom{n}{i} = 2^n$, where $2^n$ is the number of all possible subsets of $n$ elements, which is equal to the sum over $i$ going from $0$ to $n$ to select $i$ elements out of $n$. For $n \leq \dVC(\HH)$ we have $m_\HH(n) = 2^n$ and the inequality is satisfied trivially.

The proof of Theorem~\ref{thm:Sauer} slightly reminds the combinatorial proof of the binomial identity
\[
\binom{n}{k} = \binom{n-1}{k} + \binom{n-1}{k-1}.
\]
One way to count the number of ways to select $k$ elements out of $n$ on the right hand side is to take one element aside. If that element is selected, then we have $\binom{n-1}{k-1}$ possibilities to select $k-1$ additional elements out of the remaining $n-1$. If the element is not selected, then we have $\binom{n-1}{k}$ possibilities to select all $k$ elements out of remaining $n-1$. The sets including the first element are disjoint from the sets excluding it, leading to the identity above.

We need one more definition for the proof of Theorem~\ref{thm:Sauer}.
\begin{definition}
Let $B(n,d)$ be the maximal number of possible ways to label $n$ points, so that no $d+1$ points are shattered.
\end{definition}
By the definition, we have $m_\HH(n) \leq B(n, \dVC(\HH))$.

\begin{proof}[Proof of Theorem~\ref{thm:Sauer}] 
We prove by induction that
\begin{equation}
\label{eq:Bbinom}
B(n,d) \leq \sum_{i=0}^{d} \binom{n}{i}.
\end{equation}
For the induction base we have $B(n,0) = 1 = \binom{n}{0}$: if no points are shattered there is just one way to label the points. If there would be more than one way, they would differ in at least one point and that point would be shattered. By the definition of binomial coefficients, which says that for $k > n$ we have $\binom{n}{k} = 0$, we also know that for $n < d$ we have $B(n,d) = B(n,n)$. In particular, $B(0,d) = B(0,0) = 1$.

Now we proceed with induction on $d$ and for each $d$ we do an induction on $n$. We show that
\[
B(n,d) \leq B(n-1,d) + B(n-1,d-1).
\]
Let $\calS$ be a maximal set of dichotomies (labeling patterns) on $n$ points $x_1,\dots,x_n$. We take one point aside, $x_n$, and split $\calS$ into three disjoint subsets: $\calS = \calS^* \cup \calS^+ \cup \calS^-$. The set $\calS^*$ contains dichotomies on $n$ points that appear with just one sign on $x_n$, either positive or negative. The sets $\calS^+$ and $\calS^-$ contain all dichotomies that appear with both positive and negative sign on $x_n$, where the positive ones are collected in $\calS^+$ and the negative ones are collected in $\calS^-$. Thus, the sets $\calS^+$ and $\calS^-$ are identical except in their labeling of $x_n$, where in $\calS^+$ it is always labeled as $+$ and in $\calS^-$ always as $-$. By contradiction, the number of points $x_1,\dots,x_{n-1}$ that are shattered by $\calS^-$ cannot be larger than $d-1$, because otherwise the number of points that are shattered by $\calS$, which includes $\calS^+$ and $\calS^-$, would be larger than $d$, since we can use $\calS^+$ and $\calS^-$ to add $x_n$ to the set of shattered points. Therefore, $|\calS^-| \leq B(n-1,d-1)$. At the same time, the number of points $x_1,\dots,x_{n-1}$ that are shattered by $\calS^* \cup \calS^+$ cannot be larger than $d$, because the total number of points shattered by $\calS$ is at most $d$. Thus, we have $|\calS^* \cup \calS^+| \leq B(n-1,d)$. And overall
\[
B(n,d) = |\calS| = |\calS^* \cup \calS^+| + |\calS^-| \leq B(n-1,d) + B(n-1,d-1),
\]
as desired. By the induction assumption equation \eqref{eq:Bbinom} is satisfied for $B(n-1,d)$ and $B(n-1,d-1)$, and we have
\begin{align*}
B(n,d) &\leq \sum_{i=0}^d \binom{n-1}{i} + \sum_{i=0}^{d-1} \binom{n-1}{i}\\
&= 1 + \sum_{i=0}^{d-1} \lr{\binom{n-1}{i+1} + \binom{n-1}{i}}\\
&= \sum_{i=0}^d \binom{n}{i},
\end{align*}
as desired. Finally, as we have already observed, $m_\HH(n) \leq B(n,\dVC(\HH))$, completing the proof.
\end{proof}

The following lemma provides a more explicit bound on the growth function.
\begin{lemma}
\label{lem:growth-bound}
\[
\sum_{i=0}^{d} \binom{n}{i} \leq n^d + 1.
\]
\end{lemma}
The proof is based on induction, see \Cref{ex:growth-bound}.

By plugging the results of Theorem~\ref{thm:Sauer} and Lemma~\ref{lem:growth-bound} into Theorem~\ref{thm:Growth} we obtain the VC generalization bound.

\begin{theorem}[VC generalization bound]
\label{thm:VC}
Let ${\cal H}$ be a hypotheses class with VC-dimension $\dVC({\cal H}) = \dVC$. Then:
\[
\P[\exists h \in {\cal H}: \ERR(h) \geq \Err(h,S) + \sqrt{\frac{8 \ln\lr{2\lr{\lr{2 n}^{\dVC} + 1}/\delta}}{n}}] \leq \delta.
\]
\end{theorem}

For example, the VC-dimension of linear separators in $\R^d$ is $d+1$ and theorem~\ref{thm:VC} provides generalization guarantees for learning with linear separators in finite-dimensional spaces, as long as the dimension of the space $d$ is small in relation to the number of points $n$.

\section{VC Analysis of SVMs}


Kernel Support Vector Machines (SVMs) can map the data into high and potentially infinite-dimensional spaces. For example, Radial Basis Function (RBF) kernels map the data into an infinite-dimensional space. In the following we provide a more refined analysis of generalization in learning with linear separators in high-dimensional spaces. The analysis is based on the notion of \emph{separation with a margin}. We use the following definitions. 

\begin{definition}[Fat Shattering] Let $\HH_\gamma = \lrc{(\w,b): \|\w\| \leq 1/\gamma}$ be the space of hyperplanes described by $\w$ and $b$, where $\w$ is a vector in $\R^d$ (with potentially infinite dimension $d$) with $\|\w\| \leq 1/\gamma$ and $b \in \R$. We say that a set of points $\lrc{\x_1,\dots,\x_n}$ is \emph{fat-shattered} by $\HH_\gamma$ if for any set of labels $\lrc{y_1,\dots,y_n} \in \lrc{\pm 1}^n$ we have a hyperplane $(\w,b) \in \HH_\gamma$ that satisfies $y_i(\vp{\w}{\x_i} + b) \geq 1$ for all $i \in \lrc{1,\dots,n}$.
\end{definition}

Note that when $y = \sgn{\vp{\w}{\x} + b}$ the distance of a point $\x$ to a hyperplane $h$ defined by $(\w,b)$ is given by $\dist(h,\x) = \frac{y(\vp{\w}{\x} + b)}{\|\w\|}$ \citep[Page 5, Chapter 8]{AML15} and for $h=(\w,b) \in \HH_\gamma$ and $\lrc{\x_1,\dots,\x_n}$ fat-shattered by $\HH_\gamma$ we obtain $\dist(h,\x_i) \geq \gamma$ for all $i \in \lrc{1,\dots,n}$. It means that any possible labeling of $\lrc{\x_1,\dots,\x_n}$ can be achieved with margin at least $\gamma$.

\begin{definition}[Fat Shattering Dimension] We say that \emph{fat shattering dimension} $\dfat(\HH_\gamma) = d$ if $d$ is the maximal number of points that can be fat shattered by $\HH_\gamma$. (I.e., there exist $d$ points that can be fat shattered by $\HH_\gamma$ and no $d+1$ points can be fat shattered by $\HH_\gamma$.)
\end{definition}

Note that $\dfat(\HH_\gamma) \leq \dVC(\HH_\gamma) \leq d+1$, where $d$ is the dimension of $\w$. (If we can shatter $n$ points with margin $\gamma$ we can also shatter them without the margin.)

The following theorem bounds the fat shattering dimension of $\HH_\gamma$, see \citet{AML15} for a proof.

\begin{theorem}[{\citep[Theorem 8.5]{AML15}}]
\label{thm:margin}
Assume that the input space ${\cal X}$ is a ball of radius $R$ in $\R^d$ (i.e., $\|x\|\leq R$ for all $x \in {\cal X}$), where $d$ may potentially be infinite. Then:
\[
\dfat({\cal H}_\gamma) \leq \lru{R^2 / \gamma^2} + 1,
\]
where $\lru{R^2 / \gamma^2}$ is the smallest integer that is greater or equal to $R^2/\gamma^2$.
\end{theorem}

The important point is that the bound on fat shattering dimension is independent of the dimension of the space $\R^d$ that $\w$ comes from.

We define fat losses that count as error everything that falls too close to the separating hyperplane or on the wrong side of it.

\begin{definition}[Fat Losses] For $h = (\w,b)$ we define the fat losses
\begin{align*}
\lfat(h(\x),y) &= \begin{cases}0,&\text{if $y_i(\vp{\w}{\x_i} + b) \geq 1$}\\1,&\text{otherwise,}\end{cases}\\
\Lfat(h) &= \E[\lfat(h(\X),Y)],\\
\hatLfat(h,S) &= \frac{1}{n} \sum_{i=1}^n \lfat(h(\X_i),Y_i).
\end{align*}
\end{definition}

In relation to the fat losses the fat shattering dimension acts in the same way as the VC-dimension in relation to the zero-one loss. In particular, we have the following result that relates $\Lfat(h)$ to $\hatLfat(h,S)$ via $\dfat(\HH_\gamma)$ (the proof is left as an exercise).

\begin{theorem}
\label{thm:fatVC}
\[
\P[\exists h \in {\cal H}_\gamma: \Lfat(h) \geq \hatLfat(h,S) + \sqrt{\frac{8 \ln\lr{2\lr{\lr{2 n}^{\dfat(\HH_\gamma)} + 1}/\delta}}{n}}] \leq \delta.
\]
\end{theorem}

Now we are ready to analyze generalization in learning with fat linear separation. For the analysis we make a simplifying assumption that the data are contained within a ball of radius $R = 1$. The analysis for general $R$ is left as an exercise. Note that $R$ refers to the radius of the ball \emph{after} potential transformation of the data through a feature mapping / kernel function. For example, the RBF kernel maps the data into an infinite dimensional space and we consider the radius of the ball containing the transformed data in the infinite dimensional space.

\begin{theorem}
\label{thm:SVM}
Assume that the input space ${\cal X}$ is a ball of radius $R=1$ in $\R^d$, where $d$ is potentially infinite. Let ${\cal H}$ be the space of linear separators $h = (\w,b)$. Then
\[
\P[\exists h \in {\cal H}: \Lfat(h) \geq \hatLfat(h,S) + \sqrt{\frac{8 \ln \lr{2\lr{\lr{2n}^{1 + \lru{\|\w\|^2}} + 1}\lr{1 + \lru{\|\w\|^2}}\lru{\|\w\|^2} / \delta}}{n}}] \leq \delta.
\]
\end{theorem}

Observe that $L(h) \leq \Lfat(h)$ and, therefore, the theorem provides a generalization bound for $L(h)$. (If we count correct classifications within the margin as errors we only increase the loss.)

\begin{figure}[t]%
\centering
\includegraphics[width=.5\textwidth]{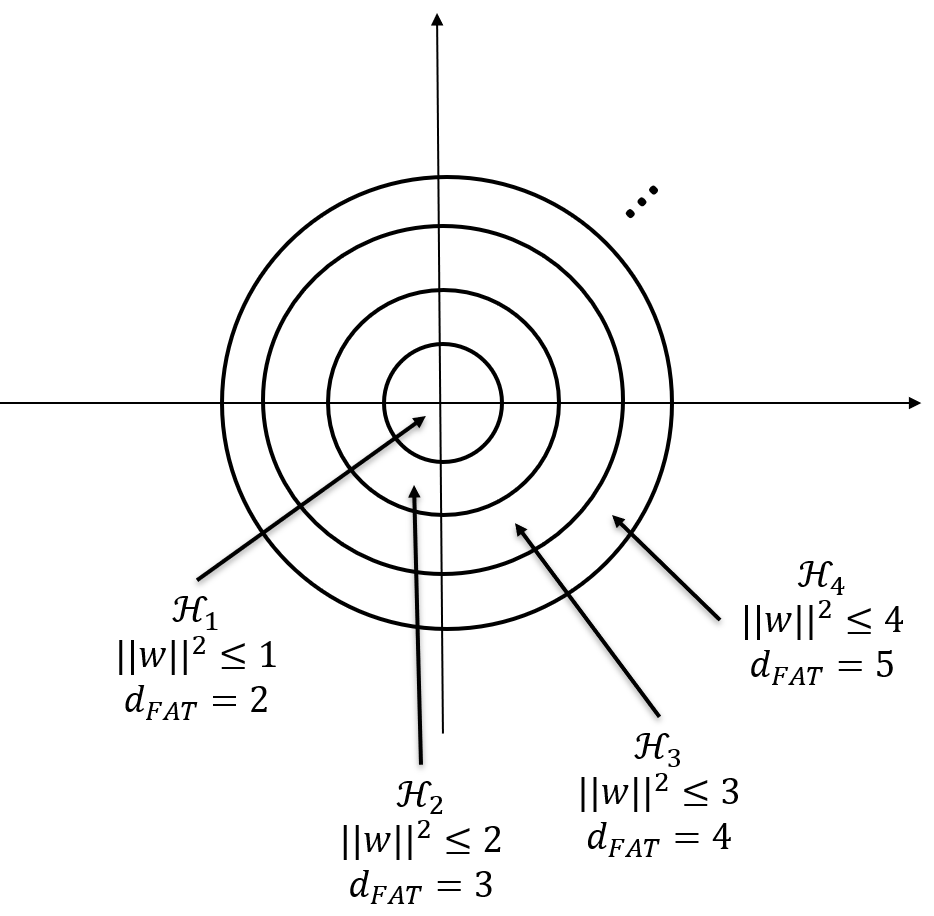}%
\caption{\textbf{Illustration for the proof of Theorem~\ref{thm:SVM}}}%
\label{fig:SVManalysis}%
\end{figure}

\begin{proof}
The proof is based on combination of VC and Occam's razor bounding techniques, see the illustration in Figure~\ref{fig:SVManalysis}. We start by noting that Theorem \ref{thm:margin} is interesting when $\lru{R^2 / \gamma^2} < d+1$, because as we have already noted $\dfat(\HH_\gamma) \leq \dVC(\HH_\gamma) \leq d+1$. We slice the hypotheses space ${\cal H}$ into a nested sequence of subspaces ${\cal H}_1 \subset {\cal H}_2 \subset \dots \subset {\cal H}_{d-1} \subset {\cal H}_d = {\cal H}$, where for all $i < d$ we define ${\cal H}_i$ to be the hypothesis space ${\cal H}_\gamma$ with $1 / \gamma^2 = i$. In other words, ${\cal H}_i = {\cal H}_{\lrc{\gamma = \frac{1}{\sqrt i}}}$ (do not let the notation to confuse you, by ${\cal H}_i$ we denote the $i$-th hypothesis space in the nested sequence of hypothesis spaces and by ${\cal H}_\gamma$ we denote the hypothesis space with $\|\w\|$ upper bounded by $1/\gamma$). By Theorem \ref{thm:margin} we have $\dfat({\cal H}_i) = i+1$ and then by Theorem \ref{thm:fatVC}:
\[
\P[\exists h \in {\cal H}_i: \Lfat(h) \geq \hatLfat(h,S) + \sqrt{\frac{8 \ln \lr{2\lr{\lr{2n}^{1 + i} + 1} / \delta_i}}{n}}] \leq \delta_i.
\]

We take $\delta_i = \frac{1}{i(i+1)}\delta$ and note that $\sum_{i=1}^\infty \frac{1}{i(i+1)} = \sum_{i=1}^\infty \lr{\frac{1}{i} - \frac{1}{i+1}} = \lr{1 - \frac{1}{2}} + \lr{\frac{1}{2} - \frac{1}{3}} + \lr{\frac{1}{3} - \frac{1}{4}} + \dots = 1$. We also note that ${\cal H} = \displaystyle \bigcup_{i=1}^d \lr{{\cal H}_i \setminus {\cal H}_{i-1}}$, where ${\cal H}_0$ is defined as the empty set and ${\cal H}_i \setminus {\cal H}_{i-1}$ is the difference between sets ${\cal H}_i$ and ${\cal H}_{i-1}$ (everything that is in ${\cal H}_i$, but not in ${\cal H}_{i-1}$). Note that the sets ${\cal H}_i \setminus {\cal H}_{i-1}$ and ${\cal H}_j \setminus {\cal H}_{j-1}$ are disjoint for $i \neq j$. Also note that $\delta_i$ is a distribution of our confidence budget $\delta$ among ${\cal H}_i \setminus {\cal H}_{i-1}$-s. Finally, note that if $h = (\w,b) \in {\cal H}_i \setminus {\cal H}_{i-1}$ then $\lru{\|\w\|^2} = i$. The remainder of the proof follows the same lines as the proof of Occam's razor bound:
\begin{align*}
&\P[\exists h \in {\cal H}: \Lfat(h) \geq \hatLfat(h,S) + \sqrt{\frac{8 \ln \lr{2\lr{\lr{2n}^{1 + \lru{\|\w\|^2}} + 1}\lr{1 + \lru{\|\w\|^2}}\lru{\|\w\|^2} / \delta}}{n}}]\\
&\qquad= \P[\exists h \in \bigcup_{i=1}^d {\cal H}_i \setminus {\cal H}_{i-1}: \Lfat(h) \geq \hatLfat(h,S) + \sqrt{\frac{8 \ln \lr{2\lr{\lr{2n}^{1 + \lru{\|\w\|^2}} + 1}\lr{1 + \lru{\|\w\|^2}}\lru{\|\w\|^2} / \delta}}{n}}]\\
&\qquad= \sum_{i=1}^d \P[\exists h \in {\cal H}_i \setminus {\cal H}_{i-1}: \Lfat(h) \geq \hatLfat(h,S) + \sqrt{\frac{8 \ln \lr{2\lr{\lr{2n}^{1 + \lru{\|\w\|^2}} + 1}\lr{1 + \lru{\|\w\|^2}}\lru{\|\w\|^2} / \delta}}{n}}]\\
&\qquad= \sum_{i=1}^d \P[\exists h \in {\cal H}_i \setminus {\cal H}_{i-1}: \Lfat(h) \geq \hatLfat(h,S) + \sqrt{\frac{8 \ln \lr{2\lr{\lr{2n}^{1 + i} + 1}\lr{1 + i}i / \delta}}{n}}]\\
&\qquad = \sum_{i=1}^d \P[\exists h \in {\cal H}_i \setminus {\cal H}_{i-1}: \Lfat(h) \geq \hatLfat(h,S) + \sqrt{\frac{8 \ln \lr{2\lr{\lr{2n}^{1 + i} + 1} / \delta_i}}{n}}]\\
&\qquad \leq \sum_{i=1}^d \P[\exists h \in {\cal H}_i: \Lfat(h) \geq \hatLfat(h,S) + \sqrt{\frac{8 \ln \lr{2\lr{\lr{2n}^{1 + i} + 1} / \delta_i}}{n}}]\\
&\qquad \leq \sum_{i=1}^d \delta_i = \sum_{i=1}^d \frac{1}{i(i+1)} \delta = \delta \sum_{i=1}^d \frac{1}{i(i+1)} \leq \delta \sum_{i=1}^\infty \frac{1}{i(i+1)} = \delta.
\end{align*}

\end{proof}

\section{VC Lower Bound}
\label{sec:lower}

In this section we show that when the VC-dimension is unbounded, it is impossible to bound the distance between $\ERR(h)$ and $\Err(h,S)$.

\begin{theorem}
\label{thm:VClower}
Let ${\cal H}$ be a hypothesis class with $d_{VC}({\cal H}) = \infty$. Then for any $n$ there exists a distribution over ${\cal X}$ and a class of target functions ${\cal F}$, such that
\[
\EEE{\sup_h \lr{\ERR(h) - \Err(h,S)}} \geq 0.25,
\]
where the expectation is over selection of a sample of size $n$ and a target function.
\end{theorem}

\begin{proof}
Pick $n$. Since $d_{VC}({\cal H}) = \infty$ we know that there exist $2n$ points that are shattered by ${\cal H}$. Let the sample space ${\cal X}_{2n} = \lrc{x_1,\dots,x_{2n}}$ be these points and let $p(x)$ be uniform on ${\cal X}_{2n}$. Let ${\cal F}$ be the set of all possible functions from ${\cal X}_{2n}$ to $\lrc{0,1}$ and let $p(f)$ be uniform over ${\cal F}$. Let $S$ be a sample of $n$ points. Let $\lrc{{\cal F}_k(S)}_k$ be maximal subsets of ${\cal F}$, such that ${\cal F} = \bigcup_k {\cal F}_k(S)$ and any $f_i, f_j \in {\cal F}_k(S)$ agree on $S$. Note that since ${\cal X}_{2n}$ is shattered by ${\cal H}$, for any $S$, any ${\cal F}_k$, and any $f_i \in {\cal F}_k$ that was used to label $S$ there exists $h^*({\cal F}_k(S),S) \in {\cal H}$, such that for any $f_i \in {\cal F}_k(S)$ the empirical error $\Err(h^*(f_i,S),S) = 0$. Let $p(k)$ and $p(i)$ be uniform. Then:
\begin{align*}
\EEE{\sup_h \lr{\ERR(h) - \Err(h,S)}} &= \mathbb E_{f \sim p(f)} \lrs{\mathbb E_{S \sim p(X)^n} \lrs{\sup_h \lr{\ERR(h) - \Err(h,S)}} \middle | f}\\
&= \mathbb E_{S \sim p(X)^n} \lrs{\mathbb E_{f \sim p(f)} \lrs{\sup_h \lr{\ERR(h) - \Err(h,S)}} \middle | S}\\
&= \mathbb E_{S \sim p(X)^n} \lrs{\mathbb E_{k \sim p(k)} \lrs{\mathbb E_{i \sim p(i)} \lrs{\sup_h \lr{\ERR(h) - \Err(h,S)}} \middle | {\cal F}_k} \middle | S}\\
&\geq \mathbb E_{S \sim p(X)^n} \lrs{\mathbb E_{k \sim p(k)} \lrs{\mathbb E_{i \sim p(i)} \lrs{\ERR(h^*({\cal F}_k, S)) - \Err(h^*({\cal F}_k, S),S)} \middle | {\cal F}_k} \middle | S}\\
&=\mathbb E_{S \sim p(X)^n} \lrs{\mathbb E_{k \sim p(k)} \lrs{\mathbb E_{i \sim p(i)} \lrs{\ERR(h^*({\cal F}_k, S))} \middle | {\cal F}_k} \middle | S}\\
&= \mathbb E_{S \sim p(X)^n} \lrs{\mathbb E_{k \sim p(k)} \lrs{0.25} \middle | S}\\
&= 0.25.
\end{align*}
\end{proof}

\begin{corollary} Under the assumptions of Theorem \ref{thm:VClower}, with probability at least 0.125, $\sup_h (\ERR(h) - \Err(h,S)) \geq 0.125$. Thus, it is impossible to have high-probability bounds on $\sup_h (\ERR(h) - \Err(h,S))$ that converge to zero as $n$ goes to infinity.
\end{corollary}

\begin{proof}
Note that $\sup_h (\ERR(h) - \Err(h,S)) \leq 1$, since $\err$ is bounded in $[0,1]$. Assume by contradiction that $\P[\sup_h \lr{\ERR(h) - \Err(h,S)} \geq 0.125] < 0.125$. Then
\[
\EEE{\sup_h \lr{\ERR(h) - \Err(h,S)}} \leq 0.125 \times 1 + (1 - 0.125) \times 0.125 < 2 \times 0.125 = 0.25,
\]
which is in contradiction with Theorem \ref{thm:VClower}.
\end{proof}

\section{PAC-Bayesian Analysis}
\label{sec:PAC-Bayes}

Occam's razor and VC analysis consider hard selection of a single hypothesis from a hypothesis class. In PAC-Bayesian analysis hard selection is replaced by a soft selection: instead of selecting a single hypothesis, it is allowed to select a distribution over the hypothesis space. When the distribution is a delta-distribution allocating all the mass to a single hypothesis, PAC-Bayes recovers the Occam's razor bound. However, the possibility of soft selection provides much more freedom and control over the approximation-estimation trade-off. 

PAC-Bayes achieves generalization by employing \emph{active avoidance of selection} when it is not necessary. For example, if two classifiers achieve the same empirical error on a sample, there is no reason to select among them. PAC-Bayes achieves this by spreading the probability of selecting classifiers uniformly over classifiers that are indistinguishable based on the sample. It leads to reduction of estimation error (because there is less selection) without affecting the approximation error.

PAC-Bayesian generalization bounds are based on \emph{change of measure} inequality, which acts as a replacement for the union bound. Change of measure inequality has two important advantages over the union bound: (1) it is tighter (see \Cref{ex:Occam-kl}) and (2) it can be applied to uncountably infinite hypothesis classes. Additionally, soft selection allows application of gradient-descent type methods to optimize the distribution over $\HH$, which in some cases leads to efficient algorithms for direct minimization of the PAC-Bayesian bounds. 

Soft selection is implemented by \emph{randomized classifiers}, which are formally defined below.
\begin{definition}[Randomized Classifier] Let $\rho$ be a distribution over ${\cal H}$. A \emph{randomized classifier} associated with $\rho$ (and named $\rho$) acts according to the following scheme. At each prediction round it:
\begin{enumerate}
\item Picks $h \in {\cal H}$ according to $\rho(h)$
\item Observes $x$
\item Returns $h(x)$
\end{enumerate}
The expected loss of $\rho$ is $\E_{h\sim\rho}\lrs{L(h)}$ and the empirical loss is $\E_{h\sim\rho}\lrs{\hat L(h,S)}$. Whenever it does not lead to confusion, we will shorten the notation to $\ErhoL$ and $\ErhoLhat$. 
\end{definition}

There is a large number of different PAC-Bayesian inequalities. We start with the classical one due to \citet{See02}.

\begin{theorem}[PAC-Bayes-kl inequality]
\label{thm:PBkl}
For any ``prior'' distribution $\pi$ over ${\cal H}$ that is independent of $S$ and $\delta \in(0,1]$\textup{:}
\begin{equation}
\label{eq:PBkl}
\P[\exists \rho: \kl\lr{\ErhoLhat\middle\|\ErhoL} \geq \frac{\KL(\rho\|\pi) + \ln \frac{2\sqrt{n}}{\delta}}{n}] \leq \delta.
\end{equation}
\end{theorem}

Another way of reading the theorem is that with probability at least $1-\delta$, for all ``posterior'' distributions $\rho$ over $\HH$:
\[
\kl\lr{\ErhoLhat\middle\|\ErhoL} \leq \frac{\KL(\rho\|\pi) + \ln \frac{2\sqrt{n}}{\delta}}{n}.
\]

The meaning of ``prior'' should be interpreted in exactly the same way as the ``prior'' in Occam's razor bound: it is any distribution over $\HH$ that sums up to one and does not depend on the sample $S$. The prior is an auxiliary construction for deriving the bound, and in contrast to Bayesian learning there is no assumption that it reflects any real-world distribution over $\HH$. Distribution $\rho$ is called a ``posterior'' distribution, because it is allowed to depend on the sample. The bound holds for all posterior distributions, including the Bayes posterior, and, therefore, it can be used to provide generalization guarantees for Bayesian learning. The posterior distribution minimizing the bound is typically \emph{not} the Bayes posterior.

Before proceeding to the proof of the theorem (given in \Cref{sec:PBkl-proof}), we provide some intuition on what the theorem tells us. Since the $\kl$ inequality is not the most intuitive divergence measure, we start by applying  Pinsker's relaxation of the $\kl$ (\Cref{cor:kl}) to obtain a more digestible (although weaker) form of the bound: with probability at least $1 - \delta$, for all distributions $\rho$
\begin{equation}
\label{eq:PBPinsker}
\ErhoL \leq \ErhoLhat + \sqrt{\frac{\KL(\rho\|\pi) + \ln \frac{2\sqrt{n}}{\delta}}{2n}}.
\end{equation}
Note that for $\rho=\pi$ the $\KL$ term is zero and we recover the generalization bound for a single hypothesis in \Cref{thm:Single} (except the minor difference of $\ln \frac{1}{\delta}$ being replaced by $\ln\frac{2\sqrt{n}}{\delta}$, which we ignore for now). Taking $\rho=\pi$ amounts to making no selection. Thus, if we start with a prior distribution $\pi$ and stay with it after observing the sample, we take no information from the sample (make no selection), and we retain the usual Hoeffding's or $\kl$ guarantee on generalization of a single prediction rule (a prediction rule that is independent of the sample). 

More generally, the amount of selection based on the sample is measured by $\KL(\rho\|\pi)$. The more the posterior distribution $\rho$ deviates from the prior distribution $\pi$ in the $\KL(\rho\|\pi)$ sense, the more information is taken from the sample, and the complexity term $\KL(\rho\|\pi)$ goes up. On the other hand, allocating higher mass $\rho$ to prediction rules with small empirical error $\hat L(h,S)$ reduces the first term of the bound, $\E_\rho[\hat L(h,S)]$. Therefore, there is a trade-off between giving preference to prediction rules with good empirical performance $\hat L(h,S)$ and keeping $\rho$ close to the prior $\pi$.

There is an important difference between the PAC-Bayes bounds and the VC bounds. In the VC bounds the complexity is measured by the VC dimension \emph{of the hypothesis class}, meaning that the complexity cost $d_{\VC}$ is paid upfront (before observing the sample), and then the algorithm is allowed to select any prediction rule in $\HH$. It can be compared to an all-you-can-eat buffet restaurant, where the customers pay at the entrance, and then get the freedom to select anything they want. Note that in the VC analysis there is no way to define complexity of individual prediction rules, because the complexity is of the hypothesis class. (It is possible to slice a hypothesis space into subclasses, and combine the VC analysis with the Occam's razor to measure the complexity at a subclass level, as we did in the analysis of SVMs, but it is much less flexible than what can be done with PAC-Bayes.) In PAC-Bayes the complexity is measured by $\KL(\rho\|\pi)$, where $\rho$ can be tuned based on the sample. Thus, it measures the actual amount of selection done based on the sample, and if there was no selection ($\rho=\pi$), the cost is zero. This can be compared to a restaurant, where the customers pay at the exit based on the dishes they have actually taken \emph{after} seeing the menu (the sample). And if they did not select any of the dishes, they exit without paying anything. Moreover, PAC-Bayes provides a possibility to define complexity $\pi(h)$ for each prediction rule individually (same as in Occam's razor, but now also for uncountably infinite sets of prediction rules) and this way incorporate prior information into learning.

For further intuition on the bound we decompose the $\KL$-divergence:
\[
\KL(\rho\|\pi) = \E_\rho \lrs{\ln\frac{\rho(h)}{\pi(h)}} =  \underbrace{\E_\rho \lrs{\ln \frac{1}{\pi(h)}}}_{\substack{\text{Average}\\\text{complexity}}} - \underbrace{\H(\rho)}_{\text{Entropy}}.
\]
The first term in the decomposition, $\E_\rho \lrs{\ln \frac{1}{\pi(h)}}$, is the average complexity $\ln \frac{1}{\pi(h)}$ of prediction rules when they are selected according to $\rho$, where the complexity is defined by the prior distribution $\pi$. The second term, $\H(\rho)$, is the entropy of the posterior distribution $\rho$. The entropy is highest when $\rho$ is close to a uniform distribution, and it is lowest when $\rho$ is close to a delta-distribution. Thus, $\H(\rho)$ can be seen as a ``bonus'' for avoidance of selection we have mentioned earlier. If $\HH$ is countable and $\rho$ places all the probability mass on a single prediction rule (implying hard selection), then $\H(\rho)=0$ and there is no ``bonus'', whereas is $\rho$ spreads the probability mass over multiple prediction rules, then $-\H(\rho) < 0$ decreases the complexity term.

For hard selection from a countable set of prediction rules, the bound in Equation \eqref{eq:PBPinsker} recovers the Occam's razor bound in \Cref{thm:Occam} (ignoring the $\ln2\sqrt{n}$ term). Note that if there are several prediction rules with identical $\hat L(h,S)$ and $\pi(h)$, then spreading $\rho$ uniformly over them yields $-\H(\rho)$ bonus in the complexity term without affecting any other terms in the bound, which once again demonstrates that avoiding selection when it is not necessary improves generalization.

A tighter relaxation of the $\kl$ divergence based on Refined Pinsker's Inequality in \Cref{cor:kl+} illustrates additional properties of the PAC-Bayes-kl inequality. By applying the relaxation to \Cref{thm:PBkl} we obtain that with probability at least $1-\delta$, for all distributions $\rho$
\begin{equation}
\label{eq:PBRefPinsker}
\E_\rho[L(h)] \leq \E_\rho[\hat L(h,S)] + \sqrt{\frac{2\E_\rho[\hat L(h,S)]\lr{\KL(\rho\|\pi)+\ln\frac{2\sqrt{n}}{\delta}}}{n}} + \frac{2\lr{\KL(\rho\|\pi)+\ln\frac{2\sqrt{n}}{\delta}}}{n}.
\end{equation}
The relaxation shows that PAC-Bayes-kl inequality exhibits ``fast convergence rate''. Namely, if $\E_\rho[\hat L(h,S)]$ is close to zero, then the distance between $\E_\rho[\hat L(h,S)]$ and $\E_\rho[L(h)]$ decreases at the rate of $\frac{1}{n}$ rather than $\frac{1}{\sqrt n}$. The ``fast convergence rate'' provides an extra advantage for placing higher mass on prediction rules with empirical loss is especially close to zero. 

We note that the ``fast convergence rate'' is a property of the $\kl$ divergence. It is possible to derive a similar ``fast'' version of the Occam's razor bound by basing it on the $\kl$ inequality, see \Cref{ex:Occam-kl}.

Finally, we note that Equations \eqref{eq:PBPinsker} and \eqref{eq:PBRefPinsker} are deterministic relaxations of the bound in \Cref{thm:PBkl}, and that the PAC-Bayes-kl inequality is always at least as tight as the best of its relaxations. And even though the $\kl$ has no analytic inverse, it is possible to invert it numerically to obtain the tightest bound.

\subsection{Relation and Differences with other Learning Approaches}

Even though it has ``Bayesian'' in its name, PAC-Bayesian analysis is a frequentist (PAC) learning framework. It has the following relation and differences with Bayesian learning and with VC analysis / Radamacher complexities.

\paragraph{Relation with Bayesian learning}
\begin{enumerate}
\item Explicit way to incorporate prior information (via $\pi(h)$).
\item Possibility of sequential updates of the prior $\pi$.
\end{enumerate}
\paragraph{Differences with Bayesian learning}
\begin{enumerate}
\item Explicit high-probability guarantee on the expected performance.
\item No belief in prior correctness (a frequentist bound).
\item Explicit dependence on the loss function. (Bayesian posterior does not depend on the choice of the loss function, whereas PAC-Bayesian posterior does.)
\item Different weighting of prior belief $\pi(h)$ vs. evidence $\hat L(h,S)$.
\item Holds for {\it any} distribution $\rho$ (including the Bayes posterior).
\end{enumerate}
\paragraph{Relation with VC analysis / Radamacher complexities}
\begin{enumerate}
\item Explicit high-probability guarantee on the expected performance.
\item Explicit dependence on the loss function.
\end{enumerate}
\paragraph{Difference with VC analysis / Radamacher complexities}
\begin{enumerate}
\item Complexity is defined individually for each $h$ via $\pi(h)$ (rather than ``complexity of a hypothesis class'').
\item Explicit way to incorporate prior knowledge.
\item The bound is defined for randomized classifiers $\rho$ (not individual $h$). For uncountably infinite sets of prediction rules it is usually necessary to spread the probability mass of $\rho$ over a measurable subset of $\HH$, because for continuous probability distributions concentrating $\rho$ on a single $h$ leads to explosion on $\KL(\rho\|\pi)$. It means that the PAC-Bayes classifiers are inherently randomized (rather than deterministic). However, it is possible to apply derandomization techniques, for example, $\rho$-weighted majority votes.
\end{enumerate}
In a sense, PAC-Bayesian analysis takes the best out of Bayesian learning and VC analysis and puts it together. Specifically, it provides a possibility to incorporate prior information through $\pi$, and it provides frequentist generalization guarantees. Additionally, it provides efficient learning algorithms, since $\KL(\rho\|\pi)$ is convex in $\rho$ and $\E_\rho[\hat L(h,S)]$ is linear in $\rho$, allowing efficient optimization of the bounds with respect to $\rho$.

\subsection{A Proof of PAC-Bayes-kl Inequality}
\label{sec:PBkl-proof}

At the basis of most of PAC-Bayesian bounds lies the change of measure inequality, which acts as replacement of the union bound for uncountably infinite sets.
\begin{theorem}[Change of measure inequality] For any measurable function $f(h)$ on $\cal H$ and any distributions $\rho$ and $\pi$:
\[
\EE_{h \sim \rho(h)} \lrs{f(h)} \leq \KL(\rho\|\pi) + \ln \EE_{h \sim \pi(h)}\lrs{e^{f(h)}}.
\]
\end{theorem}

\begin{proof}
\begin{align*}
\EE_{\rho(h)} \lrs{f(h)} &= \EE_{\rho(h)} \lrs{\ln \lr{\frac{\rho(h)}{\pi(h)} \times e^{f(h)} \times \frac{\pi(h)}{\rho(h)}}}\\
&= \KL(\rho\|\pi) + \EE_{\rho(h)} \lrs{\ln \lr{e^{f(h)} \times \frac{\pi(h)}{\rho(h)}}}\\
&\leq \KL(\rho\|\pi) + \ln \EE_{\rho(h)} \lrs{e^{f(h)} \times \frac{\pi(h)}{\rho(h)}}\\
&= \KL(\rho\|\pi) + \ln \EE_{\pi(h)} \lrs{e^{f(h)}},
\end{align*}
where the inequality in the third step is justified by Jensen's inequality (Theorem~\ref{thm:Jensen}). Note that there is nothing probabilistic in the statement of the theorem - it is a deterministic result.
\end{proof}

In the next lemma we extend $f$ to be a function of $h$ and a sample $S$ and apply a probabilistic argument to the last term of change-of-measure inequality. The lemma is the foundation for most PAC-Bayesian bounds.

\begin{lemma}[PAC-Bayes lemma] For any measurable function $f:\HH\times (\XX \times \YY)^n \to \R$ and any distribution $\pi$ on $\HH$ that is independent of the sample $S$
\[
\P[\exists \rho: \E_{h \sim \rho}\lrs{f(h,S)} \geq \KL(\rho\|\pi) + \ln \frac{\E_{h\sim\pi} \lrs{\E_S \lrs{e^{f(h,S)}}}}{\delta}] \leq \delta,
\]
where the probability is with respect to the draw of the sample $S$ and $\E_S$ is the expectation with respect to the draw of $S$.
\label{lem:PAC-Bayes}
\end{lemma}

An equivalent way of writing the above statement is 
\[
\P[\forall \rho: \E_{h \sim \rho}\lrs{f(h,S)} \leq \KL(\rho\|\pi) + \ln \frac{\E_{h\sim\pi} \lrs{\E_S \lrs{e^{f(h,S)}}}}{\delta}] \geq 1-\delta
\]
or, in words, with probability at least $1-\delta$ over the draw of $S$, for all $\rho$ simultaneously
\[
\E_\rho\lrs{f(h,S)} \leq \KL(\rho\|\pi) + \ln \frac{\E_\pi \lrs{\E_S \lrs{e^{f(h,S)}}}}{\delta}.
\]

We first present a slightly less formal, but more intuitive proof and then provide a formal one. By change of measure inequality we have
\begin{align*}
\E_\rho\lrs{f(h,S)} \hspace{.5cm} \leq \hspace{.5cm}&\KL(\rho\|\pi) + \ln \E_\pi \lrs{e^{f(h,S)}}\\
\mathrel{\mathop{\leq}\limits_{w.p.\geq 1-\delta}} &\KL(\rho\|\pi) + \ln \frac{\E_S \lrs{\E_\pi \lrs{e^{f(h,S)}}}}{\delta}\\
\hspace{.5cm} = \hspace{.5cm} &\KL(\rho\|\pi) + \ln \frac{\E_\pi \lrs{\E_S \lrs{e^{f(h,S)}}}}{\delta},
\end{align*}
where in the second line we apply Markov's inequality to the random variable $Z = \E_\pi \lrs{e^{f(h,S)}}$ (and the inequality holds with probability at least $1-\delta$) and in the last line we can exchange the order of expectations, because $\pi$ is independent of $S$. The key observation is that the change-of-measure inequality relates all posterior distributions $\rho$ to a single prior distribution $\pi$ in a deterministic way and the probabilistic argument (the Markov's inequality) is applied to a single random quantity $\E_\pi \lrs{e^{f(h,S)}}$. This way change-of-measure inequality replaces the union bound, and it holds even when $\HH$ is uncountably infinite.

Now we provide a formal proof.

\begin{proof}
\begin{align}
\P[\exists \rho: \E_\rho\lrs{f(h,S)} \geq \KL(\rho\|\pi) + \ln \frac{\E_\pi \lrs{\E_S \lrs{e^{f(h,S)}}}}{\delta}] 
&\leq \P[\E_\pi \lrs{e^{f(h,S)}} \geq \frac{\E_\pi \lrs{\E_S \lrs{e^{f(h,S)}}}}{\delta}]\label{eq:PBlem-1}\\
&= \P[\E_\pi \lrs{e^{f(h,S)}} \geq \frac{\E_S \lrs{\E_\pi \lrs{e^{f(h,S)}}}}{\delta}]\label{eq:PBlem-2}\\
&\leq \delta,\notag
\end{align}
where \eqref{eq:PBlem-1} follows by change-of-measure inequality (elaborated below), in \eqref{eq:PBlem-2} we can exchange the order of expectations, because $\pi$ is independent of $S$, and in the last step we apply Markov's inequality to the random variable $Z = \E_\pi \lrs{e^{f(h,S)}}$.

An elaboration concerning Step~\eqref{eq:PBlem-1}. By change of measure inequality, we have that $\forall \rho: \E_\rho\lrs{f(h,S)} \leq \KL(\rho\|\pi) + \ln \E_\pi \lrs{e^{f(h,S)}}$. Therefore, \emph{if} $\E_\pi \lrs{e^{f(h,S)}} \leq \frac{\E_S \lrs{\E_\pi \lrs{e^{f(h,S)}}}}{\delta}$, \emph{then} $\forall \rho: \E_\rho\lrs{f(h,S)} \leq \KL(\rho\|\pi) + \ln \frac{\E_S \lrs{\E_\pi \lrs{e^{f(h,S)}}}}{\delta}$. Let $A$ denote the event in the if-statement and $B$ denote the event in the then-statement. Then $\P[A]\leq \P[B]$ and, therefore, $\P[\bar A] \geq \P[\bar B]$, where $\bar A$ denotes the complement of event $A$. The complement of $A$ is $\E_\pi \lrs{e^{f(h,S)}} > \frac{\E_S \lrs{\E_\pi \lrs{e^{f(h,S)}}}}{\delta}$ and the complement of $B$ is $\exists \rho: \E_\rho\lrs{f(h,S)} > \KL(\rho\|\pi) + \ln \frac{\E_S \lrs{\E_\pi \lrs{e^{f(h,S)}}}}{\delta}$, which gives us the inequality in Step~\eqref{eq:PBlem-1} (as usually, we are being a tiny bit sloppy and do not trace which inequalities are strict and which are weak; with a slight extra effort this could be done, but it does not matter in practice, so we save the effort). The important point is that the change-of-measure inequality relates all posterior distributions $\rho$ to a single prior distribution $\pi$ in a deterministic way, and the probabilistic argument is applied to a single random variable $\E_\pi \lrs{e^{f(h,S)}}$, avoiding the need in taking a union bound. This way the change of measure inequality acts as a replacement of the union bound.
\end{proof}

Different PAC-Bayesian inequalities are obtained by different choices of the function $f(h,S)$. A key consideration in the choice of $f(h,S)$ is the possibility to bound $\E_S\lrs{e^{f(h,S)}}$. For example, we have done it for $f(h,S) = n\kl(\Err(h,S)\|\ERR(h))$ in \Cref{lem:klRef}, and this is the choice of $f$ in the proof of PAC-Bayes-kl inequality. Other choices of $f$ are possible. For example, Hoeffding's Lemma~\ref{lem:Hoeff} provides a bound on $\E_S\lrs{e^{f(h,S)}}$ for $f(h,S) = \lambda \lr{\ERR(h) - \Err(h,S)}$, which can be used to derive PAC-Bayes-Hoeffding inequality. We refer to \citet{SLCB+12} for more details.

\begin{proof}[Proof of Theorem \ref{thm:PBkl}]
We provide an intuitive derivation and leave the formal one (as in the proof of Lemma~\ref{lem:PAC-Bayes}) as an exercise. 

We take $f(h,S) = n \kl(\Err(h,S)\|\ERR(h))$. Then we have
\begin{align*}
n \kl \lr{\EE_\rho \lrs{\Err(h,S)} \middle \|\EE_\rho\lrs{\ERR(h)}}  
\hspace{.5cm}\leq \hspace{.5cm}&\EE_\rho \lrs{n \kl(\Err(h,S)\|\ERR(h))}\\
\mathrel{\mathop{\leq}\limits_{w.p.\geq 1-\delta}} &\KL(\rho\|\pi) + \ln \frac{\E_\pi \lrs{\E_S \lrs{e^{n \kl(\Err(h,S)\|\ERR(h))}}}}{\delta}\\
\leq \hspace{.5cm}& \KL(\rho\|\pi) + \ln \frac{\E_\pi \lrs{2\sqrt{n}}}{\delta}\\
= \hspace{.5cm}& \KL(\rho\|\pi) + \ln \frac{2\sqrt{n}}{\delta},
\end{align*}
where the first inequality is by convexity of $\kl$ (\Cref{cor:kl-convexity}), the second inequality is by the PAC-Bayes Lemma (and it holds with probability at least $1-\delta$ over the draw of $S$), and the third inequality is by Lemma \ref{lem:klRef}.
\end{proof}

\subsection{Application to SVMs}

In order to apply PAC-Bayesian bound to a given problem we have to design a prior distribution $\pi$ and then bound the $\KL$-divergence $\KL(\rho\|\pi)$ for the posterior distributions of interest. Sometimes we resort to a restricted class of $\rho$-s, for which we are able to bound $\KL(\rho\|\pi)$. You can see how this is done for SVMs in \citet[Section 5.3]{Lan05}.

\subsection{Relaxation of PAC-Bayes-kl: PAC-Bayes-$\lambda$ Inequality}

Due to its implicit form, PAC-Bayes-kl inequality is not very convenient for optimization. One way around is to replace the bound with a linear trade-off $\beta n \EE_\rho\lrs{\hat L(h,S)} + \KL(\rho\|\pi)$. Since $\KL(\rho\|\pi)$ is convex in $\rho$ and $\EE_\rho\lrs{\hat L(h,S)}$ is linear in $\rho$, for a fixed $\beta$ the trade-off is convex in $\rho$ and can be minimized. (We note that parametrization of $\rho$, for example the popular restriction of $\rho$ to a Gaussian posterior \citep{Lan05}, may easily break the convexity \citep{GLLM09}. We get back to this point in Section~\ref{sec:PBlhyp}.) The value of $\beta$ can then be tuned by cross-validation or substitution of $\rho(\beta)$ into the bound (the former usually works better).

Below we present a more rigorous approach. We prove the following relaxation of PAC-Bayes-kl inequality, which leads to a bound that can be optimized by alternating minimization.
\begin{theorem}[PAC-Bayes-$\lambda$ Inequality] For any probability distribution $\pi$ over ${\cal H}$ that is independent of $S$ and any $\delta \in (0,1)$, with probability greater than $1-\delta$ over a random draw of a sample $S$, for all distributions $\rho$ over ${\cal H}$ and all $\lambda \in (0,2)$  and $\gamma > 0$  simultaneously:
\begin{align}
\label{eq:PBlambda}
\E_\rho\lrs{L(h)} &\leq \frac{\E_\rho\lrs{\hat L(h,S)}}{1 - \frac{\lambda}{2}} + \frac{\KL(\rho\|\pi) + \ln \frac{2\sqrt{n}}{\delta}}{\lambda\lr{1-\frac{\lambda}{2}}n},\\
\E_\rho\lrs{L(h)} &\geq \lr{1 - \frac{\gamma}{2}}\E_\rho[\hat L(h,S)] - \frac{\KL(\rho\|\pi) + \ln \frac{2\sqrt{n}}{\delta}}{\gamma n}.\label{eq:PBlambda-lower}
\end{align}
\label{thm:PBlambda}
\end{theorem}
At the moment we focus on the upper bound in equation~\eqref{eq:PBlambda}. Note that the theorem holds for \emph{all} values of $\lambda \in (0,2)$ simultaneously. Therefore, we can optimize the bound with respect to $\lambda$ and pick the best one. 
\begin{proof}
We prove the upper bound in equation~\eqref{eq:PBlambda}. Proof of the lower bound \eqref{eq:PBlambda-lower} is analogous and left as an exercise. Proof of the statement that the upper and lower bounds hold simultaneously (require no union bound) is also left as an exercise.

By refined Pinsker's inequality in Corollary~\ref{cor:RefPin2}, for $p < q$
\begin{equation}
\label{eq:RefPin2kl}
\kl(p\|q) \geq (q-p)^2/(2q).
\end{equation}
By PAC-Bayes-kl inequality, Theorem~\ref{thm:PBkl}, with probability greater than $1-\delta$ for all $\rho$ simultaneously
\[
\kl\lr{\ErhoLhat\middle\|\ErhoL} \leq \frac{\KL(\rho\|\pi) + \ln \frac{2\sqrt{n}}{\delta}}{n}.
\]
By application of inequality \eqref{eq:RefPin2kl}, the above inequality can be relaxed to
\begin{equation}
\label{eq:PBsqrt}
\EE_\rho\lrs{L(h)} - \EE_\rho\lrs{\hat L(h,S)} \leq \sqrt{2 \EE_\rho\lrs{L(h)} \frac{\KL(\rho\|\pi) + \ln \frac{2\sqrt{n}}{\delta}}{n}}.
\end{equation}
We have that
\[
\min_{\lambda: \lambda > 0} \lr{\lambda x + \frac{y}{\lambda}} = 2 \sqrt{xy}
\]
(we leave this statement as a simple exercise). Thus, $\sqrt{xy} \leq \frac{1}{2}\lr{\lambda x + \frac{y}{\lambda}}$ for all $\lambda > 0$ and by applying this inequality to \eqref{eq:PBsqrt} we have that with probability at least $1-\delta$ for all $\rho$ and $\lambda > 0$
\[
\EE_\rho\lrs{L(h)} - \EE_\rho\lrs{\hat L(h,S)} \leq \frac{\lambda}{2}\EE_\rho\lrs{L(h)} + \frac{\KL(\rho\|\pi) + \ln \frac{2\sqrt{n}}{\delta}}{\lambda n}.
\]
By changing sides
\[
\lr{1 - \frac{\lambda}{2}} \EE_\rho\lrs{L(h)} \leq \EE_\rho\lrs{\hat L(h,S)} + \frac{\KL(\rho\|\pi) + \ln \frac{2\sqrt{n}}{\delta}}{\lambda n}.
\]
For $\lambda < 2$ we can divide both sides by $\lr{1-\frac{\lambda}{2}}$ and obtain the theorem statement.
\end{proof}

\subsection{Alternating Minimization of the PAC-Bayes-$\lambda$ Bound}

We use the term \emph{PAC-Bayes-$\lambda$ bound} to refer to the right hand side of PAC-Bayes-$\lambda$ inequality. A great advantage of the PAC-Bayes-$\lambda$ bound is that it can be conveniently minimized by alternating minimization with respect to $\rho$ and $\lambda$. Since $\EE_\rho\lrs{\hat L(h,S)}$ is linear in $\rho$ and $\KL(\rho\|\pi)$ is convex in $\rho$ \citep{CT06}, for a fixed $\lambda$ the bound is convex in $\rho$ and the minimum is achieved by 
\begin{equation}
\label{eq:rho}
\rho(h) = \frac{\pi(h) e^{-\lambda n \hat L(h,S)}}{\E_\pi\lrs{e^{-\lambda n \hat L(h',S)}}},
\end{equation}
where $\E_\pi\lrs{e^{-\lambda n \hat L(h',S)}}$ is a convenient way of writing the normalization factor, which covers continuous and discrete hypothesis spaces in a unified notation. In the discrete case, which will be of main interest for us, $\E_\pi\lrs{e^{-\lambda n \hat L(h',S)}} = \sum_{h'\in\HH} \pi(h') e^{-\lambda n \hat L(h',S)}$. We leave a proof of the statement that \eqref{eq:rho} defines $\rho$ which achieves the minimum of the bound as an exercise to the reader. Furthermore, for $t \in (0,1)$ and $a,b \geq 0$ the function $\frac{a}{1-t} + \frac{b}{t(1-t)}$ is convex in $t$ \citep{TS13} and, therefore, for a fixed $\rho$ the right hand side of inequality \eqref{eq:PBlambda} is convex in $\lambda$ for $\lambda \in (0,2)$ and the minimum is achieved by 
\begin{equation}
\label{eq:lambda}
\lambda = \frac{2}{\sqrt{\frac{2n \EE_\rho\lrs{\hat L(h,S)}}{\lr{\KL(\rho\|\pi) + \ln \frac{2\sqrt{n}}{\delta}}} + 1} + 1}.
\end{equation}
Note that the optimal value of $\lambda$ is smaller than 1. Alternating application of update rules \eqref{eq:rho} and \eqref{eq:lambda} monotonically decreases the bound, and thus converges.

We note that while the right hand side of inequality \eqref{eq:PBlambda} is convex in $\rho$ for a fixed $\lambda$ and convex in $\lambda$ for a fixed $\rho$, it is not simultaneously convex in $\rho$ and $\lambda$. Joint convexity would have been a sufficient, but it is not a necessary condition for convergence of alternating minimization to the global minimum of the bound. \citet{TIWS17} provide sufficient conditions under which the procedure converges to the global minimum, as well as examples of situations where this does not happen.

\subsection{Construction of a Hypothesis Space for PAC-Bayes-$\lambda$}
\label{sec:PBlhyp}

If ${\cal H}$ is infinite, computation of the partition function (the denominator in \eqref{eq:rho}) is intractable. This could be resolved by parametrization of $\rho$ (for example, restriction of $\rho$ to a Gaussian posterior), but, as we have already mentioned, this may break the convexity of the bound in $\rho$. Fortunately, things get easy when ${\cal H}$ is finite. The crucial step is to construct a sufficiently powerful finite hypothesis space ${\cal H}$. One possibility that we consider here is to construct ${\cal H}$ by training $m$ hypotheses, where each hypothesis is trained on $r$ random points from $S$ and validated on the remaining $n-r$ points. This construction resembles a cross-validation split of the data. However, in cross-validation $r$ is typically large (close to $n$) and validation sets are non-overlapping. The approach considered here works for any $r$ and has additional computational advantages when $r$ is small. We do not require validation sets to be non-overlapping and overlaps between training sets are allowed. Below we describe the construction more formally.

Let $h \in \lrc{1,\dots,m}$ index the hypotheses in ${\cal H}$. Let $S_h$ denote the training set of $h$ and $S\setminus S_h$ the validation set. $S_h$ is a subset of $r$ points from $S$, which are selected independently of their values (for example, subsampled randomly or picked according to a predefined partition of the data). We define the validation error of $h$ by $\Lval(h,S) = \frac{1}{n-r} \sum_{(X,Y) \in S\setminus S_h} \ell(h(X),Y)$. Note that the validation errors are $(n-r)$ i.i.d.\ random variables with bias $L(h)$ and, therefore, for $f(h,S) = (n-r) \kl(\Lval(h,S)\|L(h))$ we have $\EE_S\lrs{e^{f(h,S)}} \leq 2\sqrt{n-r}$. The following result is a straightforward adaptation of Theorem~\ref{thm:PBlambda} to this setting (we leave the proof as an exercise to the reader).
\begin{theorem}[PAC-Bayesian Aggregation]
\label{thm:PAC-Bayes-aggregation} Let $S$ be a sample of size $n$. Let ${\cal H}$ be a set of $m$ hypotheses, where each $h \in {\cal H}$ is trained on $r$ points from $S$ selected independently of the composition of $S$. For any probability distribution $\pi$ over ${\cal H}$ that is independent of $S$ and any $\delta \in (0,1)$, with probability greater than $1-\delta$  over a random draw of a sample S, for all distributions $\rho$ over $\cal H$ and $\lambda \in (0,2)$ simultaneously:
\begin{equation}
\label{eq:PBaggregation}
\EE_\rho\lrs{L(h)} \leq \frac{\EE_\rho\lrs{\Lval(h,S)}}{1 - \frac{\lambda}{2}} + \frac{\KL(\rho\|\pi) + \ln \frac{2\sqrt{n-r}}{\delta}}{\lambda\lr{1-\frac{\lambda}{2}}(n-r)}.
\end{equation}
\label{thm:PBaggregation}
\end{theorem}
It is natural, but not mandatory to select a uniform prior $\pi(h) = 1/m$. The bound in equation \eqref{eq:PBaggregation} can be minimized by alternating application of the update rules in equations \eqref{eq:rho} and \eqref{eq:lambda} with $n$ being replaced by $n-r$ and $\hat L$ by $\Lval$. For evaluation of the empirical performance of this learning approach see \citet{TIWS17}.

\section{PAC-Bayesian Analysis of Ensemble Classifiers}
\label{sec:PAC-Bayes-Ensembles}

So far in this chapter we have discussed various methods of selection of classifiers from a hypothesis set $\HH$. We now turn to \emph{aggregation} of predictions by multiple classifiers through a \emph{weighted majority vote}. The power of the majority vote is in the ``cancellation of errors'' effect: \emph{if} predictions of different classifiers are uncorrelated and they all predict better than a random guess (meaning that $L(h) < 1/2$), the errors tend to cancel out. This can be compared to a consultation of medical experts, which tends to predict better than the best expert in the set. Most machine learning competitions are won by strategies that aggregate predictions of multiple classifiers. The assumptions that the errors are uncorrelated and the predictions are better than random are important. For example, if we have three hypotheses with $L(h) = p$ and independent errors, the probability that a uniform majority vote $\MV_u$ makes an error equals the probability that at least two out of the three hypotheses make an error. You are welcome to verify that in this case for $p \leq 1/2$ we have $L(\MV_u) \leq \E_u\lrs{L(h)}$, where $u$ is the uniform distribution. If the errors are correlated, it can be shown that $L(\MV_\rho)$ can be larger than $\ErhoL$, but as we show below it is never larger than $2\ErhoL$. The reader is welcome to construct an example, where $L(\MV_u) > \E_u\lrs{L(h)}$.

\subsection{Ensemble Classifiers and Weighted Majority Vote}
We now turn to some formal definitions. Ensemble classifiers predict by taking a weighted aggregation of predictions by hypotheses from ${\cal H}$. In multi-class prediction (the label space $\cal Y$ is finite) $\rho$-weighted majority vote $\MV_\rho$ predicts 
\[\MV_\rho (X)= \arg\max_{Y\in{\cal Y}} \sum_{(h\in{\cal H})\wedge (h(X)=Y)} \rho(h),
\]
where $\wedge$ represents the logical ``and" operation and ties can be resolved arbitrarily. 

In binary prediction with prediction space $h(X) \in \lrc{\pm 1}$ weighted majority vote can be written as
\[
\MV_\rho(X) = \sgn{\E_\rho\lrs{h(X)}},
\]
where $\sign(x) = 1$ if $x > 0$ and $-1$ otherwise (the value of $\sign(0)$ can be defined arbitrarily). For a countable hypothesis space this becomes
\begin{equation}  
\label{eq:MVrho}
\MV_\rho(X) = \sgn{\sum_{h\in\HH} \rho(h) h(X)}.
\end{equation}

\subsection{First Order Oracle Bound for the Weighted Majority Vote}
If majority vote makes an error, we know that at least a $\rho$-weighted half of the classifiers have made an error and, therefore, $\ell(\MV_\rho(X),Y) \leq \1[\E_\rho[\1[h(X)\neq Y]] \geq 0.5]$. This observation leads to the well-known first order oracle bound for the loss of weighted majority vote.
\begin{theorem}[First Order Oracle Bound]
\label{thm:first-order}
\[
L(\MV_\rho)\leq 2\E_\rho[L(h)].
\]
\end{theorem}
\begin{proof}
We have $L(\MV_\rho) = \E_D[\ell(\MV_\rho(X),Y)] \leq \P[\E_\rho[\1[h(X)\neq Y]] \geq 0.5]$.  By applying Markov's inequality to random variable $Z = \E_\rho[\1[h(X)\neq Y]]$ we have: 
\begin{equation*}
L(\MV_\rho) \leq \P[\E_\rho[\1[h(X)\neq Y]] \geq 0.5] \leq 2\E_D[\E_\rho[\1[h(X)\neq Y]]] = 2\E_\rho[L(h)].
\end{equation*}
\end{proof}
PAC-Bayesian analysis can be used to bound $\E_\rho[L(h)]$ in Theorem~\ref{thm:first-order} in terms of $\E_\rho[\hat L(h,S)]$, thus turning the oracle bound into an empirical one. The disadvantage of the first order approach is that $\E_\rho[L(h)]$ ignores correlations of predictions, which is the main power of the majority vote.

\subsection{Second Order Oracle Bound for the Weighted Majority Vote}

Now we present a second order bound for the weighted majority vote, which is based on a second order Markov's inequality: for a non-negative random variable $Z$ and $\varepsilon > 0$, we have $\P[Z \geq \varepsilon] = \P[Z^2 \geq \varepsilon^2] \leq \varepsilon^{-2}\E[Z^2]$. We define \emph{tandem loss} of two hypotheses $h$ and $h'$ by
\[
\ell(h(X), h'(X), Y) = \1[h(X)\neq Y \wedge h'(X)\neq Y].
\]
The tandem loss counts an error on a sample $(X,Y)$ only if both $h$ and $h'$ err on $(X,Y)$. We define the expected tandem loss by
\[
L(h,h') = \E_D[\1[h(X)\neq Y \wedge h'(X)\neq Y]].
\]
The following lemma relates the expectation of the second moment of the standard loss to the expected tandem loss. We use the shorthand $\E_{\rho^2}[L(h,h')] = \E_{h\sim\rho, h'\sim\rho}[L(h,h')]$.
\begin{lemma}In multiclass classification
\label{lem:second-monent}
\[
\E_D[\E_\rho[\1[h(X) \neq Y]]^2] = \E_{\rho^2}[L(h,h')].
\]
\end{lemma}
\begin{proof}
\begin{align}
\E_D[\E_\rho[\1[h(X)\neq Y]]^2] &= \E_D[\E_\rho[\1[h(X)\neq Y]]\E_\rho[\1[h(X)\neq Y]]]\label{eq:MV-mid-bound}\\
&= \E_D[\E_{\rho^2}[\1[h(X)\neq Y]\1[h'(X)\neq Y]]]\notag\\
&= \E_D[\E_{\rho^2}[\1[h(X)\neq Y \wedge h'(X)\neq Y]]]\notag\\
&= \E_{\rho^2}[\E_D[\1[h(X)\neq Y \wedge h'(X)\neq Y]]]\notag\\
&= \E_{\rho^2}[L(h,h')].\notag   
\end{align}
\end{proof}
A combination of second order Markov's inequality with Lemma~\ref{lem:second-monent} leads to the following result.
\begin{theorem}[Second Order Oracle Bound]
\label{thm:MV-bound}
In multiclass classification
\begin{equation}
\label{eq:MV-bound}
L(\MV_\rho) \leq 4\E_{\rho^2}[L(h,h')].
\end{equation}
\end{theorem}
\begin{proof}
By second order Markov's inequality applied to $Z = \E_\rho[\1[h(X)\neq Y]]$ and Lemma~\ref{lem:second-monent}:
\begin{equation*}
L(\MV_\rho) \leq \P[\E_\rho[\1[h(X) \neq Y]] \geq 0.5] \leq 4\E_D[\E_\rho[\1[h(X)\neq Y]]^2] = 4\E_{\rho^2}[L(h,h')].
\end{equation*}
\end{proof}

\subsubsection{A Specialized Bound for Binary Classification}
We provide an alternative form of Theorem~\ref{thm:MV-bound}, which can be used to exploit unlabeled data in binary classification. We denote the \emph{expected disagreement} between hypotheses $h$ and $h'$ by $\D(h,h') = \E_D[\1[h(X)\neq h'(X)]]$ and express the tandem loss in terms of standard loss and disagreement.
\begin{lemma}
\label{lem:L2D}
In binary classification
\[
\E_{\rho^2}[L(h,h')] = \E_\rho[L(h)] - \frac{1}{2}\E_{\rho^2}[\D(h,h')].
\]
\end{lemma}
\begin{proof}[Proof of Lemma~\ref{lem:L2D}]
Picking from \eqref{eq:MV-mid-bound}, we have
\begin{align*}
\E_\rho[\1[h(X)\neq Y]]\E_\rho[\1[h(X)\neq Y]] &= \E_\rho[\1[h(X)\neq Y]](1 - \E_\rho[(1 - \1[h(X)\neq Y])]\\
&= \E_\rho[\1[h(X)\neq Y]] - \E_\rho[\1[h(X)\neq Y]]\E_\rho[\1[h(X)= Y]]\\
& = \E_{\rho}[\1[h(X)\neq Y]] - \E_{\rho^2}[\1[h(X)\neq Y\wedge h'(X)= Y]]\\
&= \E_\rho[\1[h(X)\neq Y]] - \frac{1}{2}\E_{\rho^2}[\1[h(X)\neq h'(X)]].
\end{align*}
By taking expectation with respect to $D$ on both sides and applying Lemma~\ref{lem:second-monent} to the left hand side, we obtain:
\[
\E_{\rho^2}[L(h,h')] = \E_D[\E_\rho[\1[h(X)\neq Y]] - \frac{1}{2}\E_{\rho^2}[\1[h(X)\neq h'(X)]]] = \E_\rho[L(h)] - \frac{1}{2}\E_{\rho^2}[\D(h,h')].
\]
\end{proof}

The lemma leads to the following result.
\begin{theorem}[Second Order Oracle Bound for Binary Classification]
\label{thm:MV-bound-binary}
In binary classification
\begin{equation}
\label{eq:MV-binary}
L(\MV_\rho) \leq 4\E_\rho[L(h)] - 2\E_{\rho^2}[\D(h,h')].
\end{equation}
\end{theorem}
\begin{proof}
The theorem follows by plugging the result of Lemma~\ref{lem:L2D} into Theorem~\ref{thm:MV-bound}.
\end{proof}
The advantage of the alternative way of writing the bound is the possibility of using unlabeled data for estimation of $\D(h,h')$ in binary prediction (see also \citealp{GLL+15}). We note, however, that estimation of $\E_{\rho^2}[\D(h,h')]$ has a slow convergence rate, as opposed to $\E_{\rho^2}[L(h,h')]$, which has a fast convergence rate. We discuss this point in Section~\ref{sec:fast-vs-slow}.

\subsection{Comparison of the First and Second Order Oracle Bounds}
\label{sec:FOvsSO}

From Theorems~\ref{thm:first-order} and \ref{thm:MV-bound-binary} we see that in binary classification the second order bound is tighter when $\E_{\rho^2}[\D(h,h')] > \E_\rho[L(h)]$. Below we provide a more detailed comparison of Theorems~\ref{thm:first-order} and \ref{thm:MV-bound} in the worst, the best, and the independent cases. The comparison only concerns the oracle bounds, whereas estimation of the oracle quantities, $\E_\rho[L(h)]$ and $\E_{\rho^2}[L(h,h')]$, is discussed in Section~\ref{sec:fast-vs-slow}.

\paragraph{The worst case} Since $\E_{\rho^2}[L(h,h')] \leq \E_\rho[L(h)]$ the second order bound is at most twice worse than the first order bound. The worst case happens, for example, if all hypotheses in $\cal{H}$ give identical predictions. Then $\E_{\rho^2}[L(h,h')] = \E_\rho[L(h)] = L(\MV_\rho)$ for all $\rho$.

\paragraph{The best case} Imagine that $\cal{H}$ consists of $M\geq 3$ hypotheses, such that each hypothesis errs on $1/M$ of the sample space (according to the distribution $D$) and that the error regions are disjoint. Then $L(h) = 1/M$ for all $h$ and $L(h,h') = 0$ for all $h\neq h'$ and $L(h,h)=1/M$. For a uniform distribution $\rho$ on $\cal{H}$ the first order bound is $2\E_\rho[L(h)] = 2/M$ and the second order bound is $4\E_{\rho^2}[L(h,h')] = 4/M^2$ and $L(\MV_\rho)=0$. In this case the second order bound is an order of magnitude tighter than the first order.

\paragraph{The independent case} Assume that all hypotheses in $\cal{H}$ make independent errors and have the same error rate, $L(h) = L(h')$ for all $h$ and $h'$. Then for $h\neq h'$ we have $L(h,h') = \E_D[\1[h(X)\neq Y \wedge h'(X)\neq Y]] = \E_D[\1[h(X)\neq Y]\1[h'(X)\neq Y]] = \E_D[\1[h(X)\neq Y]]\E_D[\1[h'(X)\neq Y]] = L(h)^2$ and $L(h,h)=L(h)$. For a uniform distribution $\rho$ the second order bound is $4\E_{\rho^2}[L(h,h')] = 4(L(h)^2 + \frac{1}{M}L(h)(1-L(h)))$ and the first order bound is $2\E_{\rho}[L(h)] = 2L(h)$. Assuming that $M$ is large, so that we can ignore the second term in the second order bound, we obtain that it is tighter for $L(h) < 1/2$ and looser otherwise. The former is the interesting regime, especially in binary classification.  

\subsection{Second Order PAC-Bayesian Bounds for the Weighted Majority Vote}

Now we provide an empirical bound for the weighted majority vote. We define the \emph{empirical tandem loss}
\[
\hat L(h,h',S) = \frac{1}{n}\sum_{i=1}^n \1[h(X_i)\neq Y_i \wedge h'(X_i) \neq Y_i]
\]
and provide a bound on the expected loss of $\rho$-weighted majority vote in terms of the empirical tandem losses.
\begin{theorem}
\label{thm:tandem-lambda}
For any probability distribution $\pi$ on $\cal{H}$ that is independent of $S$ and any $\delta\in(0,1)$, with probability at least $1-\delta$ over a random draw of $S$, for all distributions $\rho$ on $\cal{H}$ and all $\lambda\in(0,2)$ simultaneously:
\[
L(\MV_\rho) \leq 4\lr{\frac{\E_{\rho^2}[\hat L(h,h',S)]}{1-\lambda/2} + \frac{2\KL(\rho\|\pi) + \ln(2\sqrt n/\delta)}{\lambda(1-\lambda/2)n}}.
\]
\end{theorem}
\begin{proof}
The theorem follows by using the bound in equation~\eqref{eq:PBlambda} to bound $\E_{\rho^2}[L(h,h')]$ in Theorem~\ref{thm:MV-bound}. We note that $\KL(\rho^2\|\pi^2) = 2\KL(\rho\|\pi)$ \citep[Page 814]{GLL+15}.
\end{proof}
It is also possible to use PAC-Bayes-kl to bound $\E_{\rho^2}[L(h,h')]$ in Theorem~\ref{thm:MV-bound}, which actually gives a tighter bound, but the bound in  Theorem~\ref{thm:tandem-lambda} is more convenient for minimization. We refer the reader to \citet{MLIS20} for a procedure for bound minimization.

\subsubsection{A specialized bound for binary classification}

We define the \emph{empirical disagreement}
\[
\hat \D(h,h',S') = \frac{1}{m} \sum_{i=1}^m \1[h(X_i)\neq h'(X_i)],
\]
where $S' = \lrc{X_1,\dots,X_m}$. The set $S'$ may overlap with the labeled set $S$, however, $S'$ may include additional unlabeled data.
The following theorem bounds the loss of weighted majority vote in terms of empirical disagreements. Due to possibility of using unlabeled data for estimation of disagreements in the binary case, the theorem has the potential of yielding a tighter bound when a considerable amount of unlabeled data is available. 
\begin{theorem}
\label{thm:disagreement}
In binary classification, for any probability distribution $\pi$ on $\cal{H}$ that is independent of $S$ and $S'$ and any $\delta\in(0,1)$, with probability at least $1-\delta$ over a random draw of $S$ and $S'$, for all distributions $\rho$ on $\cal{H}$ and all $\lambda\in(0,2)$ and $\gamma > 0$ simultaneously:
\begin{align*}
L(\MV_\rho) &\leq 4\lr{\frac{\E_\rho[\hat L(h,S)]}{1-\lambda/2} + \frac{\KL(\rho\|\pi) + \ln(4\sqrt n/\delta)}{\lambda(1-\lambda/2)n}}\\
&\qquad - 2\lr{(1-\gamma/2) \E_{\rho^2}[\hat \D(h,h',S')] - \frac{2\KL(\rho\|\pi) + \ln(4\sqrt m/\delta)}{\gamma m}}.
\end{align*}
\end{theorem}
\begin{proof}
The theorem follows by using the upper bound in equation~\eqref{eq:PBlambda} to bound $\E_\rho[L(h)]$ and the lower bound in equation~\eqref{eq:PBlambda-lower} to bound $\E_{\rho^2}[\D(h,h')]$ in Theorem~\ref{thm:MV-bound-binary}. We replace $\delta$ by $\delta/2$ in the upper and lower bound and take a union bound over them.
\end{proof}
Using PAC-Bayes-kl to bound $\E_\rho[L(h)]$ and $\E_{\rho^2}[\D(h,h')]$ in Theorem~\ref{thm:MV-bound-binary} gives a tighter bound, but the bound in Theorem~\ref{thm:disagreement} is more convenient for minimisation. We refer to \citet{MLIS20} for a procedure for bound minimization.

\subsection{Ensemble Construction}

It is possible to use the same procedure as in Section~\ref{sec:PBlhyp} to construct an ensemble. Tandem losses can then be estimated on overlaps of validation sets, $(S\setminus S_h) \cap (S\setminus S_{h'})$. The sample size in Theorem~\ref{thm:tandem-lambda} should then be replaced by $\min_{h,h'} |(S\setminus S_h) \cap (S\setminus S_{h'})|$.

\subsection{Comparison of the Empirical Bounds}
\label{sec:fast-vs-slow}

We provide a high-level comparison of the empirical first order bound ($\FO$), the empirical second order bound based on the tandem loss ($\TND$, Theorem~\ref{thm:tandem-lambda}), and the new empirical second order bound based on disagreements ($\DIS$, Theorem~\ref{thm:disagreement}). The two key quantities in the comparison are the sample size $n$ in the denominator of the bounds and fast and slow convergence rates for the standard (first order) loss, the tandem loss, and the disagreements. \citet{TS13} have shown that if we optimize $\lambda$ for a given $\rho$, the PAC-Bayes-$\lambda$ bound in equation~\eqref{eq:PBlambda} can be written as
\[
\E_\rho[L(h)] \leq \E_\rho[\hat L(h,S)] + \sqrt{\frac{2\E_\rho[\hat L(h,S)]\lr{\KL(\rho\|\pi) + \ln(2\sqrt{n}/\delta)}}{n}} + \frac{2\lr{\KL(\rho\|\pi) + \ln(2\sqrt{n}/\delta)}}{n}.
\]
This form of the bound, also used by \citet{McA03}, is convenient for explanation of fast and slow rates. If $\E_\rho[\hat L(h,S)]$ is large, then the middle term on the right hand side dominates the complexity and the bound decreases at the rate of $1/\sqrt{n}$, which is known as a \emph{slow rate}. If $\E_\rho[\hat L(h,S)]$ is small, then the last term dominates and the bound decreases at the rate of $1/n$, which is known as a \emph{fast rate}. 

\paragraph{$\FO$ vs.\ $\TND$} The advantage of the $\FO$ bound is that the validation sets $S \setminus S_h$ available for estimation of the first order losses $\hat L(h,S_h)$ are larger than the validation sets $(S \setminus S_h) \cap (S \setminus S_{h'})$ available for estimation of the tandem losses. Therefore, the denominator $\nmin = \min_h |S \setminus S_h|$ in the $\FO$ bound is larger than the denominator $\nmin = \min_{h,h'}|(S \setminus S_h) \cap (S \setminus S_{h'})|$ in the $\TND$ bound. The $\TND$ disadvantage can be reduced by using data splits with large validation sets $S \setminus S_h$ and small training sets $S_h$, as long as small training sets do not overly impact the quality of base classifiers $h$. Another advantage of the $\FO$ bound is that its complexity term has $\KL(\rho\|\pi)$, whereas the $\TND$ bound has $2\KL(\rho\|\pi)$. The advantage of the $\TND$ bound is that $\E_{\rho^2}[L(h,h')] \leq E_\rho[L(h)]$ and, therefore, the convergence rate of the tandem loss is typically faster than the convergence rate of the first order loss. The 
interplay of the estimation advantages and disadvantages, combined with the advantages and disadvantages of the underlying oracle bounds discussed in Section~\ref{sec:FOvsSO}, 
depends on the data and the hypothesis space.

\paragraph{$\TND$ vs.\ $\DIS$} The advantage of the $\DIS$ bound relative to the $\TND$ bound is that in presence of a large amount of unlabeled data the disagreements $\D(h,h')$ can be tightly estimated (the denominator $m$ is large) and the estimation complexity is governed by the first order term, $\E_\rho[L(h)]$, which is "easy" to estimate, as discussed above. However, the $\DIS$ bound has two disadvantages. A minor one is its reliance on estimation of two quantities, $\E_\rho[L(h)]$ and $\E_{\rho^2}[\D(h,h')]$, which requires a union bound, e.g., replacement of $\delta$ by $\delta/2$. A more substantial one is that the disagreement term is desired to be large, and thus has a slow convergence rate. Since slow convergence rate relates to fast convergence rate as $1/\sqrt{n}$ to $1/n$, as a rule of thumb the $\DIS$ bound is expected to outperform $\TND$ only when the amount of unlabeled data is at least quadratic in the amount of labeled data, $m > n^2$.

\paragraph{} For experimental comparison of the bounds and further details we refer the reader to \citet{MLIS20}. A follow-up work by \citet{WML+21} improves the analysis by introducing a second order oracle bound for the weighted majority vote based on a parametric form of the Chebyshev-Cantelli inequality.

\section{PAC-Bayes-split-$\kl$ Inequality}

The PAC-Bayes-$\kl$ inequality in \Cref{thm:PBkl} is a good choice for binary losses, because the $\kl$ Lemma (\Cref{lem:klRef}) is tight for Bernoulli random variables. But if the loss function happens to take more than two values, then PAC-Bayes-$\kl$ might not necessarily be the best choice. In this section we focus on losses taking a finite set of values. Examples of such losses include the \emph{excess loss}, which is a difference of (potentially weighted) losses of two prediction rules, $f(h,h',X,Y) = \ell(h(X),Y) - \gamma \ell(h'(X),Y)$, where $\gamma$ is a weighting parameter. When $\ell(h(X),Y)$ is the zero-one loss, the excess loss $f(h,h',X,Y) \in\lrc{-\gamma,0,1-\gamma,1}$. We will work with such losses in Recursive PAC-Bayes in \Cref{sec:Recursive-PAC-Bayes}. Another example is a tandem loss with an offset, $\ell_\alpha(h(X),h'(X),Y) = (\ell(h(X),Y) - \alpha)(\ell(h'(X),Y)-\alpha)\in\lrc{-\alpha(1-\alpha),\alpha^2,(1-\alpha)^2}$ introduced by \citet{WML+21} in their refined analysis of the weighted majority vote. And a third example is prediction with abstention, where a learner is allowed to abstain at a fixed cost $\gamma < 0.5$, and otherwise pay the zero-one loss on the prediction.

We use the same approach as we used in split-$\kl$ inequality in \Cref{sec:split-kl}. Namely, we represent discrete random variables as a superposition of Bernoulli random variables, and then apply PAC-Bayes-$\kl$ to the decomposition.

Let $f:\HH\times\ZZ\to\lrc{b_0,\dots,b_K}$ be a $(K+1)$-valued loss function. For example, in \Cref{sec:Recursive-PAC-Bayes} we will work with excess losses $f(h,h',X,Y) = \ell(h(X),Y) - \gamma \ell(h'(X),Y)$, and there we define $\ZZ = \HH\times\XX\times\YY$, so that $h\in\HH$ is the first argument of $f$ and $Z = (h',X,Y) \in\ZZ$ is the second argument of $f$. For $j\in\lrc{1,\dots,K}$ let $f_{|j}(\cdot,\cdot) = \1[f(\cdot,\cdot) \geq b_j]$. Let $\DD_Z$ be an unknown distribution on $\ZZ$. For $h\in\HH$ let $F(h) = \E_{\DD_Z}[f(h,Z)]$ and $F_{|j}(h)=\E_{\DD_Z}[f_{|j}(h,Z)]$. Let $S = \lrc{Z_1,\dots,Z_n}$ be an i.i.d.\ sample according to $\DD_Z$ and $\hat F_{|j}(h,S)=\frac{1}{n}\sum_{i=1}^n f_{|j}(h, Z_i)$. 

\begin{theorem}[PAC-Bayes-Split-kl Inequality {\citep{WZCAS24}}]\label{thm:pac-bayes-split-kl-inequality}
For any distribution $\pi$ on $\HH$ that is independent of $S$ and any $\delta\in(0,1)$:
\[
    \P[\exists\rho\in\mathcal{P}:\E_\rho[F(h)] \geq b_0 + \sum_{j=1}^K \alpha_j \kl^{-1,+}\lr{\E_\rho[\hat F_{|j}(h,S)],\frac{\KL(\rho\|\pi)+\ln\frac{2K\sqrt{n}}{\delta}}{n}}]\leq \delta,
\]
where $\mathcal{P}$ is the set of all possible probability distributions on $\HH$ that can depend on $S$.
\end{theorem}

\begin{proof}
    We have $f(\cdot,\cdot) = b_0 + \sum_{j=1}^K\alpha_j f_{|j}(\cdot,\cdot)$ and $F(h) = b_0 + \sum_{j=1}^K \alpha_j F_{|j}(h)$. Therefore,
\begin{multline*}
    \P[\exists\rho\in\mathcal{P}:\E_\rho[F(h)] \geq b_0 + \sum_{j=1}^K \alpha_j \kl^{-1,+}\lr{\E_\rho[\hat F_{|j}(h,S)],\frac{\KL(\rho\|\pi)+\ln\frac{2K\sqrt{n}}{\delta}}{n}}]
    \\\leq\P[\exists\rho\in\mathcal{P}\text{ and }\exists j:\E_\rho[F_{|j}(h)] \geq \kl^{-1,+}\lr{\E_\rho[\hat F_{|j}(h,S)],\frac{\KL(\rho\|\pi)+\ln\frac{2K\sqrt{n}}{\delta}}{n}}]
    \leq \delta,
\end{multline*}
where the first inequality is by the decomposition of $F$ and the second inequality is by the union bound and application of \Cref{thm:PBkl} to $F_{|j}$ (note that $f_{|j}$ is a zero-one loss function).
\end{proof}

\section{Recursive PAC-Bayes}
\label{sec:Recursive-PAC-Bayes}

In this section we consider sequential processing of the data. Imagine that we have a data set $S$ that is split into $T$ non-overlapping subsets, $S = S_1\cup\dots\cup S_T$, and we process them one after the other. Sequential processing may be unavoidable if the data arrive sequentially, but as we will see, it may also be beneficial when all the data are available in advance.

In PAC-Bayesian analysis we start with a prior distribution over prediction rules and after observing the data we update it to a posterior distribution. The posterior can then be turned into a prior for processing more data. However, a challenge is that the denominator in PAC-Bayes bounds, e.g., in \Cref{thm:PBkl}, is the amount of data used for construction of the posterior, whereas information on how much data were used for constructing the prior, which reflects confidence in the prior, is lost. This leads to a leaky processing pipeline, because confidence information on the prior is lost every time a new posterior replaces the old prior.

To fix the leak, \citet{WZCAS24} have proposed \emph{Recursive PAC-Bayes}, which provides a way to preserve confidence information on the prior from one processing round to the next. In order to present the idea, we let $\pi_0^*,\pi_1^*,\dots,\pi_T^*$ be a sequence of distributions over $\HH$. The first of them, $\pi_0^*$, is an initial prior distribution that does not depend on $S$. For $t\in\lrc{1,\dots,T}$, $\pi_t^*$ is a posterior distribution over $\HH$ that is obtained by starting from a prior distribution $\pi_{t-1}^*$ and processing data subset $S_t$. After $S_t$ has been processed, $\pi_t^*$ becomes a prior for processing the next chunk, $S_{t+1}$. The final goal is to obtain a good generalization bound on the expected loss of the last posterior, $\E_{\pi_T^*}[L(h)]$.

\citet{WZCAS24} came up with the following recursive decomposition of the loss, which lays the foundation for Recursive PAC-Bayes:
\begin{equation}
\label{eq:RPB}
\E_{\pi_t}[L(h)] = \E_{\pi_t}[L(h) - \gamma_t\E_{\pi_{t-1}}[L(h')]] + \gamma_t\E_{\pi_{t-1}}[L(h')],    
\end{equation}
where $\gamma_t\in[0,1]$. 

\begin{figure}[t]
    \centering
    \includegraphics[width=\textwidth]{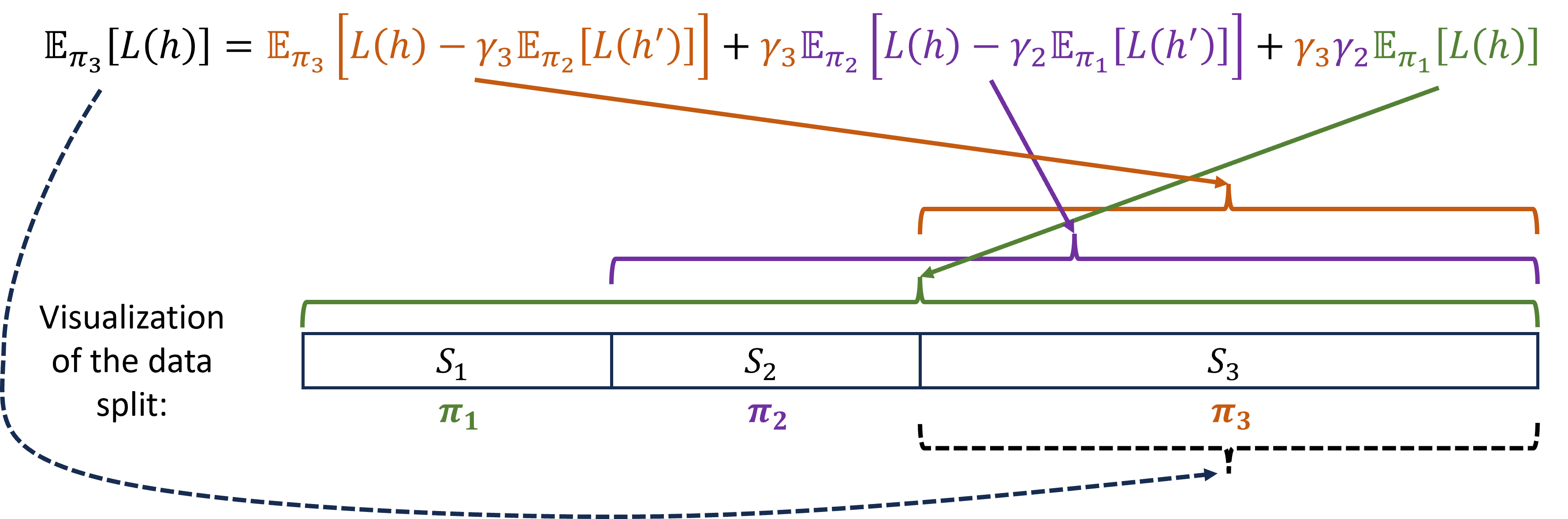}
    \caption{
\textbf{Recursive Decomposition into Three Terms.} The figure illustrates recursive decomposition of $\E_{\pi_3}[L(h)]$ into three terms based on equation \eqref{eq:RPB}, and a geometric data split. The bottom line illustrates which data are used for construction of which distribution: $S_1$ for $\pi_1$; $S_2$ for $\pi_2$; and $S_3$ for $\pi_3$. The brackets above the data show which data are used for computing PAC-Bayes bounds for which term: $S_1\cup S_2\cup S_3$ for $\E_{\pi_1}[L(h)]$; $S_2\cup S_3$ for $\E_{\pi_2}[L(h)-\gamma_2\E_{\pi_1}[L(h')]]$; and $S_3$ for $\E_{\pi_3}[L(h) - \gamma_3\E_{\pi_2}[L(h')]]$. Note that a direct computation of a PAC-Bayes bound on $\E_{\pi_3}[L(h)]$ would only use the data in $S_3$, as shown by the black dashed line. The figure illustrates that recursive decomposition provides more efficient use of the data. We also note that initially we start with poor priors, and so the $\KL(\pi_t\|\pi_{t-1})$ term for small $t$ is expected to be large, but this is compensated by a small multiplicative factor $\prod_{i=t+1}^T\gamma_i$ and availability of a lot of data $\bigcup_{i=t}^T S_i$ for computing the PAC-Bayes bound. For example, $\E_{\pi_1}[L(h)]$ is multiplied by $\gamma_3\gamma_2$ and we can use all the data for computing a PAC-Bayes bound on this term. By the time we reach higher $t$, the priors $\pi_{t-1}$ get better, and the $\KL(\pi_t\|\pi_{t-1})$ term in the bounds gets much smaller, and additionally the bounds benefit from the small variance of the excess loss. With geometric split of the data, we use little data to quickly move $\pi_t$ to a good region, and then we still have enough data for a good estimation of the later terms, like $\E_{\pi_3}[L(h) - \gamma_3\E_{\pi_2}[L(h')]]$. (The figure is borrowed from \citet{WZCAS24}.)
}
    \label{fig:3-step-decomposition}
\end{figure}

For example, for $T=3$ the decomposition becomes
\[
\E_{\pi_3}[L(h)] = \E_{\pi_3}[L(h) - \gamma_3\E_{\pi_2}[L(h')]] + \gamma_3\E_{\pi_2}[L(h) - \gamma_2\E_{\pi_1}[L(h')]] + \gamma_3\gamma_2\E_{\pi_1}[L(h')],
\] see \Cref{fig:3-step-decomposition} for a graphical illustration and discussion.

We make a few observations based on \eqref{eq:RPB}.
\begin{itemize}
    \item $\E_{\pi_{t-1}}[L(h')]$ is ``of the same kind'' as $\E_{\pi_t}[L(h)]$, which is why we can apply the decomposition recursively.
    \item The decomposition in \eqref{eq:RPB} applies to any loss function, including unbounded losses, assuming all the expectations are well-defined.
    \item $\E_{\pi_t}[L(h) - \gamma_t\E_{\pi_{t-1}}[L(h')]] = \E_{\pi_t}[\E_{\pi_{t-1},X,Y}[\ell(h(X),Y) - \gamma_t \ell(h'(X),Y)]]$ is an expected \emph{excess loss}. If $\ell(h(X),Y)$ and $\E_{\pi_{t-1}}[\ell(h'(X),Y)]$ are correlated, then the excess loss has lower variance and, therefore, potentially tighter bound than the plain loss $\E_{\pi_t}[L(h)]$. Excess loss has to be bounded using PAC-Bayes bounds that are capable of exploiting small variance, for example, PAC-Bayes-split-$\kl$. Basic PAC-Bayes bounds, such as PAC-Bayes-$\kl$, are ``blind'' to the variance and destroy the advantage of working with excess losses. 
    \item In order to bound the excess loss we are going to look at $f(h,(h',X,Y)) = \ell(h(X),Y) - \gamma_t \ell(h'(X),Y)$ and $F(h) = \E_{h',X,Y}[f(h,(h',X,Y)]$. We will construct samples of triplets $(h',X,Y)$, where $(X,Y)$ come from the sample $S$ and $h'$ is sampled according to $\pi_{t-1}$, and then proceed with PAC-Bayesian bounding of $\E_{\pi_t}[F(h)]$ using PAC-Bayes-split-$\kl$ bound.
    \item Recursive PAC-Bayes exploits the following important property of PAC-Bayes. Even though we only use $S_t$ in the construction of $\pi_t^*$, we can use $S_t, S_{t+1},\dots,S_T$ for computing empirical estimates of $F(h)$ and bounding $\E_{\pi_t^*}[F(h)]$. This is because the prior $\pi_{t-1}^*$ is independent of $S_t, S_{t+1},\dots,S_T$ and the posterior is allowed, but not required to depend on all the data. And note that even though we will use $S_{t+1},\dots,S_T$ for estimation of $\E_{\pi_t^*}[F(h)]$, $\pi_t^*$ will stay independent of $S_{t+1},\dots,S_T$, as required from a prior for $\pi_{t+1}^*$. Thus, $S_t$ is used for construction of $\pi_t^*$ and estimation of $\E_{\pi_s^*}[F(h)]$ for all $s\leq t$, yielding an efficient use of the data.
\end{itemize}

We now present a generic Recursive Bound. The bound can be seen as a generic shell that needs to be filled in with concrete bounds $\B_1(\pi_1)$ on $\E_{\pi_1}[L(h)]$ and $\Ex_t(\pi_t,\gamma_t)$ on the excess losses $\E_{\pi_t}[L(h)-\gamma_t\E_{\pi_{t-1}^*}[L(h')]]$ for $t\geq 2$. The filling of the shell will happen one theorem later.

\begin{theorem}[Recursive Bound {\citep{WZCAS24}}]
\label{thm:RPB}
    Let $S = S_1\cup\dots\cup S_T$ be an i.i.d.\ sample of size $n$ split in an arbitrary way into $T$ non-overlapping subsamples, and let $\Utrain_t = \bigcup_{s=1}^t S_s$ and $\Uval_t = \bigcup_{s=t}^T S_s$. Let $\pi_0^*,\pi_1^*,\dots,\pi_T^*$ be a sequence of distributions on $\HH$, where $\pi_t^*$ is allowed to depend on $\Utrain_t$, but not the rest of the data. Let $\gamma_2,\dots,\gamma_T$ be a sequence of coefficients, where $\gamma_t$ is allowed to depend on $\Utrain_{t-1}$, but not the rest of the data. For $t\in\lrc{1,\dots,T}$ let $\Pcal_t$ be a set of distributions on $\HH$, which are allowed to depend on $\Utrain_t$. Let $\delta\in(0,1)$. Assume there exists a function $\B_1(\pi_1)$ that satisfies
    \begin{equation}
    \label{eq:RPBbase}
    \P[\exists \pi_1\in\Pcal_1: \E_{\pi_1}[L(h)] \geq \B_1(\pi_1)]\leq \frac\delta T.
    \end{equation}
    For $t\geq 2$ assume there exist functions $\Ex_t(\pi_t,\gamma_t)$ that satisfy
    \begin{equation}
    \label{eq:RPBstep}        
    \P[\exists \pi_t\in\Pcal_t: \E_{\pi_t}[L(h) - \gamma_t\E_{\pi_{t-1}^*}[L(h')]] \geq \Ex_t(\pi_t,\gamma_t)] \leq \frac\delta T.
    \end{equation}
    Let 
    \begin{equation}
    \label{eq:RPBbound}
    \B_t(\pi_t) = \Ex_t(\pi_t,\gamma_t) + \gamma_t \B_{t-1}(\pi_{t-1}^*).
    \end{equation}
    Then
    \begin{equation}
    \label{eq:RPBstatement}        
    \P[\exists t\in\lrc{1,\dots,T}\text{ and } \pi_t \in \Pcal_t: \E_{\pi_t}[L(h)] \geq \B_t(\pi_t)] \leq \delta.
    \end{equation}
\end{theorem}

\begin{proof}
    The proof follows directly by induction, because the definition of $\B_t(\pi_t)$ in equation \eqref{eq:RPBbound} matches the recursive decomposition of the loss in equation \eqref{eq:RPB}, and so
    \begin{align*}
    &\P[\exists t\in\lrc{1,\dots,T}\text{ and } \pi_t \in \Pcal_t: \E_{\pi_t}[L(h)] \geq \B_t(\pi_t)] \\
    &\qquad\leq \P[\exists \pi_1\in\Pcal_1: \E_{\pi_1}[L(h)] \geq \B_1(\pi_1) \text{~~ OR ~~} \exists \pi_t\in\Pcal_t \text{~for~} t\geq 2: \E_{\pi_t}[L(h) - \gamma_t\E_{\pi_{t-1}^*}[L(h')]] \geq \Ex_t(\pi_t,\gamma_t)]\\
    &\qquad\leq\delta,
    \end{align*}
where the last step is by the union bound.
\end{proof}


\Cref{thm:RPB} only states that if we could control $\E_{\pi_1}[L(h)]$ and $\E_{\pi_t}[L(h)-\gamma_t\E_{\pi_{t-1}^*}[L(h')]]$ for all $t\geq 2$, then we would have $\E_{\pi_t}[L(h)]$ under control for all $t$. We can use any basic PAC-Bayes bound, for example, \Cref{thm:PBkl}, to control the plain loss $\E_{\pi_1}[L(h)]$ and any bound for non-binary losses, for example, \Cref{thm:pac-bayes-split-kl-inequality}, to control the excess losses $\E_{\pi_t}[L(h)-\gamma_t\E_{\pi_{t-1}^*}[L(h')]]$.

Next we present one concrete way of defining the bounds using PAC-Bayes-split-$\kl$ inequality. In order to apply the inequality we define
\[F_{\gamma_t,\pi_{t-1}^*}(h) = L(h) - \gamma_t \E_{\pi_{t-1}^*}[L(h')] = \E_{\pi_{t-1}^*\times\DD}[\ell(h(X),Y) - \gamma_t\ell(h'(X),Y)],
\]
where $\pi_{t-1}^*\times\DD$ is a product distribution on $\HH\times\XX\times\YY$ and $h'\in\HH$ is sampled according to $\pi_{t-1}^*$. Then $\E_{\pi_t}[L(h)-\gamma_t\E_{\pi_{t-1}^*}[L(h')]]=\E_{\pi_t}[F_{\gamma_t,\pi_{t-1}^*}(h)]$. We further define 
\[
f_{\gamma_t}(h,(h',X,Y)) = \ell(h(X),Y) - \gamma_t \ell(h'(X),Y) \in \lrc{-\gamma_t, 0, 1-\gamma_t, 1},
\]
then $F_{\gamma_t,\pi_{t-1}^*}(h) = \E_{\pi_{t-1}^*\times\DD}[f_{\gamma_t}(h,(h',X,Y))]$. In order to apply \Cref{thm:pac-bayes-split-kl-inequality}, we represent $f_{\gamma_t}$ as a superposition of binary functions. For this purpose we let $\lrc{b_{t|0},b_{t|1},b_{t|2},b_{t|3}} = \lrc{-\gamma_t, 0, 1-\gamma_t, 1}$ and define $f_{\gamma_t|j}(h,(h',X,Y)) = \1[f_{\gamma_t}(h,(h',X,Y))\geq b_{t|j}]$. We let $F_{\gamma_t,\pi_{t-1}^*|j}(h) = \E_{\pi_{t-1}^*\times\DD}[f_{\gamma_t|j}(h,(h',X,Y))]$, then $F_{\gamma_t,\pi_{t-1}^*}(h) = -\gamma_t + \sum_{j=1}^3 (b_{t|j}-b_{t|j-1})F_{\gamma_t,\pi_{t-1}^*|j}(h)$.

Now we construct an empirical estimate of $F_{\gamma_t,\pi_{t-1}^*|j}(h)$. We already have a sample of $(X,Y)$ pairs, which we need to complement with a sample of $h'$ to obtain $(h',X,Y)$ triplets. We first let $\hat \pi_{t-1}^* = \lrc{h^{\pi_{t-1}^*}_1,h^{\pi_{t-1}^*}_2,\dots}$ be a sequence of prediction rules sampled independently according to $\pi_{t-1}^*$. We define ${\hat \pi_{t-1}^* \circ \Uval_t} = \lrc{\lr{h^{\pi_{t-1}^*}_i,X_i,Y_i}:(X_i,Y_i)\in\Uval_t}$. In words, for every sample $(X_i,Y_i)\in\Uval_t$ we sample a prediction rule $h^{\pi_{t-1}^*}_i$ according to $\pi_{t-1}^*$ and place the triplet $(h^{\pi_{t-1}^*}_i,X_i,Y_i)$ in $\hat \pi_{t-1}^*\circ\Uval_t$. The triplets $(h^{\pi_{t-1}^*}_i,X_i,Y_i)$ correspond to the random variables $Z$ in \Cref{thm:pac-bayes-split-kl-inequality}. We note that $|\Uval_t| = |\hat \pi_{t-1}^*\circ\Uval_t|$, and we let $\nval_t = |\Uval_t|$. We define an empirical estimate of $F_{\gamma_t,\pi_{t-1}^*|j}(h)$ as $\hat F_{\gamma_t|j}(h,\hat \pi_{t-1}^*\circ\Uval_t) = \frac{1}{\nval_t} \sum_{(h',X,Y)\in\hat\pi_{t-1}^*\circ\Uval_t} f_{\gamma_t|j}(h,(h',X,Y))$. Note that $\E_{\pi_{t-1}^*\times\DD}[\hat F_{\gamma_t|j}(h,\hat \pi_{t-1}^*\circ\Uval_t)] = F_{\gamma_t,\pi_{t-1}^*|j}(h)$, therefore, we can use \Cref{thm:pac-bayes-split-kl-inequality} to bound $\E_{\pi_t}[F_{\gamma_t,\pi_{t-1}^*}(h)]$ using its empirical estimates. We are now ready to state the bound.

\begin{theorem}[Recursive PAC-Bayes-split-$\kl$ Inequality {\citep{WZCAS24}}]
    Let 
    \[
    \B_1(\pi_1) = \kl^{-1,+}\lr{\E_{\pi_1}[\hat L(h,S)],\frac{\KL(\pi_1\|\pi_0^*)+\ln\frac{2T\sqrt{n}}{\delta}}{n}}    
    \]
    and
    \[
    \Ex_t(\pi_t,\gamma_t) =  -\gamma_t + \sum_{j=1}^3 (b_{t|j} - b_{t|j-1})\kl^{-1,+}\lr{\E_{\pi_t}\lrs{\hat F_{\gamma_t|j}(h,\hat \pi_{t-1}^*\circ\Uval_t)}, \frac{\KL(\pi_t\|\pi_{t-1}^*)+\ln\frac{6T\sqrt{\nval_t}}{\delta}}{\nval_t}}.   
    \]
    Then $\B_1$ satisfies \eqref{eq:RPBbase} and $\Ex_t$ satisfies \eqref{eq:RPBstep}, and the statement \eqref{eq:RPBstatement} of  \Cref{thm:RPB} holds.
\end{theorem}

\begin{proof}
Statement \eqref{eq:RPBbase} follows by \Cref{thm:PBkl} and statement \eqref{eq:RPBstep} follows by \Cref{thm:pac-bayes-split-kl-inequality}.
\end{proof}

Next we dicuss practical aspects of how to construct $\pi_t^*$, how to select $\gamma_t$, and how to split the sample.

\subsubsection*{Construction of $\pi_1^*,\dots,\pi_T^*$}

An interesting point about Recursive PAC-Bayes is that $\pi_t^*$ is constructed using $S_t$, but evaluated using $\Uval_t = S_t \bigcup \lr{\bigcup_{s=t+1}^T S_s} = S_t \bigcup \Uval_{t+1}$. For the construction of $\pi_t^*$ we take the triplets $(h',X,Y)$ in $\hat \pi_{t-1}^*\circ S_t$. We could then consider a ``construction bound''
    \[
    \Ex_t^{\texttt{const}}(\pi,\gamma_t,n_t) =  -\gamma_t + \sum_{j=1}^3 (b_{t|j} - b_{t|j-1})\kl^{-1,+}\lr{\E_{\pi}\lrs{\hat F_{\gamma_t|j}(h,\hat \pi_{t-1}^*\circ S_t)}, \frac{\KL(\pi\|\pi_{t-1}^*)+\ln\frac{6T\sqrt{n_t}}{\delta}}{n_t}},
    \]
which would be a high-probability bound on $\E_\pi[L(h)-\gamma_t \E_{\pi_{t-1}^*}[L(h')]]$ based on $S_t$ alone. It involves $\E_{\pi}\lrs{\hat F_{\gamma_t|j}(h,\hat \pi_{t-1}^*\circ S_t)}$ and $n_t = |S_t|$. We could then take $\pi_t^* = \arg\min_{\pi}\Ex_t^{\texttt{const}}(\pi,\gamma_t,n_t)$, which would minimize this bound. However, we know that when we reach the evaluation stage, $\Ex_t(\pi_t,\gamma_t)$ will look at more data and involve $\E_{\pi_t}\lrs{\hat F_{\gamma_t|j}(h,\hat \pi_{t-1}^*\circ\Uval_t)}$ and $\nval_t$. So the denominator of the bound $\nval_t\geq n_t$ will be larger. But it is also likely that $\E_{\pi_t}\lrs{\hat F_{\gamma_t|j}(h,\hat \pi_{t-1}^*\circ\Uval_t)}$ will be larger than $\E_{\pi}\lrs{\hat F_{\gamma_t|j}(h,\hat \pi_{t-1}^*\circ S_t)}$, because $\E_{\pi_t}\lrs{\hat F_{\gamma_t|j}(h,\hat \pi_{t-1}^*\circ\Uval_t)}$ is a weighted average of $\E_{\pi}\lrs{\hat F_{\gamma_t|j}(h,\hat \pi_{t-1}^*\circ S_t)}$ and $\E_{\pi}\lrs{\hat F_{\gamma_t|j}(h,\hat \pi_{t-1}^*\circ\Uval_{t+1})}$. The latter is an unbiased estimate of $\E_{\pi_t^*}[L(h)-\gamma_t\E_{\pi_{t-1}^*}[L(h')]]$, but the former is an underestimate of this quantity, because $\pi_t^*$ is tailored to $S_t$. Therefore, we could exploit the knowledge that we are going to have more data at the evaluation phase and redefine the ``construction bound'' as 
    \[
    \Ex_t^{\texttt{const}}(\pi,\gamma_t,\nval_t) =  -\gamma_t + \sum_{j=1}^3 (b_{t|j} - b_{t|j-1})\kl^{-1,+}\lr{\E_{\pi}\lrs{\hat F_{\gamma_t|j}(h,\hat \pi_{t-1}^*\circ S_t)}, \frac{\KL(\pi\|\pi_{t-1}^*)+\ln\frac{6T\sqrt{\nval_t}}{\delta}}{\nval_t}}
    \]
(i.e., increase the denominator from $n_t$ to $\nval_t$ while still considering only $S_t$ in the loss estimates) and take $\pi_t^* = \arg\min_{\pi}\Ex_t^{\texttt{const}}(\pi,\gamma_t,\nval_t)$. This would allow more aggressive overfitting of $S$ making $\E_{\pi}\lrs{\hat F_{\gamma_t|j}(h,\hat \pi_{t-1}^*\circ S_t)}$ smaller, but the effect on $\E_{\pi_t}\lrs{\hat F_{\gamma_t|j}(h,\hat \pi_{t-1}^*\circ\Uval_t)}$ depends on the data. We emphasize that $\Ex_t(\pi_t,\gamma_t)$ is a valid bound for any choice of $\pi_t^*$ and both choices of the ``construction bound'' are admissible, but which of them would yield a tighter bound depends on the data. Since the bounds are conservative, empirically replacing $n_t$ with $\nval_t$ in the ``construction bound'' usually performs better.

\subsubsection*{Selection of $\gamma_2,\dots,\gamma_T$}

We naturally want to have an improvement in the bound as we proceed from one chunk of the data to the next, meaning that we want $\B_t(\pi_t^*) < \B_{t-1}(\pi_{t-1}^*)$. Substituting this inequality into \eqref{eq:RPBbound} yields $\B_{t-1}(\pi_{t-1}^*) > \B_t(\pi_t^*) = \Ex_t(\pi_t^*,\gamma_t) + \gamma_t\B_{t-1}(\pi_{t-1}^*)$, which leads to $\gamma_t < 1 - \frac{\Ex_t(\pi_t^*,\gamma_t)}{\B_{t-1}(\pi_{t-1}^*)}$. Therefore, $\gamma_t$ should be strictly smaller than 1, and it should also be non-negative. Note that $\gamma_t = 0$ recovers $\E_{\pi_t}[L(h)] = \E_{\pi_t}[L(h)]$. Since $\gamma_t$ is only used once we reach $S_t$ it can be constructed sequentially based on $\Utrain_{t-1}$. It cannot depend on $S_t$, because it would bias the estimates of $F_{\gamma_t,\pi_{t-1}^*}(h)$, but we can select $\gamma_t$ from a grid of values if we take a union bound over the selection. We note that \citet{WZCAS24} have simply taken $\gamma_t = \frac12$.

\subsubsection*{How to split the sample}

When data arrive sequentially the data split is determined by the arrival process. But when all data are available offline in advance, there is a question of how to split the data $S$ into subsets $S_1,\dots,S_T$. One could split the data uniformly, so that $|S_t| = \frac{1}{T}|S|$, but it leaves relatively little data for estimation of the final term, $\Ex_T(\pi_T^*,\gamma_t)$. In order to address this issue, \citet{WZCAS24} have proposed to work with a geometric split of the data, where $|S_T|=\frac{1}{2}|S|$, $|S_{T-1}| = \frac14 |S|$, $\dots$. This approach has two advantages. First, when we start, the prior is typically poor, and so we use little data to bring the prior to a reasonably good region. The large $\KL(\pi_t\|\pi_{t-1})$ term in the first steps is compensated by a small multiplicative factor $\prod_{s=t}^T \gamma_s$ and by the large size of $\Uval_t$, which provides a large denominator. By the time we reach later processing steps, the prior is going to be good, so that the $\KL(\pi_t\|\pi_{t-1})$ will be small, and we will still have a lot of data to compute a good bound $\Ex(\pi_T^*,\gamma_T)$. See \Cref{fig:3-step-decomposition} for an illustration.

\bigskip
\noindent
For empirical evaluation of Recursive PAC-Bayes see the work of \citet{WZCAS24}.

\section{Exercises}

\begin{exercise}[\textit{Experiment design}]~
\begin{enumerate}
	\item You are working at a hospital and you have collected an i.i.d.\ sample of 2000 patients and annotated it for presence or absence of some disease (binary annotation). You organize a competition to find a classifier for the disease. You have 20 teams that have signed up for the competition and your boss requires you to provide a confidence interval of 0.05 on the prediction accuracy of the best classifier that will hold with probability at least 95\%. In other words, with probability at least 95\% the estimate of the expected error should not underestimate the true expected zero-one error of the selected classifier by more than 0.05 (one-sided error). How many samples do you have to keep aside in order to satisfy this requirement, assuming that you accept 1 solution from each team? Provide a complete calculation, numerical answers without any derivations or explanations will not be accepted.
	
	\item You have conducted the competition above, but were not satisfied with the prediction accuracy of the winner. You decided to make another competition and were very lucky to convince your boss to support annotation of another 1000 patients. You decided to release the old 2000 patients data for training and keep the new 1000 samples for evaluating the outcome of the new competition. Your boss requires from you the same confidence interval of 0.05 with probability at least 95\%. How many teams can you accept to take part in the competition assuming that you accept only 1 solution from each team? Provide a complete calculation, numerical answers without any derivations or explanations will not be accepted.
\end{enumerate}
\end{exercise}

\begin{exercise}[\textit{Combining datasets}]
You are approached by a big and a small company, and each proposes you a classifier for a problem of interest, $h_{\texttt{big}}$ and $h_{\texttt{small}}$. They also provide you the data they have used for training their classifiers. The big company has collected a dataset $S_{\texttt{big}}$ of 10000 samples and the small company has collected a dataset $S_\texttt{small}$ of only 1000 samples. We assume that all the data are i.i.d.\ and both samples come from the same distribution. The companies provide no details on how they produced the classifiers. You have no own data to test the solutions, so instead you test $h_{\texttt{big}}$ on $S_\texttt{small}$ and $h_{\texttt{small}}$ on $S_{\texttt{big}}$. You obtain $\hat L(h_{\texttt{big}}, S_\texttt{small}) = 0.03$ and $\hat L(h_{\texttt{small}}, S_{\texttt{big}}) = 0.06$. 

You need to pick a classifier and provide a generalization bound that will hold with probability at least 95\%. Explain which of the two classifiers you will pick and provide a generalization bound for it.
\end{exercise}

\begin{exercise}[\textit{How to split data into training and test sets}]
\label{ex:data-split}

In this question you will analyze one possible approach to the question of how to split a dataset $S$ into training and test sets, $\Strain$ and $\Stest$. As we have already discussed, overly small test sets lead to unreliable loss estimates, whereas overly large test sets leave too little data for training, thus producing poor prediction models. The optimal trade-off depends on the data and the prediction model. So can we let the data speak for itself? We will give it a try.

\begin{enumerate}
\item We want to find a good balance between the sizes of $\Strain$ and $\Stest$. We consider $m$ possible splits $\lrc{\lr{\Strain_{1}, \Stest_{1}}, \dots, \lr{\Strain_{m}, \Stest_{m}}}$, where the sizes of the test sets are $n_1,\dots,n_m$, correspondingly. For example, it could be $(10\%,90\%), (20\%,80\%), \dots, (90\%,10\%)$ splits or anything else with a reasonable coverage of the possible options. We train $m$ prediction models $\hat h_1^*, \dots, \hat h_m^*$, where $\hat h_i^*$ is trained on $\Strain_{i}$. We calculate the test loss of the $i$-th model on the $i$-th test set $\hat L(\hat h_i^*, \Stest_{i})$. Derive a bound on $L(\hat h_i^*)$ in terms of $\hat L(\hat h_i^*, \Stest_{i})$ and $n_i$ that holds for all $\hat h_i^*$ simultaneously with probability at least $1-\delta$.
	
	\emph{Comment: No theorem from the book applies directly to this setting, because they all have a fixed sample size $n$, whereas here the sample sizes $n_1,\dots,n_m$ vary. You have to provide a complete derivation.}
\item We expect that most readers will treat all the splits in the previous point equally. Note, however, that models trained on more data are a-priori expected to perform better. Propose a way to give them an advantage by using a non-uniform treatment [a ``prior''] that will give preference to classifiers trained on more samples and repeat the analysis. You have to propose one explicit prior and do the analysis with that prior.
\end{enumerate}
\end{exercise}

\begin{exercise}[\textit{Efficient use of data}]
Most of the theoretical results in the book use part of the data for training prediction rules and another part for validating them. This way some data are only used for training and some data are only used for validation. But pay attention that if we would have trained a prediction rule on the validation set and validated it on the training set, we would have also gotten an unbiased estimate of the loss. (Remember: ``it's not about how you call it, it's about how you use it''!). So could we use the data more efficiently?

The approach of using part of the data for training and part for validation, and then reverting the roles, somewhat resembles cross-validation. But a warning would be in place here. Even though the standard cross-validation technique is widely used, it is a heuristic, and if it is used to validate too many prediction rules, it is prone to overfitting, in exactly the same way as the standard validation technique is prone to overfitting, unless generalization bounds are used to control the overfitting.

What you will do next is inspired by cross-validation, but it is different from the standard cross-validation approach.

So, we have a data set $S$ of size $n$ (assume that $n$ is even). We split the data set into two equal halves, $S = S_0 \cup S_1$. We train $M$ models $\lrc{h_{0,1},\dots,h_{0,M}}$ on the first half of the data and validate them on the remaining half. Let $\hat L(h_{0,i},S_1)$ for $i\in\lrc{1,\dots,M}$ be the corresponding validation losses. Then we train another $M$ models $\lrc{h_{1,1},\dots,h_{1,M}}$ on the second half of the data and validate them on the first half. Let $\hat L(h_{1,i},S_0)$ for $i\in\lrc{1,\dots,M}$ be the corresponding validation losses. Finally, we select the model $\displaystyle h_{j^*,i^*} = \arg\min_{j\in\lrc{0,1},i\in\lrc{1,\dots,M}}\hat L(h_{j,i},S_{1-j})$ with the smallest validation loss. 

Derive a high-probability generalization bound for the expected loss of $h_{j^*,i^*}$. (I.e., a bound on $L(h_{j^*,i^*})$ that holds with probability at least $1-\delta$.)

\emph{Comment: no theorem from the book directly applies to the question, because they all assume that $\hat L(h,S)$ is computed on the same $S$ for all $h$. You have to make a custom derivation, but it will not be very different from derivations you can find in the book.}
\end{exercise}

\begin{exercise}[\textit{Learning by discretization}]
We want to learn an arbitrary binary function on a unit square by discretizing the square into a uniform grid with $d^2$ cells. The hypothesis space is the space of all possible uniform grids with $d^2$ cells for $d \in \lrc{1,2,3,\dots}$, where each cell gets a binary label.

We have a sample $S$ of size $n$ to learn the function. Let $\HH_d$ be the hypothesis set of uniform grids with $d^2$ cells. Let $\HH = \bigcup_{d=1}^\infty \HH_d$ be the hypothesis set of all possible uniform grids. Let $f(h)$ denote the number of cells in the hypothesis $h$. Let $d(h) = \sqrt{f(h)}$, then $d(h) \in \lrc{1,2,3,\dots}$ and $h \in \HH_{d(h)}$.

\begin{enumerate}
    \item Derive a generalization bound for learning with $\HH_d$. (I.e., a bound on $L(h)$ that holds for all $h\in\HH_d$ with probability at least $1-\delta$: $\P[\forall h\in\HH_d: L(h)\leq \dots] \geq 1-\delta$, your task is to fill in the dots.)
    \item Derive a generalization bound for learning with $\HH$. (I.e., a bound on $L(h)$ that holds for all $h\in\HH$ with probability at least $1-\delta$.)
    \item Write down a selection rule for selecting a prediction rule $h \in \HH$ that is optimal according to the bound in the previous point. Ideally, you answer should be in a form $h^* = \dots$, where $\dots$ is a mathematical expression using the bound.
    \item What is the maximal number of cells as a function of $n$, for which your bound is non-vacuous? (It is sufficient to derive an order of magnitude, you do not need to make a precise calculation.)
    \item Explain how the density of the grid $d(h)$ affects the bound.
    Which terms in the bound (if any) increase as the density of the grid increases and which terms in the bound (if any) decrease as the density of the grid increases?
\end{enumerate}
\end{exercise}

\begin{exercise}[\textit{Early stopping}]
\label{ex:early-stopping}
Early stopping is a widely used technique to avoid overfitting in models trained by iterative methods, such as gradient descent. In particular, it is used to avoid overfitting in training neural networks.
In this question we analyze several ways of implementing early stopping. The technique sets aside a validation set $\Sval$, which is used to monitor the improvement of the training process. Let $h_1, h_2, h_3, \dots$ be a sequence of models obtained after $1, 2, 3, \dots$ epochs of training a neural network or any other prediction model (you do not need to know any details about neural networks or their training procedure to answer the question). Let $\hat L(h_1, \Sval),~ \hat L(h_2,\Sval),~ \hat L(h_3,\Sval), \dots$ be the corresponding sequence of validation errors on the validation set $\Sval$.

\begin{enumerate}
	\item Let $h_{t^*}$ be the neural network returned after training with early stopping. In which of the following cases $\hat L(h_{t^*}, \Sval)$ is an unbiased estimate of $L(h_{t^*})$ and in which cases it is not? Please, explain your answer.
	\begin{enumerate}
		\item Predefined stopping: the training procedure always stops after 100 epochs and always returns the last model $h_{t^*}=h_{100}$.
		\item Non-adaptive stopping: the training procedure is executed for a fixed number of epochs $T$, and returns the model $h_{t^*}$ with the lowest validation error observed during the training process, i.e., $\displaystyle t^* = \arg\min_{t \in \{1,\dots,T\}} \hat L(h_t, \Sval)$.
		\item Adaptive stopping: the training procedure stops when no improvement in $\hat L(h_t,\Sval)$ is observed for a significant number of epochs. It then returns the best model observed ever during training. (This procedure is proposed in \citet[Algorithm 7.1]{GBC16} or \url{https://www.quora.com/How-does-one-employ-early-stopping-in-TensorFlow}, but again, you do not need to know the details of the training procedure.) 
	\end{enumerate}
	\item\label{pnt:2} Derive a high-probability bound (a bound that holds with probability at least $1-\delta$) on $L(h_{t^*})$ in terms of $\hat L(h_{t^*}, \Sval)$, $\delta$, and the size $n$ of the validation set $\Sval$ for the three cases above. In the second case the bound may additionally depend on the total number of epochs $T$, while in the third case the bound may additionally depend on the index $t^*$ of the epoch providing the optimal model. Please, solve the last case using the series $\sum_{i=1}^\infty \frac{1}{i(i+1)} = 1$.\footnote{We have $\sum_{i=1}^\infty \frac{1}{i(i+1)} = \sum_{i=1}^\infty \lr{\frac{1}{i} - \frac{1}{i+1}} = 1$.}
	\item\label{pnt:3} The adaptive approach suggests stopping when ``no improvement in $\hat L(h_t,\Sval)$ is observed for a significant number of epochs''. A natural way of redefining the stopping criterion once we have the generalization bound is to stop when ``no improvement in the generalization bound is observed for a significant number of epochs''. The adaptive approach does not limit the number of epochs in advance, but what is the maximal number of epochs $T_\mathrm{max}$, after which it makes no sense to continue training according to the bound you derived in Point~\ref{pnt:2}? Express $T_\mathrm{max}$ in terms of the number of validation samples $n$. It is sufficient to provide an order of magnitude of $T_\mathrm{max}$ in terms of $n$, you do not have to calculate the explicit constants.
	\item How would your answer to the previous point change if you were to use the series $\sum_{i=1}^\infty \frac{1}{2^i} = 1$ for deriving the bound? (You should get that with this series you can run significantly less epochs in the adaptive approach compared to the series used in Point~\ref{pnt:2}. Thus, unlike in the case of decision trees in \Cref{sec:Occam-apps}, here the choice of the series has a significant impact.)
	\item In this question we compare the adaptive procedure with non-adaptive. Assume that the two procedures use the same initialization, so that the corresponding models at epoch $t$ are identical, and assume that the adaptive procedure has considered all the models $h_t$ for $t\in\lrc{1,\dots,T_\mathrm{max}}$. Let $t^*$ be the index of the model $h_{t^*}$ minimizing the adaptive bound and let $T^*$ be the index of the model $h_{T^*}$ minimizing the non-adaptive bound. Show that the generalization bound for adaptive stopping in Point~\ref{pnt:2} is never much worse than the generalization bound for non-adaptive stopping, but in some cases the adaptive bound can be significantly lower.
	
	Guidance: To simplify the analysis, throughout the question we assume that the confidence parameter $\delta\leq \frac{1}{2}$. For $T\geq 1$ it gives $\delta \leq \frac{1}{2}\leq \frac{T}{T+1}$. 
	\begin{enumerate}
		\item First, assume that $T \leq T_\mathrm{max}$. Let $t^*$ be the index of the epoch selected by the adaptive procedure and $T^*$ be the index of the epoch selected by the non-adaptive procedure. Since the adaptive procedure has selected $t^*$ we know that the adaptive bound for epoch $t^*$ is lower than the adaptive bound for epoch $T^*$. We also know that $T^* \leq T$, where $T$ is the number of epochs in the non-adaptive approach. Use this information and do some bounding to show that for any confidence parameter $\delta \leq \frac{1}{2}$, the adaptive bound can be at most a multiplicative factor of $\sqrt 2$ larger than the non-adaptive bound. 
		\item {[Optional]} Now consider the case $T > T_\mathrm{max}$. Show that in this case the  non-adaptive bound is at least $\frac{1}{\sqrt 2}$. Since the losses are upper bounded by 1, any bound can be truncated at 1 and still be a valid bound. In other words, for any "bound" we can define a "truncated bound" = $\max$(1, "bound") and it will still be a valid bound. So in this case the truncated adaptive bound also cannot exceed the non-adaptive bound by more than a multiplicative factor of $\sqrt 2$.
		\item You have shown that under the assumption that $\delta \leq \frac{1}{2}$ the adaptive bound never exceeds the non-adaptive bound by more than a multiplicative factor of $\sqrt 2$. Now provide \emph{two} examples of sequences of empirical losses $\hat L(h_1,\Sval),\hat L(h_2,\Sval),\dots$, for which the adaptive bound can be significantly smaller than the non-adaptive bound. In both cases you should have $T < T_\mathrm{max}$ and $\delta \leq \frac{1}{2}$.
		\item {[Optional]} Show that irrespective of the choice of $T$, there always exists a sequence of losses $\hat L(h_1,\Sval),\hat L(h_2,\Sval),\dots$, for which $\frac{\text{adaptive bound}}{\text{non adaptive bound}}\leq \lr{\frac{2\ln\frac{2}{\delta}}{n}}^\frac{1}{4}$.
	\end{enumerate}
	
	Conclusion: depending on the data, the generalization bound for adaptive stopping can be significantly smaller than the generalization bound for non-adaptive stopping, and at the same time it is guaranteed that it is never worse by more than a multiplicative factor of $\sqrt 2$.
\end{enumerate}
\end{exercise}

\begin{exercise}[\textit{Occam’s razor with $\kl$ inequality}]
\label{ex:Occam-kl}

In this exercise we derive a version of Occam's razor bound based on the $\kl$ inequality.

\begin{enumerate}

\item Prove the following theorem.

\begin{theorem}[Occam's $\kl$-razor inequality]
\label{thm:kl-Occam}
Let $S$ be an i.i.d. sample of $n$ points, let $\ell$ be a loss function bounded in the $[0,1]$ interval, let ${\cal H}$ be countable, and let $\pi(h)$ be such that it is independent of the sample $S$ and satisfies $\pi(h) \geq 0$ for all $h$ and $\displaystyle\sum_{h \in {\cal H}} \pi(h) \leq 1$. Let $\delta \in (0,1)$. Then 
\[
\P[\exists h\in\HH: \kl(\hat L(h,S)\|L(h)) \geq \frac{\ln\frac{1}{\pi(h)\delta}}{n}]\leq \delta.
\]
\end{theorem}

\noindent
You should prove the theorem directly, and not through relaxation of the PAC-Bayes-kl bound. Briefly emphasize where in your proof are you using the assumption that $\pi(h)$ is independent of $S$, and why is it necessary.

\item The bound in Theorem~\ref{thm:kl-Occam} is somewhat implicit. Prove the following corollary, which makes it more explicit, and clearly shows the ``fast convergence rate'' that it provides.

\begin{corollary}
\label{cor:Occam-kl-ref-Pinsker}
    Under the assumptions of Theorem~\ref{thm:kl-Occam}
\[
\P[\exists h\in\HH: L(h) \geq \hat L(h,S) + \sqrt{\frac{2 \hat L(h,S) \ln\frac{1}{\pi(h)\delta}}{n}} + \frac{2\ln\frac{1}{\pi(h)\delta}}{n}]\leq \delta.
\]
\end{corollary}

\item Briefly compare \Cref{cor:Occam-kl-ref-Pinsker} with Occam's razor bound in \Cref{thm:Occam}. What are the advantages and when are they most prominent? 
\end{enumerate}
\end{exercise}

\begin{exercise}[\textit{The Airline Question}]~
\begin{enumerate}
	\item An airline knows that any person making a reservation on a flight will not show up with probability of 0.05 (5 percent). They introduce a policy to sell 100 tickets for a flight that can hold only 99 passengers. Bound the probability that the number of people that show up for a flight will be larger than the number of seats (assuming they show up independently).
	
	\item An airline has collected an i.i.d.\ sample of 10000 flight reservations and figured out that in this sample 5 percent of passengers who made a reservation did not show up for the flight. They introduce a policy to sell 100 tickets for a flight that can hold only 99 passengers. Bound the probability of observing such sample and getting a flight overbooked. 
	
	There are multiple ways to approach this question. We will guide you through two options. You are asked to solve the question in both ways.
	
	\begin{enumerate}
	    \item Let $p$ be the true probability of showing up for a flight (remember that $p$ is unknown). In the first approach we consider two events: the first is that in the sample of 10000 passengers, where each passenger shows up with probability $p$, we observe 95\% of show-ups. The second event is that in the sample of 100 passengers, where each passenger shows up with probability $p$, everybody shows up. Note that these two events are independent. Bound the probability that they happen simultaneously assuming that $p$ is known. And then find the worst case $p$ (you can do it numerically). With a simple approach you can get a bound of around 0.61. If you are careful and use the right bounds you can get down to around 0.0068.
	
	    It is advised to visualize the problem (the $[0,1]$ interval with 0.95 point for the 95\% show-ups and 1 for the 100\% show-ups and $p$ somewhere in $[0,1]$). This should help you understand the problem, understand where the worst case $p$ should be, and understand what direction of inequalities you need.
	
	    Attention: This is a frequentist rather than a Bayesian question. In case you are familiar with the Bayesian approach, it cannot be applied here, because we do not provide a prior on $p$. In case you are unfamiliar with the Bayesian approach, you can safely ignore this comment.
	    
	    \item The second approach considers an alternative way of generating the two samples, using the same idea as in the proof of the VC-bound. Consider the following process of generating the two samples:
        \begin{enumerate}
	        \item We sample 10100 passenger show up events independently at random according to an unknown distribution $p$.
	        \item We then split them into 10000 passengers in the collected sample and 100 passengers booked for the 99-seats flight.
        \end{enumerate}
        Bound the probability of observing a sample of 10000 with 95\% show ups and a 99-seats flight with all 100 passengers showing up by following the above sampling protocol. If you do things right, you can get a bound of about 0.0062 (there may be some variations depending on how exactly you do the calculation).
	\end{enumerate}
\end{enumerate}    
\end{exercise}

\begin{exercise}[\textit{The Growth Function}]~
\begin{enumerate}
	\item Let ${\cal H}$ be a finite hypothesis set with $|{\cal H}| = M$ hypotheses. Prove that $m_{\cal H}(n) \leq \min\lrc{M, 2^n}$.
    \item Let $\mathcal{H}$ be a hypothesis space with 2 hypotheses (i.e., $|\mathcal{H}| = 2$). Prove that $m_{\cal H}(n) = 2$. (Pay attention that you are asked to prove an equality, $m_{\cal H}(n) = 2$, not an inequality.)

	\item\label{pnt:square} Prove that $m_{\cal H}(2n) \leq m_{\cal H}(n)^2$.
\end{enumerate}    
\end{exercise}

\begin{exercise}[\textit{The VC-dimension}]~
\label{ex:VC-dim}
\begin{enumerate}
	\item Let ${\cal H}$ be a finite hypothesis set with $|{\cal H}| = M$ hypotheses. Bound the VC-dimension of ${\cal H}$. 
	\item Let $\mathcal{H}$ be a hypothesis space with 2 hypotheses (i.e., $|\mathcal{H}| = 2$). Prove that $\dVC(\HH) = 1$. (Pay attention that you are asked to prove an equality, not an inequality.)
	\item\label{pnt:circle} Let ${\cal H}_+$ be the class of ``positive'' circles in $\R^2$ (each $h \in {\cal H}_+$ is defined by the center of the circle $c \in \R^2$ and its radius $r \in \R$; all points inside the circle are labeled positively and outside negatively). Prove that $d_{VC}({\cal H}_+) \geq 3$.
	\item Let ${\cal H} = {\cal H}_+ \cup {\cal H}_-$ be the class of ``positive'' and ``negative'' circles in $\R^2$ (the ``negative'' circles are negative inside and positive outside). Prove that $d_{VC}({\cal H}) \geq 4$.
	\item \textbf{\textit{Optional question (0 points)}} Prove the matching upper bounds $d_{VC}({\cal H}_+) \leq 3$ and $d_{VC}({\cal H}) \leq 4$. [Doing this formally is not easy, but will earn you extra honor.]
	\item What is the VC-dimension of the hypothesis space $\mathcal{H}_d$ of binary decision trees of depth $d$?
	\item What is the VC-dimension of the hypothesis space $\mathcal{H}$ of binary decision trees of unlimited depth?
\end{enumerate}
\end{exercise}
\begin{exercise}[\textit{Steps in the Proof of the VC Bound}]~
\label{ex:growth-bound}
\begin{enumerate}
    \item Prove \Cref{lem:growth-bound}. (Hint: use induction.)
	\item Verify that \Cref{thm:Growth}, \Cref{thm:Sauer}, and \Cref{lem:growth-bound} together yield \Cref{thm:VC}.
\end{enumerate}
\end{exercise}
\begin{exercise}[\textit{The VC bound}]~
\begin{enumerate}
	\item What should be the relation between $\dVC(\HH)$ and $n$ in the VC generalization bound in \Cref{thm:VC} in order for the bound to be non-trivial? [A bound on the loss that is greater than or equal to 1 is trivial, because we know that the loss is always bounded by 1. You do not have to make an exact calculation, giving an order of magnitude is sufficient.]
	\item In the case of a finite hypothesis space, $|{\cal H}| = M$, compare the generalization bound that you can obtain with \Cref{thm:VC} with the generalization bound in \Cref{thm:Finite}. In what situations which of the two bounds is tighter?
	
	\item How many samples do you need in order to ensure that the empirical loss of a linear classifier selected out of a set of linear classifiers in $\R^{10}$ does not underestimate the expected loss by more than 0.01 with 99\% confidence?

	Clarifications: (1) you are allowed to use the fact that the VC-dimension of general separating hyperplanes in $\R^d$ (not necessarily passing through the origin) is $d+1$, see \citet[Exercise 2.4]{AML12}; (2) solving the question analytically is a bit tricky, you are allowed to provide a numerical solution. In either case (numerical or analytical solution), please, explain clearly in your report what you did.
	\item You have a sample of 100,000 points and you have managed to find a linear separator that achieves $\hat L_{\texttt{FAT}}(h,S) = 0.01$ with a margin of 0.1. Provide a bound on its expected loss that holds with probability of 99\%. The input space is assumed to be within the unit ball and the hypothesis space is the space of linear separators.
	\item \textbf{The fine details of the lower bound.} We have shown that if a hypothesis space $\mathcal{H}$ has an infinite VC-dimension, it is possible to construct a worst-case data distribution that will lead to overfitting, i.e., with probability at least $\frac{1}{8}$ it will be possible to find a hypothesis for which $L(h) \geq \hat L(h,S) + \frac{1}{8}$. But does it mean that hypothesis spaces with infinite VC-dimension are always deemed to overfit? Well, the answer is that it depends on the data distribution. If the data distribution is not the worst-case for $\HH$, there may still be hope.

    Construct a data distribution $p(X,Y)$ and a hypothesis space $\HH$ with infinite VC-dimension, such that for any sample $S$ of more than 100 points with probability at least 0.95 we will have $L(h) \leq \hat L(h,S) + 0.01$ for all $h$ in $\HH$.
    
    \emph{Hint:} this can be achieved with an extremely simple example.
\end{enumerate}
\end{exercise}

\begin{exercise}[\textit{Occam's $\kl$-razor vs.\ PAC-Bayes-$\kl$}] In this question we compare Occam's $\kl$-razor inequality with the PAC-Bayes-kl inequality.
\begin{enumerate}
\item Prove the following theorem, which extends \Cref{thm:kl-Occam} from \Cref{ex:Occam-kl} to soft selection.
\begin{theorem}[Occam's $\kl$-razor inequality for soft selection]
\label{thm:kl-Occam-soft}
Under the conditions of \Cref{thm:kl-Occam}
\[
\P[\exists \rho: \kl\lr{\E_\rho\lrs{\hat L(h,S)}\middle\|\E_\rho\lrs{L(h)}} \geq \frac{\E_\rho\lrs{\ln \frac{1}{\pi(h)}} + \ln\frac{1}{\delta}}{n}] \leq \delta.
\]  
\end{theorem}

\item Compare \Cref{thm:kl-Occam-soft} to PAC-Bayes-kl inequality in \Cref{thm:PBkl}. What are the advantages of PAC-Bayes-kl and what are the disadvantages?
\end{enumerate}
\end{exercise}

\begin{exercise}[\textit{PAC-Bayesian Aggregation}]
In this question you are asked to reproduce an experiment from \citet[Section 6, Figure 2]{TIWS17} (the paper is an outcome of a master project). Figure 2 corresponds to ``the second experiment'' in Section 6 of the paper ``Experimental Results''. You are only asked to reproduce the experiment for the first dataset, Ionosphere, which you can download from the UCI repository\footnote{The dataset can be downloaded from: \url{https://archive.ics.uci.edu/ml/datasets/Ionosphere}} \citep{UCI}. You are allowed to use any SVM solver you choose. Please, document carefully what you do and clearly annotate your graphs, including legend and axis labels.

\paragraph{Comments:}
\begin{enumerate}
	\item Assuming you have read Sections~\ref{sec:PAC-Bayes} and \ref{sec:PAC-Bayes-Ensembles}, it should be sufficient to read only the ``Experimental Results'' section of the paper in order to reproduce the experiment, but you are of course welcome to read the full article.
	\item Theorem~6 in the paper corresponds to our \Cref{thm:PAC-Bayes-aggregation}.
	\item Ideally, you should repeat the experiment several times, say 10, and report the average + some form of deviation, e.g. standard deviation or quantiles, over the repetitions. We have committed a sin by not doing it in the paper. We encourage you to make a proper experiment, but in order to save time you are allowed to repeat the sin (we will not take points for that). Please, do not do it in real papers.
    \item Pay attention that you are required to find $\rho$ through alternating minimization of the bound in Theorem~6 of the paper (\Cref{thm:PAC-Bayes-aggregation} in this text), but then you report the loss of predictions by the $\rho$-weighted majority vote (defined in \eqref{eq:MVrho}) rather than the loss of the randomized classifier defined by $\rho$. A simple bound on the loss of the weighted majority vote is twice the bound on the loss of the randomized classifier (\Cref{thm:first-order}), although in practice weighted majority vote typically performs better than the randomized classifier. The paper reports the bound on the loss of the randomized classifier, which is, strictly speaking, incorrect, because the bound on the loss of the weighted majority vote is twice as large, but you are allowed to do the same.
\end{enumerate}

\paragraph{Hint:} Direct computation of the update rule for $\rho$,
\[
\rho(h) = \frac{\pi(h) e^{-\lambda (n-r) \hat L^{\mathrm{val}}(h,S)}}{\sum_{h'} \pi(h') e^{-\lambda (n-r) \hat L^{\mathrm{val}}(h',S)}},
\]
is numerically unstable, since for large $n-r$ it leads to division of zero by zero. A way to fix the problem is to normalize by $\displaystyle e^{-\lambda (n-r) \hat L^{\mathrm{val}}_{\mathrm{min}}}$, where $\hat L^{\mathrm{val}}_{\mathrm{min}} = \min_h \hat L^{\mathrm{val}}(h,S)$. This leads to
\[
\rho(h) = \frac{\pi(h) e^{-\lambda (n-r) \hat L^{\mathrm{val}}(h,S)}}{\sum_{h'} \pi(h') e^{-\lambda (n-r) \hat L^{\mathrm{val}}(h',S)}}
= \frac{\pi(h) e^{-\lambda (n-r) \lr{\hat L^{\mathrm{val}(h,S)} - \hat L^{\mathrm{val}}_{\mathrm{min}}}}}{\sum_{h'} \pi(h') e^{-\lambda (n-r) \lr{\hat L^{\mathrm{val}}(h',S) - \hat L^{\mathrm{val}}_{\mathrm{min}}}}}.
\]
Calculation of the latter expression for $\rho(h)$ does not lead to numerical instability problems.

\paragraph{Optional Add-on}
Repeat the experiment with the tandem bound on the weighted majority vote from \Cref{thm:tandem-lambda}, which corresponds to \citet[Theorem 9]{MLIS20}. You can find the details of optimization procedure for the bound in \citet[Appendix G]{MLIS20}. Tandem losses should be evaluated on overlaps of validation sets and $n$ in the bounds should be replaced with the minimal overlap size for all pairs of hypotheses. Compare the results to the first order bound.
\end{exercise}

\begin{exercise}[\textit{Majority Vote}]
In this question we illustrate a few properties of the majority vote. Let $\MV$ denote a uniformly weighted majority vote.

\begin{enumerate}
	\item Design an example of $\HH$ and decision space $\X$, where $L(\MV) = 0$ and $L(h) \geq \frac{1}{3}$ for all $h$. (Hint: three hypotheses and $|\X| = 3$ is sufficient.)
	\item Design an example of $\HH$ and $\X$, where $L(\MV) > L(h)$ for all $h$.
	\begin{enumerate}
		\item Optional: design an example, where $L(\MV) \xrightarrow[|\HH| \to \infty]{} 2 \max_h L(h)$.
	\end{enumerate}
	\item Let $\HH$ be a hypothesis space, such that $|\HH| = M$ and all $h\in \HH$ have the same expected error, $L(h) = \frac{1}{2} - \varepsilon$ for $\varepsilon > 0$, and that the hypotheses in $\HH$ make independent errors. Prove that $L(\MV) \xrightarrow[|\HH| \to \infty]{} 0$. (In words: derive a bound for $L(\MV)$ and show that as $M$ grows the bound converges to zero, even though $L(h)$ can be almost as bad as $1/2$.)
	
\end{enumerate}

\paragraph{Bottom line:} \emph{If} the errors are independent \emph{and} $L(h) < \frac{1}{2}$ for all $h \in \HH$, the majority vote improves over individual classifiers. However, if $L(h) > \frac{1}{2}$ for some $h$ the errors may get amplified, and if there is correlation it may play in either direction, depending on whether $L(h)$ is above or below $\frac{1}{2}$ and whether it is correlation or anti-correlation.    
\end{exercise}

\begin{exercise}[\textit{PAC-Bayes-Unexpected-Bernstein}]~
\paragraph{Background} The $\kl$ and PAC-Bayes-$\kl$ inequalities that we have studied in the course work well for binary random variables (the zero-one loss), but, even though they apply to any random variables bounded in the $[0,1]$ interval, they are not necessarily a good choice if a random variable is non-binary and has a high probability mass inside the interval, because the $\kl$ inequality does not exploit small variance. For example, if you have a sample of Bernoulli random variables taking values $\{0,1\}$ with probability half-half, and you have another sample of non-Bernoulli random variables from a distribution, which is concentrated on $\frac{1}{2}$ (i.e., the random variables always take the value $\frac{1}{2}$), the $\kl$ bound on the expectation will be the same in both cases, because it is only based on the empirical average $\hat p_n$, even though in the second case the random variables are much more concentrated than in the first. 

Non-binary random variables occur, for example, if the loss of false positives and false negatives is asymmetric; in learning with abstention, where an algorithm is occasionally allowed to abstain from prediction and pay an abstention cost $c\in(0,\frac{1}{2})$; in working with continuous loss functions, such as the square or the absolute loss (although the Unexpected Bernstein inequality you will derive in this question still requires that the loss is one-side bounded); and many other problems \citep{WS22}.

In this question you will derive a concentration of measure inequality belonging to the family of Bernstein's inequalities, which exploit small variance to provide tighter concentration guarantees. 

\paragraph{Guidance} The question is built step-by-step, and if you fail in one of the steps you can still proceed to the next, because the outcomes of the intermediate steps are given. While it is possible to find alternative derivations of the inequality in the literature, you are asked to follow the steps.

\begin{enumerate}
	\item\label{pnt:1} Let $Z \leq 1$ be a random variable. Show that for any $\lambda \in [0,\frac{1}{2}]$:
	\[
	\E[e^{-\lambda Z - \lambda^2 Z^2}] \leq e^{-\lambda \E[Z]}.
	\]
  	Point out where you are using the assumption that $\lambda \in [0,\frac{1}{2}]$ and where you are using the assumption that $Z \leq 1$.
	
	Hint: the following two inequalities are helpful for the proof. For any $z \geq -\frac{1}{2}$ we have $z-z^2\leq \ln(1+z)$ \citep[Lemma 1]{CBMS07}. And for any $z$, we have $1 + z \leq e^z$.
	\item Prove that for $Z\leq 1$ and $\lambda \in [0,\frac{1}{2}]$,
	\[
	\E[e^{\lambda \lr{\E[Z] - Z} - \lambda^2 Z^2}] \leq 1.
	\]
	\item Let $Z_1,\dots,Z_n$ be independent random variables upper bounded by 1. Show that for any $\lambda \in [0,\frac{1}{2}]$
	\[
	\E[e^{\lambda \sum_{i=1}^n \lr{\E[Z_i] -  Z_i} - \lambda^2 \sum_{i=1}^n Z_i^2}] \leq 1.
	\]
	\item\label{pnt:4} Let $Z_1,\dots,Z_n$ be independent random variables upper bounded by 1. Show that for any $\lambda \in (0,\frac{1}{2}]$
	\[
	\P[\E[\frac{1}{n}\sum_{i=1}^n Z_i] \geq \frac{1}{n}\sum_{i=1}^n Z_i +  \frac{\lambda}{n}\sum_{i=1}^n Z_i^2 + \frac{\ln \frac{1}{\delta}}{\lambda n}] \leq \delta.
	\]
    \item {[Unexpected Bernstein inequality]} 
    Explanation: the right hand side of the inequality inside the probability above is minimized by $\lambda^*(Z_1,\dots,Z_n) = \min\lrc{\frac{1}{2}, \sqrt{\frac{\ln \frac{1}{\delta}}{\sum_{i=1}^n Z_i^2}}}$, but we cannot plug $\lambda^*(Z_1,\dots,Z_n)$ into the bound, because it depends on the sample $Z_1,\dots,Z_n$, and if you trace the proof back to Point~\ref{pnt:1}, it assumes that $\lambda$ is independent of the sample. And, while the bound in Point~\ref{pnt:4} holds for any $\lambda$, it does not hold for all $\lambda$ simultaneously. What you will do instead is take a grid of $\lambda$ values and a union bound over the grid, and select $\lambda$ from the grid, which minimizes the bound.

    Your task: Let $\Lambda = \{\lambda_1,\dots,\lambda_k\}$ be a grid of $k$ values of $\lambda$, such that $\lambda_i\in(0,\frac{1}{2}]$ for all $i$. Prove that:
    \[
	\P[\E[\frac{1}{n}\sum_{i=1}^n Z_i] \geq \frac{1}{n}\sum_{i=1}^n Z_i + \min_{\lambda\in\Lambda} \lr{\frac{\lambda}{n}\sum_{i=1}^n Z_i^2 + \frac{\ln \frac{k}{\delta}}{\lambda n}}] \leq \delta.
	\]
    We will call the above inequality an Unexpected Bernstein inequality.
	\item {[Empirical comparison of the $\kl$ and Unexpected Bernstein inequalities.]} We compare the Unexpected Bernstein inequality with the $\kl$ inequality. Take a ternary random variable (a random variable taking three values) $Z \in \lrc{0,\frac{1}{2},1}$. Let $p_0=\P[Z=0]$, $p_{\frac{1}{2}}=\P[Z=\frac{1}{2}]$, and $p_1=\P[Z=1]$. Set $p_0=p_1=(1-p_{\frac{1}{2}})/2$, i.e., the probability of getting $Z=0$ or $Z=1$ is equal, and we have one parameter $p_\frac{1}{2}$, which controls the probability mass of the central value. We will compare the bounds as a function of $p_\frac{1}{2} \in [0,1]$. Let $p=\E[Z]$ (in the constructed example, for any value of $p_\frac{1}{2}$ we have $p=\frac{1}{2}$, because $p_0=p_1$). For each value of $p_\frac{1}{2}$ in a grid covering the $[0,1]$ interval draw a random sample $Z_1,\dots,Z_n$ from the distribution we have constructed and let $\hat p_n = \frac{1}{n} \sum_{i=1}^n Z_i$ and $\hat v_n = \frac{1}{n} \sum_{i=1}^n Z_i^2$. Generate a figure, where you plot the Unexpected Bernstein bound on $p - \hat p_n$ and the $\kl$ bound on $p-\hat p_n$ as a function of $p_\frac{1}{2}$ for $p_\frac{1}{2}\in[0,1]$. The Unexpected Bernstein bound on $p-\hat p_n$ is $\min_{\lambda\in\Lambda}\lr{\lambda \hat v_n + \frac{\ln\frac{k}{\delta}}{\lambda n}}$, and the $\kl$ bound on $p-\hat p_n$ is $\kl^{-1^+}\lr{\hat p_n,\frac{\ln\frac{1}{\delta}}{n}} - \hat p_n$; pay attention that in contrast to \Cref{ex:numerical-kl} we subtract the value of $\hat p_n$ after inversion of $\kl$ to get a bound on the difference $p-\hat p_n$ rather than on $p$. Take the following values for the comparison: $n=100$, $\delta = 0.05$, $|\Lambda| = k = \lceil \log_2(\sqrt{n/\ln(1/\delta)}/2) \rceil$, and $\Lambda = \{\frac{1}{2}, \frac{1}{2^2}, \dots, \frac{1}{2^k}\}$. Briefly comment on the result of empirical evaluation.

    Explanation: The $\kl$ inequality depends only on the empirical first moment of the sample, $\hat p_n = \frac{1}{n}\sum_{i=1}^n Z_i$ and, therefore, it is ``blind'' to the variance and cannot exploit it. The Unexpected Bernstein inequality depends on the empirical first and second moments of the sample, $\hat p_n$ and $\hat v_n=\frac{1}{n}\sum_{i=1}^n Z_i^2$. The second moment is directly linked to the variance, $\Var[Z] = \E[Z^2] - \E[Z]^2$ and, therefore, the Unexpected Bernstein inequality is able to exploit small variance.

	\item\label{it:main} Let $S$ be an i.i.d.\ sample, $h$ a prediction rule, and $\ell(y',y)$ a loss function upper bounded by 1. Define $\hat V(h,S) = \frac{1}{n}\sum_{i=1}^n \ell(h(X_i),Y_i)^2$ and $L(h)$ and $\hat L(h,S)$ as usual. Show that for any $\lambda \in [0,\frac{1}{2}]$ we have
	\[
	\E[e^{n\lr{\lambda\lr{L(h) - \hat L(h,S)} - \lambda^2 \hat V(h,S)}}] \leq 1.
	\]
     \item Let $S$ and $\ell$ be as before. Let $\HH$ be a set of prediction rules, let $\pi$ be a distribution on $\HH$ that is independent of $S$. Show that for any $\lambda \in (0,\frac{1}{2}]$
	\[
    \P[\exists \rho:\E_\rho\lrs{L(h)} \geq \E_\rho\lrs{\hat L(h,S)} + \lambda \E_\rho\lrs{\hat V(h,S)} + \frac{\KL(\rho\|\pi) + \ln \frac{1}{\delta}}{n \lambda}] \leq \delta,
	\]
    where $\rho$ denotes a distribution on $\HH$.
    
	Hint: Take $f(h,S) = n\lr{\lambda\lr{L(h) - \hat L(h,S)} - \lambda^2 \hat V(h,S)}$ and use PAC-Bayes bounding procedure and the result from the previous point.
	
	Side remark: Note that the ``optimal'' value of $\lambda$ (the one that minimizes the right hand side of the inequality inside the probability) depends on the data, and that the bound does not hold for all $\lambda$ simultaneously. In the next step we resolve this issue by taking a grid of $\lambda$ values and using the best value in the grid.
    \item {[PAC-Bayes-Unexpected-Bernstein Inequality.]} Let $\Lambda = \{\lambda_1,\dots,\lambda_k\}$ be a grid of $k$ values of $\lambda$, such that $\lambda_i\in(0,\frac{1}{2}]$ for all $i$. Let $S$, $\ell$, and $\pi$ be as before. Prove that:
	\[
    \P[\exists \rho:\E_\rho\lrs{L(h)} \geq \E_\rho\lrs{\hat L(h,S)} + \min_{\lambda \in \Lambda} \lr{\lambda \E_\rho\lrs{\hat V(h,S)} + \frac{\KL(\rho\|\pi) + \ln \frac{k}{\delta}}{n \lambda}}] \leq \delta.
	\]
\end{enumerate}    
\end{exercise}

\chapter{Supervised Learning - Regression}
\label{ch:Regression}

In this chapter we consider the regression problem, which is another special case of supervised learning with ${\cal X} = \R^d$ and ${\cal Y} = \R$. 

\section{Linear Least Squares}

Linear regression with square loss $\ell(Y',Y) = (Y' - Y)^2$ is also known as linear least squares. Let $S = \lrc{(\x_1,y_1),\dots,(\x_n,y_n)}$ be our sample. We are looking for a prediction rule of a form $h(\x) = \w^T \x$, where $\w^T \x$ is the dot-product (also known as the inner product) between a vector $\w \in \R^d$ and a data point $\x \in \R^d$. We will use $\w$ to denote the above prediction rule. Let $\X \in \R^{n\times d}$ be a matrix holding $\x_1^T,\dots,\x_n^T$ as its rows 
\[
\X = \lr{\begin{array}{c}\text{ --- }\x_1^T\text{ --- }\\\vdots\\\text{ --- }\x_n^T\text{ --- }\end{array}}
\]
and let $\y = (y_1,\dots,y_n)^T$ be the vector of labels. We are looking for $\w$ that minimizes the empirical loss $\hat L(\w,S) = \sum_{i=1}^n \ell(\w^T \x_i, y_i) = \sum_{i=1}^n \lr{\w^T \x_i - y_i}^2 = \|\X \w - \y\|^2$.

When the number of constraints $n$ (the number of points in $S$) is larger than the number of unknowns $d$ (the number of entries in $\w$), most often the linear system $\X\w=\y$ has no solutions (unless $\y$ by chance falls in the linear span of the columns of $\X$). Therefore, we are looking for the best approximation of $\y$ by a linear combination of the columns of $\X$, which means that we are looking for a \emph{projection} of $\y$ onto the column space of $\X$. There are two ways to define projections, analytical and algebraic, which lead to two ways of solving the problem. In the analytical formulation the projection is a point of a form $\X\w$ that has minimal distance to $\y$. In the algebraic formulation the projection is a vector $\X\w$ that is perpendicular to the remainder $\y - \X\w$. We present both ways in detail below.

\subsection{Analytical Approach}

We are looking for
\[
\min_{\w} \|\X \w - \y\|^2 = \min_{\w} (\X \w - \y)^T (\X \w - \y) = \min_{\w} \w^T \X^T \X \w - 2 \y^T \X \w + \y^T \y.
\]
By taking a derivative of the above and equating it to zero we have\footnote{See Appendix~\ref{app:calculus} for details on calculation of derivatives [gradients] of multidimensional functions.}
\[
\frac{d (\w^T \X^T \X \w - 2 \y^T \X \w + \y^T \y)}{d\w} = 2\X^T \X \w - 2 \X^T \y = 0.
\]
Which gives
\[
\X^T \X \w = \X^T \y.
\]
If we assume that the \emph{columns} of $\X$ are linearly independent ($dim(\X) = d$) then $\X^T \X \in \R^{d \times d}$ is invertible (see \Cref{app:lin-alg}) and we obtain
\[
\w = (\X^T \X)^{-1} \X^T \y.
\]

\subsection{Algebraic Approach - Fast Track}

The projection $\X\w$ is a vector that is orthogonal to the remainder $\y - \X \w$ (so that $\y$ is a sum of the projection and the remainder, $\y = \X \w + (\y - \X\w)$, and there is a right angle between the two). Two vectors are orthogonal if and only if their inner product is zero. Thus, we are looking for $\w$ that satisfies
\[
\lr{\X\w}^T\lr{\y - \X\w} = 0,
\]
which is equivalent to $\w^T \X^T\lr{\y - \X \w} = 0$. It is sufficient to find $\w$ that satisfies $\X^T\lr{\y - \X \w} = 0$ to solve this equation, which is equivalent to $\X^T \X \w = \X^T \y$. By multiplying both sides by $\lr{\X^T\X}^{-1}$ (which is defined, since the columns are linearly independent) we obtain a solution $\w = \lr{\X^T\X}^{-1}\X^T\y$.

This solution is, actually, unique due to independence of the columns of $\X$. Assume there is another solution $\w'$, such that $\X\w' = \y$. Then $\X\w - \X\w' = \X(\w-\w') = 0$, but since the columns of $\X$ are linearly independent the only their linear combination that yields zero is the zero vector, meaning that $\w-\w'=0$ and $\w=\w'$.

\subsection{Algebraic Approach - Complete Picture}

\begin{figure}%
\includegraphics[width=\columnwidth]{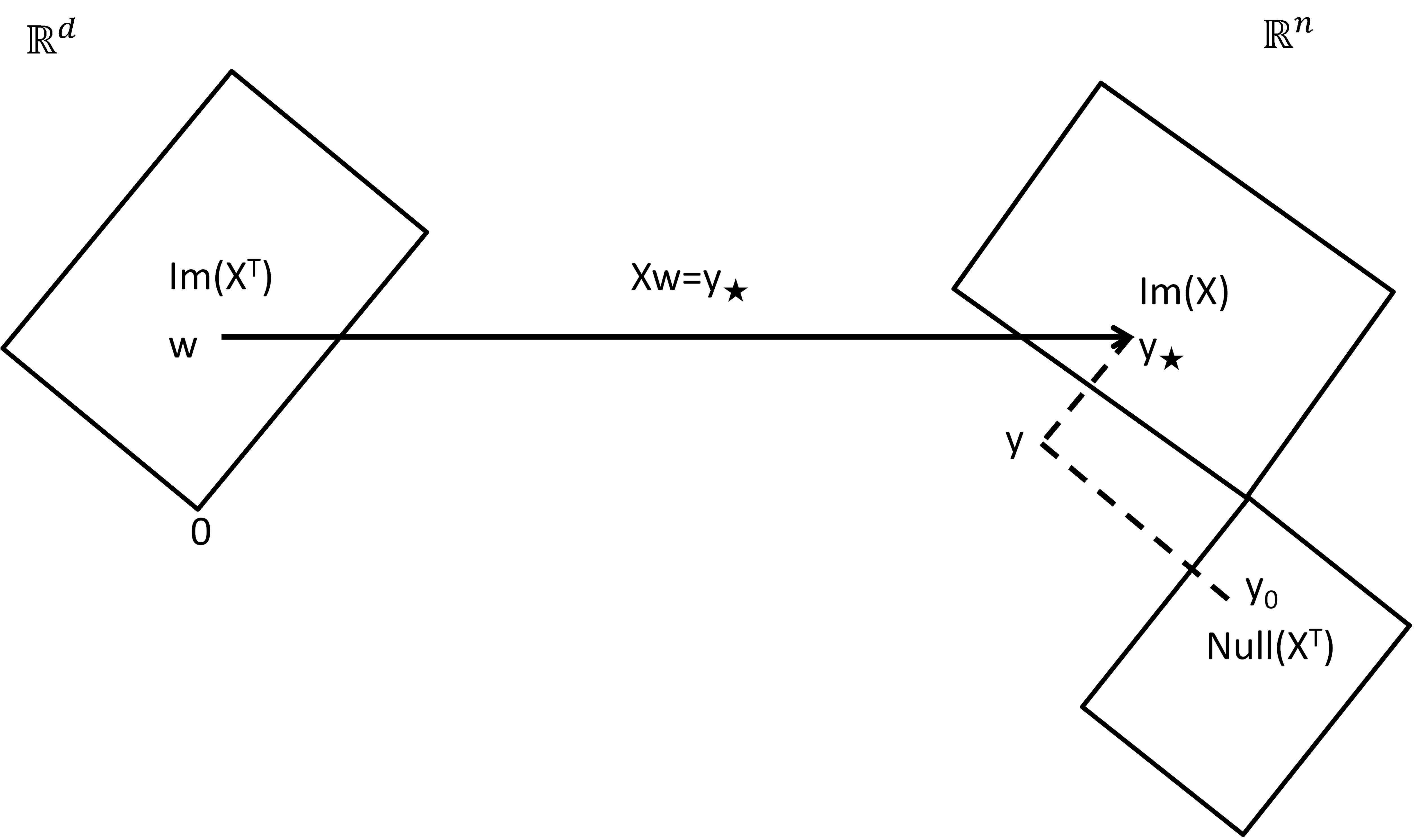}%
\caption{\textbf{Illustration of algebraic solution of linear least squares.}}%
\label{fig:LLS}%
\end{figure}

Linear Least Squares is a great opportunity to revisit a number of basic concepts from linear algebra. Once the complete picture is understood, the algebraic solution of the problem is just one line. We refer the reader to \Cref{app:lin-alg} for a quick review of basic concepts from linear algebra. We are looking for a solution of $\X \w = \y$, where $\y$ (most likely) lies outside of the column space of $\X$ and the equation has no solution. Therefore, the best we can do is to solve $\X \w = \y_\star$, where $\y_\star$ is a projection of $\y$ onto the column space of $\X$ (see Figure \ref{fig:LLS}). We assume that $dim(\X) = d$ and thus the matrix $\X^T \X$ is invertible. The projection $\y_\star$ is then given by $\y_\star = \X \lr{\X^T\X}^{-1}\X^T\y$, which means that the best we can do is to solve $\X \w = \y_\star = \X \lr{\X^T\X}^{-1}\X^T\y$ and the solution is $\w = \lr{\X^T\X}^{-1}\X^T\y$.

\subsection{Using Linear Least Squares for Learning Coefficients of Non-linear Models}

Linear Least Squares can be used for learning coefficients of non-linear models. For example, assume that we want to fit our data $S = \lrc{(x_1,y_1), \dots, (x_n,y_n)}$ (where both $x_i$-s and $y_i$-s are real numbers) with a polynomial of degree $d$. I.e., we want to have a model of a form $y = a_d x^d + a_{d-1} x^{d-1} + \dots + a_1 x + a_0$. All we have to do is to map our features $x_i$-s into feature vectors $x_i \to (x_i^d, x_i^{d-1},\dots,x_i,1)$ and apply linear least squares to the following system:
\[
\lr{\begin{array}{ccccc}x_1^d & x_1^{d-1} & \dots & x_1 & 1\\x_2^d & x_2^{d-1} & \dots & x_2 & 1\\&&\vdots&&\\x_n^d & x_n^{d-1} & \dots & x_n & 1\end{array}} \lr{\begin{array}{c}a_d\\a_{d-1}\\\vdots\\a_1\\a_0\end{array}} = \lr{\begin{array}{c}y_1\\y_2\\\vdots\\y_n\end{array}}
\]
to get the parameters vector $\lr{a_d,a_{d-1},\dots,a_1,a_0}^T$.

\chapter{Limitations and Pitfalls of the Classical Batch Learning}
\label{ch:Limitations}

This chapter is dedicated to discussion of limitations of the classical batch learning model that has been the focus of Chapters~\ref{ch:SupervisedLearning}, \ref{ch:Generalization}, and \ref{ch:Regression}. Some of the limitations are addressed by the online learning model discussed in \Cref{ch:Online}, but others are general limitations of learning based on selection.

\section{The i.i.d.\ Assumption}

The assumption that the data are i.i.d.\ and that new data come from the same distribution as the data used for training is behind everything discussed so far. (We depart from this assumption in \Cref{ch:Online}, where we introduce online learning.) As a consequence, violation of the i.i.d.\ assumption leads to break down of the guarantees derived in Chapters~\ref{ch:CoM} and \ref{ch:Generalization}. There are two ways in which the i.i.d.\ assumption can be violated. First, if the training data are not independent, the empirical error $\hat L(h,S)$ might not be converging to the expected error $L(h)$ at the same rate as in the theorems, or might not be converging at all, as in \Cref{ex:independence}. Second, if the new data are not coming from the same distribution as the training data, then the empirical estimates clearly do not represent the quantity we are interested in.

Note that the assumption that future data come from the same distribution as the training data implies that generalization guarantees derived in \Cref{ch:Generalization} are not applicable in situations, where deployment of a learning algorithm leads to feedback loops, namely, when the algorithm changes its application environment. For example, past data can be used to predict congestion level on a network link, but if the predictions are used to reroute traffic they will get invalidated.

\section{Overfitting}

The central theme of \Cref{ch:Generalization} was to derive tools to control the deviation of empirical estimates of prediction accuracy from the expected accuracy. The control is based on balancing two competing forces - \emph{concentration} and \emph{selection}. For example, for selection from a finite set of prediction rules $\HH$ with $|\HH|=M$ based on a sample of size $n$ we have
\[
\P[\exists h\in\HH: L(h) \geq \hat L(h,S) + \varepsilon] \leq \underbrace{M}_{
\text{selection}}\times \underbrace{e^{-2n\varepsilon^2}}_{\text{concentration}}.
\]
The power of concentration of each individual estimate depends on the number of points used to validate the models, whereas the power of selection depends on the richness of the class of prediction rules from which the selection is done. If the richness of selection grows (super-)exponentially with the number of points, the control over generalization may be lost, leading to overfitting. Two points are important to emphasize in this regard. 

\paragraph{``Big data'' is not a universal cure to overfitting} The abundance of data on its own does not prevent overfitting. If the amount of selection (measured in the relevant way) is (super-)exponential in the number of data points used to validate the prediction rules, overfitting may occur. In modern prediction models involving billions of parameters, the effective amount of selection may easily surpass the exponent of the number of points used for validation.

\paragraph{Internal and External selection} The second point is that selection may be \emph{internal} and \emph{external}. By internal selection we mean selection done within a particular algorithm, for example, selection of a separating hyperplane in linear separation. Since an algorithm is in direct control of the set of prediction rules $\HH$ it is selecting from, it can directly control overfitting. By external selection we mean selection happening outside of algorithms. For example, multiple research groups may apply various algorithms to a publicly available dataset, compare the results, and publish the best ones. In this case the same limited data (the publicly available dataset) are used to select from many more prediction rules than those used within any individual prediction algorithm. And this may easily lead to overfitting. In fact, many of the popular publicly available datasets are heavily overfitted. The only way to control overfitting in external selection is to turn it into internal selection. In other words, there should be a careful bookkeeping of the union of all the hypothesis classes that have been applied to a dataset and generalization bounds should be computed based on this union. Moreover, the selection of the hypothesis classes should be independent of the outcomes of preceding hypothesis classes on the validation set or, otherwise, fresh data are required for valid generalization guarantees.

\section{Human Perception of Uncertainty}

Humans are not very good at quantitative judgment of uncertainty \citep{Kah11}. Moreover, when presented with various studies they tend to ignore confidence information altogether (if it happened to be there in the first place) and focus on the average number. For example, a newspaper article might say that ``a study has shown that $X$ is $p\%$ better than $Y$'', but it might often omit the confidence interval (that would depend on the variation of outcomes and the number of subjects used in the study). Even if the confidence information is reported, it might be often ignored by the readers, who would keep in mind just the key number. Therefore, when reporting results it is important to report uncertainty information (confidence intervals), and preferably in a way that would make it hard to ignore the confidence information. For example, one may opt to plot a cloud of individual points or trajectories rather than the mean and standard deviation. Avoiding plotting the mean will make it harder to replace the outcomes with just one number, and would force the reader to give at least some consideration to uncertainty of the outcomes.

\section{Correlation $\neq$ Causation}

Machine learning provides a wealth of tools for correlation mining, and humans have a predisposition for making causal interpretations of correlation data \citep{Kah11}. Therefore, it is important to emphasize that correlation does not imply causation. For example, boxers typically do not wear glasses (a correlation), but this does not imply that taking boxing classes is likely to improve anyone's vision (a causal relation). Machine learning can reliably identify the correlation between boxing and reasonable vision based on data. Moreover, if someone practices boxing, it is a reliable predictor that they have reasonable vision, so correlations can be used for reliable predictions. But methods discussed so far assume that the data are i.i.d.\ and are not built to gauge the effect of interventions (taking boxing classes), which would violate the assumption that new data come from the same distribution as the training data. 

While the topic of causality is completely outside of the scope of this book, we note that online learning, discussed in \Cref{ch:Online}, is based on exploiting causal dependencies between actions and outcomes (rather than mere correlations). In this respect a difference between online learning and causality is that causality aims to understand the causal structure of the world, whereas online learning aims to identify actions (interventions) that lead to a desired outcomes. 

\chapter{Online Learning}
\label{ch:Online}

\begin{figure}%
\centering
\includegraphics[width=.8\columnwidth]{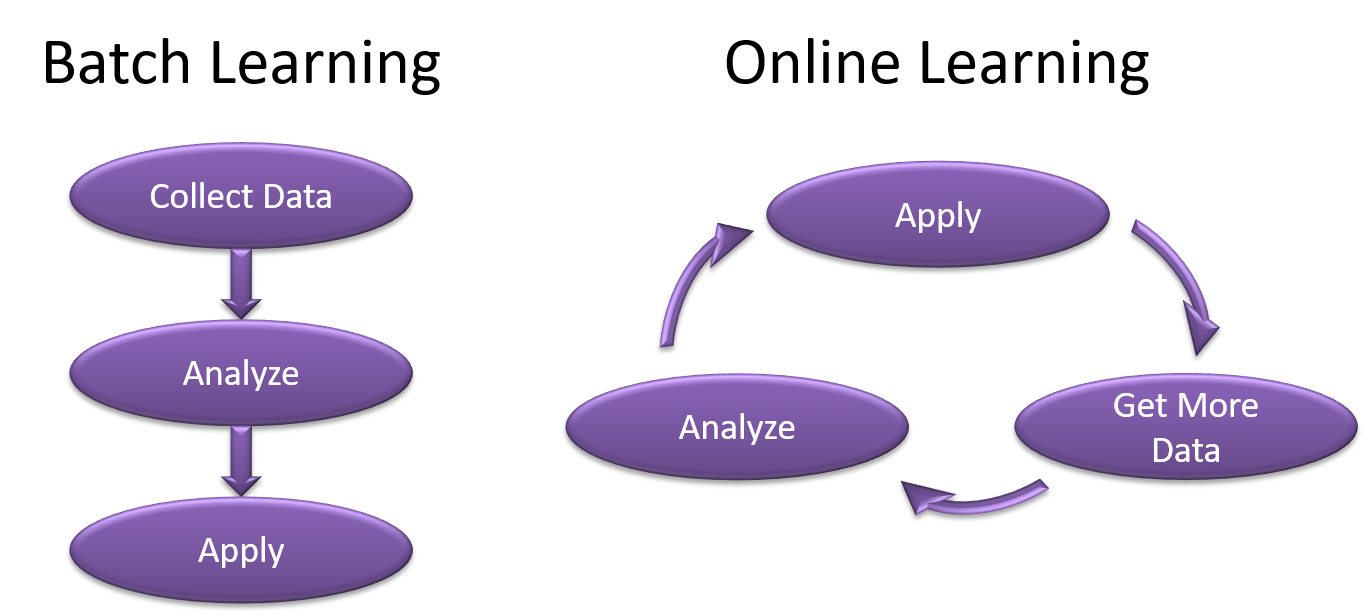}%
\caption{\textbf{Online learning vs.\ batch.}}%
\label{fig:online-vs-batch}%
\end{figure}

So far in the book has considered \emph{batch} learning. Batch learning starts with some training data, analyzes it, and then ``ships the result of the analysis into the world'' (see \Cref{fig:online-vs-batch}). ``The result'' can be a fixed classifier $h$, a distribution over classifiers $\rho$, or anything else, the important point is that it does not change from the moment the selection procedure is over. It takes no new information into account. This is also the reason why we had to assume that new samples come from the same distribution as the samples in the training set, because the classifier was not designed to adapt.

\emph{Online learning} is a learning framework, where data collection, analysis, and application of inferred knowledge are in a perpetual loop, see Figure~\ref{fig:online-vs-batch}. Examples of problems, which fit into this framework include:
\begin{itemize}
	\item Investment in the stock market.
	\item Online advertizing and personalization.
	\item Online routing.
	\item Games.
	\item Robotics.
	\item And so on ...
\end{itemize}
The recurrent nature of online learning problems makes them closely related to repeated games. They also borrow some of the terminology from the game theory, including calling the problems \emph{games} and every ``Act - Observe - Analyze'' cycle a \emph{game round}. In general, we may need online learning in the following scenarios:
\begin{itemize}
	\item Interactive learning: situations, where an algorithm continuously gets new information and taking it into account may improve the quality of future actions. For example, interaction with Internet users falls under this category -- it makes sense to adapt to user behavior as the algorithm proceeds from one user query to the next.
	\item Adversarial or game-theoretic settings, where we cannot assume that ``the future behaves similarly to the past''. For example, in spam filtering we cannot assume that new spam messages are generated from the same distribution as the old ones. Or, in playing chess we cannot assume that the moves of the opponent are sampled i.i.d..
    \item Intelligent data collection. To cite \citet{Tho33}, ``\emph{there can be no objection to the use of data, however meagre, as a guide to action required before more can be collected}''. Thompson was one of the pioneers of online learning and the theory of adaptive medical trials. In the context of adaptive medical trials the message is that it makes sense to look into the outcome of completed treatments before deciding on further treatments, as opposed to the more classical approach of A/B testing, where the size of treatment and control groups are decided upon before experimentation begins.\footnote{My other favourite quote on the topic is by \citet{Rob52}: ``Until recently, statistical theory has been restricted to the design and analysis of sampling experiments in which the size and composition of the samples are completely determined before the experimentation begins. The reasons for this are partly historical, dating back to the time when the statistician was consulted, if at all, only after the experiment was over, and partly intrinsic in the mathematical difficulty of working with anything but a fixed number of independent random variables. A major advance now appears to be in the making with the creation of a theory of the \emph{sequential design} of experiments, in which the size and composition of the samples are not fixed in advance but are functions of the observations themselves.''}
\end{itemize}
As with many other problems in computer science, having loops (as in \Cref{fig:online-vs-batch}) makes things much more challenging, but also much richer and more fun to work on. For example, online learning allows to treat adversarial environments, which is impossible to do in the batch setting.

\section{The Space of Online Learning Problems}

\begin{figure}%
\centering
\includegraphics[width=.8\columnwidth]{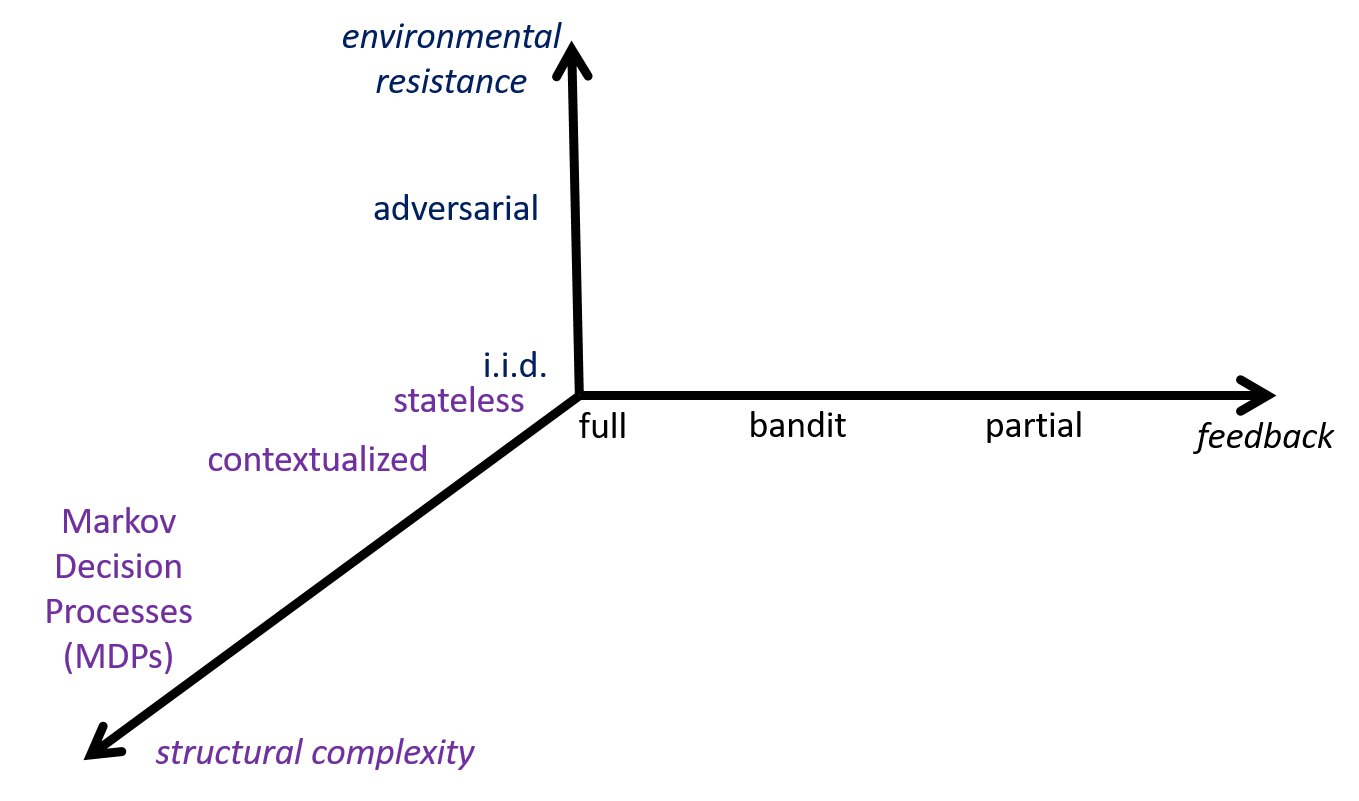}%
\caption{\textbf{The Space of Online Learning Problems.}}%
\label{fig:TheSpace}%
\end{figure}

Online learning problems are characterized by three major parameters:
\begin{enumerate}
	\item The amount of \emph{feedback} that the algorithm receives in every round of interaction with the environment.
	\item The \emph{environmental resistance} to the algorithm.
	\item The \emph{structural complexity} of a problem.
\end{enumerate}
Jointly they define \emph{the space of online learning problems}, see Figure~\ref{fig:TheSpace}. It is not really a space, but a convenient way to organize the material and get initial orientation in the zoo of online learning settings. We discuss the three axes of the space with some examples below.

\subsubsection*{Feedback}

Feedback refers to the amount of information that the algorithm receives in every round of interaction with the environment. The most basic forms of feedback are \emph{full information} and \emph{limited} (better known as \emph{bandit}\footnote{``The name derives from an imagined slot machine $\dots$. (Ordinary slot machines with one arm are one-armed bandits, since in the long run they are as effective as human
bandits in separating the victim from his money.)'' \citep{LR85}}) feedback. 

A classical example of a full information game is investment in the stock market. In every round of this game an algorithm distributes wealth over a set of stocks and the next day it observes the rates of all the stocks, which constitutes full information. With full information the algorithm can evaluate the quality of its own investment strategy, as well as any alternative investment strategy.

A classical example of a bandit feedback game are medical treatments. An algorithm has a set of \emph{actions} (in this case treatments), but it can only apply one treatment to a given patient. The algorithm observes the outcome of the applied treatment, but not of the other treatments, resulting in limited feedback. With limited feedback the algorithm only observes the quality of the selected strategy, but it gets no direct access to the quality of alternative strategies that could have been selected. Limited feedback leads to the \emph{exploration-exploitation trade-off}, which is the trade-mark signature of online learning. The essence of the exploration-exploitation trade-off is that in order to estimate the quality of actions the algorithm has to try them out (to explore). If it explores too little, it risks missing some good actions and end up performing suboptimally. However, exploration has a cost, because trying out suboptimal actions for too long is also undesirable. The goal is to balance exploration (trying new actions) with exploitation, which is taking actions, which are currently believed to be the optimal ones. The ``Act-Observe-Analyze'' cycle comes into play here, because unlike in batch learning the training set is not given, but is built by the algorithm for itself: if it does not try an action, it gets no data from it. 

There are many other problems that fall within the bandit feedback framework, with another popular example being online advertising. A simplistic way of modeling online advertising is by assuming that there is a pool of advertisements, but on every round of the game it is only allowed to show one advertisement to a user. Since the advertiser only observes feedback for the advertisement that has been presented, the problem can be formulated as an online learning problem with bandit feedback.

There are other feedback models, which we will only touch briefly. In the bandit feedback model the algorithm observes a noisy estimate of the quality of selected action, for example, whether an advertisement was clicked or not. In \emph{partial} feedback model studied under \emph{partial monitoring} the feedback has some relation to the action, but not necessarily its quality. For example, in dynamic pricing the seller only observes whether a proposed price was above or below the value of a product for a buyer, but not the value itself (the maximal price the buyer would be ready to pay for the product). Bandit feedback is a special case of partial feedback, where the observation is the value. Another example is \emph{dueling bandit} feedback, where the feedback is a relative preference over a pair of items rather than the absolute value of the items. For example, an answer to the question ``Do you prefer fish or chicken?'' is an example of dueling bandit feedback. Dueling bandit feedback model is used in information retrieval systems, since humans are much better in providing relative preferences rather than absolute utility values.

\subsubsection*{Environmental Resistance}

Environmental resistance is concerned with how much the environment resists to the algorithm. Two classical examples are i.i.d.\ (a.k.a. \emph{stochastic}) and \emph{adversarial} environments. An example of an i.i.d.\ environment is the weather. It has a high degree of uncertainty, but it does not play against the algorithm. Another example of an i.i.d.\ environment are outcomes of medical treatments. Here also there is uncertainty in the outcomes, but the patients are not playing against the algorithm. An example of an adversarial environment is spam filtering. Here the spammers are deliberately changing distribution of the spam messages in order to outplay the spam filtering algorithm. Another classical example of an adversarial environment is the stock market. Even though the stock market does not play directly against an individual investor (assuming the investments are small), it is not stationary, because if there would be regularity in the market it would be exploited by other investors and would be gone.

The environment may also be collaborative, for example, when several agents are jointly solving a common task. Yet another example are slowly changing environments, where the parameters of a distribution are slowly changing with time.

\subsubsection*{Structural Complexity}

In structural complexity we distinguish between \emph{stateless} problems, \emph{contextualized} problems (or problems with state), and \emph{Markov decision processes}. In stateless problems actions are taken without taking any additional information except the history of the outcomes into account. In contextualized problems in every round of the game the algorithm observes a context (or state) and takes an action within the observed context. An example of a context is a medical record of a patient or, in the advertising example, it could be the parameters of the user that opened a web page. 

\emph{Markov decision processes} (MDPs) are concerned with processes with evolving state. The difference between contextualized problems and Markov decision processes is that in the former the actions of the algorithm do not influence the next state, whereas in the latter they do. For example, subsequent treatments of the same patient are changing his or her state and, therefore, depend on each other. In contrast, in subsequent treatments of different patients treatment of one patient does not influence the state of the next patient and, thus, can be modeled as a contextualized problem.

Markov decision processes are studied within the field of \emph{reinforcement learning} (RL). There is no clear cut distinction between online learning and reinforcement learning, and one could be seen as a subfield of the other, or the other way around. But as a rule of thumb, problems involving evolution of states, such as Markov decision processes, are part of reinforcement learning, and problems that do not involve evolution of states are part of online learning.

\begin{figure}
    \centering
    \includegraphics[width=0.4\linewidth]{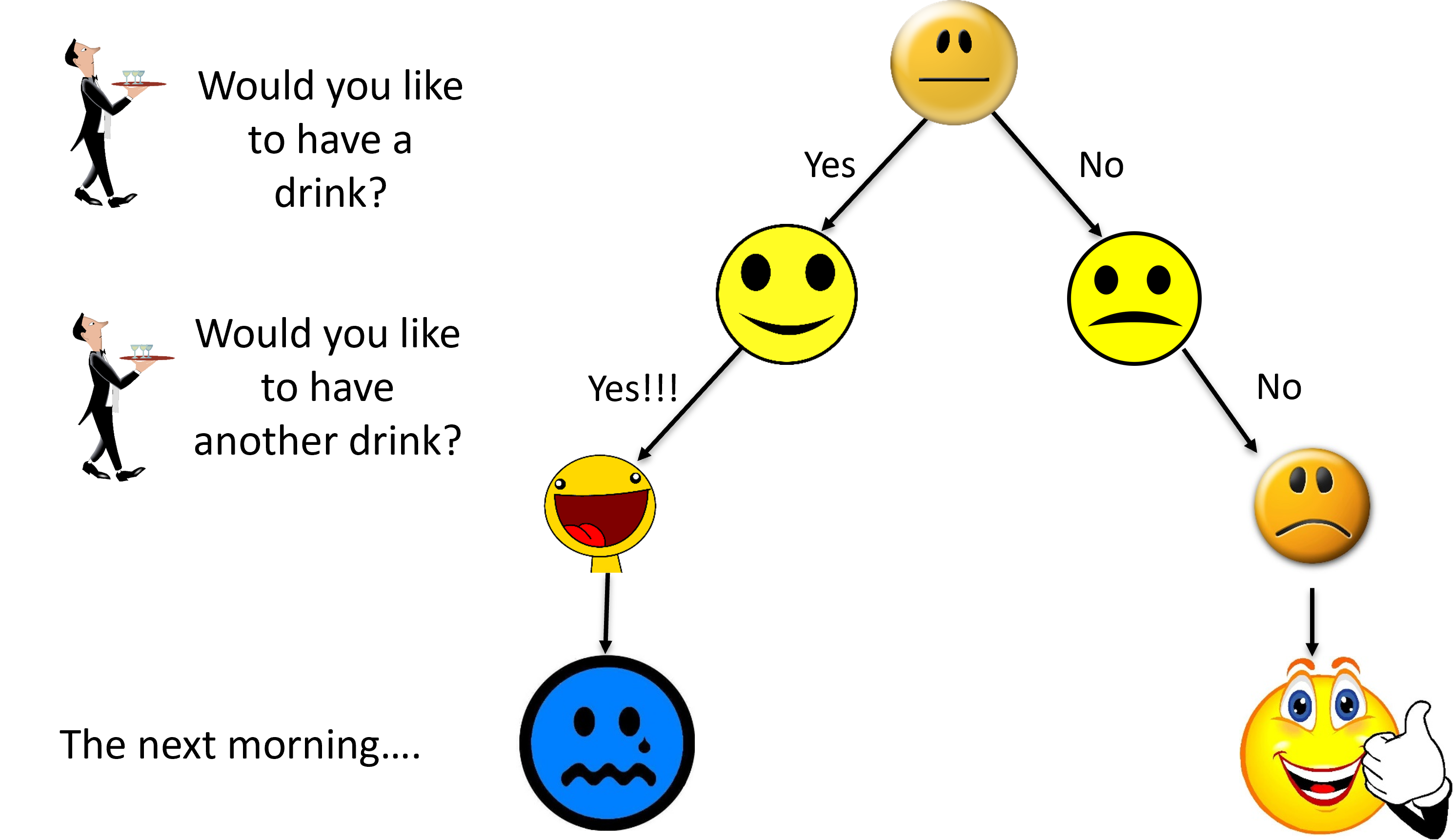} \hfill 
    \includegraphics[width=0.4\linewidth]{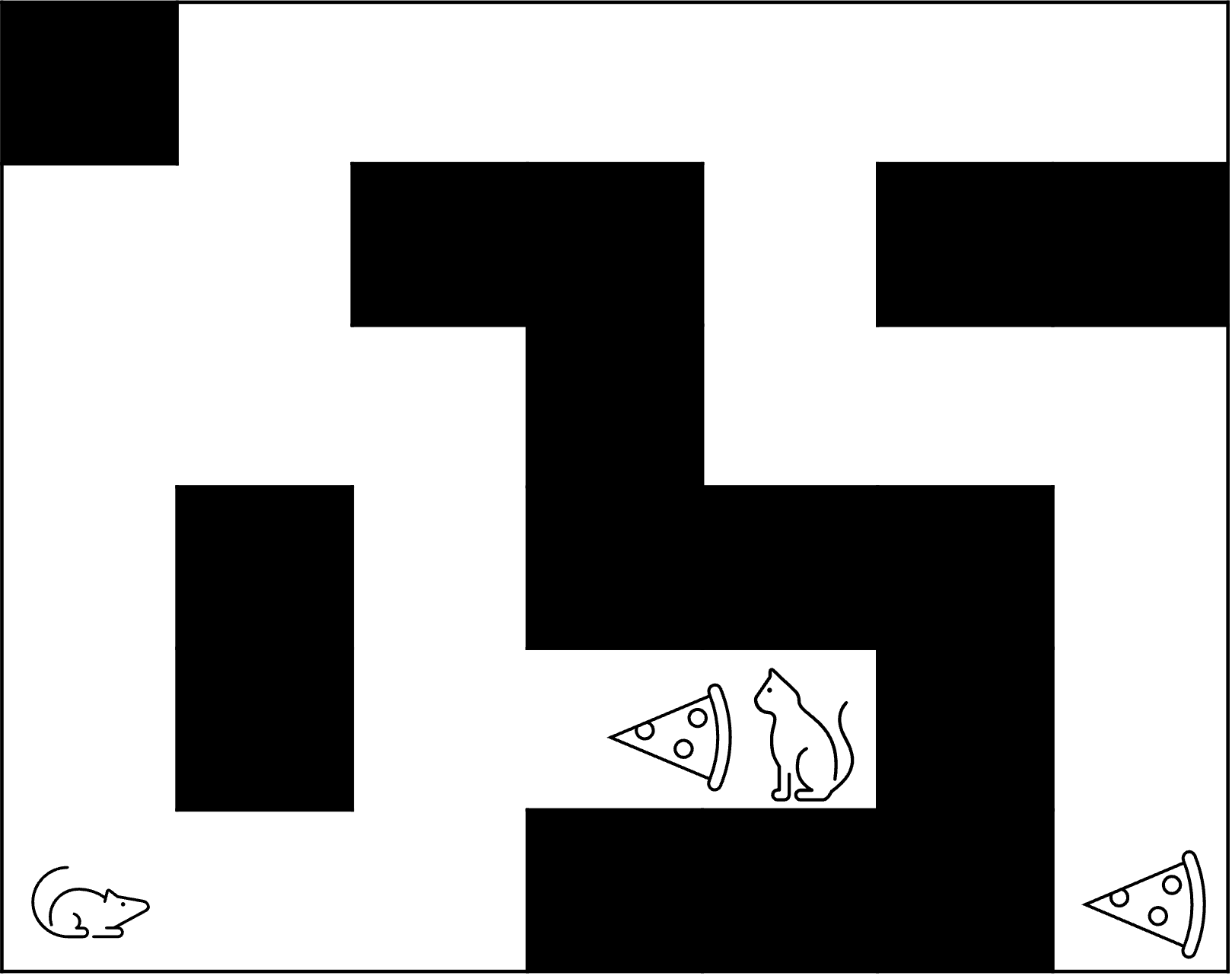}
    \caption{\textbf{Planning.} Even when the immediate outcomes are known, long-term goals require planning.}
    \label{fig:planning}
\end{figure}

When actions impact the state of the environment and the agent, they may have long-term consequences and, therefore, require \emph{planning}. For example, getting to the ``safe'' slice of pizza in \Cref{fig:planning} requires the mouse to plan a sequence of actions. Even if the immediate outcomes of every action in every state are known (there is no noise in execution of motor commands), there is still computational work to be done to infer the optimal action in each state. This is in contrast to online learning, where if the outcome of every action is known (e.g., the outcome of every treatment), inferring the best action is trivial. In many situations planning is combined with uncertainty estimation, for example, if the floor is slippery, the mouse might need to infer the relation between its actions and transitions between states. To summarize, online learning is primarily focused on uncertainty estimation, whereas reinforcement learning is focused on uncertainty estimation and planning, and the latter may be interesting and non-trivial even in absence of the former.

\noindent
\begin{center}
\begin{tabular}{|c|c|}
\hline
Online Learning & Reinforcement Learning\\
\hline
\hline
Uncertainty Estimation & Uncertainty Estimation\\
\hline
--- & Planning\\
\hline
\end{tabular}
\end{center}

There are many other online learning problems, which do not fit directly into Figure~\ref{fig:TheSpace}, but can still be discussed in terms of feedback, environmental resistance, and structural complexity. For example, in \emph{combinatorial bandits} the goal is to select a set of actions, potentially with some constraints, and the quality of the set is evaluated jointly. An instance of a combinatorial bandit problem is selection of a path in a graph, such as communication or transport network. In this case an action can be decomposed into sub-actions corresponding to selection of edges in the graph. The goal is to minimize the length of a path, which may correspond to the delay between the source and the target nodes. Various forms of feedback can be considered, including bandit feedback, where the total length of the path is observed; semi-bandit feedback, where the length of each of the selected edges is observed; cascading bandit feedback, where the lengths of the edges are observed in a sequence until a terminating node (e.g., a server that is down) or the target is reached; or a full information feedback, where the length of all edges is observed.

\subsubsection*{Summary}

To summarize, online and reinforcement learning introduce three new elements that we have not seen in batch learning: (1) incomplete feedback and \emph{exploration} to deal with it, (2) the ability to handle \emph{adversarial environments}, and (3) \emph{planning}. And there is an infinite world of novel problem formulations that can be modeled in online and reinforcement learning.

\paragraph{} In the following sections we consider in detail a number of the most basic online learning problems, and key tools for dealing with uncertainty and addressing the exploration-exploitation trade-off, and for handling i.i.d.\ and adversarial environments. (The topic of planning is left outside of the scope of the book.)

\section{A General Basic Setup}

\begin{figure}%
\centering
\includegraphics[width=.5\columnwidth]{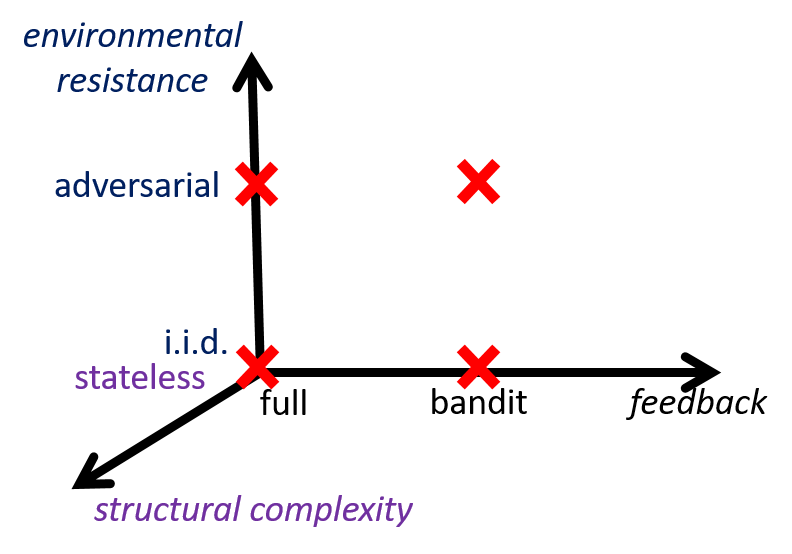}%
\caption{\textbf{The four basic online learning problems.}}%
\label{fig:4basic}%
\end{figure}

We start with four most basic games in online learning, depicted by the red crosses in \Cref{fig:4basic}: stateless i.i.d.\ full information, stateless i.i.d.\ adversarial, stateless i.i.d.\ bandit, and stateless adversarial bandit. The first setting is very easy and is studied in \Cref{ex:iidfullinfo}. The stateless i.i.d.\ adversarial setting is known as \emph{prediction with expert advice}, and the two bandit settings are known as \emph{stochastic multiarmed bandits} and \emph{adversarial multiarmed bandits}, respectively. We first provide a general setup that encompasses all the four problems, and then specialize it to each of them. 

We are given a $K \times \infty$ matrix of losses $\ell_{t,a}$, where $t \in \lrc{1,2,\dots}$ and $a \in \lrc{1,\dots,K}$ and $\ell_{t,a} \in [0,1]$. 
\[
\begin{array}{c}
\rotatebox[origin=c]{90}{\textit{Losses}}
\begin{array}{ccccc}
\ell_{1,1}, & \ell_{2,1}, & \cdots & \ell_{t,1}, & \cdots \\
\vdots   &  \vdots  & \cdots & \vdots   & \cdots\\
\ell_{1,a}, & \ell_{2,a}, & \cdots & \ell_{t,a}, & \cdots \\
\vdots   &  \vdots  & \cdots & \vdots   & \cdots\\
\ell_{1,K}, & \ell_{2,K}, & \cdots & \ell_{t,K}, & \cdots
\end{array}
\\
\\
\overrightarrow{~~~~~time~~~~~}
\end{array}
\]
The matrix is fixed before the game starts according to a protocol defined below, but not revealed to the algorithm. The game proceeds in the following way.

\paragraph{Game Protocol}~\\

\noindent
For $t = 1, 2, \dots$:
\begin{enumerate}
	\item Pick a row $A_t$
	\item Suffer $\ell_{t,A_t}$
	\item Observe $\dots$ [the observations are defined below]
\end{enumerate}

\paragraph{Definition of the four games}

There are two basic ways to generate the matrix of losses and two basic ways to define the observations, which jointly make up the four games, as summarized in the table below. 

The first way to generate the matrix is to sample $\ell_{t,a}$-s independently, so that the mean of the losses in each row is fixed, $\E[\ell_{t,a}] = \mu(a)$. The second is to generate $\ell_{t,a}$-s arbitrarily. The second model of generation of losses is known as an \emph{oblivious adversary}, since the generation happens before the game starts and does not take actions of the algorithm into account.\footnote{It is also possible to consider an \emph{adaptive adversary}, which generates losses as the game proceeds and takes past actions of the algorithm into account. We do not discuss this model in the book.}

The first way to define the observations is to reveal the full column $\ell_{t,1},\dots,\ell_{t,K}$ after the algorithm has played the row $A_t$. The second is to reveal only the selected entry $\ell_{t,A_t}$. 

Jointly the two ways of generating the matrix of losses and the two ways of defining the observations define the four variants of the game.

\noindent
\begin{tabular}{|c||c|c|}
\hline
\backslashbox{Matrix generation}{Observations} & Observe $\ell_{t,1},\dots,\ell_{t,K}$ & Observe $\ell_{t,A_t}$\\
\hline
\hline
$\ell_{t,a}$-s are sampled i.i.d. with $\E[\ell_{t,a}] = \mu(a)$ & \begin{tabular}{c}I.I.D. Prediction with\\expert advice\end{tabular} & \begin{tabular}{c}Stochastic multiarmed\\ bandits\end{tabular}\\
\hline
$\ell_{t,a}$ are selected arbitrarily (by an adversary) & \begin{tabular}{c}Prediction with expert\\advice (adversarial)\end{tabular} & \begin{tabular}{c}Adversarial multiarmed\\bandits\end{tabular}\\
\hline
\end{tabular}

\paragraph{Performance Measure} 

The goal of the algorithm is to play so that the loss it suffers will not be significantly larger than the loss of the best row in hindsight. There are several ways to formalize this goal. The basic performance measure is the \emph{regret} defined by
\[
R_T = \sum_{t=1}^T \ell_{t,A_t} - \min_a \sum_{t=1}^T \ell_{t,a}.
\]
In adversarial problems we analyze the \emph{expected regret}\footnote{It is also possible to analyze the regret, but we do not do it here.} defined by
\[
\E[R_T] = \E[\sum_{t=1}^T \ell_{t,A_t}] - \E[\min_a \sum_{t=1}^T \ell_{t,a}].
\]
If the sequence of losses is deterministic, we can remove the second expectation and obtain
\[
\E[R_T] = \E[\sum_{t=1}^T \ell_{t,A_t}] - \min_a \sum_{t=1}^T \ell_{t,a}.
\]
In stochastic problems we analyze the \emph{pseudo regret} defined by
\[
\bar R_T = \E[\sum_{t=1}^T \ell_{t,A_t}] - \min_a \E[\sum_{t=1}^T \ell_{t,a}] = \E[\sum_{t=1}^T \ell_{t,A_t}] - T \min_a \mu(a).
\]
Note that since for random variables $X$ and $Y$ we have $\E[\min\lrc{X,Y}] \leq \min \lrc{\E[X],\E[Y]}$ [it is recommended to verify this identity], we have $\bar R_T \leq \E[R_T]$. 

\paragraph{The different notions of regret} Let us briefly discuss the relations and differences between regret, expected regret, and pseudo-regret. First, we note that in the oblivious adversarial setting the losses are considered to be selected deterministically and, therefore, the expectation in the second term can be dropped, resulting in $\E[\min_a \sum_{t=1}^T \ell_{t,a}] = \min_a \E[\sum_{t=1}^T \ell_{t,a}] = \min_a \sum_{t=1}^T \ell_{t,a}$. Thus, in the oblivious adversarial setting the notions of expected regret and pseudo-regret coincide. (This is not true for the adaptive setting, but we do not delve into it here.) The difference between regret and expected/pseudo-regret in the adversarial setting is thus only in the first term -- whether we want to obtain guarantees on the expected performance of an algorithm, $\E[\sum_{t=1}^T \ell_{t,A_t}]$, or individual roll-outs of its play, $\sum_{t=1}^T \ell_{t,A_t}$. Both are valid targets, we focus on the expected performance because it is a tiny bit easier.

In the stochastic setting $\displaystyle \E[\min_a \sum_{t=1}^T \ell_{t,a}] \leq \min_a \E[\sum_{t=1}^T \ell_{t,a}] = T\min_a \mu(a)$, and so the expected regret and the pseudo-regret are not the same. In pseudo-regret the performance baseline is the expected performance of a best action, $\displaystyle T\min_a \mu(a)$, defined by $\mu(a)$, whereas in expected regret it is the expected best roll-out of any action. Imagine that there are $K$ actions and the loss of every action at every round is a Bernoulli random variable with bias $\frac12$. Then $\mu(a) = \frac12$ for all $a$ and the pseudo-regret baseline is $\frac12 T$. And for any algorithm $\E[\sum_{t=1}^T \ell_{t,A_t}] = \frac12 T$, thus $\bar R_T = 0$. However, $\E[\min_a \sum_{t=1}^T \ell_{t,a}] \leq \frac12 T$, because the baseline (the competitor of the algorithm) is allowed to select the best out of $K$ roll-outs of $T$ Bernoulli random variables with bias $\frac12$. In fact, $\E[\min_a \sum_{t=1}^T \ell_{t,a}] \approx \frac12 T - \sqrt{\frac12 T \ln K}$ (see \Cref{sec:expertslower}), leading to $\E[R_T] \approx \sqrt{\frac12 T \ln K}$. Even though the loss of any algorithm cannot be smaller than $\frac12 T$ in expectation, and the loss of any fixed row is also $\frac12 T$ in expectation, the expected regret grows as $\sqrt{\frac12 T \ln K}$, because the competitor has the advantage of being allowed to make $K$ tries of sampling $T$ Bernoulli losses and selecting the best, whereas the algorithm gets just one try. Thus, comparing the performance of an algorithm to the best row in expectation rather than the expected best roll-out is considered more reasonable. We will be able to derive bounds on pseudo-regret in the stochastic setting that grow at the rate of $\ln T$, whereas the fluctuations of $\sum_{t=1}^T\ell_{t,a}$ and $\sum_{t=1}^T\ell_{t,A_t}$ can be as large as $\sqrt T$, and so the best possible bounds on the regret and the expected regret in the stochastic setting are of order $\sqrt T$.

\paragraph{Explanation of the Names}

In the complete definition of prediction with expert advice game, in every round of the game the player gets an advice from $K$ experts and then takes an action, which may be a function of the advice. The player suffers a loss depending on the action taken, and the experts suffer losses depending on their advice. Hence the name, prediction with expert advice. If we restrict the actions of the player to following the advice of a single expert, then from the perspective of the playing strategy the actual advice does not matter and it is only the loss that defines the strategy. We consider the restricted setting, because it allows to highlight the relation to multiarmed bandits.

The name multiarmed bandits comes from the analogy with slot machines, which are one-armed bandits. In this game the actions are the ``arms'' of a slot machine.

\paragraph{Losses vs.\ Rewards}

In some games it is more natural to consider rewards (also called gains), rather than losses. In fact, in the literature on stochastic problems it is more popular to work with rewards, whereas in the literature on adversarial problems it is more popular to work with losses. There is a simple transformation $r = 1 - \ell$, which brings a losses game into a gains game and the other way around. Interestingly, in the adversarial setting working with losses leads to tighter and simpler results (see \Cref{ex:rewardsvslosses}). In the stochastic setting the choice does not matter.

\section{I.I.D.\ (stochastic) Multiarmed Bandits}
\label{sec:Stochastic-Bandits}

In this section we consider a multiarmed bandit game, where the outcomes (the sequence of losses) are generated i.i.d.\ with fixed, but unknown means. In this game there is no difference between working with losses or rewards, and since most of the literature is based on games with rewards we are going to use rewards in order to be consistent. The treatment of losses is identical - see \cite{Sel15}.

\paragraph{Notations} We are given a $K \times \infty$ matrix of rewards (or gains) $r_{t,a}$, where $t \in \lrc{1,2,\dots}$ and $a \in \lrc{1,\dots,K}$.

\[
\begin{array}{c}
\rotatebox[origin=c]{90}{\textit{Action rewards}}
\begin{array}{ccccc}
r_{1,1}, & r_{2,1}, & \cdots & r_{t,1}, & \cdots \\
\vdots   &  \vdots  & \cdots & \vdots   & \cdots\\
r_{1,a}, & r_{2,a}, & \cdots & r_{t,a}, & \cdots \\
\vdots   &  \vdots  & \cdots & \vdots   & \cdots\\
r_{1,K}, & r_{2,K}, & \cdots & r_{t,K}, & \cdots
\end{array}
\\
\\
\overrightarrow{~~~~~time~~~~~}
\end{array}
\]

We assume that $r_{t,a}$-s are in $[0,1]$ and that they are generated independently, so that $\E[r_{t,a}] = \mu(a)$. We use $\displaystyle \mu^* = \max_a \mu(a)$ to denote the expected reward of an optimal action and $\displaystyle \Delta(a) = \mu^* - \mu(a)$ to denote the \emph{suboptimality gap} (or simply the \emph{gap}) of action $a$. The suboptimality gap $\Delta(a)$ measures by how much, in expectation, playing action $a$ is worse than playing the optimal action. We use $\displaystyle a^* \in \arg \max_a \mu(a)$ to denote \emph{a best action} (note that there may be more than one best action, in such case we let $a^*$ be any of them).

\paragraph{Game Definition}~\\

\noindent
For $t = 1, 2, \dots$:
\begin{enumerate}
	\item Pick a row $A_t$
	\item Observe \& accumulate $r_{t,A_t}$
\end{enumerate}

\paragraph{Performance Measure} Let $N_t(a)$ denote the number of times action $a$ was played up to round $t$. We measure the performance using the pseudo regret and we rewrite it in the following way (note that since we are working with rewards, we subtract the expected reward of the algorithm from the expected reward of a best arm, whereas for losses it was the other way around)
\begin{align}
\bar R_T &= \max_a \E[\sum_{t=1}^T r_{t,a}] - \E[\sum_{t=1}^T r_{t,A_t}]\notag\\
&= T \mu^* - \E[\sum_{t=1}^T r_{t,A_t}]\notag\\
&= \sum_{t=1}^T \E[\mu^* - r_{t,A_t}]\notag\\
&= \sum_{t=1}^T \E[\E[\mu^* - r_{t,A_t}\middle | A_t]]\label{eq:rtoGap}\\
&= \sum_{t=1}^T \E[\mu^* - \mu(A_t)]\notag\\
&= \sum_{t=1}^T \E[\Delta(A_t)]\notag\\
&= \sum_a \Delta(a) \E[N_T(a)].\notag
\end{align}
In step \eqref{eq:rtoGap} we note that $\E[r_{t,A_t}]$ is an expectation over two random variables, the selection of $A_t$, which is based on the history of the game, and the draw of $r_{t,A_t}$, for which $\E[r_{t,A_t}|A_t] = \mu(A_t)$. We have $\E[r_{t,A_t}] =\E[\E[r_{t,A_t}|A_t]]$, where the inner expectation is with respect to the draw of $r_{t,A_t}$ and the outer expectation is with respect to the draw of $A_t$. 

Note that in the i.i.d.\ setting the reward of an algorithm is compared to the expected reward of the best action in expectation, $\displaystyle \max_a \E[\sum_{t=1}^T r_{t,a}] = T \max_a \mu(a)$, whereas in the adversarial setting the reward of an algorithm is compared to the reward of the best action in hindsight, $\displaystyle \max_a \sum_{t=1}^T r_{t,a}$.

\subsubsection*{Exploration-exploitation trade-off: A simple approach} I.I.D. multiarmed bandits is the simplest problem, where we face the exploration-exploitation trade-off. In general, the goal is to play a best arm in all the rounds, but since the identity of the best arm is unknown, it has to be identified first. In order to identify a best arm we need to explore all the arms. However, rounds used for exploration of suboptimal arms increase the regret, because every time we play a suboptmal arm $a$, we pay $\Delta(a)$ in the regret. The total regret is $\bar R = \sum_a \Delta(a) \E[N_T(a)]$, where $\E[N_T(a)]$ is the expected number of times a suboptimal action $a$ was played. At the same time, saving too much on exploration may lead to confusion between a best and a suboptimal arm, which may eventually lead to even higher regret if we start exploiting a wrong arm. 

So let us make a first attempt at quantifying this trade-off. Assume that we have just two actions, and we know the suboptimality gap $\Delta$, so that $\mu(a) = \mu(a^*) - \Delta$, but we do not know which of the two actions is the better one, so we need to figure it out. Assume that we know the time horizon $T$ we are going to play the game. A possible approach is to start with $\varepsilon T$ exploration rounds, where we play each of the two arms $\frac12 \varepsilon T$ times, followed by $(1-\varepsilon) T$ exploitation rounds, where we play the arm that yielded the highest reward by the end of the exploration phase. What should be the length of the exploration phase $\varepsilon T$ and what will be the pseudo-regret of this playing strategy?

Let $\delta(\varepsilon)$ denote the probability that we misidentify the best arm at the end of the exploration phase, namely, due to statistical fluctuations the suboptimal arm $a$ happens to yield a higher reward than the optimal arm $a^*$. The pseudo regret can be bounded by:
\[
\bar R_T \leq \underbrace{\frac{1}{2}\Delta \varepsilon T}_{\text{exploration}} + \underbrace{\delta(\varepsilon) \Delta (1 - \varepsilon) T}_{\text{exploitation}} \leq \frac{1}{2}\Delta \varepsilon T + \delta(\varepsilon) \Delta T = \lr{\frac{1}{2} \varepsilon + \delta(\varepsilon)} \Delta T,
\]
where the first term is a bound on the pseudo regret during the exploration phase and the second term is a bound on the pseudo regret during the exploitation phase in case we select a wrong arm at the end of the exploration phase. Now what is $\delta(\varepsilon)$? Let $\hat \mu_t(a)$ denote the empirical mean of observed rewards of arm $a$ up to round $t$. For the exploitation phase it is natural to select the arm that maximizes $\hat \mu_{\varepsilon T}(a)$ at the end of the exploration phase. Therefore:
\begin{align*}
\delta(\varepsilon) &= \P[\hat \mu_{\varepsilon T}(a) \geq \hat \mu_{\varepsilon T}(a^*)]\\
&\leq \P[\hat \mu_{\varepsilon T}(a) \geq \mu(a) + \frac{1}{2} \Delta] + \P[\hat \mu_{\varepsilon T}(a^*) \leq \mu^* - \frac{1}{2} \Delta]\\
&\leq 2 e^{-2 \frac{\varepsilon T}{2} \lr{\frac{1}{2} \Delta}^2} = 2 e^{- \varepsilon T \Delta^2 / 4},
\end{align*}
where the last line is by Hoeffding's inequality. By substituting this back into the regret bound we obtain:
\[
\bar R_T \leq \lr{\frac{1}{2}\varepsilon + 2 e^{- \varepsilon T \Delta^2 / 4}}\Delta T.
\]
In order to minimize $\frac{1}{2}\varepsilon + 2 e^{- \varepsilon T \Delta^2 / 4}$ we take a derivative and equate it to zero, which leads to $\varepsilon = \frac{\ln(T\Delta^2)}{T\Delta^2/4}$. It is easy to check that the second derivative is positive, confirming that this is the minimum. Note that $\varepsilon$ must be non-negative, so strictly speaking we have $\varepsilon = \max\lrc{0,\frac{\ln(T\Delta^2)}{T\Delta^2/4}}$. If we substitute this back into the regret bound we obtain:
\[
\bar R_T \leq \min\lrc{\Delta T, \lr{\frac{2\ln(T \Delta^2)}{T\Delta^2} + 2 e^{- \ln (T \Delta^2)}}\Delta T} = \min\lrc{\Delta T, \frac{2\ln(T \Delta^2)}{\Delta} + \frac{2}{\Delta}}.
\]
Note that the number of exploration rounds is $\varepsilon T = \max\lrc{0, \frac{\ln (T \Delta^2)}{\Delta^2 / 4}}$.

Pay attention that the regret bound has an inverse dependence on $\Delta$, meaning that it gets \emph{larger} as $\Delta$ gets \emph{smaller}. Although intuitively when $\Delta$ is small we do not care that much about playing a suboptimal action as opposed to the case when $\Delta$ is large, problems with small $\Delta$ are actually harder and lead to larger regret. The reason is that the number of rounds that it takes to identify the best action (the number of exploration steps $\varepsilon T$) grows with $1/\Delta^2$. Even though in each exploration round we only suffer a regret of $\Delta$, the fact that the number of exploration rounds grows with $1/\Delta^2$ makes problems with small $\Delta$ harder and makes the regret grow at the rate of $1/\Delta$. However, note that if the time horizon $T$ is very small in relation to $1/\Delta^2$, then there is not enough time to identify the best action, and the $\Delta T$ term dominates the minimum in the regret bound, see \Cref{ex:minimaxgap} for further details.

\paragraph{}
The above approach has three drawbacks: (1) it assumes knowledge of the time horizon $T$, (2) it assumes knowledge of the gap $\Delta$, and (3) if we would generalize it to more than one arm, the length of the exploration phase would depend on the smallest gap, even if there are many arms with larger gap that are much easier to eliminate. The following approach resolves all the three problems.

\subsubsection*{The Upper Confidence Bound (UCB) algorithm} 

We present the UCB1 algorithm of \citet{ACBF02}.\footnote{See \Cref{ex:ImprovedUCB1} for an improved parametrization and analysis.}

\begin{algorithm}
\caption{UCB1 \citep{ACBF02}}
\label{algo:UCB1}
\begin{algorithmic}
\State {\bf Initialization:} Play each action once.

\For{$t = K+1,K+2,...$}

\State Play $\displaystyle A_t = \arg \max_a \hat \mu_{t-1}(a) + \sqrt{\frac{3 \ln t}{2 N_{t-1}(a)}}$.

\EndFor
\end{algorithmic}
\end{algorithm}

The expression $U_t(a) = \hat \mu_{t-1}(a) + \sqrt{\frac{3 \ln t}{2 N_{t-1}(a)}}$ is called an \emph{upper confidence bound}. Why? Because $U_t(a)$ upper bounds $\mu(a)$ with high probability. UCB approach follows the \emph{optimism in the face of uncertainty principle}. That is, we take an optimistic estimate of the reward of every arm by taking the upper limit of the confidence bound. UCB1 algorithm has the following regret guarantee.

\begin{figure}%
\centering
\includegraphics[width=.5\textwidth]{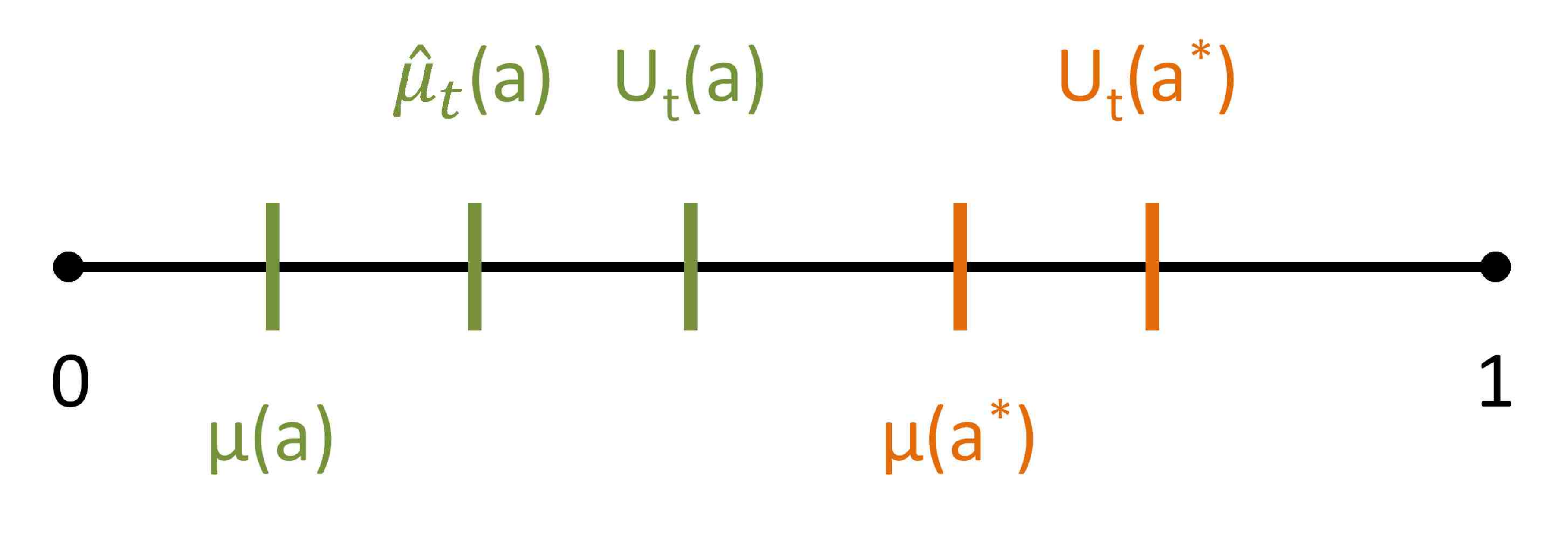}%
\caption{Illustration for UCB analysis.}%
\label{fig:UCB}%
\end{figure}

\begin{theorem}
For any time $T$ the regret of UCB1 satisfies:
\[
\bar R_T \leq 6 \sum_{a:\Delta(a)>0} \frac{\ln T}{\Delta(a)} + \lr{1 + \frac{\pi^2}{3}} \sum_a \Delta(a).
\]
\end{theorem}

\begin{proof}
For the analysis it is convenient to have the following picture in mind -- see Figure \ref{fig:UCB}. A suboptimal arm may be played if $U_t(a) \geq U_t(a^*)$. Our goal is to show that this does not happen too often. The analysis is based on the following three points, which bound the corresponding distances in Figure~\ref{fig:UCB}. 
\begin{enumerate}
	\item\label{pnt:UCB1} We show that $U_t(a^*) > \mu(a^*)$ for almost all rounds. A bit more precisely, let $F(a^*)$ be the number of rounds when $U_t(a^*) \leq \mu(a^*)$, then $\EEE{F(a^*)} \leq \frac{\pi^2}{6}$.
	\item\label{pnt:UCB2} In a similar way, we show that $\hat \mu_t(a) < \mu(a) + \sqrt{\frac{3 \ln t}{2 N_t(a)}}$ for almost all rounds. A bit more precisely, let $F(a)$ be the number of rounds when $\hat \mu_t(a) \geq \mu(a) + \sqrt{\frac{3 \ln t}{2 N_t(a)}}$, then $\EEE{F(a)} \leq \frac{\pi^2}{6}$. (Note that $L_t(a) = \hat \mu_t(a) - \sqrt{\frac{3 \ln t}{2 N_t(a)}}$ is a lower confidence bound for $\mu(a)$, meaning that with high probability $L_t(a) < \mu(a)$.)
	\item\label{pnt:UCB3} When Point \ref{pnt:UCB2} holds we have that $U_t(a) = \hat \mu_{t-1}(a) + \sqrt{\frac{3 \ln t}{2 N_{t-1}(a)}} \leq \mu(a) + 2 \sqrt{\frac{3 \ln t}{2 N_{t-1}(a)}} = \mu(a^*) - \Delta(a) + 2 \sqrt{\frac{3 \ln t}{2 N_{t-1}(a)}}$.
\end{enumerate}

Let us fix time horizon $T$ and analyze what happens by time $T$ (note that the algorithm does not depend on $T$). We have that for most rounds $t \leq T$:
\begin{align*}
U_t(a) &< \mu(a^*) - \Delta(a) + \sqrt{\frac{6 \ln t}{N_{t-1}(a)}} \leq \mu(a^*) - \Delta(a) + \sqrt{\frac{6 \ln T}{N_{t-1}(a)}},\\
U_t(a^*) &>\mu(a^*).
\end{align*}
Thus, we can play a suboptimal action $a$ only in the following cases:
\begin{itemize}
	\item Either $\sqrt{\frac{6\ln T}{N_{t-1}(a)}} \geq \Delta(a)$, which means that $N_{t-1}(a) \leq \frac{6 \ln T}{\Delta(a)^2}$. (As long as a suboptimal arm has not been played $\frac{6\ln T}{\Delta(a)^2}$ times, its confidence interval is not tight enough to reliably disambiguate it from the best arm.)
	\item Or one of the confidence intervals in Points \ref{pnt:UCB1} or \ref{pnt:UCB2} has failed.
\end{itemize}
In other words, after a suboptimal action $a$ has been played for $\lru{\frac{6 \ln T}{\Delta(a)^2}}$ rounds, it can only be played again only if one of the confidence intervals fails. Therefore,
\[
\EEE{N_T(a)} \leq \lru{\frac{6 \ln T}{\Delta(a)^2}} + \EEE{F(a^*)} + \EEE{F(a)} \leq \frac{6 \ln T}{\Delta(a)^2} + 1 + \frac{\pi^2}{3}
\]
and since $\bar R_T(a) = \sum_a \Delta(a) \EEE{N_T(a)}$ the result follows.

To complete the proof it is left to prove Points \ref{pnt:UCB1} and \ref{pnt:UCB2}. We prove Point \ref{pnt:UCB1}, the proof of Point~\ref{pnt:UCB2} is identical. We start by looking at \[\P[U_t(a^*) \leq \mu(a^*)] = \P[\hat \mu_{t-1}(a^*) + \sqrt{\frac{3 \ln t}{2N_{t-1}(a^*)}} \leq \mu(a^*)] = \P[\mu(a^*) - \hat \mu_{t-1}(a^*) \geq \sqrt{\frac{3 \ln t}{2 N_{t-1}(a^*)}}].\] The delicate point is that $N_{t-1}(a^*)$ is a random variable dependent on $\hat \mu_{t-1}(a^*)$, and thus we cannot apply Hoeffding's inequality directly. Instead, we introduce a series of random variables $X_1, X_2, \dots$, such that $X_i$-s have the same distribution as $r_{t,a^*}$-s. Let $\bar \mu_s = \frac{1}{s} \sum_{i=1}^s X_i$ be the average of the first $s$ elements of the sequence. Then we have:
\begin{align*}
\P[\mu(a^*) - \hat \mu_{t-1}(a^*) \geq \sqrt{\frac{3 \ln t}{2 N_{t-1}(a^*)}}] &\leq \P[\exists s\in\lrc{1,\dots,t}: \mu(a^*) - \bar \mu_s \geq \sqrt{\frac{3 \ln t}{2 s}}]\\
&\leq \sum_{s=1}^t \P[\mu(a^*) - \bar \mu_s \geq \sqrt{\frac{3 \ln t}{2 s}}]\\
&\leq \sum_{s=1}^t \frac{1}{t^3} = \frac{1}{t^2},
\end{align*}
where in the first line we decouple $\hat \mu_t(a^*)$-s from $N_t(a^*)$-s via the use of $\bar \mu_s$-s, and in the last line we apply Hoeffding's inequality (\Cref{cor:Hoeffding}). Note that $3 \ln t = \ln t^3$ corresponds to $\ln \frac{1}{\delta}$ in Hoeffding's inequality, and thus $\delta = \frac{1}{t^3}$. Finally, we have:
\[
\EEE{F(a^*)} = \sum_{t=1}^\infty \P[\mu(a^*) - \hat \mu_{t-1}(a^*) \geq \sqrt{\frac{3 \ln t}{2 N_{t-1}(a^*)}}] \leq \sum_{t=1}^\infty \frac{1}{t^2} = \frac{\pi^2}{6}.
\]
\end{proof}

We refer the reader to \Cref{ex:ImprovedUCB1} for an improved parametrization and a tighter regret bound for the UCB1 algorithm.

\section{Prediction with Expert Advice}
\label{sec:PredictionWithExpertAdvice}

Now we turn our attention to stateless adversarial game with full information feedback.

\paragraph{Notations} We are given a $K \times \infty$ matrix of expert losses $\ell_{t,a}$, where $t \in \lrc{1,2,\dots}$ and $a \in \lrc{1,\dots,K}$.

\[
\begin{array}{c}
\rotatebox[origin=c]{90}{\textit{Expert Losses}}
\begin{array}{ccccc}
\ell_{1,1}, & \ell_{2,1}, & \cdots & \ell_{t,1}, & \cdots \\
\vdots   &  \vdots  & \cdots & \vdots   & \cdots\\
\ell_{1,a}, & \ell_{2,a}, & \cdots & \ell_{t,a}, & \cdots \\
\vdots   &  \vdots  & \cdots & \vdots   & \cdots\\
\ell_{1,K}, & \ell_{2,K}, & \cdots & \ell_{t,K}, & \cdots
\end{array}
\\
\\
\overrightarrow{~~~~~time~~~~~}
\end{array}
\]

\paragraph{Game Definition}~\\

\noindent
For $t = 1, 2, \dots$:
\begin{enumerate}
	\item Pick a row $A_t$
	\item Observe the column $\ell_{t,1},\dots,\ell_{t,K}$ \& suffer $\ell_{t,A_t}$
\end{enumerate}

\paragraph{Performance Measure} The performance is measured by \emph{regret}
\[
R_T = \sum_{t=1}^T \ell_{t,A_t} - \min_a \lr{\sum_{t=1}^T \ell_{t,a}}.
\]
In the notes we analyze the \emph{expected regret} $\EEE{R_T}$.

\paragraph{The Hedge Algorithm} We consider the Hedge algorithm (a.k.a.\ exponential weights and weighted majority) for playing this game.

\begin{algorithm}
\caption{Hedge (a.k.a.\ Exponential Weights), \citep{Vov90, LW94}}
\label{algo:Hedge}
\begin{algorithmic}
\State {\bf Input:} Learning rates $\eta_1 \geq \eta_2 \geq \dots > 0$

\State $\forall a: L_0(a) = 0$

\For{$t = 1,2,...$}

\State $\forall a: ~ p_t(a) = \frac{e^{- \eta_t L_{t-1}(a)}}{\sum_{a'} e^{-\eta_t L_{t-1}(a')}}$

\State Sample $A_t$ according to $p_t$ and play it

\State Observe $\ell_{t,1}, \dots, \ell_{t,K}$ and suffer $\ell_{t,A_t}$

\State $\forall a: ~ L_t(a) = L_{t-1}(a) + \ell_{t,a}$
\EndFor
\end{algorithmic}
\end{algorithm}

\paragraph{Analysis} We analyze the Hedge algorithm in a slightly simplified setting, where the time horizon $T$ is known. Unknown time horizon can be handled by using the doubling trick (see \Cref{ex:doubling}) or, much more elegantly, by a more careful analysis (see, e.g., \citet{BCB12}).

The analysis is based on the following lemma.

\begin{lemma}
\label{lem:sum}
Let $\lrc{X_{1,a}, X_{2,a}, \dots}_{a \in \lrc{1,\dots,K}}$ be $K$ sequences of non-negative numbers ($X_{t,a} \geq 0$ for all $a$ and $t$). Let $L_t(a) = \sum_{s=1}^t X_{s,a}$, let $L_0(a)$ be zero for all $a$, and let $\eta > 0$. Finally, let $\displaystyle p_t(a) = \frac{e^{-\eta L_{t-1}(a)}}{\sum_{a'} e^{-\eta L_{t-1}(a')}}$. Then:
\[
\sum_{t=1}^T \sum_{a=1}^K p_t(a) X_{t,a} - \min_a L_T(a) \leq \frac{\ln K}{\eta} + \frac{\eta}{2} \sum_{t=1}^T \sum_{a=1}^K p_t(a) \lr{X_{t,a}}^2.
\]
\end{lemma}

\begin{proof}
We define $W_t = \sum_a e^{-\eta L_t(a)}$ and study how this quantity evolves. We start with an upper bound.
\begin{align}
\frac{W_t}{W_{t-1}} &= \frac{\sum_a e^{-\eta L_t(a)}}{\sum_a e^{-\eta L_{t-1}(a)}}\notag\\
&= \frac{\sum_a e^{-\eta X_{t,a}} e^{-\eta L_{t-1}(a)}}{\sum_a e^{-\eta L_{t-1}(a)}}\label{eq:a1}\\
&= \sum_a e^{-\eta X_{t,a}} \frac{e^{-\eta L_{t-1}(a)}}{\sum_{a'} e^{-\eta L_{t-1}(a')}}\notag\\
&= \sum_a e^{-\eta X_{t,a}} p_t(a)\label{eq:a2}\\
&\leq \sum_a \lr{1 - \eta X_{t,a} + \frac{1}{2} \eta^2 \lr{X_{t,a}}^2} p_t(a)\label{eq:a3}\\
&= 1 - \eta \sum_a X_{t,a} p_t(a) + \frac{\eta^2}{2} \sum_a \lr{X_{t,a}}^2 p_t(a)\notag\\
&\leq e^{- \eta \sum_a X_{t,a} p_t(a) + \frac{\eta^2}{2} \sum_a \lr{X_{t,a}}^2 p_t(a)}\label{eq:a4},
\end{align}
where in \eqref{eq:a1} we used the fact that $L_t(a) = X_{t,a} + L_{t-1}(a)$, in \eqref{eq:a2} we used the definition of $p_t(a)$, in \eqref{eq:a3} we used the inequality $e^x \leq 1 + x + \frac{1}{2}x^2$, which holds for $x \leq 0$ (this is a delicate point, because the inequality does not hold for $x > 0$ and, therefore, we must check that the condition $x \leq 0$ is satisfied; it is satisfied under the assumptions of the lemma), and inequality \eqref{eq:a4} is based on inequality $1 + x \leq e^x$, which holds for all $x$.

Now we consider the ratio $\frac{W_T}{W_0}$. On the one hand:
\[
\frac{W_T}{W_0} = \frac{W_1}{W_0} \times \frac{W_2}{W_1} \times \dots \times \frac{W_T}{W_{T-1}} \leq e^{-\eta \sum_{t=1}^T \sum_a X_{t,a} p_t(a) + \frac{\eta^2}{2} \sum_{t=1}^T \sum_a \lr{X_{t,a}}^2 p_t(a)}.
\]
On the other hand:
\[
\frac{W_T}{W_0} = \frac{\displaystyle \sum_a e^{-\eta L_T(a)}}{K} \geq \frac{\displaystyle \max_a e^{-\eta L_T(a)}}{K} = \frac{e^{\displaystyle -\eta \min_a L_T(a)}}{K},
\]
where we lower-bounded the sum by its maximal element. By taking the two inequalities together and applying logarithm we obtain:
\[
-\eta \min_a L_T(a) - \ln K \leq -\eta \sum_{t=1}^T \sum_a X_{t,a} p_t(a) + \frac{\eta^2}{2} \sum_{t=1}^T \sum_a \lr{X_{t,a}}^2 p_t(a).
\]
Finally, by changing sides and dividing by $\eta$ we get:
\[
\sum_{t=1}^T \sum_a X_{t,a} p_t(a) - \min_a L_T(a) \leq \frac{\ln K}{\eta} + \frac{\eta}{2} \sum_{t=1}^T \sum_a \lr{X_{t,a}}^2 p_t(a)
\]
\end{proof}

Now we are ready to present an analysis of the Hedge algorithm.

\begin{theorem}
The expected regret of the Hedge algorithm with a fixed learning rate $\eta$ satisfies:
\[
\E[R_T] \leq \frac{\ln K}{\eta} + \frac{\eta}{2} T.
\]
The expected regret is minimized by $\eta = \sqrt{\frac{2 \ln K}{T}}$, which leads to
\[
\E[R_T] \leq \sqrt{2 T \ln K}.
\]
\end{theorem}

\begin{proof}
We note that $\ell_{t,a}$-s are positive and apply Lemma \ref{lem:sum} to obtain:
\[
\sum_{t=1}^T \sum_{a=1}^K p_t(a) \ell_{t,a} - \min_a L_T(a) \leq \frac{\ln K}{\eta} + \frac{\eta}{2} \sum_{t=1}^T \sum_{a=1}^K p_t(a) \lr{\ell_{t,a}}^2.
\]
Note that $\sum_a p_t(a) \ell_{t,a}$ is the expected loss of Hedge at round $t$ and $\sum_{t=1}^T \sum_{a=1}^K p_t(a) \ell_{t,a}$ is the expected cumulative loss of Hedge after $T$ rounds. Thus, the left hand side of the inequality is the expected regret of Hedge. Also note that $\ell_{t,a} \leq 1$, and thus $\lr{\ell_{t,a}}^2 \leq 1$ and $\displaystyle \sum_a p_t(a) \lr{\ell_{t,a}}^2 \leq 1$. Thus, $\sum_{t=1}^T \sum_{a=1}^K p_t(a) \lr{\ell_{t,a}}^2 \leq T$. Altogether, we get that
\[
\EEE{R_T} \leq \frac{\ln K}{\eta} + \frac{\eta}{2} T.
\]
By taking the derivative of the right hand side and equating it to zero we obtain that $-\frac{\ln K}{\eta^2} + \frac{T}{2} = 0$, and thus $\eta = \sqrt{\frac{2 \ln K}{T}}$ is an extreme point. The second derivative is $\frac{2\ln K}{\eta^3}$, and since $\eta > 0$ it is positive. Thus, the extreme point is the minimum.
\end{proof}

See \Cref{ex:TightHedge} for a tighter analysis of the Hedge algorithm that matches the lower bound presented in the next section.

\subsection{Lower Bound}
\label{sec:expertslower}

A lower bound for the expected regret in prediction with expert advice is based on the following construction. We draw a $K \times \infty$ matrix of losses with each loss drawn according to a Bernoulli distribution with bias $1/2$. In this game the expected loss of any algorithm after $T$ rounds is $T/2$, irrespective of what the algorithm is doing. However, the loss of the best action in hindsight is lower, because we are selecting the ``best'' out of $K$ rows. For each individual row the expected loss is $T/2$, but the expectation of the minimum of the losses is lower. The reduction is quantified in the following theorem, see \citet{CBL06} for a proof.
\begin{theorem}
\label{thm:fullinfolower}
Let $\ell_{t,a}$ be i.i.d.\ Bernoulli random variables with bias $1/2$, then
\[
\lim_{T\to\infty} \lim_{K\to\infty} \frac{T/2 - \E[\min_a \sum_{t=1}^T \ell_{t,a}]}{\sqrt{\frac{1}{2}T\ln K}} = 1.
\]
\end{theorem}
Note that the numerator in the above expression, $T/2 - \E[\min_a \sum_{t=1}^T \ell_{t,a}]$, is the expectation of the expected regret with respect to generation of the matrix of losses. Thus, if the adversary generates the matrix of losses according to the construction described above, then in expectation with respect to generation of the matrix and in the limit of $K$ and $T$ going to infinity the expected regret cannot be smaller than $\sqrt{\frac{1}{2}T\ln K}$. 

The lower bound in \Cref{thm:fullinfolower} matches up to constants the refined upper bound on the expected regret of Hedge provided in \Cref{ex:TightHedge}, which is an extremely rare case. It shows that Hedge is a minimax optimal algorithm for this game.

\section{Adversarial Multiarmed Bandits}

We proceed to stateless adversarial game with bandit feedback.

\paragraph{Game definition} We are working with the same matrix of losses as in prediction with expert advice, but now at each round of the game we are allowed to observe only the loss of the row that we have played:

\noindent
For $t = 1, 2, \dots$:
\begin{enumerate}
	\item Pick a row $A_t$
	\item Observe \& suffer $\ell_{t,A_t}$. ($\ell_{t,a}$-s for $a \neq A_t$ remain unobserved) 
\end{enumerate}

\paragraph{Importance-Weighted loss estimates: Emulating full information under bandit feedback}

In the bandit setting an algorithm only observes the loss $\ell_{t,A_t}$ of the action $A_t$ played at round $t$. An elegant way to convert bandit feedback into an imaginary full information feedback is to use importance-weighted loss estimates defined by
\[
\tilde \ell_{t,a} = \frac{\ell_{t,a} \1[A_t = a]}{p_t(a)} = \begin{cases} \frac{\ell_{t,a}}{p_t(a)}, & \text{if $A_t = a$}\\0, & \text{otherwise}\end{cases}\quad,
\]
where 
\[
\1[A_t = a] = \begin{cases} \frac{\ell_{t,a}}{p_t(a)}, & \text{If $A_t = a$}\\0, & \text{otherwise}\end{cases}
\] 
is the indicator function. Note that $\tilde \ell_{t,a}$ is well-defined for all $a$, even though we do not observe $\ell_{t,a}$ for $a\neq A_t$. As we will show in a moment, it is also an unbiased estimate of $\ell_{t,a}$, namely, $\E[\tilde \ell_{t,a}] = \ell_{t,a}$, although it has a higher variance. 

\paragraph{The EXP3 Algorithm} The importance-weighted loss estimates can be used as a substitute for the true losses in running the Hedge algorithm. This leads to the EXP3 algorithm proposed by \citet{ACB+02} and summarized in \Cref{algo:EXP3} display.\footnote{We note that the original algorithm of \citet{ACB+02} was formulated for the gains game. Here we present an improved and simplified version of the algorithm for the losses game \citep{Sto05,Bub10}. We refer to \Cref{ex:rewardsvslosses} for the difference between the two.} Its name, EXP3, stands for EXPonential EXPloration-EXPloitation.

\begin{algorithm}
\caption{EXP3 \citep{ACB+02}} 
\label{algo:EXP3}
\begin{algorithmic}
\State {\bf Input:} Learning rates $\eta_1 \geq \eta_2 \geq \dots > 0$

\State $\forall a: \tilde L_0(a) = 0$

\For{$t = 1,2,...$}

\State $\forall a: ~ p_t(a) = \frac{e^{- \eta_t \tilde L_{t-1}(a)}}{\sum_{a'} e^{-\eta_t \tilde L_{t-1}(a')}}$

\State Sample $A_t$ according to $p_t$ and play it

\State Observe and suffer $\ell_{t,A_t}$

\State Set $\tilde \ell_{t,a} = \frac{\ell_{t,a} \1[A_t = a]}{p_t(a)} = \begin{cases} \frac{\ell_{t,a}}{p_t(a)}, & \text{If $A_t = a$}\\0, & \text{otherwise}\end{cases}$

\State $\forall a: ~ \tilde L_t(a) = \tilde L_{t-1}(a) + \tilde \ell_{t,a}$
\EndFor
\end{algorithmic}
\end{algorithm}

\paragraph{Properties of importance-weighted loss estimates} Before analyzing the EXP3 algorithm we discuss a number of important properties of importance-weighted loss estimates.

\begin{enumerate}
	\item The samples $\tilde \ell_{t,a}$ are not independent in two ways. First, for a fixed $t$, the set $\lrc{\tilde \ell_{t,1},\dots,\tilde \ell_{t,K}}$ is dependent (if we know that one of $\tilde \ell_{t,a}$-s is non-zero, we automatically know that all the rest are zero). And second, $\tilde \ell_{t,a}$ depends on all $\tilde \ell_{s,{a'}}$ for $s < t$ and all $a'$ since $p_t(a)$ depends on $\lrc{\tilde \ell_{s,a}}_{1 \leq s < t, a \in \lrc{1,\dots,K}}$, which is the history of the game up to round $t$. In other words, $p_t(a)$ itself is a random variable.
	\item Even though $\tilde \ell_{t,a}$-s are not independent, they are unbiased estimates of the true losses. Specifically,
	\begin{align*}
	\E[\tilde \ell_{t,a}] &= \E[\frac{\ell_{t,a} \1[A_t = a]}{p_t(a)}]\\
	&= \E[\E[\frac{\ell_{t,a} \1[A_t = a]}{p_t(a)} \middle | A_1,\dots,A_{t-1}]]\\
	&= \E[\frac{\ell_{t,a}}{p_t(a)}\E[\1[A_t = a] \middle | A_1,\dots,A_{t-1}]]\\
	&= \E[\frac{\ell_{t,a}}{p_t(a)}p_t(a)]\\
	&= \ell_{t,a}.
	\end{align*}
	The first expectation above is with respect to $A_1,\dots,A_t$. In the nested expectations, the external expectation is with respect to $A_1,\dots,A_{t-1}$ and the internal is with respect to $A_t$. Note that $p_t(a)$ is a random variable depending on $A_1,\dots,A_{t-1}$, thus after the conditioning on $A_1,\dots,A_{t-1}$ it is deterministic.
	\item Since $\ell_{t,a} \in [0,1]$, we have $\tilde \ell_{t,a} \in \lrs{0, \frac{1}{p_t(a)}}$.
	\item What is important is that the second moment of $\tilde \ell_{t,a}$-s is by an order of magnitude smaller than the second moment of a general random variable in the corresponding range. This is because the expectation of $\tilde \ell_{t,a}$-s is in the $[0,1]$ interval. Specifically:
	\begin{align*}
	\E[\lr{\tilde \ell_{t,a}}^2] &= \E[\lr{\frac{\ell_{t,a} \1[A_t=a]}{p_t(a)}}^2]\\
	&= \E[\frac{\lr{\ell_{t,a}}^2 \lr{\1[A_t=a]}^2}{p_t(a)^2}]\\
	&= \E[\frac{\lr{\ell_{t,a}}^2 \1[A_t=a]}{p_t(a)^2}]\\
	&\leq \E[\frac{\1[A_t=a]}{p_t(a)^2}]\\
	&= \E[\E[\frac{\1[A_t=a]}{p_t(a)^2} \middle | A_1,\dots,A_{t-1}]]\\
	&= \E[\frac{1}{p_t(a)^2}\E[\1[A_t=a] \middle | A_1,\dots,A_{t-1}]]\\
	&= \E[\frac{1}{p_t(a)}],
	\end{align*}
	where we have used $\lr{\1[A_t = a]}^2 = \1[A_t = a]$ and $\lr{\ell_{t,a}}^2 \leq 1$ (since $\ell_{t,a} \in [0,1]$).
\end{enumerate}

\paragraph{Analysis} Now we are ready to present an analysis of the algorithm.

\begin{theorem}
The expected regret of the EXP3 algorithm with a fixed learning rate $\eta$ satisfies:
\[
\EEE{R_T} \leq \frac{\ln K}{\eta} + \frac{\eta}{2} KT.
\]
The expected regret is minimized by $\eta = \sqrt{\frac{2 \ln K}{KT}}$, which leads to
\[
\EEE{R_T} \leq \sqrt{2 K T \ln K}.
\]
\end{theorem}

Note that the extra payment for being able to observe just one entry rather than the full column is the multiplicative $\sqrt{K}$ factor in the regret bound.

\begin{proof}
The proof of the theorem is based on Lemma \ref{lem:sum}. We note that $\tilde \ell_{t,a}$-s are all non-negative and, thus, by Lemma \ref{lem:sum} we have:
\[
\sum_{t=1}^T \sum_a p_t(a) \tilde \ell_{t,a} - \min_a \tilde L_T(a) \leq \frac{\ln K}{\eta} + \frac{\eta}{2}\sum_{t=1}^T \sum_a p_t(a) \lr{\tilde \ell_{t,a}}^2.
\]
By taking expectation of the two sides of the inequality we obtain:
\[
\EEE{\sum_{t=1}^T \sum_a p_t(a) \tilde \ell_{t,a}} - \EEE{\min_a \tilde L_T(a)} \leq \frac{\ln K}{\eta} + \frac{\eta}{2}\EEE{\sum_{t=1}^T \sum_a p_t(a) \lr{\tilde \ell_{t,a}}^2}.
\]
We note that $\EEE{\min\lrs{\cdot}} \leq \min\lrs{\EEE{\cdot}}$ and thus:
\[
\EEE{\sum_{t=1}^T \sum_a p_t(a) \tilde \ell_{t,a}} - \min_a \EEE{\tilde L_T(a)} \leq \frac{\ln K}{\eta} + \frac{\eta}{2}\EEE{\sum_{t=1}^T \sum_a p_t(a) \lr{\tilde \ell_{t,a}}^2}.
\]
And now we consider the three expectation terms in this inequality.
\[
\EEE{\sum_{t=1}^T \sum_a p_t(a) \tilde \ell_{t,a}} = \EEE{\sum_{t=1}^T \sum_a \EEE{p_t(a) \tilde \ell_{t,a} \middle | A_1,\dots,A_{t-1}}} = \EEE{\sum_{t=1}^T \sum_a p_t(a) \ell_{t,a}},
\]
which is the expected loss of EXP3.
\[
\EEE{\tilde L_T(a)} = \EEE{\sum_{t=1}^T \tilde \ell_{t,a}} = \sum_{t=1}^T \ell_{t,a},
\]
which is the cumulative loss of row $a$ up to time $T$.
And, finally,
\[
\EEE{\sum_{t=1}^T \sum_a p_t(a) \lr{\tilde \ell_{t,a}}^2} = \EEE{\sum_{t=1}^T \sum_a \EEE{p_t(a) \lr{\tilde \ell_{t,a}}^2 \middle | A_1,\dots,A_{t-1}}} \leq \EEE{\sum_{t=1}^T \sum_a p_t(a) \frac{1}{p_t(a)}} = KT.
\]
This step is known as ``the bandit magic''. We have shown earlier that the second moment of importance-weighted loss estimates satisfies $\E[\lr{\tilde \ell_{t,a}}^2]\leq \E[\frac{1}{p_t(a)}]$, which can still be a large number if $p_t(a)$ is small (and it is expected to be small for suboptimal $a$). However, when the actions are sampled according to $p_t(a)$, the weighted second moment satisfies $\E[\sum_a p_t(a) \lr{\tilde \ell_{t,a}}^2] \leq K$, so it is perfectly under control. Recall that in the full information setting we had $\sum_a p_t(a)\lr{\ell_{t,a}}^2 \leq 1$, and so $K$ is the estimator's variance price that we pay for having limited rather than full information feedback.

Plugging the bounds on the three expectation terms back into the inequality we obtain the first statement of the theorem. And, as before, we find $\eta$ that minimizes the bound.
\end{proof}

\subsection{Lower Bound}
\label{sec:AdvBandLower}

The lower bound is based on construction of $K+1$ games. In the 0-th game all the losses are sampled from Bernoulli distribution with bias 1/2. In the $i$-th game for $i \in \lrc{1,\dots,K}$ all the losses are Bernoulli with bias 1/2 except the losses of the $i$-th arm, which are Bernoulli with bias $1/2-\varepsilon$ for $\varepsilon = \sqrt{cK/T}$, where $c$ is a properly selected constant. With $T/K$ pulls it is impossible to distinguish between Bernoulli distribution with bias 1/2 and Bernoulli distribution with bias $1/2-\sqrt{K/T}$, because they induce indistinguishable distributions over sequences of length $T/K$. As a result, within $T$ pulls the player cannot distinguish between the 0-th game and the $i$-th games. Therefore, if the adversary picks an $i$-th game at random, the player's regret will on average (with respect to the adversary's and the players choices) be at least $\Omega\lr{\varepsilon T} = \Omega\lr{\sqrt{KT}}$. For the details of the proof see \citet{CBL06} or \citet{BCB12}.

\paragraph{} 
Note that there is a gap of $\sqrt{\ln K}$ factor between the lower bound and the upper bound for the EXP3 algorithm. There exists a different algorithm, Tsallis-INF, which closes this gap and achieves $O(\sqrt{KT})$ regret upper bound. The algorithm was proposed by \citet{AB09,AB10} and refined by \citet{ZS21}.

\section{Adversarial Multiarmed Bandits with Expert Advice}

In this section we move outside the stateless plane and introduce a contextual setting.

\paragraph{Game setting} We are, again, working with the same matrix of losses as in prediction with expert advice. But now in every round of the game we get advice of $N$ experts indexed by $h$ in a form of a distribution over the $K$ arms. More formally:

\noindent
For $t = 1, 2, \dots$:
\begin{enumerate}
	\item Observe $q_{t,1},\dots,q_{t,N}$, where $q_{t,h}$ is a probability distribution over $\lrc{1,\dots,K}$.
	\item Pick a row $A_t$.
	\item Observe \& suffer $\ell_{t,A_t}$. ($\ell_{t,a}$-s for $a \neq A_t$ remain unobserved) 
\end{enumerate}

\paragraph{Performance measure} We compare the expected loss of the algorithm to the expected loss of the best expert in hindsight, where the expectation of the loss of expert $h$ is taken with respect to its advice vector $q_h$. Specifically:
\[
\EEE{R_T} = \sum_{t=1}^T \sum_a p_t(a) \ell_{t,a} - \min_h \sum_{t=1}^T \sum_a q_{t,h}(a) \ell_{t,a}.
\]

\paragraph{The EXP4 Algorithm} One approach to playing this game, proposed by \citet{ACB+02}, is quite similar to the EXP3 algorithm.\footnote{As with the EXP3 algorithm, we present a slightly improved version of the algorithm for the game with losses. The original algorithm was designed for the game with rewards.} Its name, EXP4, stands for EXPonential EXPloration-EXPloitation with EXPert advice. Note that now $\tilde L_t(h)$ tracks cumulative importance-weighted estimates of expert losses, rather than of individual arm losses.

\begin{algorithm}
\caption{EXP4 \citep{ACB+02}} 
\begin{algorithmic}
\State {\bf Input:} Learning rates $\eta_1 \geq \eta_2 \geq \dots > 0$

\State $\forall h: \tilde L_0(h) = 0$

\For{$t = 1,2,...$}

\State $\forall h$: $w_t(h) = \frac{e^{- \eta_t \tilde L_{t-1}(h)}}{\sum_{h'} e^{-\eta_t \tilde L_{t-1}(h')}}$

\State Observe $q_{t,1},\dots,q_{t,N}$

\State $\forall a: ~ p_t(a) = \sum_h w_t(h) q_{t,h}(a)$

\State Sample $A_t$ according to $p_t$ and play it

\State Observe and suffer $\ell_{t,A_t}$

\State Set $\tilde \ell_{t,a} = \frac{\ell_{t,a} \1[A_t = a]}{p_t(a)} = \begin{cases} \frac{\ell_{t,a}}{p_t(a)}, & \text{if $A_t = a$}\\0, & \text{otherwise}\end{cases}$

\State Set $\tilde \ell_{t,h} = \sum_a q_t(h) \tilde \ell_{t,a}$

\State $\forall h: ~ \tilde L_t(h) = \tilde L_{t-1}(h) + \tilde \ell_{t,h}$
\EndFor
\end{algorithmic}
\end{algorithm}

\paragraph{Analysis} The EXP4 algorithm satisfies the following regret guarantee.

\begin{theorem}
The expected regret of the EXP4 algorithm with a fixed learning rate $\eta$ satisfies:
\[
\EEE{R_T} \leq \frac{\ln N}{\eta} + \frac{\eta}{2} KT.
\]
The expected regret is minimized by $\eta = \sqrt{\frac{2 \ln N}{KT}}$, which leads to
\[
\EEE{R_T} \leq \sqrt{2 K T \ln N}.
\]
\end{theorem}

Note that the $\ln N$ term plays the role of complexity of the class of experts in a very similar way to the complexity terms we saw earlier in supervised learning (specifically, in \Cref{thm:Finite}).

\begin{proof}
The analysis is quite similar to the analysis of the EXP3 algorithm. We note that $\tilde \ell_{t,h}$-s are all non-negative and that $w_t$ is a distribution over $\lrc{1,\dots,N}$ defined in the same way as $p_t$ in Lemma \ref{lem:sum}. Thus, by Lemma \ref{lem:sum} we have:
\[
\sum_{t=1}^T \sum_h w_t(h) \tilde \ell_{t,h} - \min_h \tilde L_T(h) \leq \frac{\ln N}{\eta} + \frac{\eta}{2}\sum_{t=1}^T \sum_h w_t(h) \lr{\tilde \ell_{t,h}}^2.
\]
By taking expectations of the two sides of this expression we obtain:
\[
\EEE{\sum_{t=1}^T \sum_h w_t(h) \tilde \ell_{t,h}} - \EEE{\min_h \tilde L_T(h)} \leq \frac{\ln N}{\eta} + \frac{\eta}{2}\EEE{\sum_{t=1}^T \sum_h w_t(h) \lr{\tilde \ell_{t,h}}^2}.
\]
As before, $\EEE{\min\lrs{\cdot}} \leq \min \lrs{\EEE{\cdot}}$ and thus:
\[
\EEE{\sum_{t=1}^T \sum_h w_t(h) \tilde \ell_{t,h}} - \min_h \EEE{\tilde L_T(h)} \leq \frac{\ln N}{\eta} + \frac{\eta}{2}\EEE{\sum_{t=1}^T \sum_h w_t(h) \lr{\tilde \ell_{t,h}}^2}.
\]
And now we consider the three expectation terms in this inequality.
\begin{align*}
\EEE{\sum_{t=1}^T \sum_h w_t(h) \tilde \ell_{t,h}} &= \EEE{\sum_{t=1}^T \sum_h w_t(h) \sum_a q_{t,h}(a) \tilde \ell_{t,a}}\\
&= \EEE{\sum_{t=1}^T \sum_a \lr{\sum_h w_t(h) q_{t,h}(a)} \tilde \ell_{t,a}}\\
&= \EEE{\sum_{t=1}^T \sum_a p_t(a) \tilde \ell_{t,a}}\\
&= \EEE{\sum_{t=1}^T \sum_a p_t(a) \ell_{t,a}},
\end{align*}
where the first equality is by the definition of $\tilde \ell_{t,h}$ and the last equality is due to unbiasedness of $\tilde \ell_{t,a}$. Thus, the first expectation is the expected loss of EXP4.
\[
\EEE{\tilde L_T(h)} = \EEE{\sum_{t=1}^T \tilde \ell_{t,h}} = \EEE{\sum_{t=1}^T \sum_a q_t(a) \tilde \ell_{t,a}} = \EEE{\sum_{t=1}^T \sum_a q_t(a) \ell_{t,a}},
\]
where we can remove tilde due to unbiasedness of $\tilde \ell_{t,a}$, and we obtain the expected cumulative loss of expert $h$ over $T$ rounds.
And, finally,
\begin{align*}
\EEE{\sum_{t=1}^T \sum_h w_t(h) \lr{\tilde \ell_{t,h}}^2} &= \EEE{\sum_{t=1}^T \sum_h w_t(h) \lr{\sum_a q_{t,h}(a) \tilde \ell_{t,a}}^2}\\
&\leq \EEE{\sum_{t=1}^T \sum_h w_t(h) \sum_a q_{t,h}(a) \lr{\tilde \ell_{t,a}}^2}\\
&= \EEE{\sum_{t=1}^T \sum_a \lr{\sum_h w_t(h) q_{t,h}(a)} \lr{\tilde \ell_{t,a}}^2}\\
&= \EEE{\sum_{t=1}^T \sum_a p_t(a) \lr{\tilde \ell_{t,a}}^2}\\
&\leq KT,
\end{align*}
where the first inequality is by Jensen's inequality (\Cref{thm:Jensen}) and convexity of $x^2$, and the last inequality is along the same lines as the analogous inequality in the analysis of EXP3. By substituting the three expectations back into the inequality we obtain the first statement of the theorem. And, as before, we find $\eta$ that minimizes the bound.
\end{proof}

\subsection{Lower Bound}

It is relatively easy to show that the regret of adversarial multiarmed bandits with expert advice must be at least $\Omega\lr{\sqrt{KT\frac{\ln N}{\ln K}}}$. The lower bound is based on construction of $\frac{\ln N}{\ln K}$ independent bandit problems, each according to the construction of the lower bound for multiarmed bandits in Section~\ref{sec:AdvBandLower}, and construction of expert advice, so that for every possible selection of best arms for the subproblems there is an expert that recommends that selection. For details of the proof see \citet{ADK12, SL16}. 

With a bit more work it is possible to derive a tight lower bound of $\Omega\lr{\sqrt{KT\ln\frac{N}{K}}}$ \citep{CIM25}. And by replacing EXP3-style approach with Tsallis-INF-style techniques it is possible to derive an algorithm with matching $O\lr{\sqrt{KT\ln\frac{N}{K}}}$ regret upper bound \citep{Kal14}.

\section{Exercises}

\begin{exercise}[\textit{Find an online learning problem from real life}] Find two examples of real life problems that fit into the online learning framework (online, not reinforcement!). For each of the two examples explain what is the set of actions an algorithm can take, what are the losses (or rewards) and what is the range of the losses/rewards, whether the problem is stateless or contextual, and whether the problem is i.i.d.\ or adversarial, and with full information or bandit feedback.
\end{exercise}

\begin{exercise}[\textit{Follow The Leader (FTL) algorithm for i.i.d.\ full information games}]
\label{ex:iidfullinfo}
Follow the leader (FTL) is a playing strategy that at round $t$ plays the action that was most successful up to round $t$ (``the leader''). Derive a bound for the pseudo regret of FTL in i.i.d.\ full information games with $K$ possible actions and outcomes bounded in the $[0,1]$ interval (you can work with rewards or losses, as you like). You can use the following guidelines (which assume a game with rewards):
\begin{enumerate}
	\item You are allowed to solve the problem for $K=2$. (The guidelines are not limited to $K=2$.)
	\item It may be helpful to write the algorithm down explicitly. For the analysis it does not matter how you decide to break ties.
	\item Let $\mu(a)$ be expected reward of action $a$ and let $\hat \mu_t(a)$ be empirical estimate of the reward of action $a$ at round $t$ (the average of rewards observed so far). Let $a^*$ be an optimal action (there may be more than one optimal action, but then things only get better [convince yourself that this is true], so we can assume that there is a single $a^*$). Let $\Delta(a) = \mu(a^*) - \mu(a)$. FTL may play $a \neq a^*$ at rounds $t$ for which $\hat \mu_{t-1}(a) \geq \max_{a'} \hat \mu_{t-1}(a')$ (in the case of two arms it means $\hat \mu_{t-1}(a) \geq \hat \mu_{t-1}(a^*)$). So you should analyze how often this may happen.
	\item Note that the number of times an action $a$ was played can be written as $N_T(a) = \sum_{t=1}^T \1[A_t = a]$, and that $\E[{\1[A_t=a]}] \leq \P[\hat \mu_{t-1}(a) \geq \max_{a'} \hat \mu_{t-1}(a')] \leq \P[\hat \mu_{t-1}(a) \geq \hat \mu_{t-1}(a^*)]$, where $\1$ is the indicator function.
	\item Bound $\P[\hat \mu_{t-1}(a) \geq \hat \mu_{t-1}(a^*)]$.
	\item At some point in the proof you will need to sum up a geometric series. A geometric series is a series of a form $\sum_{t=0}^\infty r^t$, and for $r < 1$ we have $\sum_{t=0}^\infty r^t = \frac{1}{1-r}$. In your case $r$ will be an exponent $r = e^\alpha$ for some constant $\alpha$.
	\item At the end you should get a bound of a form $\bar R_T \leq \sum_{a:\Delta(a)>0} \frac{c}{1-\exp\lr{-\Delta(a)^2/2}} \Delta(a)$, where $c$ is a constant.
\end{enumerate}
\emph{Important observations to make:
\begin{enumerate}
	\item Note that in the full information i.i.d.\ setting the regret does not grow with time!!! (Since the bound is independent of $T$.)
	\item Note that even though you have used $\Delta(a)$ in the analysis of the algorithm, you do not need to know it in order to define the algorithm! I.e., you can run the algorithm even if you do not know $\Delta(a)$.
\end{enumerate}
}
\end{exercise}

\begin{exercise}[\textit{Decoupling exploration and exploitation in i.i.d.\ multiarmed bandits}]
Assume an i.i.d.\ multiarmed bandit game, where the observations are not coupled to the actions. Specifically, we assume that at each round of the game the player is allowed to observe the reward of a single arm, but it does not have to be the same arm that was played at that round (and if it is not the same arm, the player does not observe its own reward, but instead observes the reward of the alternative option).

Derive a playing strategy and a regret bound for this game. (You should solve the problem for a general $K$ and you should get that the regret does not grow with time.)

\emph{Remark: note that in this setting the exploration is ``for free'', because we do not have to play suboptimal actions in order to test their quality. And if we contrast this with the standard multiarmed bandit setting we observe that the regret stops growing with time instead of growing logarithmically with time. Actually, the result that you should get is much closer to the regret bound for FTL with full information than to the regret bound for multiarmed bandits. Thus, it is not the fact that we have just a single observation that makes i.i.d.\ multiarmed bandits harder than full information games, but the fact that this single observation is linked to the action. In adversarial multiarmed bandits the effect of decoupling is more involved \citep{AMS12,SBCA14,RS20}.}
\end{exercise}

\begin{exercise}[\textit{The worst case gap for a fixed $T$}]
\label{ex:minimaxgap}
``Exploration-exploitation trade-off: A simple approach'' in \Cref{sec:Stochastic-Bandits} shows that problems with small gap $\Delta$ are harder (have higher pdeudo-regret) than problems with large gap $\Delta$. This is true if the time horizon is unlimited, but for a limited time horizon there is actually a limit on how large the pseudo-regret can be.

\begin{enumerate}
    \item Argue why in the two-arms case the regret never exceeds $\Delta T$.
    \item Show that the worst-case gap (the gap that leads to the highest pseudo-regret) is of order $\Delta = \theta\lr{\sqrt{\frac{\ln T}{T}}}$.
    \item Show that the regret satisfies $\bar R_T \leq \theta\lr{\sqrt{T\ln T}}$
\end{enumerate}

\noindent
\emph{Conclusion: Indeed, as $\Delta$ decreases the regret increases, but if the time horizon is fixed $T$, it only increases up to a limit of order $\sqrt{T\ln T}$, and thereafter starts decreasing. Bounds depending on $\Delta$ (as in \Cref{sec:Stochastic-Bandits}) are known as \emph{instance-dependent bounds} (they depend on the instance of a problem with gap $\Delta$), whereas bounds that are independent of $\Delta$ and only depend on $T$ (as the one you derived above) are known as \emph{worst-case bounds}.}    
\end{exercise}

\begin{exercise}[\textit{Improved Parametrization of UCB1}]
\label{ex:ImprovedUCB1}
In this question we refine the upper confidence bound for the UCB1 algorithm from \Cref{sec:Stochastic-Bandits} and derive a tighter regret bound.

\begin{enumerate}
\item Show that if we replace the upper confidence bounds in UCB1 with
\[
U_t(a) = \hat \mu_{t-1}(a) + \sqrt{\frac{\ln t}{N_{t-1}(a)}}
\]
then its pseudo-regret satisfies
\[
\bar R_T \leq 4 \sum_{a:\Delta(a) > 0} \frac{\ln T}{\Delta(a)} + (2\ln(T) + 3) \sum_a \Delta(a).
\]

\emph{Hint: The $T$-th harmonic number, $\sum_{t=1}^T \frac{1}{t}$, satisfies $\sum_{t=1}^T \frac{1}{t} \leq \ln(T) + 1$}.

\item Write a simulation to compare numerically the performance of UCB1 from \Cref{sec:Stochastic-Bandits} with performance of UCB1 with modified confidence bounds proposed above. Instructions for the simulation:
\begin{itemize}
    \item Generate Bernoulli rewards for two actions, $a^*$ and $a$, so that $\E[r_{t,a^*}] = \frac{1}{2} + \frac{1}{2}\Delta$ and $\E[r_{t,a}] = \frac{1}{2} - \frac{1}{2}\Delta$. (The rewards may be generated dynamically as you run the algorithms and, actually, you only need them for the actions that are played by the algorithms.)
    \item Run the experiment with $\Delta = \frac{1}{4}$, $\Delta = \frac{1}{8}$, and $\Delta = \frac{1}{16}$. (Three different experiments.)
    \item Take $T = 100000$. (In general, the time horizon should be large in relation to $\frac{1}{\Delta^2}$.)
    \item Plot the empirical pseudo regret defined by $\hat R_t = \sum_{s=1}^t \Delta(A_s)$ for the two algorithms as a function of time for $1\leq t\leq T$. (To remind you: $A_s$ is the action taken by the algorithm in round $s$ and $\Delta(a) = \max_{a'} \E[r_{s,a'}] - \E[r_{s,a}] = \mu(a^*) - \mu(a)$.) To make the plot you should make 20 runs of each algorithm and plot the average pseudo regret over the 20 runs and the average pseudo regret + one standard deviation over the 20 runs. Do not forget to add a legend to your plot.
    \item Answer the following questions:
    \begin{itemize}
        \item Which values of $\Delta$ lead to higher regret? 
        \item What can you say about the relative performance of the two parametrizations?
    \end{itemize}
\end{itemize}

\end{enumerate}
\emph{Comment: The UCB1 algorithm in \Cref{sec:Stochastic-Bandits} takes confidence intervals $\sqrt{\frac{3\ln t}{2N_{t-1}(a)}} = \sqrt{\frac{\ln t^3}{2N_{t-1}(a)}}$, corresponding to confidence parameter $\delta = \frac{1}{t^3}$. The modified UCB1 algorithm in this question takes confidence intervals $\sqrt{\frac{2\ln t}{2N_{t-1}(a)}} = \sqrt{\frac{\ln t^2}{2N_{t-1}(a)}}$, corresponding to confidence parameter $\delta = \frac{1}{t^2}$. The original algorithm and analysis by \citet{ACBF02} uses confidence intervals $\sqrt{\frac{4\ln t}{2N_{t-1}(a)}} = \sqrt{\frac{\ln t^4}{2N_{t-1}(a)}}$, corresponding to confidence parameter $\delta = \frac{1}{t^4}$, due to one unnecessary union bound. The reason we can move from $\delta = \frac{1}{t^3}$ to $\delta = \frac{1}{\delta^2}$ is not due to elimination of additional union bounds (we still need two of them), but due to a compromise on the last term in the regret bound.} 
\end{exercise}

\begin{exercise}[\textit{Introduction of New Products}]
Imagine that we have an established product on the market, which sells with probability 0.5. We have received a new product, which sells with an unknown probability $\mu$. Assume that at every sales round we can offer only one product, so we have to choose between offering the established or the new product. Propose a strategy for maximizing the number of sales and analyze its pseudo-regret. Write your answer in terms of the gap $\Delta = 0.5 - \mu$. (The regret bound for $\Delta > 0$ is different from the regret bound for $\Delta < 0$, i.e., you should obtain two separate answers for positive and negative gap.)

\emph{Hint}: The solution is \emph{not} an application of an existing algorithm. You should design a \emph{new} algorithm tailored for the problem. The new algorithm will not be very different from something we have studied, but you have to make a small adaptation; this is the whole point of the question.

Pay attention that if $\mu > 0.5$ (the new product is better than the old one) then $\Delta < 0$, and if $\mu < 0.5$ (the new product is worse than the old one) then $\Delta > 0$. If you do things correctly, for $\Delta < 0$ the regret of your algorithm should be bounded by a constant that is independent of the number of game rounds $T$, and for $\Delta > 0$ it should grow logarithmically with $T$. The algorithm should not know whether $\Delta$ is positive or negative, but the analysis of the two cases can be done separately.    
\end{exercise}

\begin{exercise}[\textit{Empirical evaluation of algorithms for adversarial environments}]
Is it possible to evaluate experimentally the quality of algorithms for adversarial environments? If yes, how would you design such an experiment? If no, explain why it is not possible. 

\emph{Hint: Think what kind of experiments can certify that an algorithm for an adversarial environment is good and what kind of experiments can certify that the algorithm is bad? How easy or hard is it to construct the corresponding experiments?}    
\end{exercise}

\begin{exercise}[\textit{A tighter analysis of the Hedge algorithm}]~
\label{ex:TightHedge}
\begin{enumerate}
\item Apply Hoeffding's lemma (\Cref{lem:Hoeff}) in order to derive a better parametrization and a tighter bound for the expected regret of the Hedge algorithm (\Cref{algo:Hedge} in \Cref{sec:PredictionWithExpertAdvice}). [Do not confuse Hoeffding's lemma (\Cref{lem:Hoeff}) with Hoeffding's inequality (\Cref{thm:Hoeffding})!] Guidance:
\begin{enumerate}
	\item Traverse the analysis of the Hedge algorithm that we did in class. There will be a place where you will have to bound expectation of an exponent of a function ($\sum_a p_t(a) e^{-\eta X_{t,a}}$). Instead of going the way we did, apply Hoeffding's lemma.
	\item Find the value of $\eta$ that minimizes the new bound. (You should get $\eta = \sqrt{\frac{8 \ln K}{T}}$ - please, prove this formally.)
	\item At the end you should obtain $\E[R_T] \leq \sqrt{\frac{1}{2} T \ln K}$. (I.e., you will get an improvement by a factor of 2 compared to what we did in class.)
\end{enumerate}
\emph{Remark: Note that the regret upper bound matches the lower bound up to the constants. This is an extremely rare case.}
\item Explain why the same approach cannot be used to tighten the regret bound of the EXP3 algorithm.
\end{enumerate}
\end{exercise}

\begin{exercise}[\textit{Empirical comparison of FTL and Hedge}]
Assume that you have to predict a binary sequence $X_1,X_2,\dots$ and that you know that $X_i$-s are i.i.d.\ Bernoulli random variables with an unknown bias $\mu$. (You know that $X_i$-s are i.i.d., but you do not know the value of $\mu$.) At every round you can predict ``0'' or ``1'' (i.e., you have two actions - ``predict 0'' or ``predict 1'') and your loss is the zero-one loss depending on whether your prediction matches the outcome. The regret is computed with respect to the performance of the best out of the two possible actions.
\begin{enumerate}
	\item Write a simulation that compares numerically the performance of Follow The Leader (FTL) algorithm with performance of the Hedge algorithm (\Cref{algo:Hedge} in \Cref{sec:PredictionWithExpertAdvice}) with $\eta = \sqrt{\frac{2 \ln K}{T}}$, and performance of the reparametrized Hedge algorithm from \Cref{ex:TightHedge}, with $\eta = \sqrt{\frac{8 \ln K}{T}}$. The Hedge algorithm should operate with the aforementioned two actions. To make things more interesting we will add an ``anytime'' version of Hedge to the comparison. ``Anytime'' algorithm is an algorithm that does not depend on the time horizon. Let $t$ be a running time index ($t = 1, \dots, T$). Anytime Hedge corresponding to the simple analysis uses $\eta_t = \sqrt{\frac{\ln K}{t}}$ and anytime Hedge corresponding to the tighter analysis in \Cref{ex:TightHedge} uses $\eta_t = 2 \sqrt{\frac{\ln K}{t}}$ (the learning rate $\eta_t$ of anytime Hedge changes with time and does not depend on the time horizon). Some instructions for the simulation:
	\begin{itemize}
		\item Take time horizon $T = 2000$. (In general, the time horizon should be large in comparison to $\frac{1}{\lr{\mu - \frac{1}{2}}^2}$.)
		\item Test several values of $\mu$. We suggest $\mu = \frac{1}{2} - \frac{1}{4}$, $\mu = \frac{1}{2} - \frac{1}{8}$, $\mu = \frac{1}{2} - \frac{1}{16}$.
		\item Plot the empirical pseudo regret defined by $\hat R_t = \sum_{s=1}^t \Delta(A_s)$ of the five algorithms with respect to the best out of ``0'' and ``1'' actions as a function of $t$ for $1\leq t\leq T$ and for the different values of $\mu$ (make a separate plot for each $\mu$). 
Make 10 runs of each algorithm and report the average empirical pseudo regret over the 10 runs and the average empirical pseudo regret + one standard deviation over the 10 runs. (Generate a new sequence $X_1,X_2,\dots$ for each run of the algorithm.) Do not forget to add a legend to your plot.
	\end{itemize}
	\item Which values of $\mu$ lead to higher regret? How the relative performance of the algorithms evolves with time and does it depend on $\mu$?
	\item Design an adversarial (non-i.i.d.) sequence, which makes the FTL algorithm perform poorly. Ideally, your solution should not depend on the tie breaking approach of FTL. (If you need to make assumptions about tie breaking, please, state them clearly. We may take a few points, because it would make the problem easier.) Explain the design of your adversarial sequence and report a plot with a simulation, where you compare the performance of FTL with the different versions of Hedge. 
As before, make 10 repetitions of the experiment and report the average regret (in this case you should use regret and not pseudo regret) and the average + one standard deviation. Comment on your observations.
\end{enumerate}
\end{exercise}

\begin{exercise}[\textit{The doubling trick}]
\label{ex:doubling}
Consider the following forecasting strategy (the ``doubling trick''): time is divided into periods $(2^m, \dots , {2^{m+1}-1})$, where $m = 0, 1, 2, \dots$. (In other words, the periods are $(1), (2,3), (4,\dots,7), (8,\dots,15), \dots$.) In period $(2^m,\dots, 2^{m+1} - 1)$ the strategy uses the optimized Hedge forecaster (from the previous question) initialized at time $2^m$ with parameter $\eta_m = \sqrt{\frac{8 \ln K}{2^m}}$. Thus, the Hedge forecaster is reset at each time instance that is an integer power of 2 and restarted with a new value of $\eta$. By the analysis of optimized Hedge we know that with $\eta_m = \sqrt{\frac{8 \ln K}{2^m}}$ its expected regret within the period $(2^m, \dots , 2^{m+1} - 1)$ is bounded by $\sqrt{\frac{1}{2} 2^m \ln K}$.
\begin{enumerate}
	\item Prove that for any $T = 2^m - 1$ the overall expected regret (considering the time period $(1,\dots,T)$) of this forecasting strategy satisfies
	\[
	\E[R_T] \leq \frac{1}{\sqrt 2 - 1} \sqrt{\frac{1}{2} T \ln K}.
	\]
	\emph{(Hint: at some point in the proof you will have to sum up a geometric series.)}
	\item Prove that for any arbitrary time $T$ the expected regret of this forecasting strategy satisfies
	\[
	\E[R_T] \leq \frac{\sqrt 2}{\sqrt 2 - 1} \sqrt{\frac{1}{2} T \ln K}.
	\]
\end{enumerate}
\emph{Remark: The expected regret of ``anytime'' Hedge with $\eta_t = 2 \sqrt{\frac{\ln K}{t}}$ satisfies $\E[R_T] \leq \sqrt{T \ln K}$ for any $T$. For comparison, $\frac{1}{\sqrt 2 (\sqrt 2 - 1)} \approx 1.7$ and $\frac{1}{\sqrt 2 - 1} \approx 2.4$. Thus, anytime Hedge is both more elegant and more efficient than Hedge with the doubling trick.}
\end{exercise}

\begin{exercise}[\textit{Regularization by relative entropy and the Gibbs distribution}]
In this question we will show that regularization by relative entropy leads to solutions in a form of the Gibbs distribution. Let's assume that we have a finite hypothesis class $\mathcal H$ of size $m$ and we want to minimize
\[
\mathcal{F}(\rho) = \alpha \E_\rho\lrs{\hat L(h,S)} + \KL(\rho\|\pi) = \alpha \sum_{h=1}^m \rho(h) \hat L(h,S) + \sum_{h=1}^m \rho(h) \ln \frac{\rho(h)}{\pi(h)}
\]
with respect to the distribution $\rho$. This objective is closely related to the objective of PAC-Bayes-$\lambda$ inequality when $\lambda$ is fixed and this sort of minimization problem appears in many other places in machine learning. Let's slightly simplify and formalize the problem. Let $\rho = (\rho_1, \dots, \rho_m)$ be the posterior distribution, $\pi = (\pi_1,\dots,\pi_m)$ the prior distribution, and $L = (L_1,\dots,L_m)$ the vector of losses. You should solve
\begin{align}
\min_{\rho_1,\dots,\rho_m} &\qquad   \alpha \sum_{h=1}^m \rho_h L_h+ \sum_{h=1}^m \rho_h \ln \frac{\rho_h}{\pi_h}\label{eq:min}\\
s.t. ~~~ & \qquad \sum_{h=1}^m \rho_h = 1\notag\\
     & \qquad \forall h: \rho_h \geq 0\notag
\end{align}
and show that the solution is $\rho_h = \frac{\pi_h e^{-\alpha L_h}}{\sum_{h'=1}^m \pi_{h'} e^{-\alpha L_{h'}}}$. Distribution of this form is known as the Gibbs distribution.

\paragraph{Guidelines:}
\begin{enumerate}
	\item Take a shortcut. Instead of solving minimization problem \eqref{eq:min}, solve the following minimization problem
\begin{align}
\min_{\rho_1,\dots,\rho_m} & \qquad  \alpha \sum_{h=1}^m \rho_h L_h + \sum_{h=1}^m \rho_h \ln \frac{\rho_h}{\pi_h}\label{eq:min-2}\\
s.t. ~~~ & \qquad \sum_{h=1}^m \rho_h = 1,\notag
\end{align}
i.e., drop the last constraint in \eqref{eq:min}.
\item Use the method of Lagrange multipliers to show that the solution of the above problem has a form of $\rho_h = \pi_h e^{-\alpha L_h + \texttt{something}}$, where $\texttt{something}$ is something involving the Lagrange multiplier.
\item Show that $\rho_h \geq 0$ for all $h$. (This is trivial. But it gives us that the solutions of \eqref{eq:min} and \eqref{eq:min-2} are identical.)
\item Finally, $e^\texttt{something}$ should be such that the constraint $\sum_{h=1}^m \rho_h = 1$ is satisfied. So you can easily get the solution. You even do not have to compute the Lagrange multiplier explicitly.
\end{enumerate}    
\end{exercise}

\begin{exercise}[\textit{Empirical comparison of UCB1 and EXP3 algorithms}]
Implement and compare the performance of UCB1 with improved parametrization derived in \Cref{ex:ImprovedUCB1} and EXP3 in the i.i.d.\ multiarmed bandit setting. For EXP3 take time-varying $\eta_t = \sqrt{\frac{\ln K}{t K}}$. 
\begin{enumerate}
\item Use the following settings:
\begin{itemize}
	\item Time horizon $T = 10000$.
	\item Take a single best arm with Bernoulli distribution with bias $\mu^* = 0.5$.
	\item Take $K-1$ suboptimal arms for $K = 2, 4, 8, 16$.
	\item For suboptimal arms take Bernoulli distributions with bias $\mu = \mu^* - \frac{1}{4}$, $\mu = \mu^* - \frac{1}{8}$, $\mu = \mu^* - \frac{1}{16}$ (in total 12 different experiments corresponding to the four values of $K$ and three values of $\mu$, with all suboptimal arms in each experiment sharing the same $\mu$).
\end{itemize}
Make 20 repetitions of each experiment and for each experiment plot the average empirical pseudo regret (over the 20 repetitions) as a function of time and the average empirical pseudo regret + one standard deviation (over the 20 repetitions). The empirical pseudo regret is defined as $\hat R_T = \sum_{t=1}^T \Delta(A_t)$.

Important: do not forget to add a legend and labels to the axes.

\item \textbf{\textit{Break UCB1}} ~~Now design an adversarial sequence for which you expect UCB1 to suffer linear regret. The sequence should be oblivious (i.e., it should be independent of the actions taken by UCB1). Ideally, your solution should not depend on the tie breaking approach of UCB1. Explain how you design the sequence, and execute UCB1 and EXP3 on it and report the observations (in a form of a plot). You should make several repetitions of the experiment (say, 20), because there is internal randomness in the algorithms. 

\emph{Hint: there will always be a tie in the very first round (i.e., in the initial ``play each action once'' phase, if UCB1 starts in random order), so you need to find a way how to start. After that ties are unlikely, and you can avoid them altogether by using rewards in $[0,1]$ rather than $\{0,1\}$. You need to explain how.}
\end{enumerate}


You are welcome to try other settings (not for submission). For example, check what happens when $\mu^*$ is close to 1 or 0. Or what happens when the best arm is not static, but switches between rounds. Even though you can break UCB1, it is actually pretty robust, unless you design adversarial sequences that exploit the knowledge of the algorithm.
\end{exercise}

\begin{exercise}[\textit{Rewards vs.\ losses}]
\label{ex:rewardsvslosses}
The original EXP3 algorithm for multiarmed bandits was designed for the game with rewards rather than losses (see \citet[Page 6]{ACB+02}). In the game with rewards we have an infinite matrix of rewards $\lrc{r_t^a}_{a \in \lrc{1,\dots,K}, t \geq 1}$, where $r_t^a \in [0,1]$. On each round of the game the algorithm plays an action $A_t$ and accumulates and observed reward $r_t^{A_t}$. The remaining rewards $r_t^a$ for $a \neq A_t$ remain unobserved.

On the one hand, we can easily convert rewards into losses by taking $\ell_t^a = 1 - r_t^a$ and apply the EXP3 algorithm we saw in class. On the other hand, a bit surprisingly, working directly with rewards (as Auer et al.\ did) turns to be more cumbersome and less efficient. The high-level reason is that the games with rewards and losses have a different dynamics. In the rewards game when an action is played its relative quality (expressed by the cumulative reward) increases. Therefore, we need explicit exploration to make sure that we do not get locked on a suboptimal action. In the losses game when an action is played its relative quality (expressed by the cumulative loss) decreases. Therefore, we never get locked on any particular action and exploration happens automatically without the need to add it explicitly (sometimes this is called implicit exploration). The low-level reason when it comes down to the analysis of the algorithm is that it is easier to upper bound the exponent of $x$ for negative $x$ as opposed to positive $x$.

The original EXP3 algorithm for the rewards game looks as follows, where the most important difference with the algorithm for the losses game is highlighted in red and two additional minor differences are highlighted in blue (we explicitly emphasize that the sign in the exponent changes from ``-'' to ``+''). $\tilde R_t$ is used to denote cumulative importance-weighted rewards.

\begin{algorithm}[ht!]
\caption{The EXP3 Algorithm for the game with rewards and fixed time horizon}
\begin{algorithmic}[1]
\State $\forall a: \tilde R_0(a) = 0$\;
%
\For{$t = 1,\dots,T$}
\State $\displaystyle \forall a: ~ p_t(a) = {\color{blue}{(1 - \eta)}} \frac{e^{{\color{blue}{+}}\eta \tilde R_{t-1}(a)}}{\sum_{a'} e^{{\color{blue}{+}}\eta \tilde R_{t-1}(a')}} {\color{red}{+ \frac{\eta}{K}}}$\;
%
\State Sample $A_t$ according to $p_t$ and play it\;
%
\State Observe $r_{t,{A_t}}$\;
%
\State $\forall a: ~ \tilde r_{t,a} = \frac{r_{t,a} \1[A_t = a]}{p_t(a)} = \begin{cases}\frac{r_{t,a}}{p_t(a)}, &\text{if } A_t = a\\~~0, &\text{otherwise} \end{cases}$\;
%
\State $\forall a: ~ \tilde R_t(a) = \tilde R_{t-1}(a) + \tilde r_{t,a}$\;
\EndFor
\end{algorithmic}
\end{algorithm}

\begin{enumerate}
	\item Explain why the analysis of the EXP3 algorithm for the rewards game without the addition of explicit exploration $\frac{\eta}{K}$ in Line 3 of the algorithm would not work. More specifically - if you would try to follow the lines of the analysis of EXP3 with losses, at which specific point you would get stuck and why?
	\item How the addition of explicit exploration term $\frac{\eta}{K}$ in Line 3 of the algorithm allows the analysis to go through? (You can check the analysis of the algorithm in \citet[Page 7]{ACB+02}.)
	\item By how much the expected regret guarantee for EXP3 with rewards is weaker than the expected regret guarantee for EXP3 with losses? (Check \citet[Corollary 3.2]{ACB+02} and assume that $g$ takes its worst-case value, which is $T$.)
	\item You are mostly welcome to experiment and see whether theoretical analysis reflects the performance in practice. I.i.d.\ experiments will be the easiest to construct, but you are also welcome to try adversarial settings.
\end{enumerate}
\end{exercise}

\begin{exercise}[\textit{Offline Evaluation of Bandit Algorithms}]~

\begin{enumerate}
	\item Evaluation of algorithms for online learning with limited feedback in real life (as opposed to simulations) is a challenging topic. The straightforward way is to implement an algorithm, execute it ``live'', and measure the performance, but most often this is undesirable. Give two reasons why.

	\item There are two alternative offline evaluation methods: importance-sampling and rejection sampling. In both cases we have to know the distribution that was used for collecting the data. Note that we only observe the reward (or loss) for an action taken by the algorithm when the action matches the action of the logging policy. In the importance-sampling approach we reweigh the reward by inverse probability of the action being taken by the logging policy when there is a match and assign zero reward otherwise. The rejection sampling approach requires that the logging policy samples all actions uniformly at random. At the evaluation phase rejection sampling scrolls through the log of events until the first case where the action of the logging policy matches the action of the evaluated policy. The corresponding reward is assigned and all events that were scrolled over are discarded. You can read more about the rejection sampling approach in \citet{LCLW11}. The importance-sampling approach is more versatile, because it does not require a uniform logging policy. With importance-sampling it is possible to take data collected by an existing policy and evaluate new policies, as long as the logging distribution is known and strictly positive for all actions.
	
	\paragraph{\textit{The Theoretical Part of the Task}} Our theoretical aim is to modify the UCB1 and EXP3 algorithms to be able to apply them to logged data using the importance-sampling approach. For simplicity, we assume that the logging policy used uniform sampling. Pay attention that importance weighting in offline evaluation based on uniform sampling (the one you are asked to analyze) changes the range of the rewards from $[0,1]$ to $[0,K]$, where $K$ is the number of actions. Recall that the original versions of UCB1 and EXP3 assumed that the rewards are bounded in the $[0,1]$ interval. Your task is to modify the two algorithms accordingly. 
	Pay attention that the variance of the importance weighted estimates is ``small'' (of order $K$ rather than of order $K^2$) and if you do the analysis carefully, you should be able to exploit it in the modified EXP3, but not in UCB1.
	
	\begin{enumerate}
	    \item Modify the UCB1 algorithm with improved parametrization from \Cref{ex:ImprovedUCB1} to work with importance-weighted losses generated by a logging policy based on uniform sampling. Provide a pseudo-code of the modified algorithm (at the same level as the UCB1 pseudo-code in \Cref{algo:UCB1}) and all the necessary calculations supporting your modification, including a pseudo-regret bound. You do not need to redo the full derivation, it is sufficient to highlight the key points where you make changes and how they affect the regret bound, assuming you do it clearly.
	    \item Briefly explain why you are unable to exploit the small variance in the modified UCB1.
	    \item Modify the EXP3 algorithm with a fixed learning rate $\eta$ and known time horizon $T$ to work with importance-weighted losses generated by a logging policy based on uniform sampling. Provide a pseudo-code of the modified algorithm (at the same level as the EXP3 pseudo-code in \Cref{algo:Hedge}) and all the necessary calculations supporting your modification, including an expected regret bound. As already mentioned, with a careful analysis you should be able to exploit the small variance of importance-weighted losses.
	    \item Anytime modification: In order to transform the fixed-horizon EXP3 from the previous task to an anytime EXP3 (an EXP3 that assumes no knowledge of the time horizon) you should replace the time horizon $T$ in the learning rate by the running time $t$ and reduce the learning rate by a factor of $\sqrt{2}$. The regret bound of anytime EXP3 is larger by a factor of $\sqrt{2}$ compared to the regret bound of EXP3 tuned for a specific $T$. In return, the bound holds for all $t$ and not just for one specific time $T$ the algorithm was tuned for. All you need to do for this point is to write the new learning rate and the new regret bound, you do not need to prove anything. You can find more details on the  derivation in \citep{BCB12}, if you want. In the experiments you should use the anytime version of the algorithm and the anytime expected regret bound.
	\end{enumerate}
	
	\paragraph{\textit{The Practical Part of the Task}} Now you should evaluate the modified UCB1 algorithm and the modified anytime EXP3 algorithm on the data. 

	In this question you will work with a preprocessed subset of R6B Yahoo! Front Page Today Module User Click Log Dataset\footnote{\url{https://webscope.sandbox.yahoo.com/catalog.php?datatype=r}}. The data is given in \texttt{data\_preprocessed\_features} file as space-separated numbers. There are 701682 rows in the file. Each row has the following format:
	\begin{enumerate}
		\item First comes the ID of the action that was taken (the ID of an article displayed to a user). The subset has 16 possible actions, indexed from 0 to 15, corresponding to 16 articles.
		\item Then comes the click indicator (0 = no click = no reward; 1 = click = reward). You may notice that the clicks are very sparse.
		\item And then you have 10 binary features for the user, which you can ignore. (Optionally, you can try to use the features to improve the selection strategy, but this is not for submission.)
	\end{enumerate}
	You are given that the actions were selected uniformly at random and you should work with importance-weighted approach in this question.
 
 In the following we refer to the quality of arms by their cumulative reward at the final time step $T=701682$. Provide plots as described in the next two points for the following subsets of arms:
	\begin{itemize}
		\item[i.] All arms.
		\item[ii.] Extract rounds with the best and the worst arm (according to the reward at $T$) and repeat the experiment with just these two arms. Pay attention that after the extraction you can assume that you make offline evaluation with just two arms that were sampled uniformly at random [out of two arms]. The time horizon will get smaller.
		\item[iii.] The same with the best and two worst arms.
		\item[iv.] The best and three worst arms.
		\item[v.] The best, the median, and the worst arm. (Since the number of arms is even, there are two median arms, the ``upper'' and the ``lower'' median; you can pick any of the two.)
	\end{itemize}
    
    \begin{enumerate}[resume]
		\item Provide one plot per each setup described above, where you report the estimate of the \emph{regret} of EXP3 and UCB1 based on offline importance-weighted samples as a function of time $t$. For each of the algorithms you should make 10 repetitions and report the mean and the mean $\pm$ one standard deviation over the repetitions. Pay attention that the regret at running time $t$ should be computed against the action that is the best at time $t$, not the one that is the best at the final time $T$!
		\item The same plot, where you add the expected regret bound for EXP3 and the regret of a random strategy, which picks actions uniformly at random. (We leave you to think why we are not asking to provide a bound for UCB1.)
		\item Discussion of the results.
	\end{enumerate}
	
	Optional, not for submission (for those who have taken ``Machine Learning B'' course): since the mean rewards are close to zero and have small variance, algorithms based on the $\kl$-inequality, such as $\kl$-UCB \citep{CGM+13}, or algorithms that are able to exploit small variance, for example, by using the Unexpected Bernstein inequality, are expected to be able to exploit the small variance of importance-weighted losses and perform much better than Hoeffding-based UCB1. 
\end{enumerate}  
\end{exercise}

\appendix

\chapter{Set Theory Basics}

In this chapter we provide a number of basic definitions and notations from the set theory that are used in the notes.

\paragraph{Countable and Uncountable sets} A set is called \emph{countable} if its elements can be counted or, in other words, if every element in a set can be associated with a natural number. For example, the set of integer numbers is countable and the set of rational numbers (ratios of two integers) is also countable. Finite sets are countable as well. A set is called \emph{uncountable} if its elements cannot be enumerated. For example, the set of real numbers $\R$ is uncountable and the set of numbers in a $[0,1]$ interval is also uncountable.

\paragraph{Relations between sets} For two sets $A$ and $B$ we use $A \subseteq B$ to denote that $A$ is a subset of $B$.

\paragraph{Operations on sets} For two sets $A$ and $B$ we use $A \cup B$ to denote the union of $A$ and $B$; $A \cap B$ the intersection of $A$ and $B$; and $A \setminus B$ the difference of $A$ and $B$ (the set of elements that are in $A$, but not in $B$). 

\paragraph{The empty set} We use $\emptyset$ or $\lrc{}$ to denote the empty set.

\paragraph{Disjoint sets} Two sets $A$ and $B$ are called \emph{disjoint} if $A \cap B = \emptyset$.

\chapter{Probability Theory Basics}
\label{app:prob-theory}

This chapter provides a number of basic definitions and results from the probability theory. It is partially based on \citet{MU05}. 

\section{Axioms of Probability}

We start with a definition of a probability space.

\begin{definition}[Probability space] A probability space is a tuple $\lr{\Omega, \F, \P}$, where
\begin{itemize}
	\item $\Omega$ is a \emph{sample space}, which is the set of all possible outcomes of the random process modeled by the probability space.
	\item $\F$ is a family of sets representing the allowable events, where each set in $\F$ is a subset of the sample space $\Omega$.
	\item $\P$ is a probability function $\P:\F\to[0,1]$ satisfying Definition~\ref{def:prob}.
\end{itemize}
\end{definition}

Elements of $\Omega$ are called \emph{simple} or \emph{elementary} events.

\begin{example} For coin flips the sample space is $\Omega = \lrc{H, T}$, where $H$ stands for ``heads'' and $T$ for ``tails''.

In dice rolling the sample space is $\Omega = \lrc{1, 2, 3, 4, 5, 6}$, where 1,\dots,6 label the sides of a dice (you should consider them as labels rather than numerical values, we get back to this later in Example~\ref{ex:labels}).

If we simultaneously flip a coin and roll a dice the sample space is\\
$\Omega = \lrc{(H,1), (T,1), (H,2), (T,2), \dots , (H,6), (T,6)}$.
\end{example}

If $\Omega$ is countable (including finite), the probability space is \emph{discrete}. In discrete probability spaces the family $\F$ consists of all subsets of $\Omega$. In particular, $\F$ always includes the empty set $\emptyset$ and the complete sample space $\Omega$. If $\Omega$ is uncountably infinite (for example, the real line or the $[0,1]$ interval) a proper definition of $\F$ requires concepts from the measure theory, which go beyond the scope of these notes.

\begin{example}
In the coin flipping experiment $\F = \lrc{\emptyset, \lrc{H}, \lrc{T}, \lrc{H,T}}$.
\end{example}

\begin{definition}[Probability Axioms] A probability function is any function $\P:\F\to\R$ that satisfies the following conditions
\begin{enumerate}
	\item For any event $E \in \F$, $0 \leq \P[E] \leq 1$.
	\item $\P[\Omega] = 1$.
	\item For any finite or countably infinite sequence of mutually disjoint events $E_1,E_2, \dots$
	\[
	\P[\bigcup_{i\geq 1} E_i] = \sum_{i\geq 1}\P[E_i].
	\]
\end{enumerate}
\label{def:prob}
\end{definition}

We now consider a number of basic properties of probabilities. 

\begin{lemma}[Monotonicity]
\label{lem:p-monoton}
Let $A$ and $B$ be two events, such that $A \subseteq B$. Then
\[
\P[A] \leq \P[B].
\]
\end{lemma}

\begin{proof}
We have that $B = A \cup (B \setminus A)$ and the events $A$ and $B \setminus A$ are disjoint. Thus,
\[
\P[B] = \P[A] + \P[B \setminus A] \geq \P[A],
\]
where the equality is by the third axiom of probabilities and the inequality is by the first axiom of probabilities, since $\P[B \setminus A] \geq 0$.
\end{proof}

The next simple, but very important result is known as the \emph{union bound}.

\begin{lemma}[The union bound]
For any finite or countably infinite sequence of events $E_1, E_2, \dots$,
\[
\P[\bigcup_{i\geq 1} E_i] \leq \sum_{i\geq 1} \P[E_i].
\]
\end{lemma}

\begin{proof}
We have
\[
\bigcup_{i\geq 1} E_i = E_1 \cup \lr{E_2 \setminus E_1} \cup \lr{E_3 \setminus \lr{E_1 \cup E_2}} \cup \dots = \bigcup_{i\geq 1} F_i,
\]
where the events $F_i = E_i \setminus \bigcup_{j=1}^{i-1} E_j$ are disjoint, $F_i \subseteq E_i$, and $\bigcup_{i\geq 1} F_i = \bigcup_{i\geq 1} E_i$. Therefore,
\[
\P[\bigcup_{i\geq 1} E_i] = \P[\bigcup_{i\geq 1} F_i] = \P[\bigcup_{i\geq 1} F_i]  = \sum_{i\geq 1} \P[F_i] \leq \sum_{i\geq 1} \P[E_i],
\]
where the second equality is by the third axiom of probabilities and the inequality is by monotonicity of the probability (Lemma~\ref{lem:p-monoton}).
\end{proof}

\begin{example} Let $E_1 = \lrc{1, 3, 5}$ be the event that the outcome of a dice roll is odd and $E_2 = \lrc{1, 2, 3}$ be the event that the outcome is at most 3. Then $\P[E_1 \cup E_2] = \P[1,2,3,5] \leq \P[E_1] + \P[E_2]$. Note that this is true irrespective of the choice of the probability measure $\P$. In particular, this is true irrespective of whether the dice is fair or not.
\end{example}

\begin{definition}[Independence]
Two events $A$ and $B$ are called \emph{independent} if and only if 
\[
\P[A \cap B] = \P[A] \cdot \P[B].
\]
\end{definition}

\begin{definition}[Pairwise independence] Events $E_1,\dots,E_n$ are called \emph{pairwise independent} if and only if for any pair $i,j$
\[
\P[E_i \cap E_j] = \P[E_i] \P[E_j].
\]
\end{definition}

\begin{definition}[Mutual independence] Events $E_1,\dots,E_n$ are called \emph{mutually independent} if and only if for any subset of indices $I \subseteq \lrc{1,\dots,n}$
\[
\P[\bigcap_{i \in I} E_i] = \prod_{i\in I} \P[E_i].
\]
\end{definition}

Note that pairwise independence does not imply mutual independence. Take the following example: assume we roll a fair tetrahedron (a three-dimensional object with four faces) with faces colored in red, blue, green, and the fourth face colored in all three colors, red, blue, and green. Let $E_1$ be the event that we observe red color, $E_2$ be the even that we observe blue color, and $E_3$ be the event that we observe green color. Then for all $i$ we have $\P[E_i] = \frac{1}{2}$ and for any pair $i\neq j$ we have $\P[E_i \cap E_j] = \frac{1}{4} = \P[E_i] \P[E_j]$. However, $\P[E_1 \cap E_2 \cap E_3] = \frac{1}{4} \neq \P[E_1] \P[E_2] \P[E_3]$ and, thus, the events are pairwise independent, but not mutually independent. If we say that events $E_1,\dots,E_n$ are independent without further specifications we imply mutual independence.

\begin{definition}[Conditional probability] The \emph{conditional probability} that event $A$ occurs given that event $B$ occurs is 
\[
\P[A|B] = \frac{\P[A \cap B]}{\P[B]}.
\]
The conditional probability is well-defined only if $\P[B] > 0$.
\end{definition}

By the definition we have that $\P[A \cap B] = \P[B] \P[A|B] = \P[A] \P[B|A]$.

\begin{example} For a fair dice let $A = \lrc{1, 6}$ and $B = \lrc{1, 2, 3, 4}$. Then
\begin{align*}
\P[A] &= \frac{1}{3},\\
\P[B] &= \frac{2}{3},\\
A \cap B &= \lrc{1},\\
\P[A \cap B] &= \frac{1}{6},\\
\P[A|B] &= \frac{\frac{1}{6}}{\frac{2}{3}} = \frac{1}{4}.
\end{align*}

\end{example}

\begin{lemma}[The law of total probability] Let $E_1, E_2, \dots, E_n$ be mutually disjoint events, such that $\bigcup_{i=1}^n E_n = \Omega$. Then
\[
\P[A] = \sum_{i=1}^n \P[A \cap E_i] = \sum_{i=1}^n \P[A|E_i] \P[E_i].
\]
\end{lemma}

\begin{proof}
Since the $E_i$-s are disjoint and cover the entire space it follows that $A = \bigcup_{i=1}^n (A \cap E_i)$ and the events $A \cap E_i$ are mutually disjoint. Therefore,
\[
\P[A] = \P[\bigcup_{i=1}^n (A \cap E_i)] = \sum_{i=1}^n \P[A \cap E_i] = \sum_{i=1}^n \P[A|E_i] \P[E_i].
\]
\end{proof}

\section{Discrete Random Variables}

We now define another basic concept in probability theory, a \emph{random variable}.

\begin{definition}
A \emph{random variable} $X$ on a sample space $\Omega$ is a real-valued function on $\Omega$, that is $X:\Omega\to\R$. A \emph{discrete} random variable is a random variable that takes on only a finite or countably infinite number of values.
\end{definition}

\begin{example}\label{ex:labels} For a coin we can define a random variable $X$, such that $X(H) = 1$ and $X(T) = 0$. We can also define another random variable $Y$, such that $Y(H) = 1$ and $Y(T) = -1$. 

For a dice we can define a random variable $X$, such that $X(1) = 1, X(2) = 2, X(3) = 3, X(4) = 4, X(5) = 5, X(6) = 6$. We can also define a random variable $Y$, such that $Y(1) = 3, Y(2) = 2.4, Y(3) = -6, Y(4) = 8, Y(5) = 8, Y(6) = 0$. This example emphasizes the difference between labeling of events and assignment of numerical values to events. Note that the random variable $Y$ does not distinguish between faces 4 and 5 of the dice, even though they are separate events in the probability space.

Functions of random variables are also random variables. In the last example, a random variable $Z = X^2$ takes values $Z(1) = 1, Z(2) = 4, Z(3) = 9, \dots, Z(6) = 36$.
\end{example}

\begin{definition}[Independence of random variables] Two random variables $X$ and $Y$ are \emph{independent} if and only if 
\[
\P[(X = x) \cap (Y = y)] = \P[X = x] \P[Y = y].
\]
for all values $x$ and $y$.
\end{definition}

\begin{definition}[Pairwise independence] Random variables $X_1,\dots,X_n$ are \emph{pairwise independent} if and only if for any pair $i,j$ and any values $x_i, x_j$
\[
\P[(X_i = x_i) \cap (X_j = x_j)] = \P[X_i = x_i] \P[X_j = x_j].
\]
\end{definition}

\begin{definition}[Mutual independence] Random variables $X_1,\dots,X_n$ are \emph{mutually independent} if for any subset of indices $I \subseteq \lrc{1,\dots,n}$ and any values $x_i$, $i \in I$
\[
\P[\bigcap_{i\in I} (X_i = x_i)] = \prod_{i \in I} \P[X_i = x_i].
\]
\end{definition}

Similar to the example given earlier, pairwise independence of random variables does not imply their mutual independence. If we say that random variables are independent without further specifications we imply mutual independence.

\section{Expectation}

\emph{Expectation} is the most basic characteristic of a random variable.

\begin{definition}[Expectation]
Let $X$ be a discrete random variable and let ${\cal X}$ be the set of all possible values that it can take. The \emph{expectation} of $X$, denoted by $\E[X]$, is given by
\[
\E[X] = \sum_{x \in {\cal X}} x \P[X = x].
\]
The expectation is finite if $\sum_{x \in {\cal X}} |x| \P[X=x]$ converges; otherwise the expectation is unbounded.
\end{definition}

\begin{example} For a fair dice with faces numbered 1 to 6 let $X(i) = i$ (the $i$-th face gets value $i$). Then
\[
\E[X] = \sum_{i=1}^6 i \frac{1}{6} = \frac{7}{2}.
\]
Take another random variable $Z = X^2$ then
\[
\E[Z] = \E[X^2] = \sum_{i=1}^6 i^2 \frac{1}{6} = \frac{91}{6}.
\]
\end{example}

Expectation satisfies a number of important properties (these properties also hold for continuous random variables). We leave a proof of these properties as an exercise.
\begin{lemma}[Multiplication by a constant]
For any constant $c$
\[
\E[cX] = c \E[X].
\]
\end{lemma}

\begin{theorem}[Linearity]
\label{thm:linexp}
For any pair of random variables $X$ and $Y$, not necessarily independent,
\[
\E[X+Y] = \E[X] + \E[Y].
\]
\end{theorem}

\begin{theorem}
\label{thm:independent-product}
If $X$ and $Y$ are independent random variables, then
\[
\E[XY] = \E[X] \E[Y].
\]
We emphasize that in contrast with Theorem~\ref{thm:linexp}, this property does not hold in the general case (if $X$ and $Y$ are not independent).
\end{theorem}

\section{Variance}

\emph{Variance} is the second most basic characteristic of a random variable.

\begin{definition}[Variance]
The \emph{variance} of a random variable $X$ (discrete or continuous), denoted by $\Var[X]$, is defined by
\[
\Var[X] = \E[\lr{X - \E[X]}^2] = \E[X^2] - \lr{\E[X]}^2.
\]
\end{definition}

We invite the reader to prove that $\E[\lr{X - \E[X]}^2] = \E[X^2] - \lr{\E[X]}^2$.

\begin{example} For a fair dice with faces numbered 1 to 6 let $X(i) = i$ (the $i$-th face gets value $i$). Then
\[
\Var[X] = \E[X^2] - \lr{\E[X]}^2 = \frac{91}{6} - \frac{49}{4} = \frac{35}{12}.
\]
\end{example}

\begin{theorem}
\label{thm:independent-var}
If $X_1,\dots,X_n$ are independent random variables then
\[
\Var[\sum_{i=1}^n X_i] = \sum_{i=1}^n \Var[X_i].
\]
\end{theorem}

The proof is based on Theorem~\ref{thm:independent-product} and the result does not necessarily hold when $X_i$-s are not independent. We leave the proof as an exercise.

\section{The Bernoulli and Binomial Random Variables}

Two most basic discrete random variables are Bernoulli and binomial.

\begin{definition}[Bernoulli random variable]
A random variable $X$ taking values $\lrc{0,1}$ is called a \emph{Bernoulli random variable}. The parameter $p = \P[X = 1]$ is called the \emph{bias} of $X$.
\end{definition}

Bernoulli random variable has the following property (which does not hold in general):
\[
\E[X] = 0 \cdot (1 - p) + 1 \cdot p = p = \P[X = 1].
\]

\begin{definition}[Binomial random variable]
A \emph{binomial random variable} $Y$ with parameters $n$ and $p$, denoted by $B(n,p)$, is defined by the following probability distribution on $k \in \lrc{0,1,\dots,n}$:
\[
\P[Y = k] = \binom{n}{k} p^k(1-p)^{n-k}.
\]
\end{definition}

Binomial random variable can be represented as a sum of independent identically distributed Bernoulli random variables.

\begin{lemma}
Let $X_1,\dots,X_n$ be independent Bernoulli random variables with bias $p$. Then $Y = \sum_{i=1}^n X_i$ is a binomial random variable with parameters $n$ and $p$.
\end{lemma}

A proof of this lemma is left as an exercise to the reader.

\section{Jensen's Inequality}

Jensen's inequality is one of the most basic in probability theory.

\begin{theorem}[Jensen's inequality]
\label{thm:Jensen}
If $f$ is a convex function and $X$ is a random variable, then
\[
\E[f(X)] \geq f\lr{\E[X]}.
\]
\end{theorem}

For a proof see, for example, \citet{MU05} or \citet{CT06}.

\chapter{Linear Algebra}
\label{app:lin-alg}

\begin{figure}%
\includegraphics[width=\columnwidth]{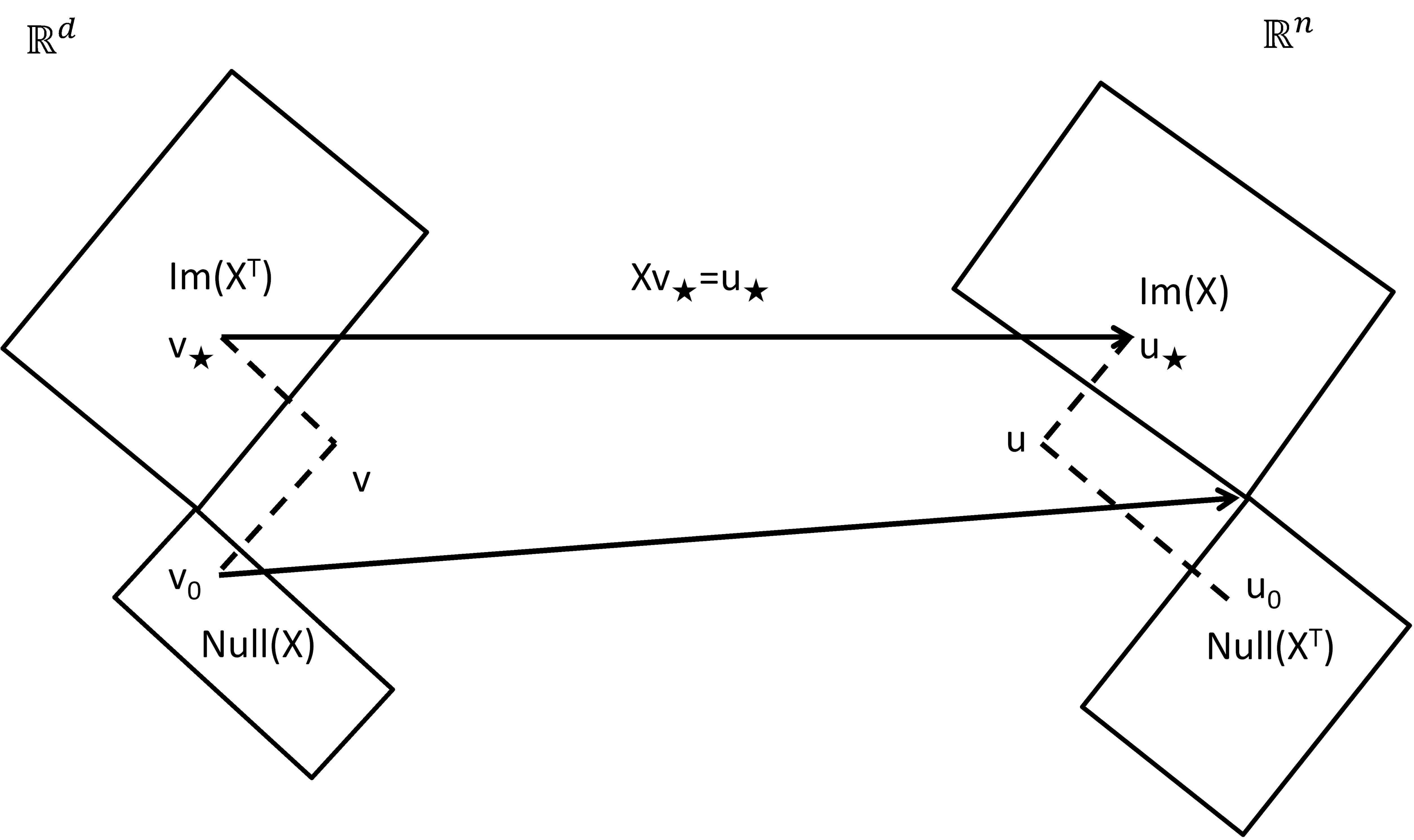}%
\caption{\textbf{The four fundamental subspaces of a matrix $\X$.} There is a right angle between $Im(\X)$ and $Null(\X^T)$, as well as between $Im(\X^T)$ and $Null(\X)$.}%
\label{fig:4spaces}%
\end{figure}

We revisit a number of basic concepts from linear algebra. This is only a brief revision of the main concepts that we are using in the book. For more details, please, refer to \cite{Str09} or some other textbook on linear algebra.

\paragraph{Vectors}
\begin{itemize}
    \item We use $\u = \lr{\begin{array}{c}u_1\\\vdots\\u_d\end{array}}$ to denote a \emph{vector} in $\R^d$. By default, vectors are column vectors.
    
    \item For two vectors $\u = \lr{\begin{array}{c}u_1\\\vdots\\u_d\end{array}}$ and $\v = \lr{\begin{array}{c}v_1\\\vdots\\v_d\end{array}}$ in $\R^d$, the \emph{inner product} is defined by $\u^T \v = \sum_{i=1}^d u_i v_i$. The same quantity is also known as the \emph{scalar product}, and the \emph{dot product}. An alternative notation is $\langle \u, \v \rangle = \u^T \v$. Note that $\u^T \v = \v^T \u$ and that the result is a scalar (a number). 

    \item Two vectors $\u$ and $\v$ are perpendicular, $\u \perp \v$, if and only if their inner product $\u^T \v = 0$.

    \item The \emph{outer product} is defined as $\u \v^T = \lr{\begin{array}{ccc}u_1 v_1 & \dots & u_1 v_d\\\vdots & \ddots & \vdots\\u_d v_1 & \dots & u_d v_d\end{array}}$. Note that the outer product is a matrix in $\R^{d\times d}$, and that $\u^T \v \neq \u \v^T$, unless $\u^T \v$ is a symmetric matrix. Also note that $\u^T \v$ is only defined when $\u$ and $\v$ have the same dimension, whereas $\u\v^T$ is defined also when the dimensions are not the same.

    \item If you consider $\u$ and $\v$ as matrices in $\R^{d\times 1}$, then it is easy to see that the definition of the inner and the outer product follow directly from the rules of matrix multiplication.
    
    \item The \emph{Euclidean norm} of a vector $\u$, which corresponds to the length of $\u$, is denoted by $\|\u\| = \sqrt{\sum_{i=1}^d u_i^2}$. Note that the square norm satisfies $\|\u\|^2 = \u^T\u$.
\end{itemize}

\paragraph{Matrices} A matrix $\X \in \R^{n \times d}$ takes vectors in $\R^d$ and maps them into $\R^n$. There are two fundamental subspaces associated with a matrix $\X$. The \emph{image} of $\X$, denoted $Im(\X) \subseteq \R^n$, is the space of all vectors $\v \in \R^n$ that can be obtained through multiplication of $\X$ with a vector $\w$. The image $Im(\X)$ is a linear subspace of $\R^n$ and it is also called a \emph{column space} of $\X$. The second subspace is the \emph{nullspace} of $\X$, denoted $Null(\X) \subseteq \R^d$, which is the space of all vectors $\w$ for which $\X \w = 0$. The nullspace is a linear subspace of $\R^d$. The subspaces are illustrated in Figure \ref{fig:4spaces}.

\paragraph{Matrix transpose} Matrix transpose $\X^T$ takes vectors in $\R^n$ and maps them into $\R^d$. The corresponding subspaces are $Im(\X^T)$, the \emph{row space} of $\X$, and $Null(\X^T)$.

\paragraph{Orthogonality of the fundamental subspaces $Im(\X) \perp Null(\X^T)$ and $Im(\X^T) \perp Null(\X)$} There is an important and extremely beautiful relation between the four fundamental subspaces associated with a matrix $\X$ and its transpose. Namely, the image of $\X$ is orthogonal to the nullspace of $\X^T$ and the image of $\X^T$ is orthogonal to the nullspace of $\X$. It means that if we take any two vectors $\u \in Im(\X)$ and $\v \in Null(\X^T)$ then $\u^T \v = 0$ (and the same for the second pair of subspaces). The proof of this fact is short and elegant. Any vector in $\u \in Im(\X)$ can be represented as a linear combination of the rows of $\X$, meaning that $\u = \X \z$. At the same time, by definition of a nullspace, if $\v \in Null(\X^T)$ then $\X^T \v = 0$. By putting these two facts together we obtain:
\[
\u^T \v = (\X \z)^T \v = \z^T \X^T \v = \z^T (\X^T \v) = 0.
\]

\paragraph{Complete relation between $Im(\X)$, $Im(\X^T)$, $Null(\X)$, and $Null(\X^T)$} Not only the pairs $Im(\X)$ with $Null(\X^T)$ and $Im(\X^T)$ with $Null(\X)$ are orthogonal, they also complement each other. Let $dim(\A)$ denote dimension of a matrix $\A$. The dimension is equal to the number of independent columns, which is equal to the number of independent rows (this fact can be shown by bringing $\A$ to a diagonal form). Then we have the following relations:
\begin{enumerate}
	\item $dim(Im(\X)) = dim(Im(\X^T)) = dim(\X)$.
	\item $dim(Null(\X)) = d - dim(Im(\X^T))$ and $dim(Null(\X^T)) = n - dim(Im(\X))$.
	\item $Im(\X) \perp Null(\X^T)$ and $Im(\X^T) \perp Null(X)$.
\end{enumerate}
Together these properties mean that a combination of bases for $Im(\X^T)$ and $Null(\X)$ makes a basis for $\R^d$ and a combination of bases for $Im(\X)$ and $Null(\X^T)$ make a basis for $\R^n$. It means that any vector $\v \in \R^d$ can be represented as $\v = \v_\star + \v_0$, where $\v_\star \in Im(\X^T)$ belongs to the row space of $\X$ and $v_0 \in Null(\X)$ belongs to the nullspace of $\X$. 

\paragraph{The mapping between $Im(\X^T)$ and $Im(\X)$ is one-to-one and, thus, invertible} Every vector $\u$ in the column space comes from one and only one vector in the row space $\v$. The proof of this fact is also simple. Assume that $\u = \X \v = \X \v'$ for two vectors $\v, \v' \in Im(\X^T)$. Then $\X(\v-\v') = 0$ and the vector $\v - \v' \in Null(\X)$. But $Null(\X)$ is perpendicular to $Im(\X^T)$, which means that $\v - \v'$ is orthogonal to itself and, therefore, must be the zero vector.

\paragraph{$\X^T \X$ is invertible if and only if $\X$ has linearly independent columns} $(\X^T \X)^{-1}$ is a very important matrix. We show that $\X^T \X$ is invertible if and only if $\X$ has linearly independent columns, meaning that $dim(X) = d$. We show this by proving that $\X$ and $\X^T \X$ have the same nullspace. Let $\v \in Null(\X)$, then $\X \v = 0$ and, therefore, $\X^T \X \v = 0$ and $\v \in Null(\X^T \X)$. In the other direction, let $\v \in Null(\X^T \X)$. Then $\X^T \X \v = 0$ and we have:
\[
\|\X \v\|^2 = (\X \v)^T (\X \v) = \v^T \X^T \X \v = \v^T (\X^T \X \v) = 0.
\]
Since $\|\X \v\|^2 = 0$ if and only if $\X \v = 0$, we have $\v \in Null(\X)$.

$\X^T \X$ is a $d \times d$ square matrix, therefore $dim(\X^T \X) = d - dim(Null(\X^T \X)) = d - dim(Null(\X))$ and matrix $\X^T \X$ is invertible if and only if the dimension of the nullspace of $\X$ is zero, meaning that $\X$ has linearly independent columns. (Note that unless $n = d$, $\X$ itself is a rectangular matrix and that inverses are not defined for rectangular matrices.)

\paragraph{Projection onto a line} A line in direction $\u$ is described by $\alpha \u$ for $\alpha \in \R$. Projection of vector $\v$ onto vector $\u$ means that we are looking for a projection vector $\p = \alpha \u$, such that the remainder $\v - \p$ is orthogonal to the projection. So we have:
\begin{align*}
(\v - \alpha \u)^T \alpha \u &= 0,\\
\alpha \v^T \u &= \alpha^2 \u^T \u,\\
\alpha &= \frac{\v^T \u}{\u^T \u} = \frac{\u^T \v}{\u^T \u}.
\end{align*}
Thus, the projection $\p = \alpha \u = \frac{\u^T \v}{\u^T \u} \u$. Note that $\frac{\u^T \v}{\u^T \u}$ is a scalar, thus 
\[
\p = \frac{\u^T \v}{\u^T \u} \u = \u \frac{\u^T \v}{\u^T \u} = \frac{\u \u^T}{\u^T \u} \v.
\]
The matrix $\PP = \frac{\u \u^T}{\u^T \u}$ is a \emph{projection matrix}. For any vector $\v$ the matrix $\PP$ projects $\v$ onto $u$.

\paragraph{Projection onto a subspace} A subspace can be described by a set of linear combinations $\A \z$, where the columns of matrix $\A$ span the subspace. Projection of a vector $\v$ onto a subspace described by $\A$ means that we are looking for a projection $\p = \A \z$, such that the remainder $\v - \p$ is perpendicular to the projection. The projection $\p = \A \z$ belongs to the image of $\A$, $Im(\A)$. Thus, the remainder must be in the nullspace of $\A^T$, meaning that $\A^T(\v - \p) = 0$. Assuming that the columns of $A$ are independent, we have:
\begin{align*}
\A^T(\v - \A \z) &= 0,\\
\A^T \v &= \A^T \A \z,\\
\z &= (\A^T \A)^{-1} \A^T \v,
\end{align*}
where we used independence of the columns of $\A$ in the last step to invert $\A^T \A$. The projection is $\p = \A \z = \A (\A^T \A)^{-1} \A^T \v$ and the projection matrix is $\PP = \A (\A^T \A)^{-1} \A^T$. The projection matrix $\PP$ maps any vector $\v$ onto the space spanned by the columns of $\A$, $Im(\A)$. Note how $(\A^T \A)^{-1}$ plays the role of $\frac{1}{\u^T \u}$ in projection onto a line.

\paragraph{Projection matrices} Projection matrices satisfy a number of interesting properties:
\begin{enumerate}
	\item If $\PP$ is a projection matrix then $\PP^2 = \PP$ (the second projection does not change the vector).
	\item If $\PP$ is a projection matrix projecting onto a subspace described by $\A$ then $\I - \PP$ is also a projection matrix. It projects onto a subspace that is perpendicular to the subspace described by $\A$.
\end{enumerate}

\chapter{Calculus}
\label{app:calculus}

We revisit some basic concepts from calculus.

\section{Gradients} Gradients are vectors of partial derivatives. For a vector $\x = \left(\begin{array}{c}x_1\\\vdots\\x_d\end{array}\right)$ and a function $f(\x)$ the gradient of $f$ is defined as 
\[
\nabla f(\x) = \left(\begin{array}{c}\frac{\partial f}{\partial x_1}\\\vdots\\\frac{\partial f}{\partial x_d}\end{array}\right).
\]

\subsection*{Gradient of a multivariate quadratic function $f(\x) = \x^T A \x$} Let $A$ be a matrix with entries $a_{ij}$. Then
\[
f(\x) = \x^T A \x = (x_1,\cdots,x_d) \left(\begin{array}{ccc}a_{11} & \cdots &a_{1d}\\\vdots & & \vdots\\a_{d1} & \cdots & a_{dd}\end{array}\right) \left(\begin{array}{c}x_1\\\vdots\\x_d\end{array}\right) = (x_1,\cdots,x_d) \left(\begin{array}{c}\sum_{j=1}^d a_{1j}x_j\\\vdots\\\sum_{j=1}^d a_{dj}x_j\end{array}\right) = \sum_{i=1}^d \sum_{j=1}^d a_{ij} x_i x_j.
\]
The partial derivative $\frac{\partial f}{\partial x_k}$ then becomes:
\[
\frac{\partial f}{\partial x_k} = \frac{\partial\left( \sum_{i=1}^d \sum_{j=1}^d a_{ij} x_i x_j\right)}{\partial x_k} = \sum_{j = 1}^d a_{kj} x_j + \sum_{i=1}^d a_{ik} x_i,
\]
where the first sum corresponds to the first element in the product $x_i x_j$ being $x_k$ and the second sum corresponds to the second element in the product $x_i x_j$ being $x_k$. Putting all the derivatives together we obtain:
\begin{align*}
\nabla f(\x) &= \left(\begin{array}{c}\sum_{j = 1}^d a_{1j} x_j + \sum_{i=1}^d a_{i1} x_i \\ \vdots \\ \sum_{j = 1}^d a_{d1} x_j + \sum_{i=1}^d a_{id} x_i \end{array}\right)\\
& = \left(\begin{array}{ccc}a_{11} & \cdots &a_{1d}\\\vdots & & \vdots\\a_{d1} & \cdots & a_{dd}\end{array}\right) \left(\begin{array}{c}x_1\\\vdots\\x_d\end{array}\right) + \left ((x_1,\cdots,x_d) \left(\begin{array}{ccc}a_{11} & \cdots &a_{1d}\\\vdots & & \vdots\\a_{d1} & \cdots & a_{dd}\end{array}\right)\right)^T\\
& = A \x + A^T \x\\
& = (A + A^T) \x.
\end{align*}
A matrix $A$ is called \emph{symmetric} if $A^T = A$. For a symmetric matrix we have $\nabla f(\x) = 2 A \x$ and for a general matrix we have $\nabla f(\x) = (A + A^T) \x$. Note the similarity and dissimilarity with the derivative of a univariate quadratic function $f(x) = a x^2$, which is $f'(x) = 2ax$.

\subsection*{Gradient of a linear function $f(\x) = \mathbf{b}^T \x$}
Let $\mathbf{b} = \left(\begin{array}{c}b_1\\\vdots\\b_d\end{array}\right)$ be a vector and let $f(\x) = \mathbf{b}^T \x = \sum_{i=1}^d b_i x_i$. We leave it as an exercise to prove that the gradient $\nabla f(\x) = \left(\begin{array}{c}\frac{\partial f}{\partial x_1}\\\vdots\\\frac{\partial f}{\partial x_d}\end{array}\right)=\mathbf{b}$.

\chapter{Vectorized Implementation of the $K$ Nearest Neighbors Algorithm}
\label{app:KNN}

\newcommand{\xtrain}{\x^{\texttt{train}}}
\newcommand{\xtest}{\x^{\texttt{targ}}}
\newcommand{\Xtrain}{\X^{\texttt{train}}}
\newcommand{\Xtest}{\X^{\texttt{targ}}}

Python, as well as other interpreter programming languages, provides built-in precompiled functions for vector and matrix operations, making it more efficient to work with vector operations rather than explicitly loop through vectors in the code. The appendix explains how to implement the $K$-NN algorithm using vector and matrix operations. Please, check the ``Vectors'' paragraph in \Cref{app:lin-alg} on Linear Algebra for a definition and basic facts about vectors (you do not need the rest of the appendix for this section).

For two points $\x$ and $\z$ in $\R^d$, the \emph{Euclidean distance} between $\x$ and $\z$ is given by the norm of the vector from $\x$ to $\z$, $\dist(\x,\z) = \|\z - \x\|$. The square distance can then be written as
    \begin{equation}
    \label{eq:dist}
    \dist(\x,\z)^2 = \|\x-\z\|^2 = (\x - \z)^T(\x-\z) = \x^T\x - 2 \x^T\z + \z^T \z.
    \end{equation}
Next we show how to exploit the above vectorized representation of a distance between two points for vectorized computation of pairwise distances across sets of multiple points.

\paragraph{Efficient Computation of Distances}
In order to label a target point $\x$, the $K$-NN algorithm requires sorting training points by their distance to the target point. Note that sorting the points by distance is equivalent to sorting them by the square distance, which allows to save the computation of the square root in the definition of the Euclidean norm. We show how to use the square distance representation in Equation \eqref{eq:dist} for efficient vectorized computation of distances from all training points to all target points without using any for-loops. Let $\Xtrain = \lr{\lr{\begin{array}{c}|\\\xtrain_1\\|\end{array}},\cdots,\lr{\begin{array}{c}|\\\xtrain_m\\|\end{array}}} \in \R^{d\times m}$ be a set of $m$ training points and $\Xtest = \lr{\lr{\begin{array}{c}|\\\xtest_1\\|\end{array}},\cdots,\lr{\begin{array}{c}|\\\xtest_n\\|\end{array}}} \in \R^{d\times n}$ be a set of $n$ target points, written as column vectors in the corresponding matrices. Let $\xtrain_i$ be the $i$-th training point and $\xtest_j$ be the $j$-th target point, then 
\[
\dist(\xtrain_i,\xtest_j) = (\xtrain_i)^T\xtrain_i - 2(\xtrain_i)^T\xtest_j + (\xtest_j)^T\xtest_j.
\]
Our aim is to compute the matrix 
\[
D = \lr{\begin{array}{ccc}\dist(\xtrain_1,\xtest_1) & \dots & \dist(\xtrain_1 \xtest_n)\\\vdots & \ddots & \vdots\\\dist(\xtrain_m,\xtest_1) & \dots & \dist(\xtrain_m \xtest_n)\end{array}}\in \R^{m\times n}
\]
of all pairwise distances. 

\begin{itemize}
    \item For a square matrix $M = \lr{\begin{array}{ccc}m_{1,1} & \dots & m_{1,d}\\\vdots & \ddots & \vdots\\ m_{d,1} & \dots & m_{d,d}\end{array}} \in \R^{d\times d}$ let $\diag(M) = \lr{\begin{array}{c}m_{1,1} \\\vdots\\ m_{d,d}\end{array}}$ be a column vector of the diagonal elements of $M$. (You can find a built-in Python function for extracting the diagonal.)
    \item Let $\ones^d = \lr{\begin{array}{c}1 \\\vdots\\ 1\end{array}} \in \R^d$ be a vector of $d$ ones.
    \item Verify (convince yourself) that
    \[
    D = \diag\lr{(\Xtrain)^T \Xtrain} (\ones^n)^T - 2 (\Xtrain)^T \Xtest + \ones^m \lr{\diag\lr{(\Xtest)^T \Xtest}}^T.
    \]
    It should be a good idea to visualize the relevant vectors and matrices and their products to convince yourself. Appreciate the power of linear algebra --- the above expression provides distances from all the training points to all the target points in just one line without any for-loops!
\end{itemize}
\paragraph{Additional Guidance}
\begin{itemize}
	\item Note that for a single data point you can compute the output of $K$-NN for all $K$ in one shot using vector operations. No need in for-loops! And with a bit extra effort you should be able to do it without for-loops for the whole dataset.
	\item You may find the following functions useful:
	\begin{itemize}
		\item Built-in sorting functions for sorting the distances.
		\item Built-in functions for computing a cumulative sum of elements of a vector \texttt{v} (for computing the predictions of $K$-NN for all $K$ at once).
	\end{itemize}
	\item It may be a good idea to debug your code with a small subset of the data.
\end{itemize}

\backmatter

\bibliography{bibliography}

\end{document}